\documentclass{article} 
\usepackage{iclr2023_conference,times}

\usepackage{amsmath,amsfonts,bm}

\def\eqref#1{equation~\ref{#1}}

\def\1{\bm{1}}

\DeclareMathAlphabet{\mathsfit}{\encodingdefault}{\sfdefault}{m}{sl}
\SetMathAlphabet{\mathsfit}{bold}{\encodingdefault}{\sfdefault}{bx}{n}

\usepackage{hyperref}
\usepackage{url}

\usepackage{algorithm}
\usepackage{algorithmic}
\usepackage{tabularx}
\usepackage{amsmath}
\usepackage{amssymb}
\usepackage{bm}
\usepackage{graphicx}
\usepackage{booktabs}  
\usepackage{threeparttable} \usepackage{multirow}
\usepackage{cite}
\usepackage{subfigure}

\newtheorem{theorem}{\textbf{Theorem}}
\newtheorem{definition}{\textbf{Definition}}
\newtheorem{assumption}{\textbf{Assumption}}

\newtheorem{lemma}{\textbf{Lemma}}

\title{Asynchronous Distributed Bilevel Optimization}

\author{Yang Jiao  \\
Tongji University \\
\And
Kai Yang\thanks{Corresponding author.}\\
Tongji University\\
\And
Tiancheng Wu \\
Tongji University\\
\And
Dongjin Song \\
University of Connecticut\\
\And
Chengtao Jian \\
Tongji University
}

\iclrfinalcopy 
\begin{document}

\maketitle

\begin{abstract}
Bilevel optimization plays an essential role in many machine learning tasks, ranging from hyperparameter optimization to meta-learning. Existing studies on bilevel optimization, however, focus on either centralized or synchronous distributed setting. The centralized bilevel optimization approaches require collecting a massive amount of data to a single server, which inevitably incur significant communication expenses and may give rise to data privacy risks. Synchronous distributed bilevel optimization algorithms, on the other hand, often face the straggler problem and will immediately stop working if a few workers fail to
respond. As a remedy, we propose \textbf{A}synchronous \textbf{D}istributed \textbf{B}ilevel \textbf{O}ptimization (ADBO) algorithm. The proposed ADBO can tackle bilevel optimization problems with both nonconvex upper-level and lower-level objective functions, and its convergence is theoretically guaranteed. Furthermore, it is revealed through theoretical analysis that the iteration complexity of ADBO to obtain the $\epsilon$-stationary point is upper bounded by $\mathcal{O}(\frac{1}{{{\epsilon ^2}}})$. Thorough empirical studies on public datasets have been conducted to elucidate the effectiveness and efficiency of the proposed ADBO.
\end{abstract}

\section{Introduction}
Recently, bilevel optimization has emerged due to its popularity in various machine learning applications, \textit{e.g.}, hyperparameter optimization~\citep{khanduri2021near,liu2021value}, meta-learning~\citep{likhosherstov2021debiasing,ji2020convergence}, reinforcement learning~\citep{hong2020two,zhou2022romfac}, and neural architecture search \citep{jiang2020sp,jiao2022timeautoad}. In bilevel optimization, one optimization problem is embedded or nested with another. Specifically, the outer optimization problem is called the upper-level optimization problem and the inner optimization problem is called the lower-level optimization problem. A general form of the bilevel optimization problem can be written as,
\begin{equation}
\label{eq:1_new}
\begin{array}{l}
\mathop {\min } \; \;\; \; F({\boldsymbol{x}},{\boldsymbol{y}})\\
{\rm{s.t.}}\; \; \; {\boldsymbol{y}} = \mathop {\arg \min }\limits_{{\boldsymbol{y}’}} f({\boldsymbol{x}},{\boldsymbol{y}’})\\
{\rm{var.}}\; \; \; \; \; \; \boldsymbol{x}, \boldsymbol{y},
\end{array}
\end{equation}
where $F$ and $f$ denote the upper-level and lower-level objective functions, respectively. ${\boldsymbol{x}} \! \in \!  {\mathbb{R}^n}$ and ${\boldsymbol{y}} \! \in \! {\mathbb{R}^m}$ are variables. Bilevel optimization can be treated as a special case of constrained optimization since the lower-level optimization problem can be viewed as a constraint to the upper-level optimization problem \citep{sinha2017review}.

The proliferation of smartphones and Internet of Things (IoT) devices has generated a plethora of data in various real-world applications. Centralized bilevel optimization approaches require collecting a massive amount of data from distributed edge devices and passing them to a centralized server for model training. These methods, however,  may give rise to data privacy risks~\citep{subramanya2021centralized} and encounter communication bottlenecks ~\citep{subramanya2021centralized}. To tackle these challenges, recently, distributed algorithms have been developed to solve the decentralized
bilevel optimization problems~\citep{yang2022decentralized,chen2022decentralized,lu2022decentralized}. \citet{tarzanagh2022fednest} and \citet{li2022local} study the bilevel optimization problems under a federated setting. Specifically, the distributed bilevel optimization problem can be given by
\begin{align}
\label{eq:new_18}
\begin{array}{l}
\mathop {\min } {\rm{ }}\,\; F(\boldsymbol{x},\boldsymbol{y}) = \sum\limits_{i = 1}^N {{G_i}(\boldsymbol{x},\boldsymbol{y})} \\
{\rm{s.t.}}\; \boldsymbol{y} = \mathop {\arg \min }\limits_{{\boldsymbol{y}'}} f(\boldsymbol{x},{\boldsymbol{y}'}) = \sum\limits_{i = 1}^N {{g_i}(\boldsymbol{x},{\boldsymbol{y}'})}\\
{\rm{var.}}\; \; \; \; \;\; \; \; \;\; \boldsymbol{x}, \boldsymbol{y},
\end{array}
\end{align}
where $N$ is the number of workers (devices), ${G_i}$ and $g_i$ denote the local upper-level and lower-level objective functions, respectively. Although existing approaches have shown their success in resolving distributed bilevel optimization problems, they only focus on the synchronous distributed setting. Synchronous distributed methods may encounter the straggler problem~\citep{jiang2021asynchronous} and its speed is limited by the worker with maximum delay~\citep{chang2016asynchronous}. Moreover, synchronous distributed method
will immediately stop working if a few workers fail to respond~\citep{zhang2014asynchronous} (which is common in large-scale distributed systems). The aforementioned issues give rise to the following question:

$\qquad$ $\quad$ \textbf{\textit{Can}} \textbf{\textit{we}} \textbf{\textit{design}} \textbf{\textit{an}} \textbf{\textit{asynchronous}} \textbf{\textit{distributed}} \textbf{\textit{algorithm}} \textbf{\textit{for}} \textbf{\textit{bilevel}} \textbf{\textit{optimization?}}

To this end, we develop an \textbf{A}synchronous \textbf{D}istributed \textbf{B}ilevel \textbf{O}ptimization (ADBO) algorithm which is a single-loop algorithm and computationally efficient. The proposed ADBO regards the lower-level optimization problem as a constraint to the upper-level optimization problem, and utilizes cutting planes to approximate this constraint. Then, the approximate problem is solved in an \textbf{\textit{asynchronous}} \textbf{\textit{distributed}} \textbf{\textit{manner}} by the proposed ADBO.  We prove that even if both the upper-level and lower-level objectives are \emph{nonconvex}, the proposed ADBO is guaranteed to converge. The iteration complexity of ADBO is also theoretically derived. To facilitate the comparison, we not only present a centralized bilevel optimization algorithm in Appendix \ref{appendix:centralized BO}, but also compare the convergence results of ADBO to state-of-the-art bilevel optimization algorithms with both centralized and distributed settings in Table \ref{tab:convergence rate compare}.

\textbf{Contributions.} Our contributions can be summarized as:

1. We propose a novel algorithm, ADBO, to solve the bilevel optimization problem in an \textbf{\textit{asynchronous}} \textbf{\textit{distributed}} \textbf{\textit{manner}}. ADBO is a single-loop algorithm and is computationally efficient. To the best of our knowledge, it is the first work in tackling asynchronous distributed bilevel optimization problem.

2. We demonstrate that the proposed ADBO can be applied to bilevel optimization with \emph{nonconvex} upper-level and lower-level objectives \emph{with} \emph{constraints}. We also theoretically derive that the iteration complexity for the  proposed ADBO to obtain the $\epsilon$-stationary point is upper bounded by $\mathcal{O}(\frac{1}{{{\epsilon ^2}}})$.

3. Our thorough empirical studies justify the superiority of the proposed ADBO over the existing state-of-the-art methods.

    

    

\vspace{-1mm}
\section{Related Work}\label{Related Work}
\textbf{Bilevel optimization:} The bilevel optimization problem was firstly introduced by \citet{bracken1973mathematical}. In recent years, many approaches have been developed to solve this problem and they can be divided into three categories~\citep{gould2016differentiating}. The first type of approaches assume there is an analytical solution to the lower-level optimization problem (\textit{i.e.}, $\phi({\boldsymbol{x}}) = \mathop {\arg \min }_{{\boldsymbol{y}'}} f({\boldsymbol{x}},{\boldsymbol{y}'})$) \citep{zhang2021revisiting}. In this case, the bilevel optimization problem can be simplified to a single-level optimization problem (\textit{i.e.}, $\mathop {\min }_{{\boldsymbol{x}} } F\left({\boldsymbol{x}},\phi({\boldsymbol{x}})\right)$. Nevertheless, finding the analytical solution for the lower-level optimization problem is often very difficult, if not impossible. The second type of approaches replace the lower-level optimization problem with the sufficient conditions for optimality (\textit{e.g.}, KKT conditions) \citep{biswas2019literature,sinha2017review}. Then, the bilevel program can be reformulated as a single-level constrained optimization problem. However, the resulting problem could be hard to solve since it often involves a large number of constraints~\citep{ji2021bilevel,gould2016differentiating}. The third type of approaches are gradient-based methods~\citep{ghadimi2018approximation,hong2020two,liao2018reviving} that compute the hypergradient (or the estimation of hypergradient), \textit{i.e.}, $\frac{{\partial F({\boldsymbol{x}},{\boldsymbol{y}})}}{{\partial {\boldsymbol{x}}}} + \frac{{\partial F({\boldsymbol{x}},{\boldsymbol{y}})}}{{\partial {\boldsymbol{y}}}}\frac{{\partial {\boldsymbol{y}}}}{{\partial {\boldsymbol{x}}}}$, and use gradient descent to solve the bilevel optimization problems. Most of the existing bilevel optimziation methods focus on centralized settings and require collecting a massive amount of data from distributed edge devices (workers). This may give rise to data privacy risks~\citep{subramanya2021centralized} and encounter communication bottlenecks ~\citep{subramanya2021centralized}.

\textbf{Asynchronous distributed optimization:} 
To alleviate the aforementioned issues in the centralized setting, various distributed optimization methods can be employed. Distributed optimization methods can be generally divided into synchronous distributed methods and asynchronous distributed methods \citep{assran2020advances}. For synchronous distributed methods \citep{boyd2011distributed}, the master needs to wait for the updates from all workers before it proceeds to the next iteration. Therefore, it may suffer from the straggler problem and the speed is limited by the worker with maximum delay \citep{chang2016asynchronous}. \textcolor{black}{There are several advanced techniques have been proposed to make the synchronous algorithm more efficient, such as large batch size, warmup and so on \citep{goyal2017accurate,you2019large,huo2021large,liu2022communication,wang2020large}.} For asynchronous distributed methods \citep{chen2020asynchronous,matamoros2017asynchronous}, the master can update its variables once it receives updates from $S$ workers, \textit{i.e.}, active workers ($1 \! \le S \! \le N$, where $N$ is the number of all workers). The asynchronous distributed algorithm is strongly preferred for large scale distributed systems in practice since it does not suffer from the straggler problem \citep{jiang2021asynchronous}. \textcolor{black}{Asynchronous distributed methods \citep{wu2017decentralized,liu2017asynchronous} have been employed for many real-world applications, such as Google's DistBelief system \citep{dean2012large}, the training of  10 million YouTube videos \citep{le2013building}, federated learning for edge computing \citep{lu2019differentially,liu2021blockchain}.} Since the action orders of each worker are different in the asynchronous distributed setting, which will result in complex interaction dynamics \citep{jiang2021asynchronous}, the theoretical analysis for asynchronous distributed algorithms is usually more challenging than that of the synchronous distributed algorithms. \textcolor{black}{In summary, the synchronous and asynchronous algorithm have different application scenarios. When the delay of each worker is not much different, the synchronous algorithm suits better. While there are stragglers in the distributed system, the asynchronous algorithm is more preferred.} So far, existing works for distributed bilevel optimization only focus on the synchronous setting \citep{tarzanagh2022fednest,li2022local,chen2022decentralized}, how to design an asynchronous algorithm for distributed bilevel optimization remains under-explored. To the best of our knowledge, this is the first work that designs an asynchronous algorithm for distributed bilevel optimization.

\vspace{-1mm}

\section{Asynchronous Distributed Bilevel Optimization}

\vspace{-1mm}

In this section, we propose \textbf{A}synchronous \textbf{D}istributed \textbf{B}ilevel \textbf{O}ptimization (ADBO) to solve the distributed bilevel optimization problem in an asynchronous manner. First, we reformulate problem in Eq. (\ref{eq:new_18}) as a consensus problem \citep{matamoros2017asynchronous,chang2016asynchronous},
\vspace{-1mm}
\begin{align}
\label{eq:new_19}
\begin{array}{l}
\mathop {\min } \; \;  {\rm{ }}F(\{ {\boldsymbol{x}_i}\} ,\{ {\boldsymbol{y}_i}\} ,\boldsymbol{v},\boldsymbol{z}) = \sum\limits_{i = 1}^N {{G_i}(\boldsymbol{x}_i,\boldsymbol{y}_i)} \\
{\rm{s.t.}} \quad\, {\boldsymbol{x}_i} = \boldsymbol{v},{\rm{ }}i = 1, \cdots ,N\\
\qquad \;\{ {\boldsymbol{y}_i}\} ,\boldsymbol{z} = \mathop {\arg \min }\limits_{\{ {\boldsymbol{y}_i'}  \} ,{\boldsymbol{z}'}} f(\boldsymbol{v},\{ {\boldsymbol{y}_i'}  \}, \boldsymbol{z}' ) = \sum\limits_{i = 1}^N {{g_i}(\boldsymbol{v},{\boldsymbol{y}_i'})} \\
\qquad \qquad \qquad \quad {\boldsymbol{y}_i'}   = {\boldsymbol{z}'},i = 1, \cdots ,N\\
{\rm{var.}}\; \; \; \; \;\; \; \; \;\; \; \{{\boldsymbol{x}_i}\}, \{{\boldsymbol{y}_i}\},\boldsymbol{v}, \boldsymbol{z},
\end{array}
\end{align}
where ${\boldsymbol{x}_i} \! \in \! \mathbb{R}^{n}$ and ${\boldsymbol{y}_i} \! \in \! \mathbb{R}^{m}$ are local variables in $i^{\rm{th}}$ worker, $\boldsymbol{v}\! \in \! \mathbb{R}^{n} $ and $\boldsymbol{z}\! \in \! \mathbb{R}^{m} $ are the consensus variables in the master node. \textcolor{black}{The reformulation given in Eq. (\ref{eq:new_19}) is a consensus problem which allows to develop distributed training algorithms for bilevel optimization based on the parameter server architecture \citep{assran2020advances}. As shown in Figure \ref{fig:PS}, in parameter server architecture, the
communication is centralized around the master, and workers pull the consensus variables $\boldsymbol{v},\boldsymbol{z}$ from and send their local variables ${\boldsymbol{x}_i},{\boldsymbol{y}_i}$ to the master.   Parameter server training is a well-known data-parallel approach for scaling up machine learning model training on a multitude of machines \citep{verbraeken2020survey}.} Most of the existing bilevel optimization works in machine learning only consider the bilevel programs without upper-level and lower-level constraints~\citep{franceschi2018bilevel,yang2021provably,chen2022single} or bilevel programs with only upper-level (or lower-level) constraint~\citep{zhang2022revisiting,mehra2021penalty}. On the contrary, we focus on the bilevel programs (\textit{i.e.}, Eq. (\ref{eq:new_19})) with \textit{both} lower-level and upper-level constraints, which is more challenging. By defining $\phi({\boldsymbol{v}}) = \mathop {\arg \min }\limits_{\{ {{{\boldsymbol{y}}}_i'}\} ,{\boldsymbol{z}'}} \{ \sum_{i = 1}^N {{g_i}(\boldsymbol{v},{\boldsymbol{y}_i'})}: {{\boldsymbol{y}_i'}} \! = \! {\boldsymbol{z}'},i \! = \! 1, \!\cdots\! ,N\} $ and $h(\boldsymbol{v},\{ {\boldsymbol{y}_i}\} ,\boldsymbol{z})=||\left[ \begin{array}{l}
\{ {\boldsymbol{y}_i}\} \\
\boldsymbol{z}
\end{array} \right]- \phi ({\boldsymbol{v}})|{|^2}$, we can reformulate problem in Eq. (\ref{eq:new_19}) as:
\vspace{-1.5mm}
\begin{align}
\label{eq:new_20}
\begin{array}{l}
\mathop {\min }  \; \; \;\;   F(\{ {\boldsymbol{x}_i}\} , \!\{ {\boldsymbol{y}_i}\} ,\boldsymbol{v},\boldsymbol{z})\! =\! \sum\limits_{i = 1}^N {{G_i}({\boldsymbol{x}_i},{\boldsymbol{y}_i})} \\
{\rm{s.t.}}\qquad  {\boldsymbol{x}_i} = \boldsymbol{v}, i = 1, \cdots ,N\\
\qquad \quad \;  h(\boldsymbol{v},\{ {\boldsymbol{y}_i}\} ,\boldsymbol{z}) =0 \\
{\rm{var.}}\; \; \;\; \; \;  \{{\boldsymbol{x}_i}\}, \{{\boldsymbol{y}_i}\},\boldsymbol{v}, \boldsymbol{z}.
\end{array}
\end{align}
\textcolor{black}{To better clarify how ADBO works, we sketch the procedure of ADBO. Firstly, ADBO computes the estimate of the solution to lower-level optimization problem. Then, inspired by cutting plane method, a set of cutting planes is utilized to approximate the feasible region of the upper-level bilevel optimization problem. Finally, the asynchronous algorithm for solving the resulting problem and how to update cutting planes are proposed. The remaining contents are divided into four parts, \textit{i.e.}, estimate of solution to lower-level optimization problem, polyhedral approximation, asynchronous algorithm, updating cutting planes.}

\textcolor{black}{\textbf{3.1 Estimate of Solution to Lower-level Optimization Problem}}

A consensus problem, \textit{i.e.}, the lower-level optimization problem in Eq. (\ref{eq:new_19}), needs to be solved in a distributed manner if an exact $\phi ({\boldsymbol{v}})$ is desired.  Following existing works \citep{li2022local,gould2016differentiating,yang2021provably} for bilevel optimization, instead of pursuing the exact $\phi ({\boldsymbol{v}})$, an estimate of $\phi ({\boldsymbol{v}})$ could be utilized. For this purpose, we first obtain the first-order Taylor approximation of $g_i(\boldsymbol{v},\{ {\boldsymbol{y}_i'}  \} )$ with respect to $\boldsymbol{v}$, \textit{i.e.}, for a given point $\overline{\boldsymbol{v}}$,  $\widetilde{g_i}(\boldsymbol{v},\{ {\boldsymbol{y}_i'}  \} )=g_i(\overline{\boldsymbol{v}},\{ {\boldsymbol{y}_i'}  \} ) + {\nabla _{\boldsymbol{v}}}g_i(\overline{\boldsymbol{v}},\{ {\boldsymbol{y}_i'}  \} )^\top ({\boldsymbol{v}} - \overline{\boldsymbol{v}})$. Then, similar to many works that use $K$ steps of gradient descent (GD) to approximate the optimal solution of lower-level optimization problem \citep{ji2021bilevel,yang2021provably,liu2021towards}, we utilize the results after $K$ communication rounds between workers and master to approximate $\phi({\boldsymbol{v}})$. Specifically, given $\widetilde{g_i}(\boldsymbol{v},\{ {\boldsymbol{y}_i'}  \} )$, the augmented Lagrangian function of the lower-level optimization problem in Eq. (\ref{eq:new_19}) can be expressed as,
\vspace{-1mm}
\begin{equation}
\label{eq:97_22}
g_p(\boldsymbol{v},\{ \boldsymbol{y}_i'  \}, \boldsymbol{z}', \{\boldsymbol{\varphi} _i\}) = \sum_{i = 1}^N \left( {\widetilde{g_i}(\boldsymbol{v},\boldsymbol{y}_i'})+ \boldsymbol{\varphi} _i ^\top(\boldsymbol{y}_i' -  \boldsymbol{z}')+\frac{\mu}{2} ||\boldsymbol{y}_i' -  \boldsymbol{z}'||^2 \right),
\end{equation}
where $\boldsymbol{\varphi} _i \! \in \! \mathbb{R}^m$ is the dual variable, and $\mu\!>\!0$ is a penalty parameter. In $(k+1)^{{\rm{th}}}$ iteration, we have,

(1) Workers update their local variables as follows,
\begin{align}
\label{eq:97_35}
\boldsymbol{y}_{i,k+1}' = {\boldsymbol{y}_{i,k}'}   - {\eta _{\boldsymbol{y}}}{\nabla _{{\boldsymbol{y}_i}}}g_p(\boldsymbol{v},\{ \boldsymbol{y}_{i,k}'  \}, \boldsymbol{z}_{k}', \{\boldsymbol{\varphi} _{i,k}\}),
\end{align}
where ${\eta _{\boldsymbol{y}}}$ is the step-size. Then, workers transmit the local variables  $\boldsymbol{y}_{i,k+1}'$ to the master.

(2) After receiving updates from workers, the master updates variables as follows,
\begin{align}
\label{eq:97_36}
\boldsymbol{z}_{k+1}' = {\boldsymbol{z}_{k}'}   - {\eta _{\boldsymbol{z}}}{\nabla _{{\boldsymbol{z}}}}g_p(\boldsymbol{v},\{ \boldsymbol{y}_{i,k}'  \}, \boldsymbol{z}_{k}', \{\boldsymbol{\varphi} _{i,k}\}),
\end{align}
\vspace{-4.5mm}
\begin{align}
\label{eq:97_37}
\boldsymbol{\varphi}_{i, k+1} = {\boldsymbol{\varphi}_{i, k}}   + {\eta _{\boldsymbol{\varphi}}}{\nabla _{{\boldsymbol{\varphi}_i}}}g_p(\boldsymbol{v},\{ \boldsymbol{y}_{i,k+1}'  \}, \boldsymbol{z}_{k+1}', \{\boldsymbol{\varphi} _{i,k}\}),
\end{align}
where ${\eta _{\boldsymbol{z}}}$ and ${\eta _{\boldsymbol{\varphi}}}$ are step-sizes.  Next, the master broadcasts $\boldsymbol{z}_{k+1}'$ and $\boldsymbol{\varphi}_{i,k+1}$ to workers. 

\vspace{-0.5mm}

As mentioned above, we utilize the results after $K$ communication rounds to approximate $\phi({\boldsymbol{v}})$, \textit{i.e.}, 
\vspace{-2mm}
\begin{align}
\label{eq:97_23}
\phi({\boldsymbol{v}}) = \left[ \begin{array}{l}
\{ {\boldsymbol{y}_{i,0}'} - \sum_{k = 0}^{K - 1} {{\eta _{\boldsymbol{y}}}{\nabla _{{\boldsymbol{y}_i}}}g_p(\boldsymbol{v},\{ \boldsymbol{y}_{i,k}'  \}, \boldsymbol{z}_{k}', \{\boldsymbol{\varphi} _{i,k}\})} \} \vspace{2mm} \\ 
{\boldsymbol{z}_0'} - \sum_{k = 0}^{K - 1} {{\eta _{\boldsymbol{z}}}{\nabla _{{\boldsymbol{z}}}}g_p(\boldsymbol{v},\{ \boldsymbol{y}_{i,k}'  \}, \boldsymbol{z}_{k}', \{\boldsymbol{\varphi} _{i,k}\} )} 
\end{array} \right].
\end{align}

\textcolor{black}{\textbf{3.2 Polyhedral Approximation}}

Considering $\phi({\boldsymbol{v}})$ in Eq. (\ref{eq:97_23}), the relaxed problem with respect to the problem in Eq. (\ref{eq:new_20}) is,
\vspace{-0.5mm}
\begin{align}
\label{eq:new_21}
\begin{array}{l}
\mathop {\min }  \; \; \;\;   F(\{ {\boldsymbol{x}_i}\} , \!\{ {\boldsymbol{y}_i}\} ,\boldsymbol{v},\boldsymbol{z})\! =\! \sum\limits_{i = 1}^N {{G_i}({\boldsymbol{x}_i},{\boldsymbol{y}_i})} \\
{\rm{s.t.}}\qquad  {\boldsymbol{x}_i} = \boldsymbol{v}, i = 1, \cdots ,N\\
\qquad \quad \;  h(\boldsymbol{v},\{ {\boldsymbol{y}_i}\} ,\boldsymbol{z}) \le \varepsilon \\
{\rm{var.}}\; \; \;\; \; \;  \{{\boldsymbol{x}_i}\}, \{{\boldsymbol{y}_i}\},\boldsymbol{v}, \boldsymbol{z},
\end{array}
\end{align}
where $\varepsilon>0$ is a constant. Assuming that $h(\boldsymbol{v},\{ {\boldsymbol{y}_i}\} ,\boldsymbol{z})$ is a convex function with respect to $(\boldsymbol{v},\{ {\boldsymbol{y}_i}\} ,\boldsymbol{z})$, which is always satisfied when we set $K=1$ in Eq. (\ref{eq:new_21}) according to the operations  that preserve convexity \citep{boyd2004convex}. Since the sublevel set of a convex function is convex \citep{boyd2004convex}, the feasible set with respect to constraint $h(\boldsymbol{v},\{ {\boldsymbol{y}_i}\} ,\boldsymbol{z}) \! \le \! \varepsilon$ is a convex set. In this paper, inspired by the cutting plane method \citep{boyd2007localization,michalka2013cutting,franc2011cutting,yang2014distributed}, a set of cutting planes is utilized to approximate the feasible region with respect to constraint $h(\boldsymbol{v},\{ {\boldsymbol{y}_i}\} ,\boldsymbol{z}) \! \le \! \varepsilon$ in Eq. (\ref{eq:new_21}). The set of cutting planes forms a polytope, let ${\boldsymbol{\mathcal{P}}^{t}  }$ denote the polytope in $(t+1)^{{\rm{th}}}$ iteration, which can be expressed as,
\vspace{-2mm}
\begin{equation}
{\boldsymbol{\mathcal{P}}^{t}} = \{ {\boldsymbol{a}_l}\!^\top\!  \boldsymbol{v} + \sum\limits_{i = 1}^N {{\boldsymbol{b}_{i,l}}\!^\top \! {\boldsymbol{y}_i}}  +  {\boldsymbol{c}_l}\!^\top \! \boldsymbol{z} + {\kappa_l} \le 0,\, l \! =\! 1, \!\cdots\! ,|\boldsymbol{\mathcal{P}}^{t}|\},
\end{equation}

\vspace{-3mm}

where ${\boldsymbol{a}_l} \! \in \! \mathbb{R}^n$, ${\boldsymbol{b}_{i,l}} \! \in \! \mathbb{R}^m$, ${\boldsymbol{c}_l} \! \in \! \mathbb{R}^m$ and ${\kappa_l} \! \in \! \mathbb{R}^1$ are the parameters in $l^{\rm{th}}$ cutting plane, and $|{\boldsymbol{\mathcal{P}}^{t}  }|$ denotes the number of cutting planes in ${\boldsymbol{\mathcal{P}}^{t}  }$. Thus, the approximate problem in $(t+1)^{{\rm{th}}}$ iteration can be expressed as follows,
\vspace{-3mm}
\begin{align}
\label{eq:new_22}
\begin{array}{l}
\mathop {\min } \;\;\; {\rm{ }}F(\{ {\boldsymbol{x}_i}\} ,\{ {\boldsymbol{y}_i}\} ,\boldsymbol{v},\boldsymbol{z}) = \sum\limits_{i = 1}^N {{G_i}({\boldsymbol{x}_i},{\boldsymbol{y}_i})} \\
 {\rm{s.t.}}  \qquad  {\boldsymbol{x}_i} = \boldsymbol{v},{\rm{ }}i = 1, \cdots ,N\\
\qquad \quad \; \, {\boldsymbol{a}_l}\!^\top \! \boldsymbol{v} \!+\! \sum\limits_{i = 1}^N {{\boldsymbol{b}_{i,l}}\!^\top \! {\boldsymbol{y}_i}} \! + \! {\boldsymbol{c}_l}\!^\top \! \boldsymbol{z} \!+\! {\kappa_l}\! \le\! 0,l \! =\! 1, \!\cdots\! ,|\boldsymbol{\mathcal{P}}^{t}| \\
{\rm{var.}}\; \; \;\; \; \;  \{{\boldsymbol{x}_i}\}, \{{\boldsymbol{y}_i}\},\boldsymbol{v}, \boldsymbol{z},
\end{array}
\end{align}
The cutting planes will be updated to refine the approximation, details are given in Section 3.4.

\textcolor{black}{\textbf{3.3 Asynchronous Algorithm}}

In the proposed ADBO, we solve the distributed bilevel optimization problem in an asynchronous manner. The Lagrangian function of Eq. (\ref{eq:new_22}) can be written as:
\vspace{-2mm}
\begin{equation}
\label{eq:new_23}
\begin{array}{l}
 {L_p} =  \sum\limits_{i = 1}^N  {{G_i}({\boldsymbol{x}_i},{\boldsymbol{y}_i})}  + \sum\limits_{l = 1}^{|\boldsymbol{\mathcal{P}}^{t}|}  {{\lambda _l}} \left( {{\boldsymbol{a}_l} ^\top  \boldsymbol{v} +\sum\limits_{i = 1}^N {{\boldsymbol{b}_{i,l}} ^\top   {\boldsymbol{y}_i}}  + {\boldsymbol{c}_l}  ^\top   \boldsymbol{z} + {\kappa_l}}  \right) 
  + \sum\limits_{i = 1}^N {{\boldsymbol{\theta }_i} ^\top  ({\boldsymbol{x}_i} - \boldsymbol{v})} ,
\end{array}
\end{equation}
where ${\lambda _l}\! \in\! \mathbb{R}^1$, ${\boldsymbol{\theta }_i}\! \in\! \mathbb{R}^n$ are dual variables, $L_p$ is  simplified form of ${L_p}(\{ {\boldsymbol{x}_i}\} , \! \{ {\boldsymbol{y}_i}\} ,\! \boldsymbol{v},\! \boldsymbol{z},\! \{ {\lambda _l}\} ,\!\{ {\boldsymbol{\theta }_i}\} )$. The regularized version \citep{xu2020unified} of Eq. (\ref{eq:new_23}) is employed to update all variables as follows,
\vspace{-2mm}
\begin{equation}
\label{eq:new_24}
{\widetilde{L}_p}(\{ {\boldsymbol{x}_i}\} ,\!\{ {\boldsymbol{y}_i}\} ,\boldsymbol{v},\boldsymbol{z},\!\{ {\lambda _l}\} ,\!\{ {\boldsymbol{\theta }_i}\} ) = {L_p} - \sum\limits_{l = 1}^{|\boldsymbol{\mathcal{P}}^{t}|} \frac{c_1^t  }{2} || {\lambda _l}||^2   -  \sum\limits_{i = 1}^N  \frac{c_2^t  }{2} || {\boldsymbol{\theta }_i}||^2 ,
\end{equation}

\vspace{-4mm}

where $c_1^t  $ and $c_2^t  $ denote the regularization terms in $(t+1)^{\rm{th}}$ iteration. In each iteration, we set that ${|\boldsymbol{\mathcal{P}}^{t}|} \!\le\!M,\forall t$.  ${c_1^t} = \frac{1}{{{{\eta _{{\lambda}}}}{(t+1)^{\frac{1}{4}}}}} \! \ge \! \underline{c}_1$, ${c_2^t} = \frac{1}{{{{\eta _{{\boldsymbol{\theta }}}}}{(t+1)^{\frac{1}{4}}}}} \! \ge \! \underline{c}_2$ are two nonnegative non-increasing sequences, where ${\eta _{{\lambda}}}$ and ${\eta _{{\boldsymbol{\theta }}}}$ are positive constants, and constants $\underline{c}_1$, $\underline{c}_2$  meet that $0\!<\!\underline{c}_1\!\le\!1/{\eta _{{\lambda}}}c$, $0\!<\!\underline{c}_2\!\le\!1/{\eta _{{\boldsymbol{\theta }}}}c$, $c\!=\!({({{4M\alpha_3}} / {{{{\eta _{\lambda}}}^2}} \!+\! {{4N\alpha_4}}/{{{{\eta _{\boldsymbol{\theta }}}}^2}})^2}1/\epsilon^2  \! +\! 1)^{\frac{1}{4}}$ ($\epsilon$, $\alpha_3$, $\alpha_4$ are introduced in Section \ref{section:discussion}). 

\textcolor{black}{Following \citep{zhang2014asynchronous}, to alleviate the staleness issue in ADBO,  we set that master updates its variables once it receives updates from $S$ active workers at every iteration and every worker has to communicate with the master at least once every $\tau$ iterations.} In $(t+1)^{\rm{th}}$ iteration, let ${\boldsymbol{\mathcal{Q}}^{t + 1}}$ denote the index set of active workers, the proposed algorithm proceeds as follows,

\vspace{-1mm}

(1) \emph{Active workers} update the local variables as follows,
\vspace{-1mm}
\begin{align}
\label{eq:new_25}
{\boldsymbol{x}_i^{t + 1}} = \left\{ \begin{array}{l}
{\boldsymbol{x}_i^{t}  } - {\eta _{\boldsymbol{x}}}{\nabla _{{\boldsymbol{x}_i}}} {\widetilde{L}_p}(\{ {\boldsymbol{x}_i^{\hat{t}_i}}\} ,\!\{ {\boldsymbol{y}_i^{\hat{t}_i}}\} ,\!\boldsymbol{v}^{\hat{t}_i}, \boldsymbol{z}^{\hat{t}_i}, \!\{ {\lambda _l^{\hat{t}_i}}\} ,\!\{ {\boldsymbol{\theta }_i^{\hat{t}_i}}\} ),i \in {\boldsymbol{\mathcal{Q}}^{t + 1}}\\
{\boldsymbol{x}_i^{t}  },i \notin {\boldsymbol{\mathcal{Q}}^{t + 1}}
\end{array} \right.,
\end{align}
\vspace{-4mm}
\begin{align}
\label{eq:new_26}
\boldsymbol{y}_i^{t + 1} = \left\{ \begin{array}{l}
{\boldsymbol{y}_i^{t}}   - {\eta _{\boldsymbol{y}}}{\nabla _{{\boldsymbol{y}_i}}}{\widetilde{L}_p}(\{ {\boldsymbol{x}_i^{\hat{t}_i}}\} ,\!\{ {\boldsymbol{y}_i^{\hat{t}_i}}\} ,\!\boldsymbol{v}^{\hat{t}_i}, \boldsymbol{z}^{\hat{t}_i},\! \{ {\lambda _l^{\hat{t}_i}}\} ,\!\{ {\boldsymbol{\theta }_i^{\hat{t}_i}}\} ),i \in {\boldsymbol{\mathcal{Q}}^{t + 1}}\\
{\boldsymbol{y}_i^{t}}  ,i \notin {\boldsymbol{\mathcal{Q}}^{t + 1}}
\end{array} \right.,
\end{align}

\vspace{-4mm}

where ${\hat{t}_i}$ denotes the last iteration during which worker $i$ was active, ${\eta _{\boldsymbol{x}}}$ and ${\eta _{\boldsymbol{y}}}$ are step-sizes. Then, the active workers transmit the local variables ${\boldsymbol{x}_i^{t + 1}}$ and ${\boldsymbol{y}_i^{t + 1}}$ to the master.

\vspace{-1mm}

(2) After receiving the updates from active workers, the \emph{master} updates the variables as follows,
\begin{align}
\label{eq:new_27}
{\boldsymbol{v}^{t + 1}} = {\boldsymbol{v}^{t}  } - {\eta _{\boldsymbol{v}}}{\nabla _{{\boldsymbol{v}}}}{\widetilde{L}_p}(\{ \boldsymbol{x}_i^{t + 1}\} ,\!\{ \boldsymbol{y}_i^{t + 1}\} ,\!{\boldsymbol{v}^{t}  },{\boldsymbol{z}^{t}  },\!\{ \lambda _l^{t}  \} ,\!\{ \boldsymbol{\theta }_i^{t}  \} ),
\end{align}
\vspace{-5mm}
\begin{align}
\label{eq:new_28}
{\boldsymbol{z}^{t + 1}} = {\boldsymbol{z}^{t}  } - {\eta _{\boldsymbol{z}}}{\nabla _{{\boldsymbol{z}}}}{\widetilde{L}_p}(\{ \boldsymbol{x}_i^{t + 1}\} ,\!\{ \boldsymbol{y}_i^{t + 1}\} ,\!{\boldsymbol{v}^{t+1}},{\boldsymbol{z}^{t}  },\!\{ {\lambda _l}^{t}  \} ,\!\{ \boldsymbol{\theta }_i^{t}  \} ),
\end{align}
\vspace{-5mm}
\begin{align}
\label{eq:new_29}
\lambda _l^{t + 1} = \lambda _l^{t}   + {\eta _{\lambda}}{\nabla _{\lambda_l}}{\widetilde{L}_p}(\{ \boldsymbol{x}_i^{t + 1}\} ,\!\{ \boldsymbol{y}_i^{t + 1}\} ,\!{\boldsymbol{v}^{t + 1}},{\boldsymbol{z}^{t + 1}},\!\{ \lambda _l^{t}\} ,\!\{ \boldsymbol{\theta }_i^{t}  \} ),
\end{align}
\vspace{-2.5mm}
\begin{align}
\label{eq:new_30}
\boldsymbol{\theta }_i^{t + 1} = \left\{ \begin{array}{l}
{\boldsymbol{\theta }_i^{t}}   + {\eta _{\boldsymbol{\theta}}}{\nabla _{{\boldsymbol{\theta}_i}}}{\widetilde{L}_p}(\{ \boldsymbol{x}_i^{t + 1}\} ,\!\{ \boldsymbol{y}_i^{t + 1}\} ,\!{\boldsymbol{v}^{t + 1}},{\boldsymbol{z}^{t + 1}},\!\{ {\lambda _l}^{t+1}  \} ,\!\{ \boldsymbol{\theta }_i^{t}  \} ),i \in {\boldsymbol{\mathcal{Q}}^{t + 1}}\\
{\boldsymbol{\theta }_i^{t}}  ,i \notin {\boldsymbol{\mathcal{Q}}^{t + 1}}
\end{array} \right.,
\end{align}
where ${\eta _{\boldsymbol{v}}}$, ${\eta _{\boldsymbol{z}}}$, ${\eta _{\lambda}}$ and ${\eta _{\boldsymbol{\theta}}}$ are step-sizes. Next, the master broadcasts ${\boldsymbol{v}^{t + 1}}, {\boldsymbol{z}^{t + 1}}, \boldsymbol{\theta }_i^{t + 1}$ and $\{\lambda _l^{t + 1} \}$ to worker $i, i \in {\boldsymbol{\mathcal{Q}}^{t + 1}}$ (\textit{i.e.}, active workers). Details are summarized in Algorithm \ref{algorithm_asyn}.

\textcolor{black}{\textbf{3.4 Updating Cutting Planes}}

Every $k_{\rm{pre}}$ iterations ($k_{\rm{pre}}\!>\!0$ is a pre-set constant, which can be controlled flexibly), the cutting planes are updated based on the following two steps (a) and (b) when $t<T_1$: 

\textbf{(a) Removing the inactive cutting planes,}
\begin{equation}
\label{eq:97_new_35}
{{\boldsymbol{\mathcal{P}}}^{t + 1}} = \left\{ \begin{array}{l}
{\rm{Drop(}}{{\boldsymbol{\mathcal{P}}}^{t}  },c{p_l}{\rm{),  if \;  }}{\lambda _l^{t + 1}} \;{\rm{and}} \; {\lambda _l^{t}} = 0\\
{{\boldsymbol{\mathcal{P}}}^{t}  },{\rm{otherwise}}
\end{array} \right.,
\end{equation}
where $cp_l$ represents the $l^{\rm{th}}$ cutting plane in ${\boldsymbol{\mathcal{P}}}^{t}  $ and ${\rm{Drop(}}{{\boldsymbol{\mathcal{P}}}^{t}  },c{p_l})$ represents the $l^{\rm{th}}$ cutting plane $cp_l$ is removed  from ${\boldsymbol{\mathcal{P}}}^{t}  $. The dual variable set $\{ {\lambda ^{t+1}  }\}$ will be updated as follows,
\begin{equation}
\label{eq:97_new_36}
\{ {\lambda ^{t + 1}}\} \! =\! \left\{ \begin{array}{l}
{\rm{Drop(}}\{ {\lambda ^{t}  }\} ,{\lambda _l}{\rm{),  if \;  }}{\lambda _l^{t + 1}} \;{\rm{and}} \; {\lambda _l^{t}} = 0\\
\{ {\lambda ^{t}  }\} ,{\rm{otherwise}}
\end{array} \right.,
\end{equation}
where $\{ {\lambda ^{t + 1}}\} $ and $\{ {\lambda ^{t}}\} $ represent the dual variable set in $(t+1)^{{\rm{th}}}$ and $t^{{\rm{th}}}$ iterations, respectively. ${\rm{Drop(}}\{ {\lambda ^{t}  }\} ,{\lambda _l})$ represents that ${\lambda _l}$ is removed from the dual variable set $\{ {\lambda ^{t}  }\}$.

\vspace{-1mm}

\textbf{(b) Adding new cutting planes.} Firstly, we investigate whether $(\boldsymbol{v}^{t+1},\!\{ {\boldsymbol{y}_i^{t+1}}\} ,\boldsymbol{z}^{t+1})$ is feasible for the constraint $h(\boldsymbol{v},\{ {\boldsymbol{y}_i}\} ,\boldsymbol{z}) \! \le\! \varepsilon$. We can obtain $h(\boldsymbol{v}^{t+1},\!\{ {\boldsymbol{y}_i^{t+1}}\} ,\boldsymbol{z}^{t+1})$ according to $\phi(\boldsymbol{v}^{t+1})$ in Eq. (\ref{eq:97_23}). If $(\boldsymbol{v}^{t+1},\!\{ {\boldsymbol{y}_i^{t+1}}\} ,\boldsymbol{z}^{t+1})$ is not a feasible solution to the original problem (Eq. (\ref{eq:new_21})), new cutting plane $cp^{t+1}_{new}$ will be generated to separate the point $(\boldsymbol{v}^{t+1},\!\{ {\boldsymbol{y}_i^{t+1}}\} ,\boldsymbol{z}^{t+1})$ from the feasible region of constraint $h(\boldsymbol{v},\{ {\boldsymbol{y}_i}\} ,\boldsymbol{z})\! \le\! \varepsilon$. Thus, the \textit{valid} cutting plane \citep{boyd2007localization} ${\boldsymbol{a}_l}\!^\top\!  \boldsymbol{v} + \sum_{i = 1}^N {{\boldsymbol{b}_{i,l}}\!^\top \! {\boldsymbol{y}_i}}  +  {\boldsymbol{c}_l}\!^\top \! \boldsymbol{z} + {\kappa_l} \le  0$  must satisfy that,
\vspace{0.5mm}
\begin{equation}
\label{eq:915_23}
\left\{ \begin{array}{l}
 {\boldsymbol{a}_l}\!^\top\!  \boldsymbol{v} + \sum_{i = 1}^N {{\boldsymbol{b}_{i,l}}\!^\top \! {\boldsymbol{y}_i}}  +  {\boldsymbol{c}_l}\!^\top \! \boldsymbol{z} + {\kappa_l} \le 0, \forall (\boldsymbol{v},\!\{ {\boldsymbol{y}_i}\} ,\boldsymbol{z}) \; {\rm{satisfies}} \; h(\boldsymbol{v},\!\{ {\boldsymbol{y}_i}\} ,\boldsymbol{z}) \!\le\! \varepsilon\\
{\boldsymbol{a}_l}\!^\top\!  \boldsymbol{v}^{t+1} + \sum_{i = 1}^N {{\boldsymbol{b}_{i,l}}\!^\top \! {\boldsymbol{y}_i^{t+1}}}  +  {\boldsymbol{c}_l}\!^\top \! \boldsymbol{z}^{t+1} + {\kappa_l}  > 0
\end{array} \right..
\end{equation}

Since $h(\boldsymbol{v},\{ {\boldsymbol{y}_i}\} ,\boldsymbol{z})$ is a convex function, we have that,
\vspace{-1mm}
\begin{equation}
\label{eq:915_24}
\begin{array}{l}
\! h(\boldsymbol{v},\{ {\boldsymbol{y}_i}\} ,\boldsymbol{z})  \!\ge\!  h(\boldsymbol{v}^{t+1},\{ {\boldsymbol{y}_i^{t+1}}\} ,\boldsymbol{z}^{t+1}) \! + \! {\left[ \begin{array}{l}
\frac{{\partial h(\boldsymbol{v}^{t+1},\{ {\boldsymbol{y}_i^{t+1}}\} ,\boldsymbol{z}^{t+1})}}{{\partial \boldsymbol{v}}}\\
\!\{ \frac{{\partial h(\boldsymbol{v}^{t+1},\{ {\boldsymbol{y}_i^{t+1}}\} ,\boldsymbol{z}^{t+1})}}{{\partial \boldsymbol{y}_i}} \}\\
\frac{{\partial h(\boldsymbol{v}^{t+1},\{ {\boldsymbol{y}_i^{t+1}}\} ,\boldsymbol{z}^{t+1})}}{{\partial \boldsymbol{z}}}
\end{array} \right]^\top }\!\left( {\left[ \begin{array}{l}
\boldsymbol{v}\\
\!\{ \boldsymbol{y}_i \} \\
\boldsymbol{z}\\
\end{array} \right]\! -\! \left[ \begin{array}{l}
\boldsymbol{v}^{t + 1}\\
\!\{\boldsymbol{y}_{i}^{t + 1}\}\\
\boldsymbol{z}^{t + 1}
\end{array} \right]} \right)\end{array}\!.
\end{equation}
Combining Eq. (\ref{eq:915_24}) with Eq. (\ref{eq:915_23}), we have that a valid cutting plane (with respect to point $(\boldsymbol{v}^{t+1},\{ {\boldsymbol{y}_i^{t+1}}\} ,\boldsymbol{z}^{t+1})$) can be expressed as,
\vspace{-1mm}
\begin{equation}
    h(\boldsymbol{v}^{t+1},\{ {\boldsymbol{y}_i^{t+1}}\} ,\boldsymbol{z}^{t+1}) + {\left[ \begin{array}{l}
\frac{{\partial h(\boldsymbol{v}^{t+1},\{ {\boldsymbol{y}_i^{t+1}}\} ,\boldsymbol{z}^{t+1})}}{{\partial \boldsymbol{v}}}\\
\{ \frac{{\partial h(\boldsymbol{v}^{t+1},\{ {\boldsymbol{y}_i^{t+1}}\} ,\boldsymbol{z}^{t+1})}}{{\partial \boldsymbol{y}_i}} \}\\
\frac{{\partial h(\boldsymbol{v}^{t+1},\{ {\boldsymbol{y}_i^{t+1}}\} ,\boldsymbol{z}^{t+1})}}{{\partial \boldsymbol{z}}}
\end{array} \right]^\top  }\!\left( {\left[ \begin{array}{l}
\boldsymbol{v}\\
\{ \boldsymbol{y}_i \} \\
\boldsymbol{z}\\
\end{array} \right]\! -\! \left[ \begin{array}{l}
\boldsymbol{v}^{t + 1}\\
\{\boldsymbol{y}_{i}^{t + 1}\}\\
\boldsymbol{z}^{t + 1}
\end{array} \right]} \right) \le \varepsilon.
\label{eq:cuts}
\end{equation}

For brevity, we utilize $cp^{t+1}_{new}$ to denote the new added cutting plane (\textit{i.e.}, Eq. (\ref{eq:cuts})). Thus the polytope ${{\boldsymbol{\mathcal{P}}}^{t + 1}}$ will be updated as follows,
\begin{equation}
\label{eq:97_39}
    {{\boldsymbol{\mathcal{P}}}^{t + 1}} = \left\{ \begin{array}{l}
{\rm{Add(}}{{\boldsymbol{\mathcal{P}}}^{t + 1}},cp_{new}^{t + 1}),{\rm{  if \; }}h(\boldsymbol{v}^{t+1},\!\{ {\boldsymbol{y}_i^{t+1}}\} ,\boldsymbol{z}^{t+1}) > \varepsilon \\
{{\boldsymbol{\mathcal{P}}}^{t + 1}},{\rm{otherwise}}
\end{array} \right.,
\end{equation}
where ${\rm{Add(}}{{\boldsymbol{\mathcal{P}}}^{t + 1}},cp_{new}^{t + 1})$ represents that new cutting plane  $cp^{t+1}_{new}$ is added to polytope ${{\boldsymbol{\mathcal{P}}}^{t + 1}}$. The dual variable set $\{ {\lambda ^{t + 1}}\}$ is updated as follows,
\begin{equation}
\label{eq:97_40}
    \{ {\lambda ^{t + 1}}\}  = \left\{ \begin{array}{l}
{\rm{Add(}}\{ {\lambda ^{t + 1}}\} ,{\lambda _{|{{\boldsymbol{\mathcal{P}}}^{t + 1}}|}^{t + 1}}{\rm{),  if \; }}h(\boldsymbol{v}^{t+1},\!\{ {\boldsymbol{y}_i^{t+1}}\} ,\boldsymbol{z}^{t+1}) > \varepsilon \\
\{ {\lambda ^{t + 1}}\} ,{\rm{otherwise}}
\end{array} \right.,{\rm{  }}
\end{equation}
where ${\rm{Add(}}\{ {\lambda ^{t + 1}}\} ,{\lambda _{|{{\boldsymbol{\mathcal{P}}}^{t + 1}}|}^{t + 1}})$ represents that dual variable ${\lambda _{|{{\boldsymbol{\mathcal{P}}}^{t + 1}}|}^{t + 1}}$ is added to the dual variable set $\{ {\lambda ^{t+1}  }\}$. Finally, master broadcasts the updated ${{\boldsymbol{\mathcal{P}}}^{t + 1}}$ and $\{ {\lambda ^{t + 1}}\}$ to all workers. The details of the proposed algorithm are summarized in Algorithm \ref{algorithm_asyn}.

{\linespread{1}
\begin{algorithm*}[t]
   \caption{ADBO: Asynchronous Distributed Bilevel Optimization}
   \label{algorithm_asyn}
\begin{algorithmic}
   \STATE {\bfseries Initialization:}  master iteration $t\! =\! 0$, variables $\{{{\boldsymbol{x}}_i^0}\}$, $\{{{\boldsymbol{y}}_i^0}\}$, ${\boldsymbol{v}^{0}}, {\boldsymbol{z}^{0}}, \{ \lambda _l^{0} \}, \{ \boldsymbol{\theta }_i^{0}\}$ and polytope ${{\boldsymbol{\mathcal{P}}}^0}$.
   
   \REPEAT
   
   \FOR{\emph{active worker}}
   \STATE updates variables ${\boldsymbol{x}_i^{t+1}}$, ${\boldsymbol{y}_i^{t+1}}$ according to Eq. (\ref{eq:new_25}) and (\ref{eq:new_26});
   \ENDFOR
   
   \STATE Active workers transmit their local variables to master;
   
   \FOR{\emph{master}}
   \STATE updates variables ${\boldsymbol{v}^{t + 1}}, {\boldsymbol{z}^{t + 1}}, \{ \lambda _l^{t + 1} \}, \{ \boldsymbol{\theta }_i^{t + 1}\}$  according to Eq. (\ref{eq:new_27}), (\ref{eq:new_28}), (\ref{eq:new_29}) and (\ref{eq:new_30});
   \ENDFOR
   \STATE master broadcasts variables to active workers;

   \IF{$(t+1)$ mod $k_{\rm{pre}}$ $==0$ and $t<T_1$}
   \STATE   master computes $\phi(\boldsymbol{v}^{t + 1})$ according to Eq.  (\ref{eq:97_23});
   \STATE   master updates ${{\boldsymbol{\mathcal{P}}}^{t + 1}}$ and $\{ {\lambda ^{t + 1}}\}$ according to Eq. (\ref{eq:97_new_35}), (\ref{eq:97_new_36}), (\ref{eq:97_39}) and (\ref{eq:97_40});

   \STATE master broadcasts ${{\boldsymbol{\mathcal{P}}}^{t + 1}}$ and $\{ {\lambda ^{t + 1}}\}$ to all workers;
   \ENDIF

   \STATE $t =t+1$;
   \UNTIL{termination.}
\end{algorithmic}
\end{algorithm*}}

\section{Discussion}
\label{section:discussion}
\begin{theorem}
\label{theorem: convergence}
(\textbf{Convergence}) As the cutting plane continues to be added to the polytope, the optimal objective value of approximate problem in Eq. (\ref{eq:new_22}) converges monotonically.
\end{theorem}
\vspace{-3mm}

The proof of Theorem \ref{theorem: convergence} is presented in Appendix \ref{appendix:theorem1}.

\vspace{-2mm}

\begin{definition}
\label{definition:1}
{\textbf{(Stationarity gap)}} Following \citep{xu2020unified,lu2020hybrid,jiao2022distributed}, the \textit{stationarity} \textit{gap} of our problem at $t^{{th}}$ iteration is defined as:
{\begin{equation}
\label{eq:definition1}
\nabla G^t = \left[ \begin{array}{l}
\{ {\nabla _{{{\boldsymbol{x}}_i}}}{L_p}(\{ {\boldsymbol{x}_i^t}\} , \! \{ {\boldsymbol{y}_i^t}\} ,\! \boldsymbol{v}^t,\! \boldsymbol{z}^t,\! \{ {\lambda _l^t}\} ,\!\{ {\boldsymbol{\theta }_i^t}\} )\}\\
\{ {\nabla _{{{\boldsymbol{y}}_i}}}{L_p}(\{ {\boldsymbol{x}_i^t}\} , \! \{ {\boldsymbol{y}_i^t}\} ,\! \boldsymbol{v}^t,\! \boldsymbol{z}^t,\! \{ {\lambda _l^t}\} ,\!\{ {\boldsymbol{\theta }_i^t}\} )\} \\
\,{\nabla _{{{\boldsymbol{v}}}}}{L_p}(\{ {\boldsymbol{x}_i^t}\} , \! \{ {\boldsymbol{y}_i^t}\} ,\! \boldsymbol{v}^t,\! \boldsymbol{z}^t,\! \{ {\lambda _l^t}\} ,\!\{ {\boldsymbol{\theta }_i^t}\})\\
\,{\nabla _{{{\boldsymbol{z}}}}}{L_p}(\{ {\boldsymbol{x}_i^t}\} , \! \{ {\boldsymbol{y}_i^t}\} ,\! \boldsymbol{v}^t,\! \boldsymbol{z}^t,\! \{ {\lambda _l^t}\} ,\!\{ {\boldsymbol{\theta }_i^t}\})\\
\{ {\nabla _{\lambda _l}}{L_p}(\{ {\boldsymbol{x}_i^t}\} , \! \{ {\boldsymbol{y}_i^t}\} ,\! \boldsymbol{v}^t,\! \boldsymbol{z}^t,\! \{ {\lambda _l^t}\} ,\!\{ {\boldsymbol{\theta }_i^t}\} )\}\\
\{ {\nabla _{\boldsymbol{\theta }_i}}{L_p}(\{ {\boldsymbol{x}_i^t}\} , \! \{ {\boldsymbol{y}_i^t}\} ,\! \boldsymbol{v}^t,\! \boldsymbol{z}^t,\! \{ {\lambda _l^t}\} ,\!\{ {\boldsymbol{\theta }_i^t}\} )\}
\end{array} \right].
\end{equation}}
\end{definition}

\vspace{-3mm}

\begin{definition}
\label{definition:e stationary point}
{\textbf{($\epsilon$-stationary point)}}  $(\{ \boldsymbol{x}_i^{t}\} ,\!\{ \boldsymbol{y}_i^{t}\} ,\!{\boldsymbol{v}^{t}  },{\boldsymbol{z}^{t}  },\!\{ \lambda _l^{t}  \} ,\!\{ \boldsymbol{\theta }_i^{t}  \} )$ is an $\epsilon$-stationary point ($\epsilon  \ge 0$) of a differentiable function ${L_p}$,  if $\,||\nabla G^t||^2 \le \epsilon $.  $T(\epsilon )$ is the first iteration index such that $||\nabla G^t||^2   \le  \epsilon$, \textit{i.e.}, $T(\epsilon )  =  \min \{ t \ |\; ||\nabla G^t||^2   \le  \epsilon \}  $.
\end{definition}
\vspace{-3mm}

\begin{assumption}\label{assumption:1}
\textbf{{{(Smoothness/Gradient Lipschitz})}} Following \citep{ji2021bilevel}, we assume that
$L_p$ has Lipschitz continuous gradients, i.e., for any $\boldsymbol{\omega}, \boldsymbol{\omega}'$, we assume that there exists $L>0$ satisfying that,
\begin{equation}
    \begin{array}{l}
||{\nabla}{L_p}(\boldsymbol{\omega})  -  {\nabla}{L_p}(\boldsymbol{\omega}')|| \le L||\boldsymbol{\omega}-\boldsymbol{\omega}'||,
\end{array}
\end{equation}
\end{assumption}
\vspace{-3mm}

\begin{assumption}\label{assumption:2}
{\textbf{(Boundedness)}} Following \citep{qian2019robust}, we assume that variables are bounded, i.e.,  $||\boldsymbol{x}_i||^2  \!\le \! \alpha_1, ||\boldsymbol{v}||^2 \! \le\! \alpha_1, ||\boldsymbol{y}_i||^2 \!\le \! \alpha_2, ||\boldsymbol{z}||^2  \!\le \! \alpha_2, ||\lambda_l||^2  \!\le \! \alpha_3, ||\boldsymbol{\theta }_i||^2  \!\le \! \alpha_4$. And we assume that before obtaining the $\epsilon$-stationary point (i.e., $t\!\le\! T(\epsilon )\!-\!1$), the variables in master satisfy that $||\boldsymbol{v}^{t+1} \!-\! \boldsymbol{v}^t|{|^2}\!+\!||\boldsymbol{z}^{t+1} \!-\! \boldsymbol{z}^t|{|^2}\!+\!\sum\nolimits_l||{\lambda _l^{t + 1}} \!-\! {\lambda _l^t}|{|^2} \ge \vartheta $, where $\vartheta >0$ is a relative small constant. The change of the variables in master is upper bounded within $\tau$ iterations:
\begin{equation}
\begin{array}{*{20}{l}}
{||\boldsymbol{v}^t - \boldsymbol{v}^{t - k}|{|^2} \! \le \! \tau{k_1}\vartheta}, \;\;
{||\boldsymbol{z}^t - \boldsymbol{z}^{t - k}|{|^2} \! \le \! \tau{k_1}\vartheta},\;\;
{\sum\nolimits_l||{\lambda _l^t} - {\lambda _l^{t - k}}|{|^2} \! \le \! \tau{k_1}\vartheta}, {\forall 1 \! \le \! k \! \le \! \tau },
\end{array}
\end{equation}
where $k_1 >0$ is a constant.
\end{assumption}
\vspace{-3mm}

\begin{theorem}
\label{theorem 2}
{\textbf{(Iteration complexity)}} Suppose Assumption \ref{assumption:1} and \ref{assumption:2} hold, we set the step-sizes as
${\eta _{\boldsymbol{x}}}\! = \! {\eta _{\boldsymbol{y}}} \!=\! {\eta _{\boldsymbol{v}}} \!=\! {\eta _{\boldsymbol{z}}} \! =\! \frac{2}{{L + {{\eta _{\lambda}}}M{L^2} + {{\eta _{\boldsymbol{\theta }}}}N{L^2} + 8(\frac{{M\gamma {L^2}}}{{{{\eta _{\lambda}}}{\underline{c}_1}^2}} + \frac{{N\gamma {L^2}}}{{{{\eta _{\boldsymbol{\theta }}}}{\underline{c}_2}^2}})}}$,   ${{\eta _{\lambda}}} < \min \{\frac{ 2}{{L + 2c_1^0}}  ,\frac{1}{{30\tau{k_1}N{L^2}}}\} $ and $ {\eta _{\boldsymbol{\theta }}} \le \frac{2}{{L + 2c_2^0}} $. For a given $\epsilon $, we have:
\vspace{-1mm}
\begin{equation}
T( \epsilon ) \! \sim  \! \mathcal{O}(\max  \{ {(\frac{{4M\alpha_3}}{{{{\eta _{\lambda}}}^2}}\! +\! \frac{{4N\alpha_4}}{{{{\eta _{\boldsymbol{\theta }}}}^2}})^2}\frac{1}{{{\epsilon ^2}}}, 
{(\frac{{4{{{(d_7 + \frac{{{{\eta _{\boldsymbol{\theta }}}}(N  -  S){{L}^2}}}{2})}}} (\mathop d\limits^ -  +  k_d\tau(\tau  -  1))  {d_6}}}{{{\epsilon}}} + (T_1 + 2)^{\frac{1}{2}})^2}\}), 
\end{equation}

\vspace{-3mm}

where ${\alpha_3}$, ${\alpha_4}$, $\gamma$,  $k_d$, $T_1$, $\mathop d\limits^ -  $, ${d_6}$ and ${d_7}$ are constants. The detailed proof is given in Appendix \ref{appendix: theorem 2}.
\end{theorem}

\section{Experiment}
In this section, experiments\footnote{Codes are available in \url{https://github.com/ICLR23Submission6251/adbo}.} are conducted on two hyperparameter optimization tasks (\textit{i.e.}, data hyper-cleaning task and regularization coefficient optimization task) in the distributed setting to evaluate the performance of the proposed ADBO. The proposed ADBO is compared with the state-of-the-art distributed bilevel optimization method FEDNEST \citep{tarzanagh2022fednest}. In data hyper-cleaning task, experiments are carried out on MNIST \citep{lecun1998gradient} and Fashion MNIST \citep{xiao2017fashion} datasets. In coefficient optimization task, following \citep{chen2022single}, experiments are conducted on Covertype \citep{blackard1999comparative}  and IJCNN1 \citep{prokhorov2001ijcnn}  datasets.

\subsection{Data Hyper-Cleaning}\label{subsec:exp1}
\textcolor{black}{Following \citep{ji2021bilevel,yang2021provably}}, we compare the performance of the proposed ADBO and distributed bilevel optimization method FEDNEST on the distributed data hyper-cleaning task \citep{chen2022decentralized} on MNIST and Fashion MNIST datasets. Data hyper-cleaning involves training a classifier in a contaminated environment where each training data label is changed to a random class number with a probability (\textit{i.e.}, the corruption rate). In the experiment, the distributed data hyper-cleaning problem is considered, whose formulation can be expressed as,
\vspace{-1mm}
\begin{align}
\label{eq:917_27}
\begin{array}{l}
\mathop {\min } {\rm{ }}\, F(\boldsymbol{\psi},\boldsymbol{w}) = \sum\limits_{i = 1}^N\! {\frac{1}{{|\mathcal{D}_i^{{\rm{val}}}|}}\!\sum\limits_{({{\bf{x}}_j},{y_j}) \in \mathcal{D}_i^{{\rm{val}}}}\! {\mathcal{L}({{\bf{x}}_j}^\top \boldsymbol{w},{y_j})} }  \\
{\rm{s.t.}}\; \boldsymbol{w} \!=\! \mathop {\arg \min }\limits_{{\boldsymbol{w}'}} f(\boldsymbol{\psi},\boldsymbol{w}') \!=\! \sum\limits_{i = 1}^N\! {\frac{1}{{|\mathcal{D}_i^{{\rm{tr}}}|}}\!\sum\limits_{({{\bf{x}}_j},{y_j}) \in \mathcal{D}_i^{{\rm{tr}}}} \!{\sigma ({{\psi_j}})\mathcal{L}({{\bf{x}}_j}^\top \boldsymbol{w}',{y_j}) + {C_r}||\boldsymbol{w}'|{|^2}} } \\
{\rm{var.}}\; \; \;\; \; \; \;\; \; {\boldsymbol{\psi}}, {\boldsymbol{w}},
\end{array}
\end{align}
where $\mathcal{D}_i^{{\rm{tr}}}$ and $\mathcal{D}_i^{{\rm{val}}}$ denote the training and validation datasets on $i^{{\rm{th}}}$ worker, respectively. $({{\bf{x}}_j},{y_j})$ denote the $j^{{\rm{th}}}$ data and label. $\sigma(.)$ is the sigmoid function, $\mathcal{L}$ is the cross-entropy loss, ${C_r}$ is a  regularization parameter and $N$ is the number of workers in the distributed system.  \textcolor{black}{In MNIST and Fashion MNIST datasets, we set $N=18$, $S=9$ and $\tau=15$.} According to \citet{cohen2021asynchronous}, we assume that the communication delay of each worker obeys the heavy-tailed distribution. The proposed ADBO is compared with the state-of-the-art distributed bilevel optimization method FEDNEST and SDBO (Synchronous Distributed Bilevel Optimization, \textit{i.e.}, ADBO without asynchronous setting). The test accuracy versus time is shown in Figure \ref{fig:hyperclean_acc}, and the test loss versus  time is shown in Figure \ref{fig:hyperclean_loss}.  We can observe that the proposed ADBO is the most efficient algorithm since 1) the asynchronous setting is considered in ADBO, the master can update its variables once it receives updates from $S$ active workers instead of all workers; and 2) ADBO is a single-loop algorithm and only gradient descent/ascent is required at each iteration, thus ADBO is computationally more efficient.

\vspace{-1mm}

\subsection{Regularization Coefficient Optimization}
\textcolor{black}{Following \citep{chen2022single},} we compare the proposed ADBO with baseline algorithms FEDNEST and SDBO on the regularization coefficient optimization task with Covertype and IJCNN1 datasets. The distributed regularization coefficient optimization problem is given by,
\vspace{-1mm}
\begin{align}
\label{eq:917_28}
\begin{array}{l}
\mathop {\min } {\rm{ }}\, F(\boldsymbol{\psi},\boldsymbol{w}) = \sum\limits_{i = 1}^N\! {\frac{1}{{|\mathcal{D}_i^{{\rm{val}}}|}}\!\sum\limits_{({{\bf{x}}_j},{y_j}) \in \mathcal{D}_i^{{\rm{val}}}}\! {\mathcal{L}({{\bf{x}}_j}^\top \boldsymbol{w},{y_j})} }  \\
{\rm{s.t.}}\; \boldsymbol{w} \!=\! \mathop {\arg \min }\limits_{{\boldsymbol{w}'}} f(\boldsymbol{\psi},\boldsymbol{w}') \!=\! \sum\limits_{i = 1}^N\! {\frac{1}{{|\mathcal{D}_i^{{\rm{tr}}}|}}\!\sum\limits_{({{\bf{x}}_j},{y_j}) \in \mathcal{D}_i^{{\rm{tr}}}} \!{\mathcal{L}({{\bf{x}}_j}^\top \boldsymbol{w}',{y_j}) + {\sum\limits_{j = 1}^n \psi_j ({w_j'})^2 }}}\\
{\rm{var.}}\; \; \;\; \; \; \;\; \; {\boldsymbol{\psi}}, {\boldsymbol{w}},
\end{array}
\end{align}
where ${\boldsymbol{\psi}} \! \in \! {\mathbb{R}}^n$, ${\boldsymbol{w}} \! \in \! {\mathbb{R}}^n$ and $\mathcal{L}$ respectively denote the regularization coefficient, model parameter, and logistic loss, and ${\boldsymbol{w}'}\!=\![w_1',\dots,w_n']$. In Covertype and IJCNN1 datasets, we set $N=18$, $S=9$, \textcolor{black}{$\tau = 15$} and $N=24$, $S=12$, \textcolor{black}{$\tau = 15$}, respectively. We also assume that the delay of each worker obeys the heavy-tailed distribution. Firstly, we compare the performance of the proposed ADBO, SDBO and FEDNEST in terms of test accuracy and test loss on Covertype and IJCNN1 datasets, which are shown in Figure \ref{fig:regularizer_acc} and \ref{fig:regularizer_loss}. It is seen that the proposed ADBO is more efficient because of the same two reasons we gave in Section~\ref{subsec:exp1}. 

\vspace{-1mm}

\makeatletter\def\@captype{figure}\makeatother 
\begin{minipage}{0.48\textwidth}  
\setlength{\abovecaptionskip}{-1mm} 
\subfigure[MNIST] 
{\begin{minipage}[t]{0.49\linewidth}
	\centering      
	\includegraphics[scale=0.23]{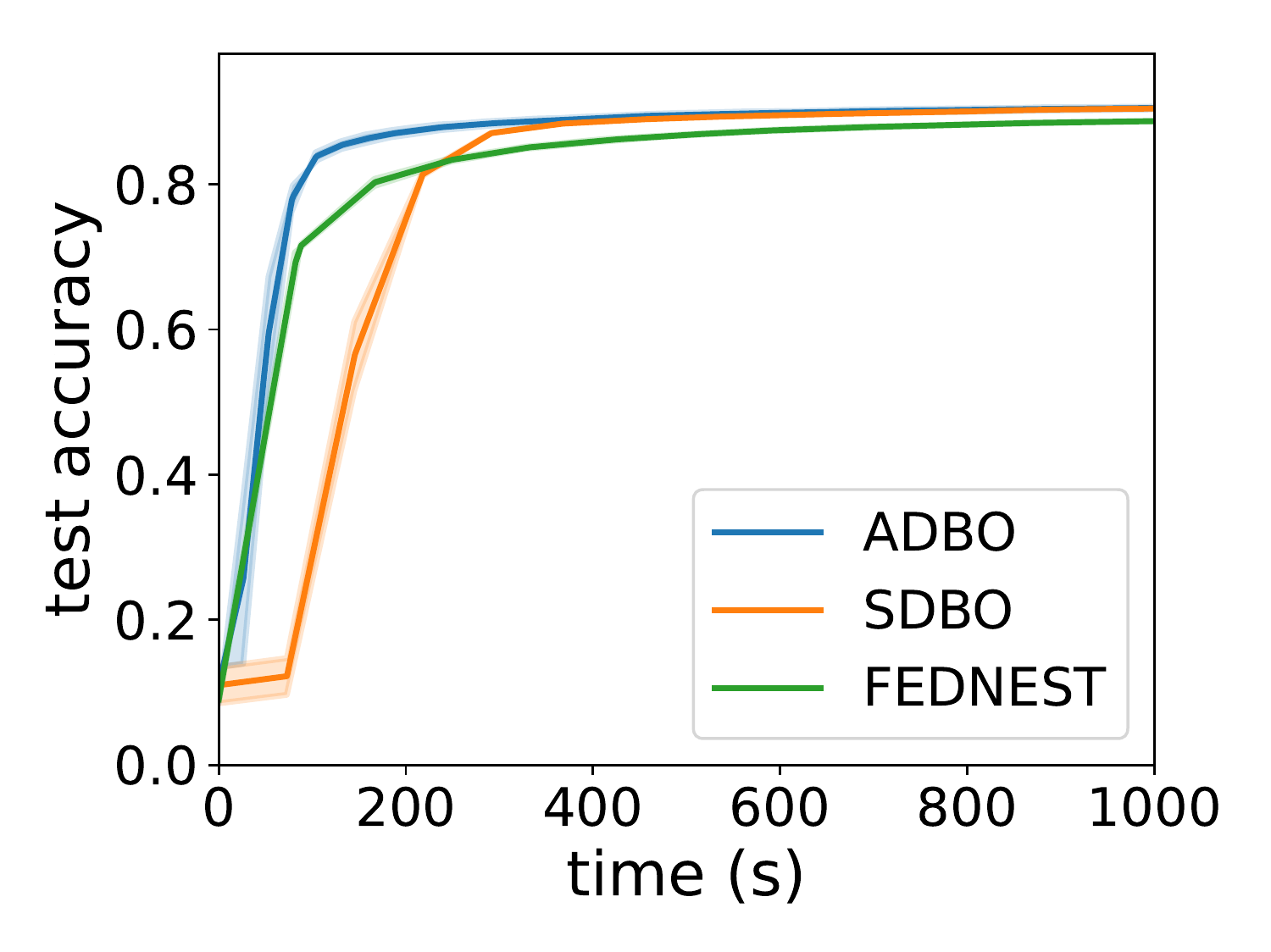}   
	\end{minipage}}
\subfigure[Fashion MNIST] 
{\begin{minipage}[t]{0.49\linewidth}
	\centering      
	\includegraphics[scale=0.23]{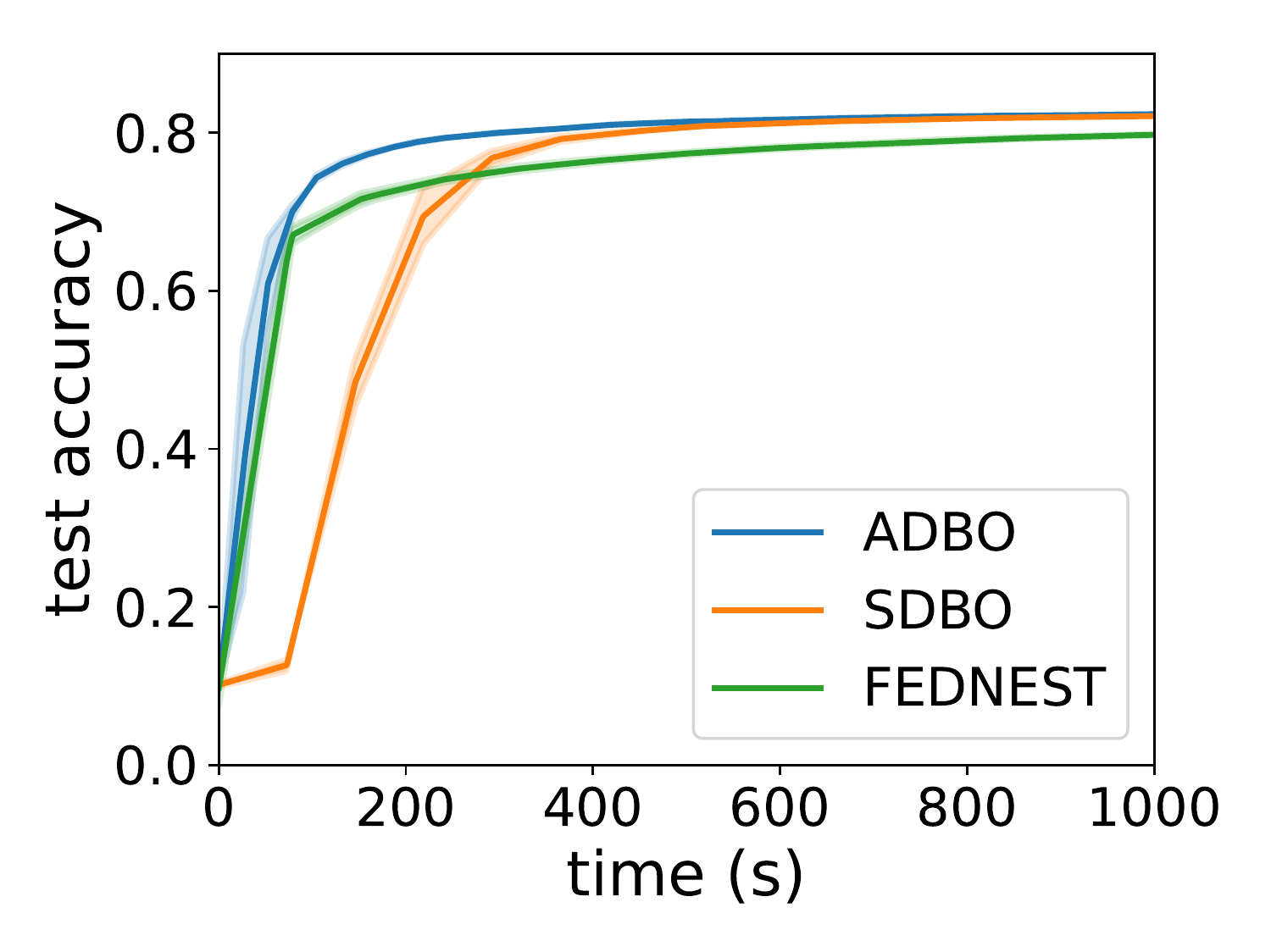}   
	\end{minipage}}
\caption{Test accuracy vs time on (a) MNIST and (b) Fashion MNIST datasets.} 
\label{fig:hyperclean_acc}  
\end{minipage} $\;$
\makeatletter\def\@captype{figure}\makeatother 
\begin{minipage}{0.48\textwidth} 
\setlength{\abovecaptionskip}{-1mm} 
\subfigure[MNIST] 
{\begin{minipage}[t]{0.49\linewidth}
	\centering      
	\includegraphics[scale=0.23]{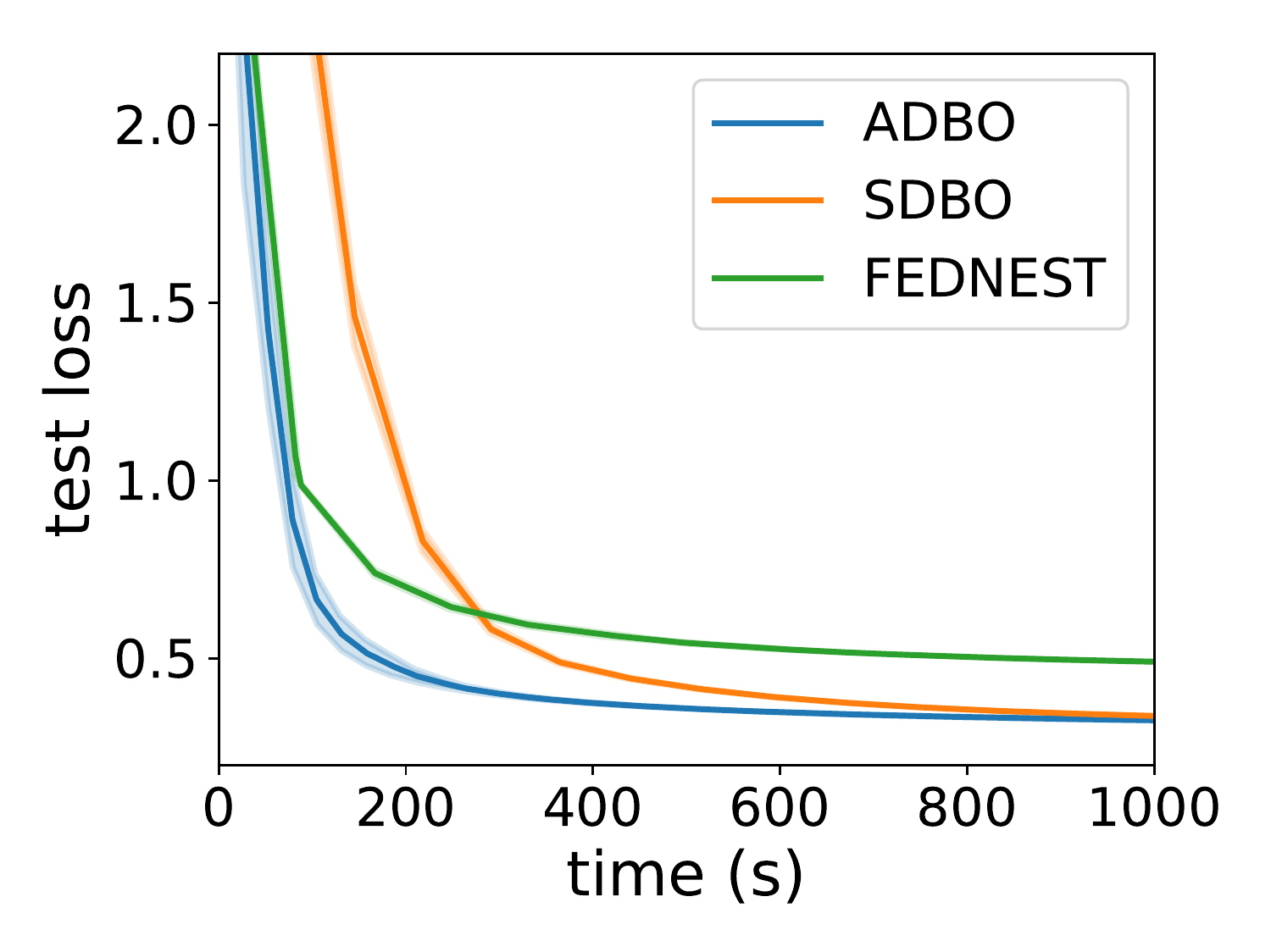}   
	\end{minipage}}
\subfigure[Fashion MNIST] 
{\begin{minipage}[t]{0.49\linewidth}
	\centering      
	\includegraphics[scale=0.23]{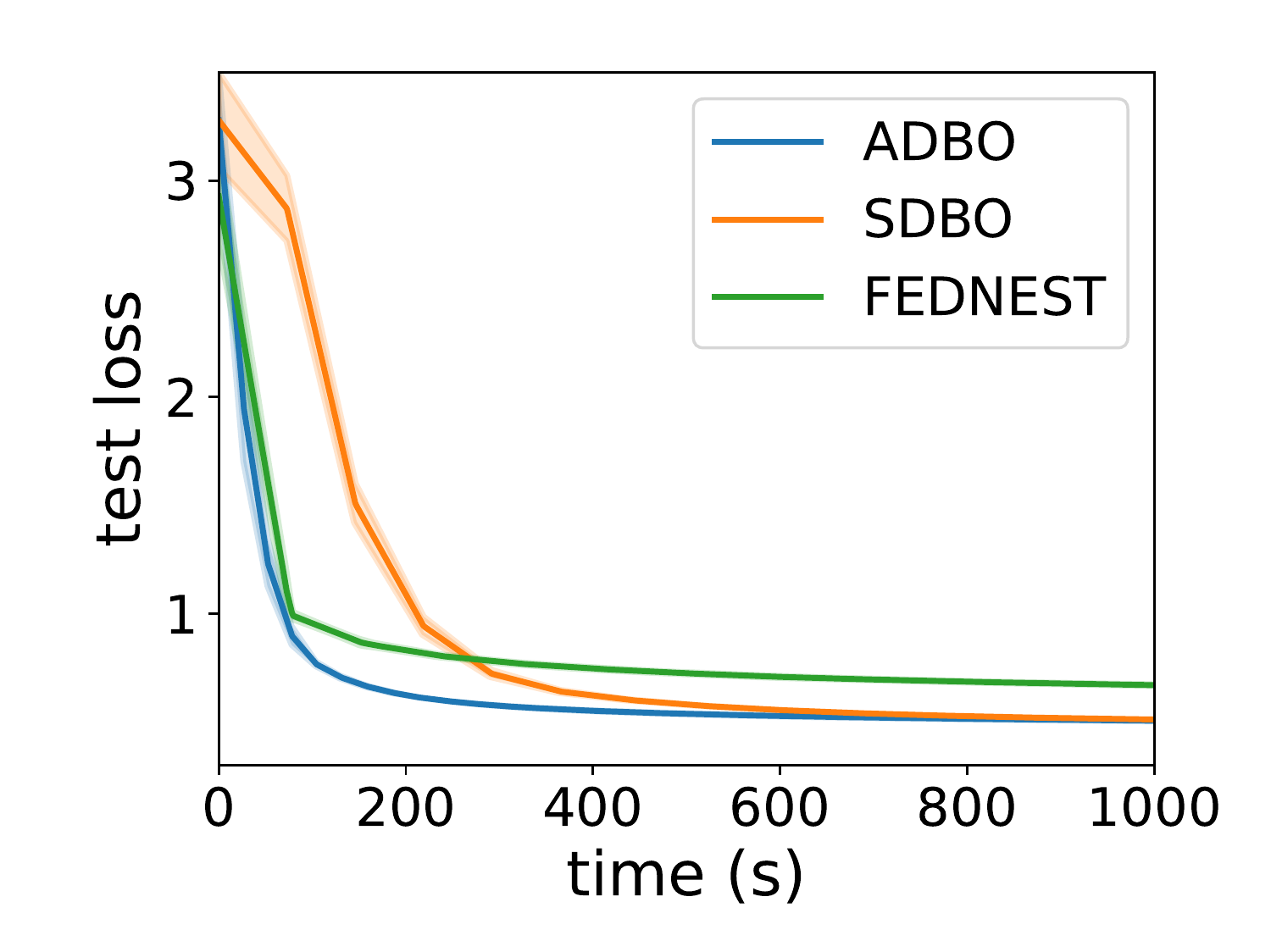}   
	\end{minipage}}
\caption{Test loss vs time on (a) MNIST and (b) Fashion MNIST datasets.} 
\label{fig:hyperclean_loss}  
\end{minipage}

\makeatletter\def\@captype{figure}\makeatother 
\begin{minipage}{0.48\textwidth}  
\setlength{\abovecaptionskip}{-1mm} 
\subfigure[Covertype] 
{\begin{minipage}[t]{0.49\linewidth}
	\centering      
	\includegraphics[scale=0.23]{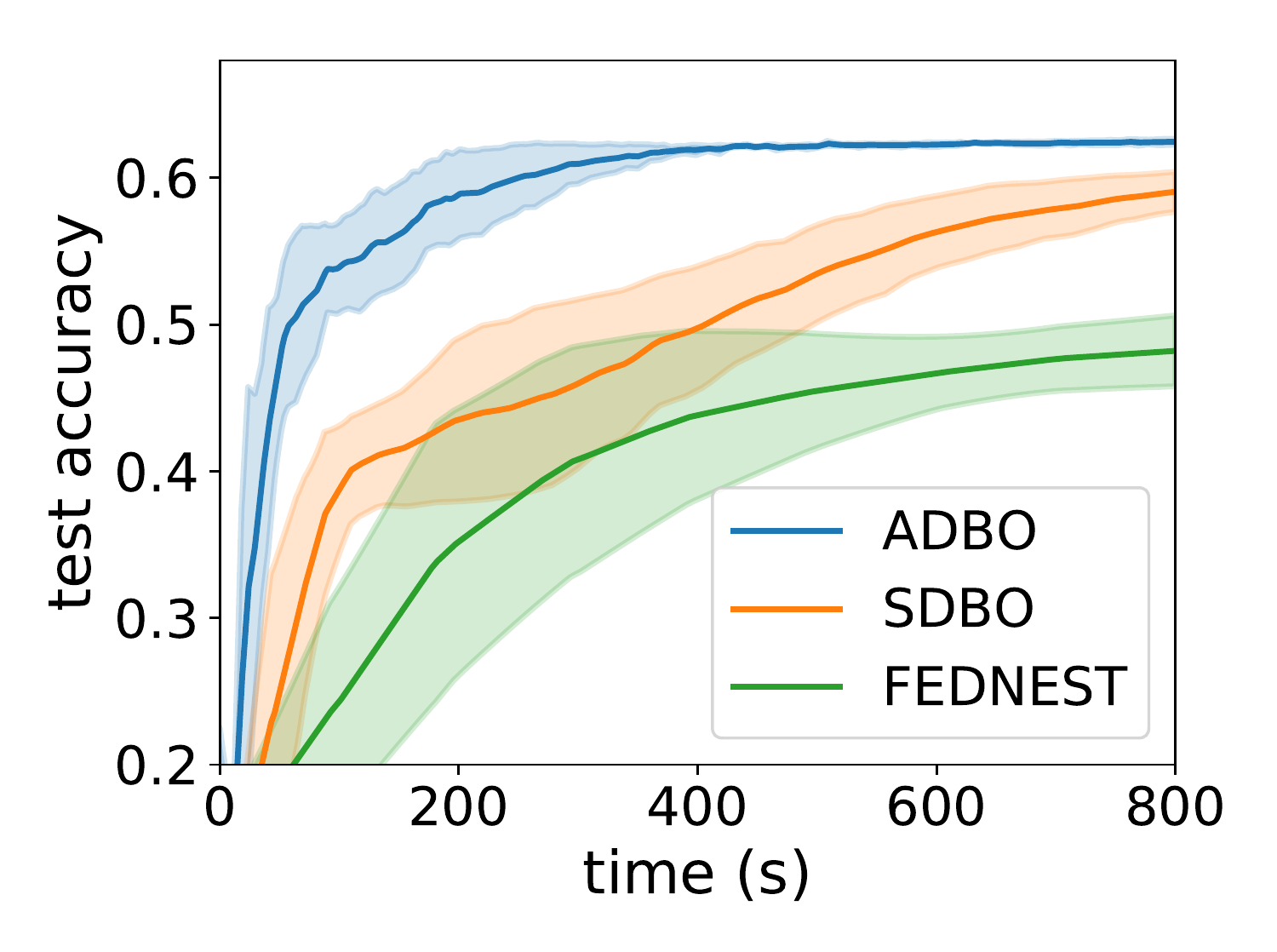}   
	\end{minipage}}
\subfigure[IJCNN1] 
{\begin{minipage}[t]{0.49\linewidth}
	\centering      
	\includegraphics[scale=0.23]{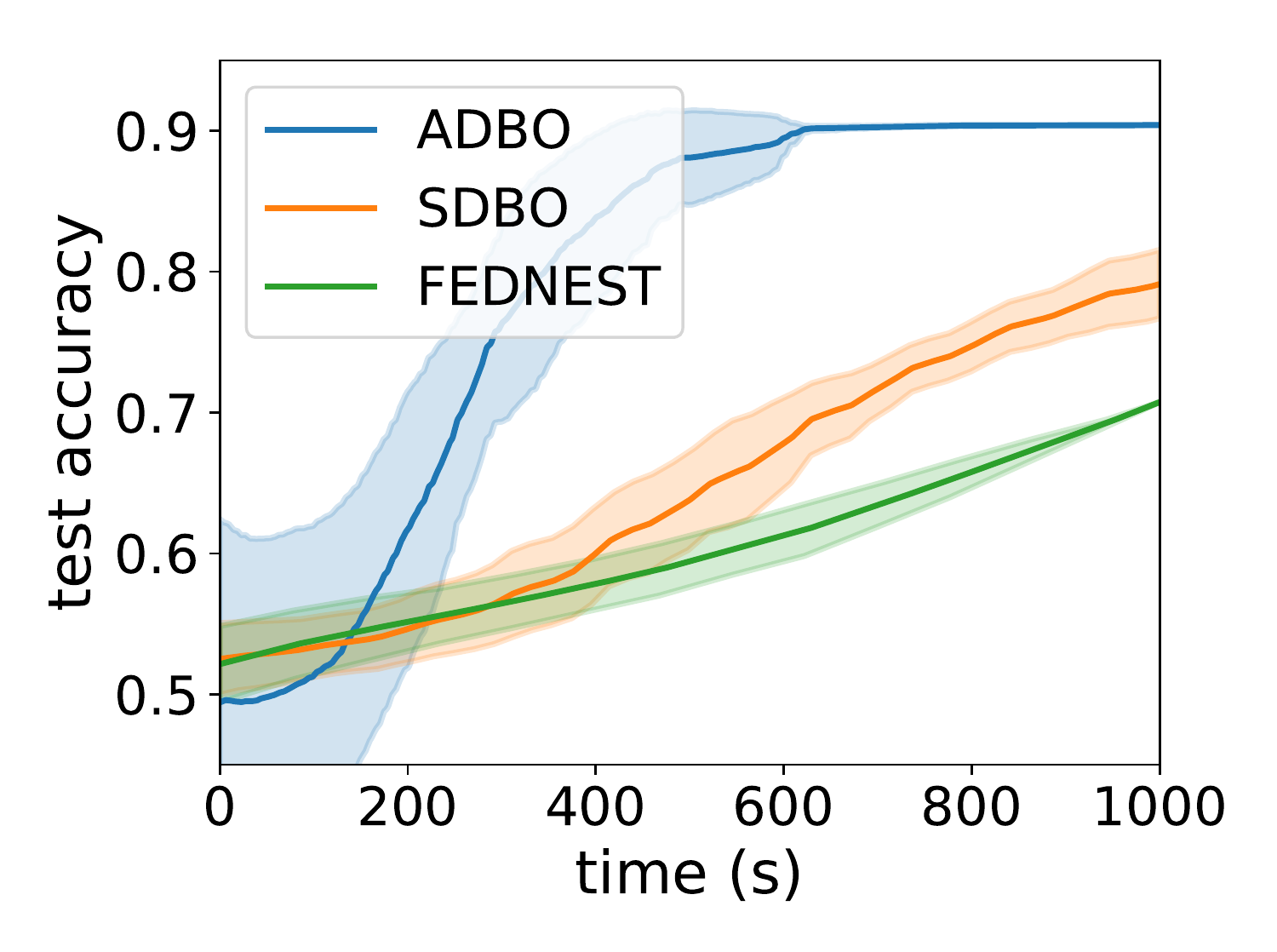}   
	\end{minipage}}
\caption{Test accuracy vs time on (a) Covertype and (b) IJCNN1 datasets.} 
\label{fig:regularizer_acc}  
\end{minipage} $\;$
\makeatletter\def\@captype{figure}\makeatother 
\begin{minipage}{0.48\textwidth} 
\setlength{\abovecaptionskip}{-1mm} 
\subfigure[Covertype] 
{\begin{minipage}[t]{0.49\linewidth}
	\centering      
	\includegraphics[scale=0.23]{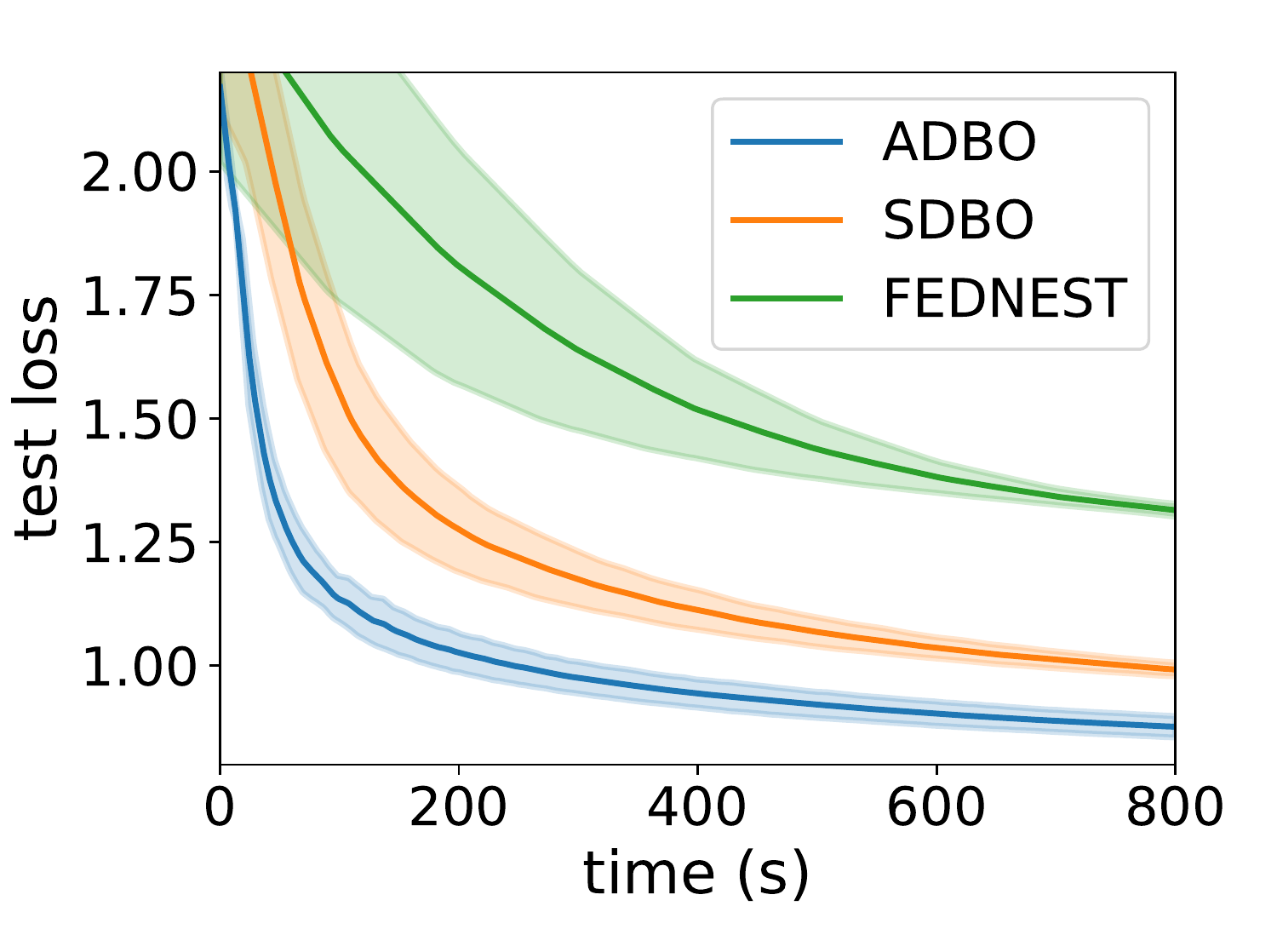}   
	\end{minipage}}
\subfigure[IJCNN1] 
{\begin{minipage}[t]{0.49\linewidth}
	\centering      
	\includegraphics[scale=0.23]{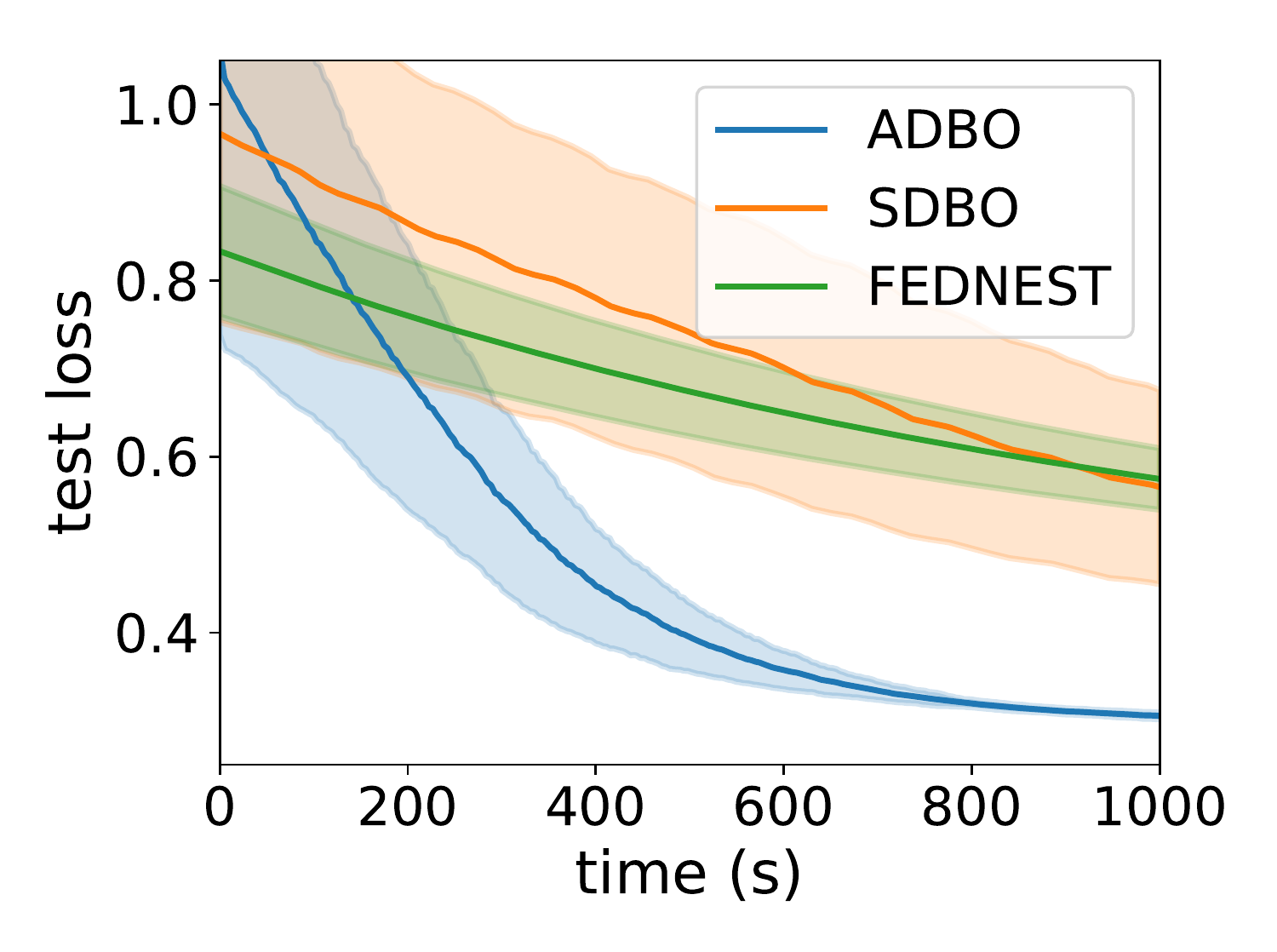}   
	\end{minipage}}
\caption{Test loss vs time on (a) Covertype and (b) IJCNN1 datasets.} 
\label{fig:regularizer_loss}  
\end{minipage}

\makeatletter\def\@captype{figure}\makeatother 
\begin{minipage}{0.48\textwidth}  
\setlength{\abovecaptionskip}{-1mm} 
\subfigure[Covertype] 
{\begin{minipage}[t]{0.49\linewidth}
	\centering      
	\includegraphics[scale=0.23]{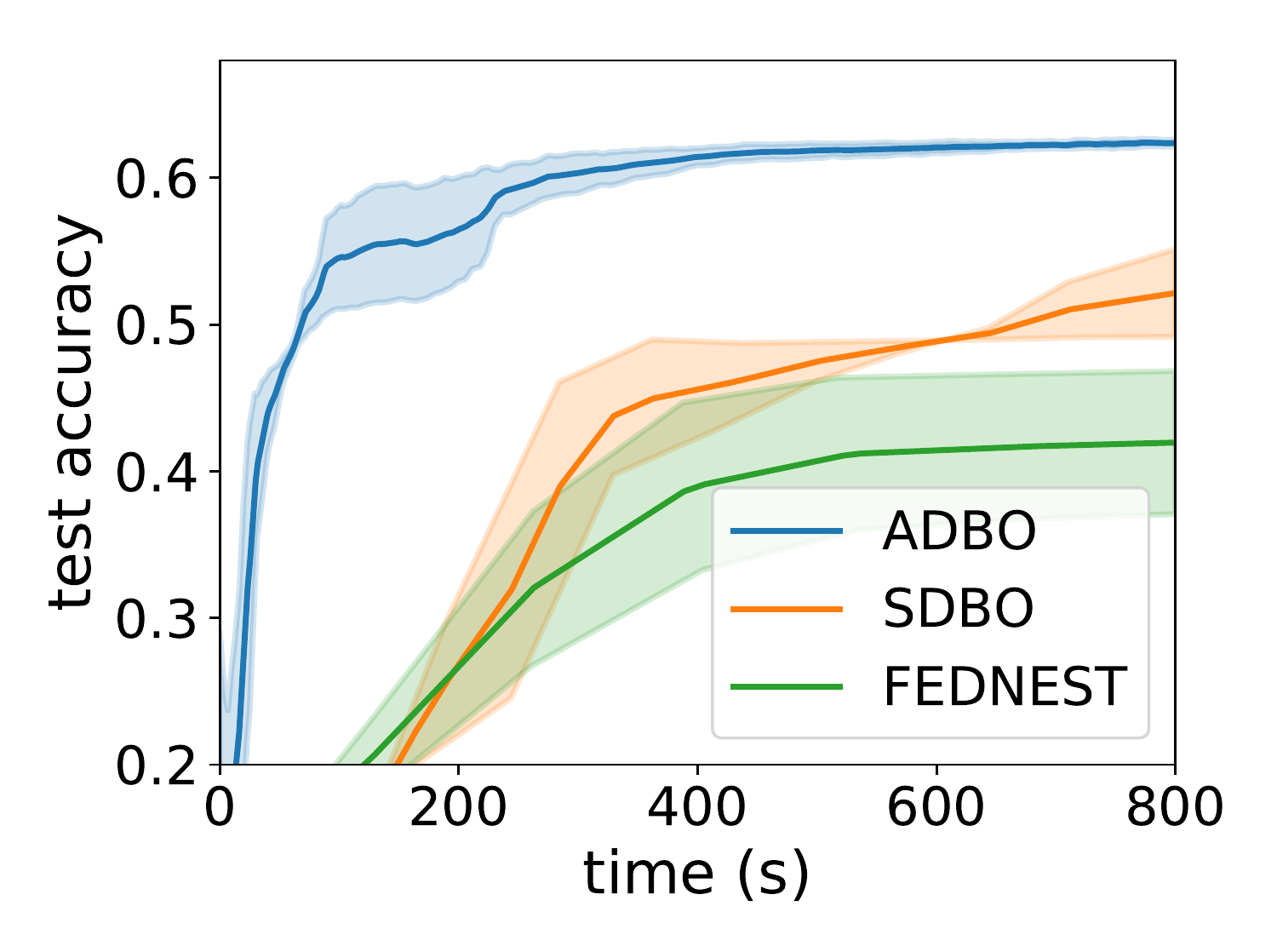}   
	\end{minipage}}
\subfigure[IJCNN1] 
{\begin{minipage}[t]{0.49\linewidth}
	\centering      
	\includegraphics[scale=0.23]{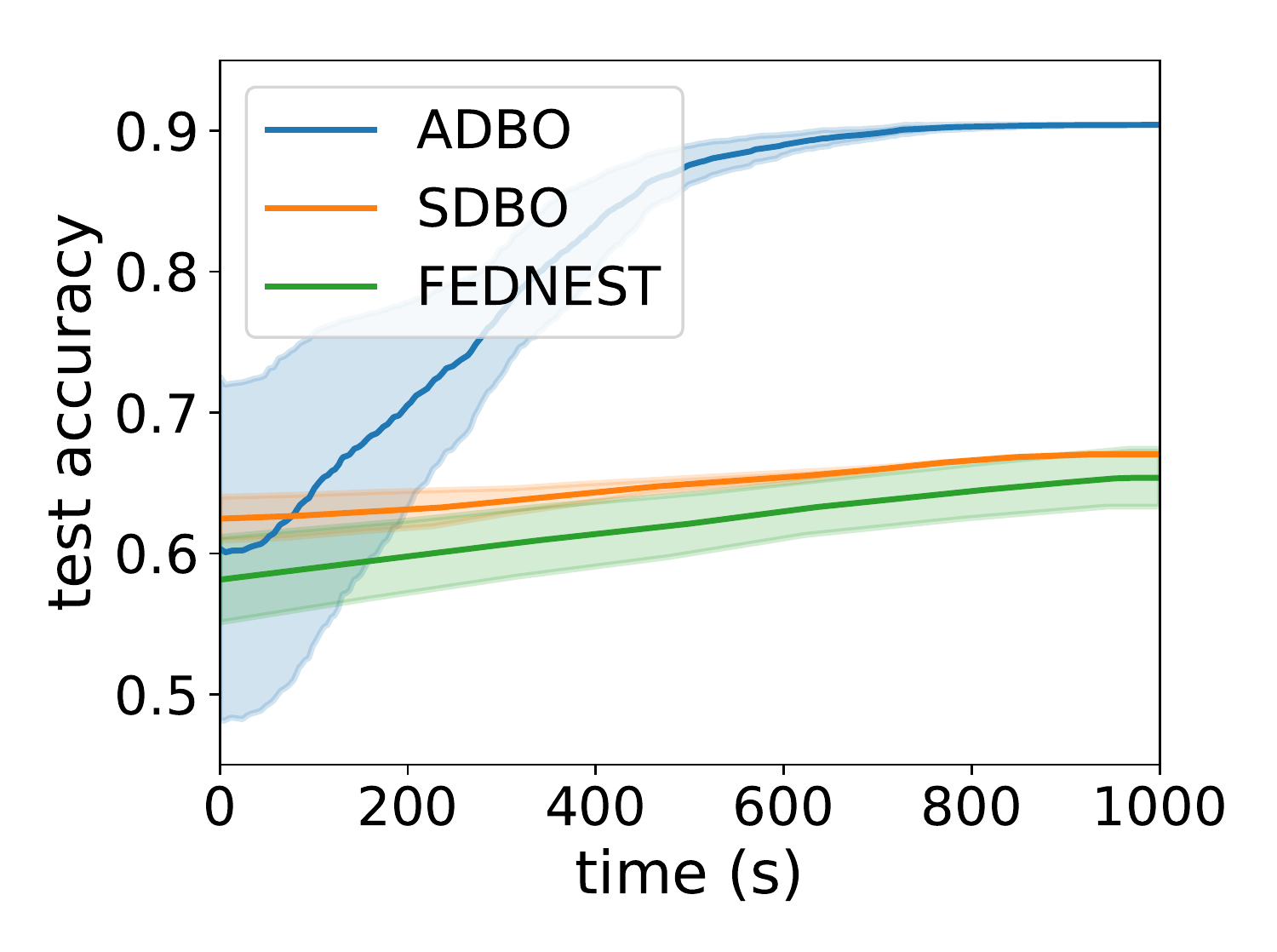}   
	\end{minipage}}
\caption{Test accuracy vs time on (a) Covertype and (b) IJCNN1 datasets when there are stragglers in distributed system.} 
\label{fig:regularizer_acc_latency}  
\vspace{2.5mm}
\end{minipage} $\;$
\makeatletter\def\@captype{figure}\makeatother 
\begin{minipage}{0.48\textwidth} 
\setlength{\abovecaptionskip}{-1mm} 
\subfigure[Covertype] 
{\begin{minipage}[t]{0.49\linewidth}
	\centering      
	\includegraphics[scale=0.23]{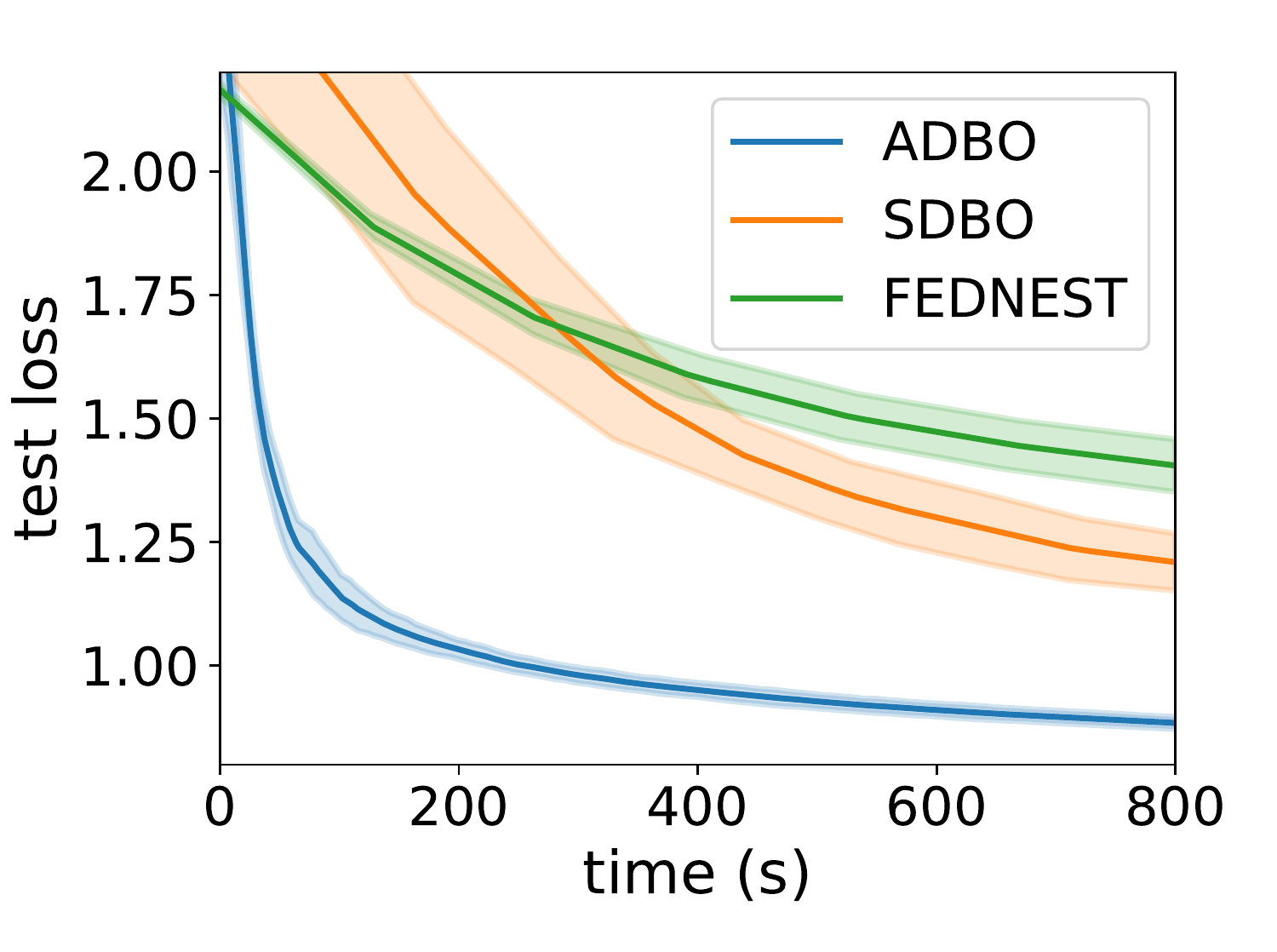}   
	\end{minipage}}
\subfigure[IJCNN1] 
{\begin{minipage}[t]{0.49\linewidth}
	\centering      
	\includegraphics[scale=0.23]{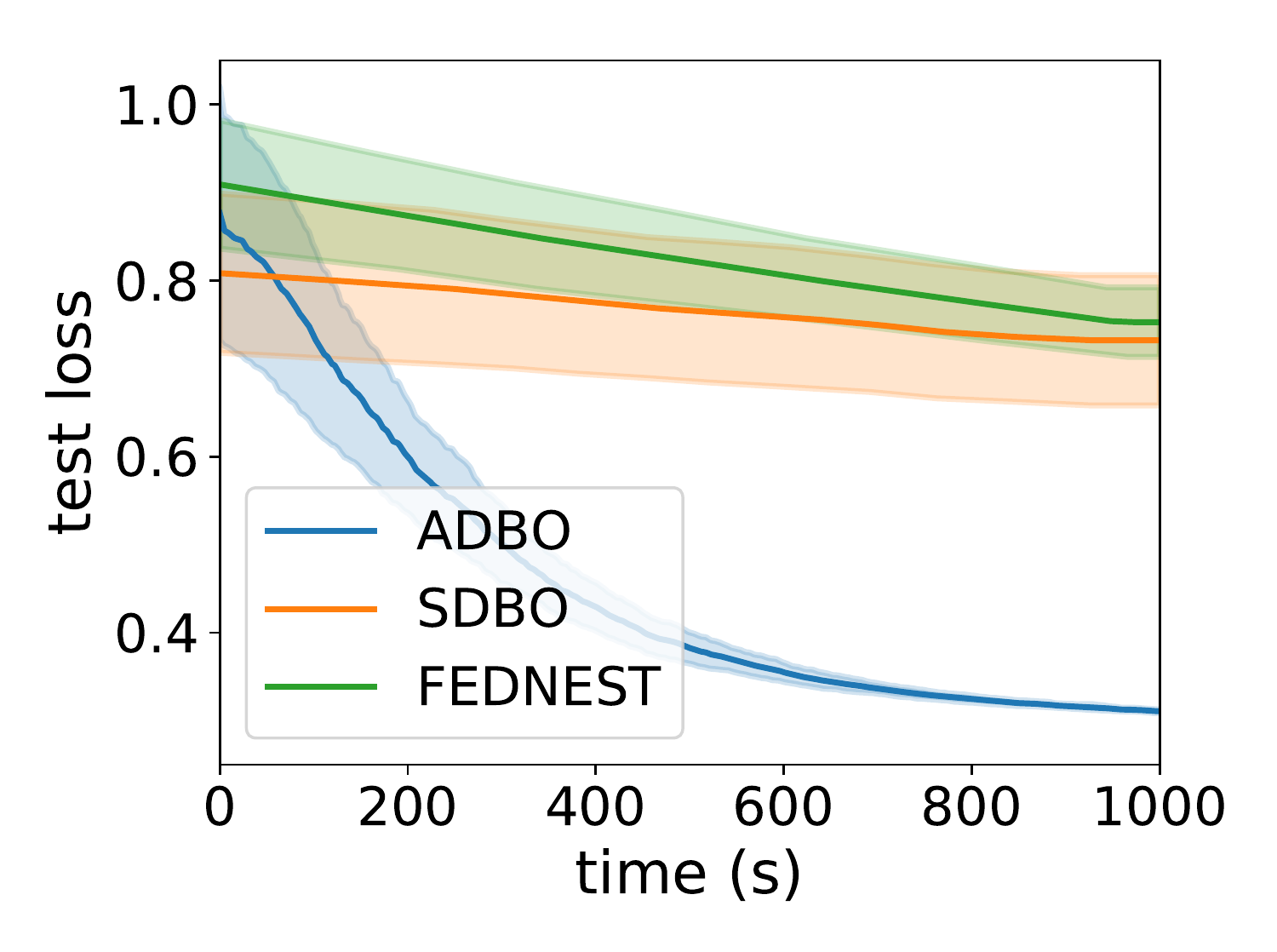}   
	\end{minipage}}
\caption{Test loss vs time on (a) Covertype and (b) IJCNN1 datasets when there are stragglers in distributed system.} 
\label{fig:regularizer_loss_latency}  
\vspace{2.5mm}
\end{minipage}

We also consider the straggler problem, \textit{i.e.}, there exist workers with high delays (stragglers) in the distributed system. In this case, the efficiency of the bilevel optimization method with the synchronous distributed setting will be affected heavily. In the experiment, we assume there are three stragglers in the distributed system, and the mean of (communication + computation) delay of stragglers is four times the delay of normal workers. The results on Covertype and IJCNN1 datasets are reported in Figure \ref{fig:regularizer_acc_latency} and \ref{fig:regularizer_loss_latency}. It is seen that the efficiency of the synchronous distributed algorithms (FEDNEST and SDBO) will be significantly affected, while the proposed ADBO does not suffer from the straggler problem since it is an asynchronous method and is able to only consider active workers.

\vspace{-2mm}

\section{Conclusion}

\vspace{-2mm}

Existing bilevel optimization works focus either on the centralized or synchronous distributed setting, which will give rise to data privacy risks and suffer from the straggler problem. As a remedy, we propose ADBO in this paper to solve the bilevel optimization problem in an asynchronous distributed manner. To our best knowledge, this is the first work that devises the asynchronous distributed algorithm for bilevel optimization. We demonstrate that the proposed ADBO can effectively tackle bilevel optimization problems with both nonconvex upper-level and lower-level objective functions. Theoretical analysis has also been conducted to analyze the convergence properties and iteration complexity of ADBO. Extensive empirical studies on real-world datasets demonstrate the efficiency and effectiveness of the proposed ADBO.


\bibliography{iclr2023_conference}
\bibliographystyle{iclr2023_conference}

\newpage

\appendix
\section{Cutting Plane Method for Bilevel Optimization} 
\label{appendix:centralized BO}
In this section, a cutting plane method, named CPBO, is proposed for bileve optimization. Defining $\phi({\boldsymbol{x}}) = \mathop {\arg \min }\limits_{{\boldsymbol{y}'}} f({\boldsymbol{x}},{\boldsymbol{y}'})$ and $h({\boldsymbol{x}},{\boldsymbol{y}})=||{\boldsymbol{y}} - \phi ({\boldsymbol{x}})|{|^2}$, we can reformulate problem in Eq. (\ref{eq:1_new}) as:
\begin{equation}
\label{eq:2_new}
\begin{array}{l}
\mathop {\min } {\rm{ }}\;\;\, F({\boldsymbol{x}},{\boldsymbol{y}})\\

\, {\rm{s.t.}} \; \quad  h({\boldsymbol{x}},{\boldsymbol{y}})\! =\! 0\\
 
\, {\rm{var.}} \; \quad {\boldsymbol{x}}, {\boldsymbol{y}}.
\end{array}
\end{equation}

Following the previous works \citep{li2022local,gould2016differentiating,yang2021provably} in bilevel optimization, it is not necessary to get the exact $\phi ({\boldsymbol{x}})$, and the approximate $\phi ({\boldsymbol{x}})$ is given as follows. Firstly, as many work do \citep{ji2021bilevel,yang2021provably}, we utilize the $K$ steps of gradient descent (GD) to  approximate $\phi({\boldsymbol{x}})$. And the first-order Taylor approximation of $f({\boldsymbol{x}},{{\boldsymbol{y}'}})$ with respect to ${\boldsymbol{x}}$ is considered, \textit{i.e.}, for a given point $\overline{\boldsymbol{x}}$,  $\widetilde{f}({\boldsymbol{x}},{{\boldsymbol{y}'}})=f(\overline{\boldsymbol{x}},{{\boldsymbol{y}'}}) + {\nabla _{\boldsymbol{x}}}f{(\overline{\boldsymbol{x}},{{\boldsymbol{y}'}})^\top  }({\boldsymbol{x}} - \overline{\boldsymbol{x}})$. Thus, we have,
\begin{equation}
\label{eq:3_new_11_16}
\phi({\boldsymbol{x}}) = {{\boldsymbol{y}}_0'} - \sum\nolimits_{k = 0}^{K - 1} {\eta {\nabla _{\boldsymbol{y}}}\widetilde{f}({\boldsymbol{x}},{{\boldsymbol{y}}_k'})} ,
\end{equation}
where $\eta$ is the step-size. Considering the estimated $\phi({\boldsymbol{x}})$ in Eq. (\ref{eq:3_new_11_16}), the relaxed problem with respect to problem in Eq. (\ref{eq:2_new}) is considered as follows,
\begin{equation}
\label{eq:3_new}
\begin{array}{l}
\mathop {\min } {\rm{ }}\;\;\, F({\boldsymbol{x}},{\boldsymbol{y}})\\

\, {\rm{s.t.}} \; \quad  h({\boldsymbol{x}},{\boldsymbol{y}})\! \le\! \varepsilon\\
 
\, {\rm{var.}} \; \quad {\boldsymbol{x}}, {\boldsymbol{y}}.
\end{array}
\end{equation}

Assuming that $h({\boldsymbol{x}},{\boldsymbol{y}})$ is a convex function with respect to $({\boldsymbol{x}},{\boldsymbol{y}})$, which is always satisfied when we set $K=1$ in Eq. (\ref{eq:3_new_11_16}) according to the operations that preserve convexity \citep{boyd2004convex}. Since the sublevel set of a convex function is convex, we have that the feasible set of $({\boldsymbol{x}},{{\boldsymbol{y}}})$, \textit{i.e.},
\begin{equation}
\label{eq:5_new}
{{{\boldsymbol{Z}}}^{relax}} = \{ ({{\boldsymbol{x}}},{{\boldsymbol{y}}}) \in \mathbb{R}^n \!\times\! \mathbb{R}^m|{\rm{ }}h({{\boldsymbol{x}}},{{\boldsymbol{y}}}) \le {\epsilon}\} ,
\end{equation}
is a convex set. We utilize a set of cutting plane constraints (\textit{i.e.}, linear constraints) to approximate the feasible set ${{{\boldsymbol{Z}}}^{relax}}$. The set of cutting plane constraints forms a polytope, which can be expressed as follows,
\begin{equation}
\label{eq:6_new}
{\boldsymbol{\mathcal{P}}} = \{ ({\boldsymbol{x}},{\boldsymbol{y}}) \in \mathbb{R}^n \!\times\! \mathbb{R}^m|{{\boldsymbol{a}}_l}^\top  {\boldsymbol{x}} + {{\boldsymbol{b}}_l}^\top  {\boldsymbol{y}} + {\kappa_l} \le 0,\;l = 1, \cdots ,L\},
\end{equation}
where $\boldsymbol{a}_l \! \in \! \mathbb{R}^n$, $\boldsymbol{b}_l \! \in \! \mathbb{R}^m$ and ${\kappa_l} \! \in \! \mathbb{R}^1$ are parameters in $l^{\rm{th}}$ cutting plane, and $L$ represents the number of cutting planes in ${\boldsymbol{\mathcal{P}}}$. Considering the approximate problem,  which can be expressed as follows,
\begin{align}
\label{eq:b103}
&\mathop {\min } \quad F({\boldsymbol{x}}, {\boldsymbol{y}}) \nonumber\\
&\; {\rm{{\rm{s.t.}}}} \quad {{\boldsymbol{a}}_l}^\top  {\boldsymbol{x}} \! + \! {{\boldsymbol{b}}_l}^\top  {\boldsymbol{y}} \! + \! {\kappa_l} \le 0, \;  l\! = \! 1,\! \cdots \!,  |{\boldsymbol{\mathcal{P}}^{t}  }|\\
& \; {\rm{var.}} \quad \; {\boldsymbol{x}}, {\boldsymbol{y}},  \nonumber
\end{align}
where ${\boldsymbol{\mathcal{P}}^{t}  }$ is the polytope in $(t+1)^{{\rm{th}}}$ iteration, and $|{\boldsymbol{\mathcal{P}}^{t}  }|$ denotes the number of cutting planes in ${\boldsymbol{\mathcal{P}}^{t}  }$. Then, the Lagrangian function of Eq. (\ref{eq:b103}) can be written as,
\begin{equation}
\renewcommand{\theequation}{\arabic{equation}}
\label{eq:B104}
 {L_p} (\boldsymbol{x},\boldsymbol{y},\{ {\lambda _l}\} ) = F(\boldsymbol{x},\boldsymbol{y}) + \sum\limits_{l = 1}^{|{{\boldsymbol{\mathcal{P}}}^{t}  }|} {{\lambda _l}({{\boldsymbol{a}}_l}^\top  {\boldsymbol{x}} \! + \! {{\boldsymbol{b}}_l}^\top  {\boldsymbol{y}} \! + \! {\kappa_l})},
\end{equation}
where ${\lambda _l}$ is the dual variable. The proposed algorithm proceeds as follows in $(t+1)^{\rm{th}}$ iteration:

If $t < T_1$, the variables are updated as follows,
\begin{equation}
\renewcommand{\theequation}{\arabic{equation}}
\label{eq:B106}
{\boldsymbol{x}}^{t+1} = {\boldsymbol{x}}^{t} - {\eta _{\boldsymbol{x}}}{\nabla _{{\boldsymbol{x}}}} {{L}_p} (\boldsymbol{x}^{t},\boldsymbol{y}^{t}, \{ {\lambda _l^{t}}\} ),
\end{equation}
\begin{equation}
\renewcommand{\theequation}{\arabic{equation}}
\label{eq:B107}
{\boldsymbol{y}}^{t+1} = {\boldsymbol{y}}^{t} - {\eta _{\boldsymbol{y}}}{\nabla _{{\boldsymbol{y}}}} {{L}_p} (\boldsymbol{x}^{t+1},\boldsymbol{y}^{t}, \{ {\lambda _l^{t}}\} ),
\end{equation}
\begin{equation}
\renewcommand{\theequation}{\arabic{equation}}
\label{eq:B108}
{\lambda _l^{t+1}} = \lambda _l^{t} + {\eta _{\lambda _l}}{\nabla _{{\lambda _l}}} {{L}_p} (\boldsymbol{x}^{t+1},\boldsymbol{y}^{t+1}, \{ {\lambda _l^{t}}\} ),\;  l\! = \! 1,\! \cdots \!,  |{{\boldsymbol{\mathcal{P}}}^{t}  }|,
\end{equation}
where ${\eta _{\boldsymbol{x}}}$, ${\eta_{\boldsymbol{y}}}$ and ${\eta _{\lambda _l}}$ are the step-sizes.

\renewcommand\arraystretch{1.2}
\renewcommand\tabcolsep{5pt}
\begin{table*}[t]
\renewcommand{\thetable}{\arabic{table}}
\caption{Convergence results of bilevel optimization algorithms (with centralized and distributed setting).}
{
\scalebox{0.85}{
\begin{tabular}{l|c|c|c}
\toprule
Method    & Centralized     & Synchronous (Distributed) & Asynchronous (Distributed)      \\ \hline
AID-BiO \citep{ghadimi2018approximation} & $\mathcal{O}(\frac{1}{{{\epsilon ^{1.25}}}})$  & NA & NA \\ 
AID-BiO \citep{ji2021bilevel} & $\mathcal{O}(\frac{1}{{{\epsilon ^1}}})$  & NA & NA \\ 
ITD-BiO \citep{ji2021bilevel} & $\mathcal{O}(\frac{1}{{{\epsilon ^1}}})$  & NA & NA \\ 
STABLE \citep{chen2022single} & $\mathcal{O}(\frac{1}{{{\epsilon ^2}}})^1$  & NA & NA \\ 
stocBio \citep{ji2021bilevel} & $\mathcal{O}(\frac{1}{{{\epsilon ^2}}})^1$  & NA & NA \\
VRBO \citep{yang2021provably} & $\mathcal{O}(\frac{1}{{{\epsilon ^{1.5}}}})^1$  & NA & NA \\
FEDNEST \citep{tarzanagh2022fednest} &  NA  & $\mathcal{O}(\frac{1}{{{\epsilon ^{2}}}})^1$ & NA \\
SPDB \citep{lu2022decentralized}   & NA &$\mathcal{O}(\frac{1}{{{\epsilon ^{2}}}})^1$ & NA \\
DSBO \citep{yang2022decentralized}   & NA &$\mathcal{O}(\frac{1}{{{\epsilon ^{2}}}})^1$ & NA \\\hline
\textbf{Proposed Method} & $\mathcal{O}(\frac{1}{{{\epsilon ^1}}})$ & NA & $\mathcal{O}(\frac{1}{{{\epsilon ^2}}})$\\
\bottomrule  
\end{tabular}}
\label{tab:convergence rate compare}}
\\\footnotesize{$^1$ Stochastic optimization algorithm.}
\vspace{-2mm}
\end{table*}

And every $k_{\rm{pre}}$ iterations ($k_{\rm{pre}}\!>\!0$ is a pre-set constant, which can be controlled flexibly) the cutting planes will be updated based on the following two steps: 

(a) Removing the inactive cutting planes, that is,
\begin{equation}
\label{eq:13_new}
{{\boldsymbol{\mathcal{P}}}^{t + 1}} = \left\{ \begin{array}{l}
{\rm{Drop(}}{{\boldsymbol{\mathcal{P}}}^{t}  },c{p_l}{\rm{),  if \;  }}{\lambda _l^{t + 1}} \;{\rm{and}} \; {\lambda _l^{t}} = 0\\
{{\boldsymbol{\mathcal{P}}}^{t}  },{\rm{otherwise}}
\end{array} \right.,
\end{equation}
where $cp_l$ represents the $l^{\rm{th}}$ cutting plane in ${\boldsymbol{\mathcal{P}}}^{t}  $, and ${\rm{Drop(}}{{\boldsymbol{\mathcal{P}}}^{t}  },c{p_l})$ represents removing the $l^{\rm{th}}$ cutting plane $cp_l$ from ${\boldsymbol{\mathcal{P}}}^{t}  $. And the dual variable set $\{ {\lambda ^{t}  }\}$ will be updated as follows,
\begin{equation}
\label{eq:14_new}
\{ {\lambda ^{t + 1}}\} \! =\! \left\{ \begin{array}{l}
{\rm{Drop(}}\{ {\lambda ^{t}  }\} ,{\lambda _l^{t}}{\rm{),  if \;  }}{\lambda _l^{t + 1}} \;{\rm{and}} \; {\lambda _l^{t}} = 0\\
\{ {\lambda ^{t}  }\} ,{\rm{ otherwise }}
\end{array} \right.,
\end{equation}
where $\{ {\lambda ^{t + 1}}\} $ and $\{ {\lambda ^{t}}\} $ respectively represent the dual variable set in $(t+1)^{{\rm{th}}}$ and $t^{{\rm{th}}}$ iteration. And ${\rm{Drop(}}\{ {\lambda ^{t}  }\} ,{\lambda _l^{t}})$ represents that ${\lambda _l^{t}}$ is removed from the dual variable set $\{ {\lambda ^{t}  }\}$. 

(b) Adding new cutting planes. Firstly, we investigate whether $({\boldsymbol{x}}^{t+1}, {\boldsymbol{y}}^{t+1})$ is a feasible solution to the original problem in Eq. (\ref{eq:3_new}).  If $({\boldsymbol{x}}^{t+1}, {\boldsymbol{y}}^{t+1})$ is not a feasible solution to the original problem, that is $h({\boldsymbol{x}}^{t+1}, {\boldsymbol{y}}^{t+1}) > \varepsilon $, new cutting plane is generated to separate the point $({\boldsymbol{x}}^{t+1}, {\boldsymbol{y}}^{t+1})$ from ${{{\boldsymbol{Z}}}^{relax}}$, that is, the \textit{valid} cutting plane ${{\boldsymbol{a}}_l}^\top  {\boldsymbol{x}} \! + \! {{\boldsymbol{b}}_l}^\top  {\boldsymbol{y}} \! + \! {\kappa_l} \le 0$ must satisfy that,
\begin{equation}
\label{eq:15_new}
\left\{ \begin{array}{l}
 {{\boldsymbol{a}}_l}^\top  {\boldsymbol{x}} \! + \! {{\boldsymbol{b}}_l}^\top  {\boldsymbol{y}} \! + \! {\kappa_l} \le 0, \forall (\boldsymbol{x}, \boldsymbol{y}) \in {{{\boldsymbol{Z}}}^{relax}}\\
{{\boldsymbol{a}}_l}^\top  {\boldsymbol{x}^{t+1}} \! + \! {{\boldsymbol{b}}_l}^\top  {\boldsymbol{y}^{t+1}} \! + \! {\kappa_l} > 0
\end{array} \right..
\end{equation}

Since $h(\boldsymbol{x}, \boldsymbol{y})$ is a convex function, we have that,
\begin{equation}
\begin{array}{l}
\label{eq:16_new}
 h(\boldsymbol{x},\boldsymbol{y})\ge  h(\boldsymbol{x}^{t + 1},\boldsymbol{y}^{t + 1})  + {\left[ \begin{array}{l}
\frac{{\partial h(\boldsymbol{x}^{t + 1},\boldsymbol{y}^{t + 1})}}{{\partial \boldsymbol{x}}}\\
\frac{{\partial h(\boldsymbol{x}^{t + 1},\boldsymbol{y}^{t + 1})}}{{\partial \boldsymbol{y}}}
\end{array} \right]^\top  }\!\left( {\left[ \begin{array}{l}
\boldsymbol{x}\\
\boldsymbol{y}
\end{array} \right]\! -\! \left[ \begin{array}{l}
\boldsymbol{x}^{t + 1}\\
\boldsymbol{y}^{t + 1}
\end{array} \right]} \right)\end{array}.
\end{equation}

According to Eq. (\ref{eq:16_new}), $h(\boldsymbol{x}^{t + 1},\boldsymbol{y}^{t + 1}) + {\left[ \begin{array}{l}
\frac{{\partial h(\boldsymbol{x}^{t + 1},\boldsymbol{y}^{t + 1})}}{{\partial \boldsymbol{x}}}\\
\frac{{\partial h(\boldsymbol{x}^{t + 1},\boldsymbol{y}^{t + 1})}}{{\partial \boldsymbol{y}}}
\end{array} \right]^\top  }\!\left( {\left[ \begin{array}{l}
\boldsymbol{x}\\
\boldsymbol{y}
\end{array} \right]\! -\! \left[ \begin{array}{l}
\boldsymbol{x}^{t + 1}\\
\boldsymbol{y}^{t + 1}
\end{array} \right]} \right) \le \varepsilon$ is a valid cutting plane at point $({\boldsymbol{x}}^{t+1}, {\boldsymbol{y}}^{t+1})$ which satisfies Eq. (\ref{eq:15_new}). For brevity, we utilize $cp^{t+1}_{new}$ to denote this cutting plane. Thus, we have that,
\begin{equation}
\label{eq:17_new}
    {{\boldsymbol{\mathcal{P}}}^{t + 1}} = \left\{ \begin{array}{l}
{\rm{Add(}}{{\boldsymbol{\mathcal{P}}}^{t + 1}},cp_{new}^{t + 1}),{\rm{  if \; }}h(\boldsymbol{x}^{t + 1},\boldsymbol{y}^{t + 1}) > \varepsilon \\
{{\boldsymbol{\mathcal{P}}}^{t + 1}},{\rm{  if \; }}h(\boldsymbol{x}^{t + 1},\boldsymbol{y}^{t + 1}) \le \varepsilon 
\end{array} \right.,
\end{equation}
where ${\rm{Add(}}{{\boldsymbol{\mathcal{P}}}^{t + 1}},cp_{new}^{t + 1})$ represents that new cutting plane  $cp^{t+1}_{new}$ is added to polytope ${{\boldsymbol{\mathcal{P}}}^{t + 1}}$. And the dual variable set is updated as follows,
\begin{equation}
\label{eq:18_new}
    \{ {\lambda ^{t + 1}}\}  = \left\{ \begin{array}{l}
{\rm{Add(}}\{ {\lambda ^{t + 1}}\} ,{\lambda _{|{{\boldsymbol{\mathcal{P}}}^{t + 1}}|}^{t + 1}}{\rm{),  if \; }}h(\boldsymbol{x}^{t + 1},\boldsymbol{y}^{t + 1}) > \varepsilon \\
\{ {\lambda ^{t + 1}}\} ,{\rm{  if \; }}h(\boldsymbol{x}^{t + 1},\boldsymbol{y}^{t + 1}) \le \varepsilon
\end{array} \right.,
\end{equation}
where ${\rm{Add(}}\{ {\lambda ^{t + 1}}\} ,{\lambda _{|{{\boldsymbol{\mathcal{P}}}^{t + 1}}|}^{t + 1}})$ represents that new dual variable ${\lambda _{|{{\boldsymbol{\mathcal{P}}}^{t + 1}}|}^{t + 1}}$ is added to $\{ {\lambda ^{t + 1}}\}$.

Else if $t \ge T_1$, the polytope ${{\boldsymbol{\mathcal{P}}}^{T_1}}$ and dual variables will be fixed. Variables $\boldsymbol{x}, \boldsymbol{y}$ will be updated as follows,
\begin{equation}
\renewcommand{\theequation}{\arabic{equation}}
\label{eq:96_18}
{\boldsymbol{x}}^{t+1} = {\boldsymbol{x}}^{t} - {\eta _{\boldsymbol{x}}}{\nabla _{{\boldsymbol{x}}}} {\hat{L}_p} (\boldsymbol{x}^{t},\boldsymbol{y}^{t} ),
\end{equation}
\begin{equation}
\renewcommand{\theequation}{\arabic{equation}}
\label{eq:96_19}
{\boldsymbol{y}}^{t+1} = {\boldsymbol{y}}^{t} - {\eta _{\boldsymbol{y}}}{\nabla _{{\boldsymbol{y}}}} {\hat{L}_p} (\boldsymbol{x}^{t+1},\boldsymbol{y}^{t} ),
\end{equation}
where ${\hat{L}_p}(\boldsymbol{x}, \boldsymbol{y})= F(\boldsymbol{x},\boldsymbol{y}) + \sum\limits_{l = 1}^{|{{\boldsymbol{\mathcal{P}}}^{T_1}}|} {{\lambda _l}[\max\{0,{{\boldsymbol{a}}_l}^\top  {\boldsymbol{x}} \! + \! {{\boldsymbol{b}}_l}^\top  {\boldsymbol{y}} \! + \! {\kappa_l}\}]^2} $. And details of the proposed algorithm are summarized in Algorithm \ref{algorithm:centralized}. The comparison about the convergence results between the proposed method and state-of-the-art methods are summarized in Table \ref{tab:convergence rate compare}. 

{\linespread{1}
\begin{algorithm*}[t]
   \caption{CPBO: Cutting Plane Method for Bilevel Optimization}
   \label{algorithm:centralized}
\begin{algorithmic}
   \STATE {\bfseries Initialization:}  iteration $t = 0$, variables ${{\boldsymbol{x}}^0}$, ${{\boldsymbol{y}}^0}$, $\{{\lambda _l^0}\}$ and polytope ${{\boldsymbol{\mathcal{P}}}^0}$.
   
   \REPEAT
   \IF{$t<T_1$}

   \STATE updating variables ${\boldsymbol{x}^{t+1}}$, ${\boldsymbol{y}^{t+1}}$ and ${\lambda _l^{t+1}}$  according to Eq. (\ref{eq:B106}), (\ref{eq:B107}) and (\ref{eq:B108});

   \IF{$(t+1)$ mod $k_{\rm{pre}}$ $==$ 0}
   \STATE   updating the polytope ${{\boldsymbol{\mathcal{P}}}^{t + 1}}$ according to Eq. (\ref{eq:13_new}) and (\ref{eq:17_new});

\STATE   updating the dual variable set $\{ {\lambda ^{t + 1}}\}$ according to Eq. (\ref{eq:14_new}) and (\ref{eq:18_new});

   \ENDIF
   
   \ELSE
    \STATE updating variables ${\boldsymbol{x}^{t+1}}$ and ${\boldsymbol{y}^{t+1}}$ according to Eq. (\ref{eq:96_18}) and (\ref{eq:96_19});
   
   \ENDIF

   \STATE $t =t+1$;
   \UNTIL{termination.}
\end{algorithmic}
\end{algorithm*}}

\begin{figure*}[t]
\centering    
\subfigure[test accuracy vs time] 
{\begin{minipage}[t]{0.38\linewidth}
	\centering      
	\includegraphics[scale=0.32]{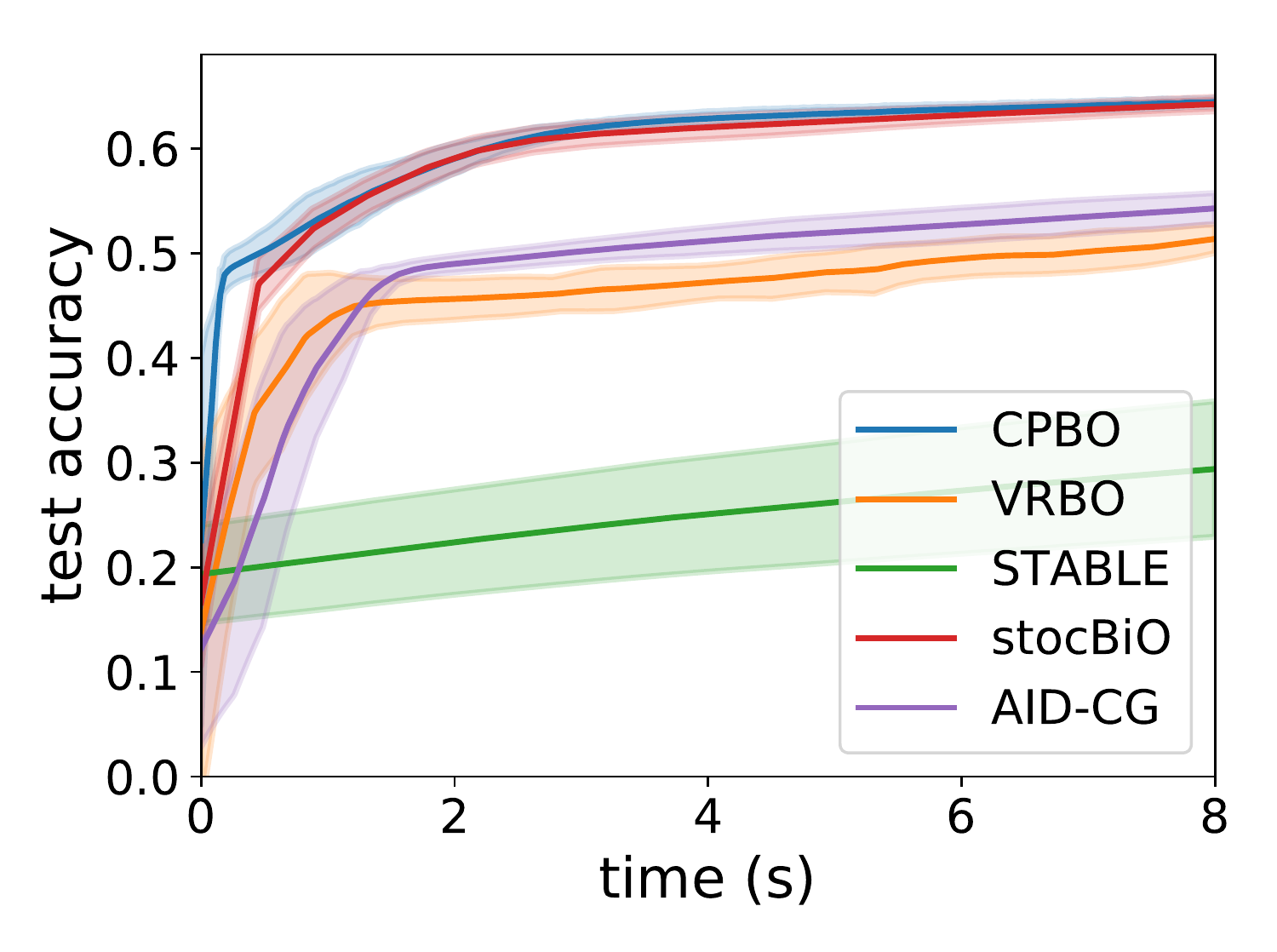}   
	\end{minipage}}
\subfigure[test loss vs time] 
{\begin{minipage}[t]{0.38\linewidth}
	\centering      
	\includegraphics[scale=0.32]{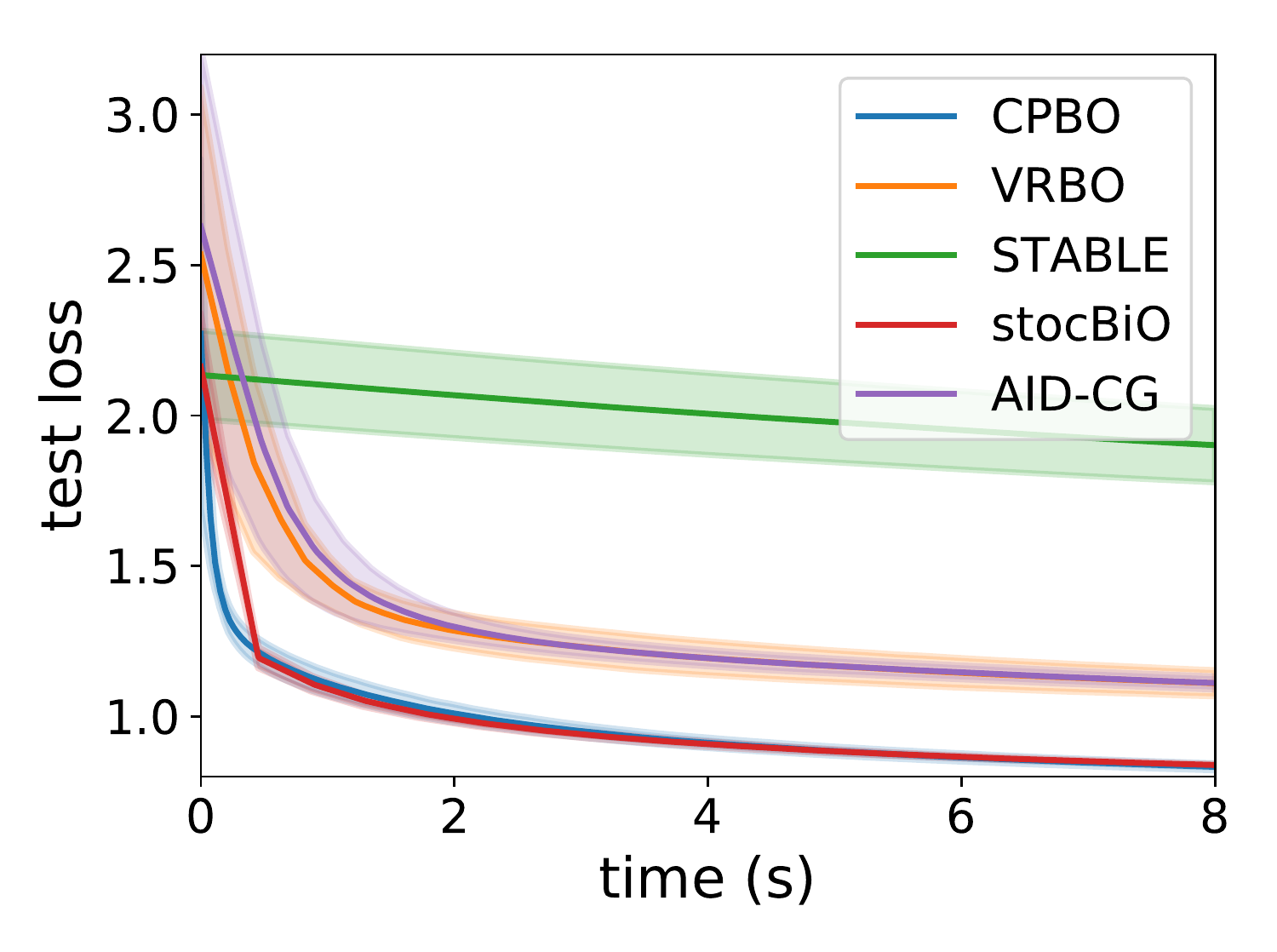}   
	\end{minipage}}
\caption{Comparison of (a) test accuracy vs time, (b) test loss vs time on Covertype dataset.} 
\label{fig:appendix covertype} 
\end{figure*}

\begin{figure*}[t]
\centering    
\subfigure[test accuracy vs time] 
{\begin{minipage}[t]{0.38\linewidth}
	\centering      
	\includegraphics[scale=0.32]{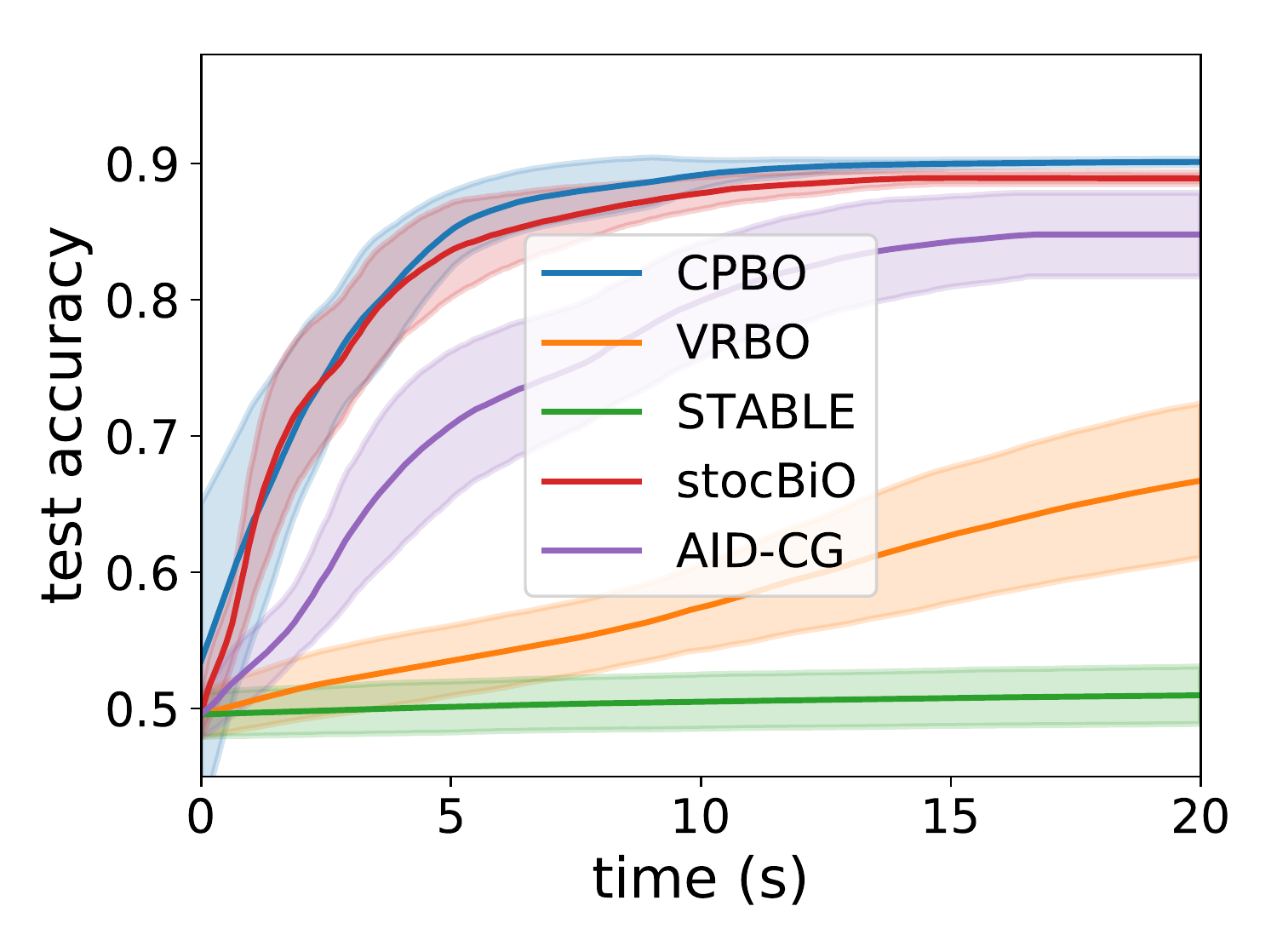}   
	\end{minipage}}
\subfigure[test loss vs time] 
{\begin{minipage}[t]{0.38\linewidth}
	\centering      
	\includegraphics[scale=0.32]{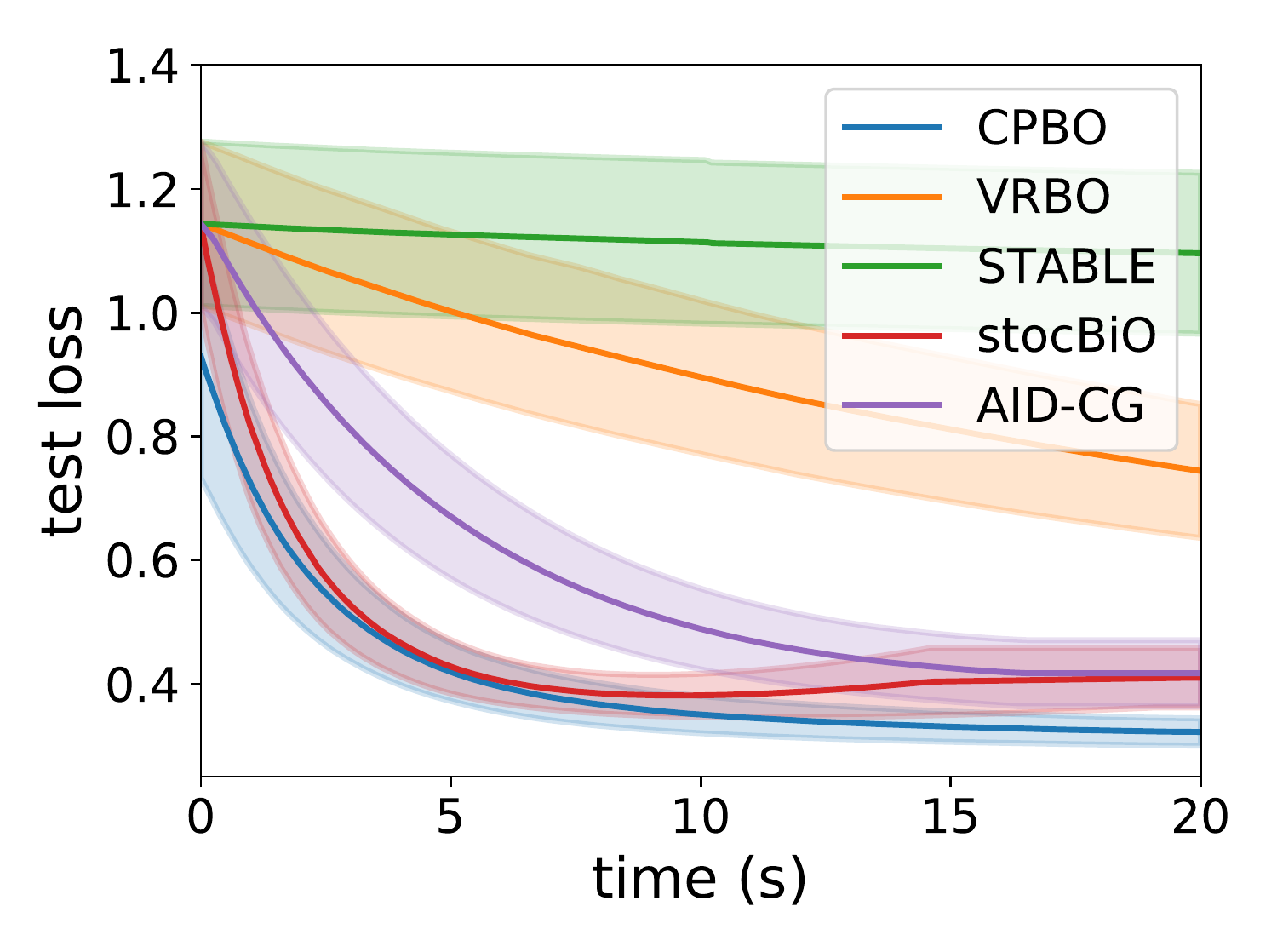}   
	\end{minipage}}
\caption{Comparison of (a) test accuracy vs time, (b) test loss vs time on IJCNN1 dataset.} 
\label{fig:appendix ijcnn1} 
\end{figure*}

\subsection{Experiment}
\begin{figure*}[t]
\centering    
\subfigure[test accuracy vs time] 
{\begin{minipage}[t]{0.38\linewidth}
	\centering      
	\includegraphics[scale=0.32]{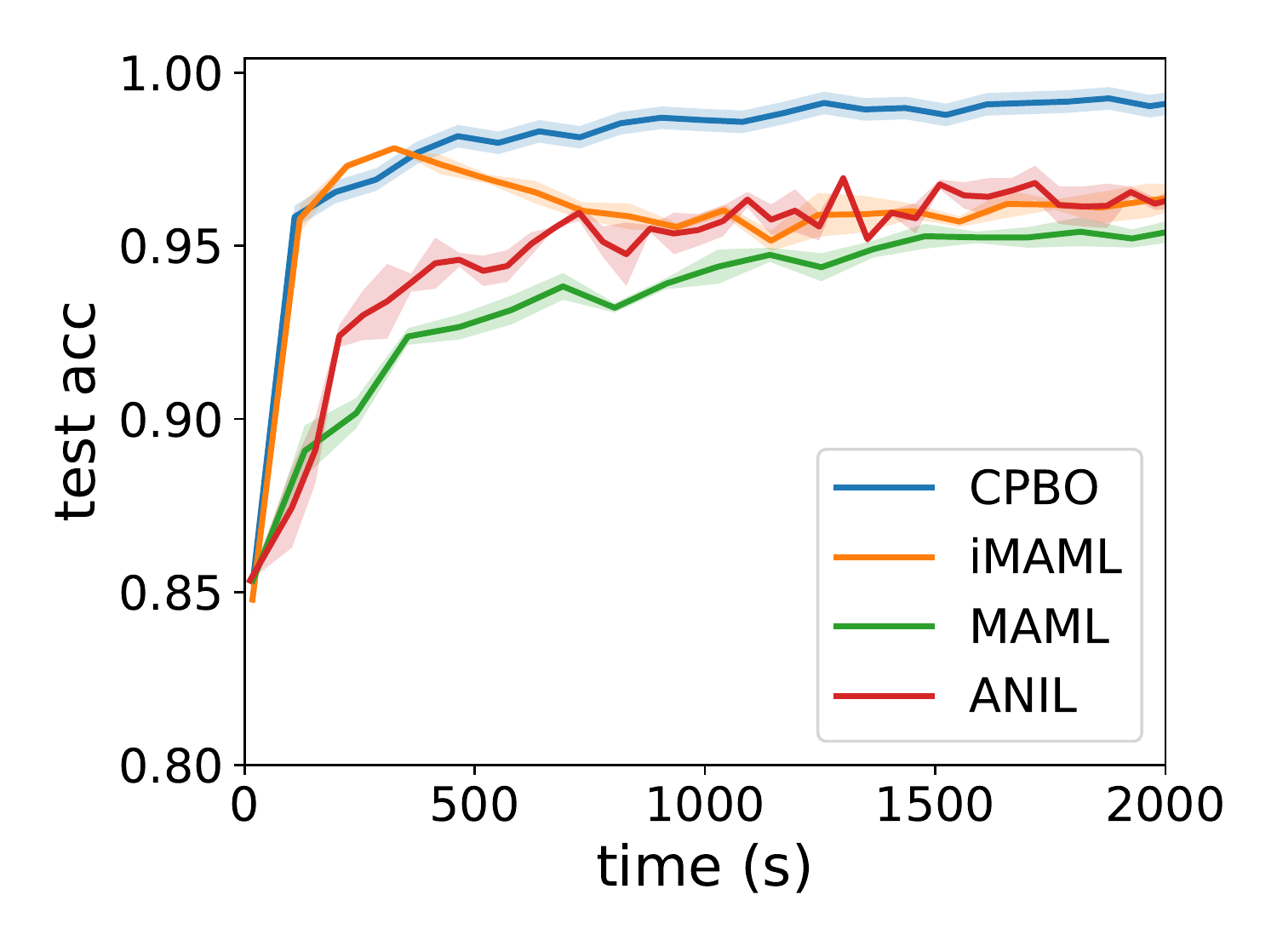}   
	\end{minipage}}
\subfigure[test loss vs time] 
{\begin{minipage}[t]{0.38\linewidth}
	\centering      
	\includegraphics[scale=0.32]{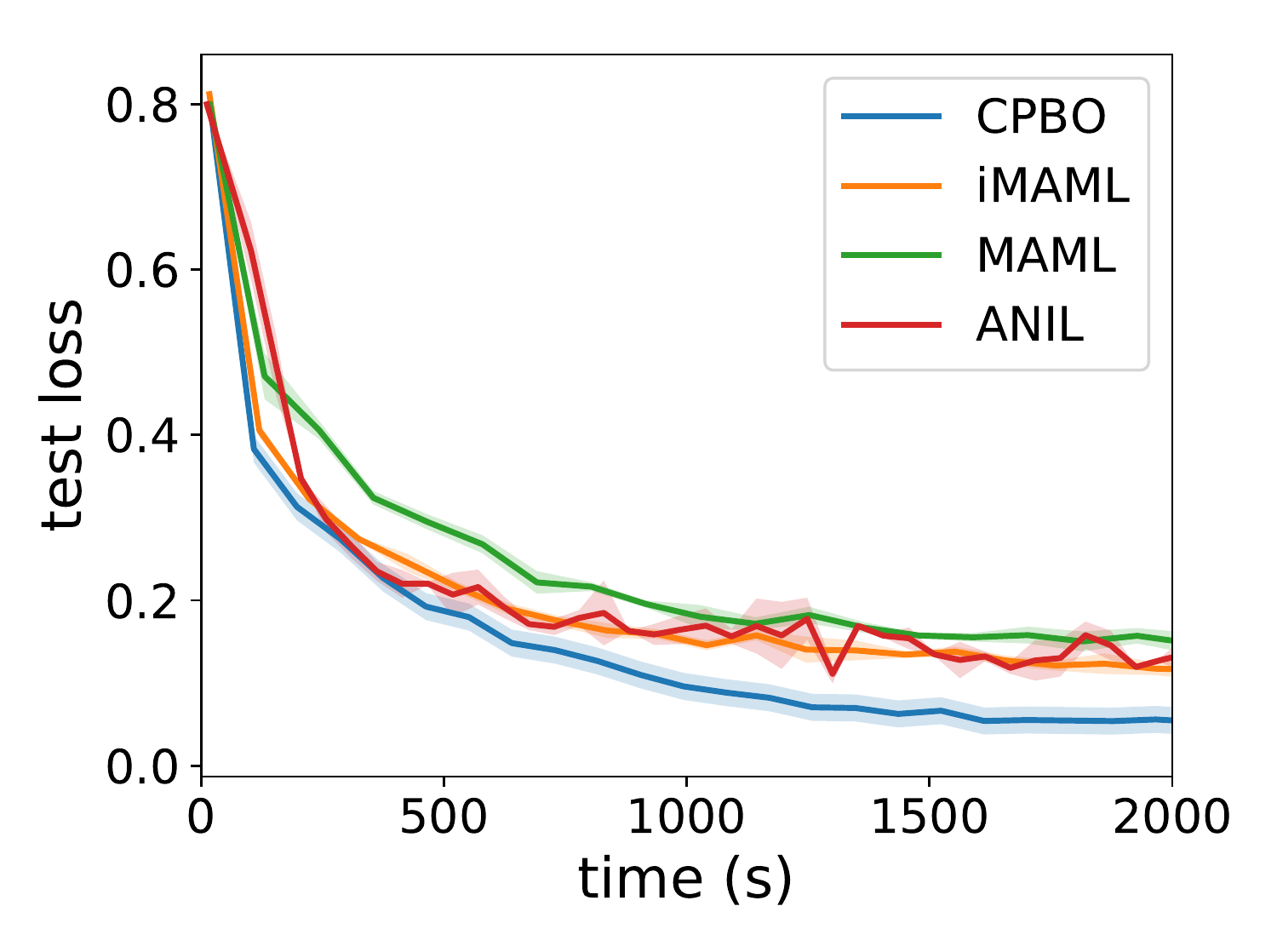}   
	\end{minipage}}
\caption{Comparison of (a) test accuracy vs time, (b) test loss vs time on Omniglot dataset.} 
\label{fig:appendix omniglot} 
\end{figure*}

\begin{figure*}[t]
\centering    
\subfigure[test accuracy vs time] 
{\begin{minipage}[t]{0.38\linewidth}
	\centering      
	\includegraphics[scale=0.32]{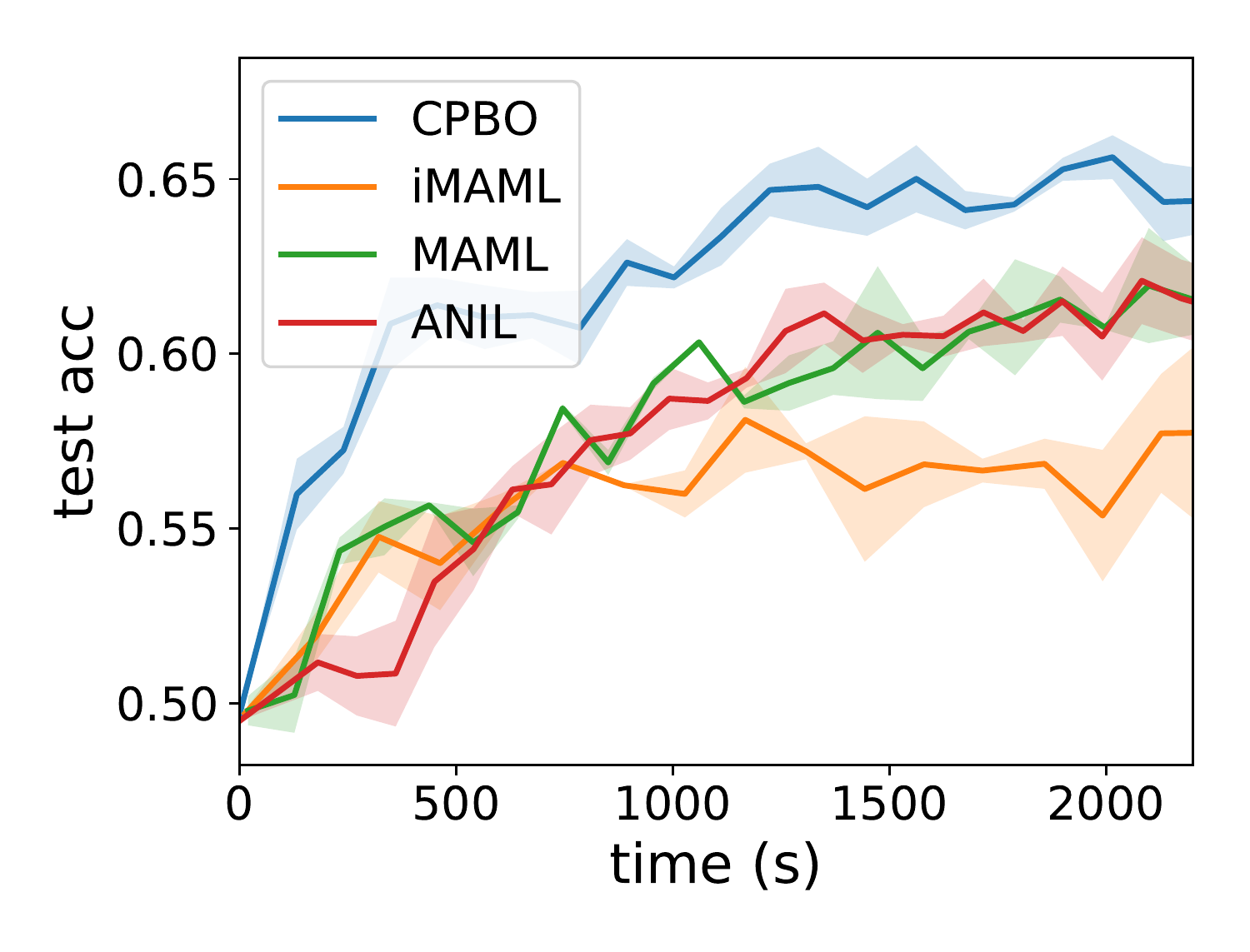}   
	\end{minipage}}
\subfigure[test loss vs time] 
{\begin{minipage}[t]{0.38\linewidth}
	\centering      
	\includegraphics[scale=0.32]{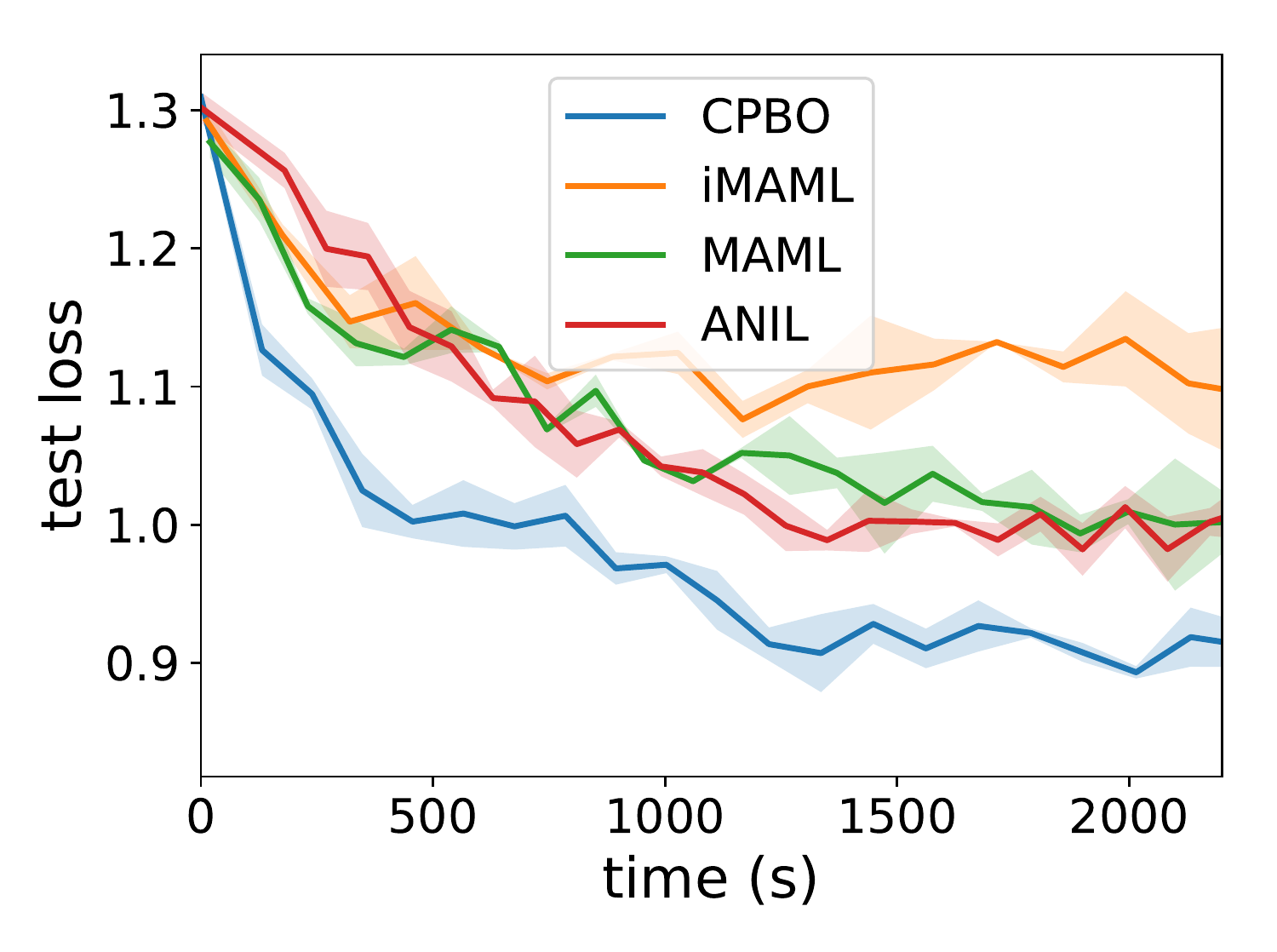}   
	\end{minipage}}
\caption{Comparison of (a) test accuracy vs time, (b) test loss vs time on CIFAR-FS dataset.} 
\label{fig:appendix cifar} 
\end{figure*}

\textcolor{black}{To evaluate the performance of the proposed CPBO, experiments are carried out on two applications: 1) hyperparameter optimization, 2) meta-learning. In hyperparameter optimization, we compare CPBO with baseline algorithms stocBio \citep{ji2021bilevel}, STABLE \citep{chen2022single}, VRBO \citep{yang2021provably}), and AID-CG \citep{grazzi2020iteration} on the regularization coefficient optimization task \citep{chen2022single} with Covertype \citep{blackard1999comparative} and IJCNN1 \citep{prokhorov2001ijcnn} datasets. We compare the performance of the proposed CPBO with all competing algorithms in terms of both the test accuracy and the test loss, which are shown in Figure \ref{fig:appendix covertype} and \ref{fig:appendix ijcnn1}. In meta-learning, we focus on the bilevel optimization problem in \citep{rajeswaran2019meta}. And we compare the proposed CPBO with baseline algorithms MAML \citep{finn2017model}, iMAML \citep{rajeswaran2019meta}, and ANIL \citep{raghu2019rapid} on Omniglot  \citep{lake2015human} and CIFAR-FS  \citep{bertinetto2018meta} datasets. And the comparison between the proposed method with the baseline algorithms are shown in Figure \ref{fig:appendix omniglot} and \ref{fig:appendix cifar}.
It is seen that the proposed CPBO can achieve relatively fast convergence rate among all competing algorithms since 1) the iteration complexity of the proposed method is not high; 2) every step in CPBO is computationally efficient.}

\subsection{Discussion}
\setcounter{definition}{0}
\renewcommand{\thedefinition}{A.\arabic{definition}}
\begin{definition}
\label{defi:A1}
$(\boldsymbol{x},\boldsymbol{y})$ is an $\epsilon$-stationary point of a differentiable function ${{\hat{L}_p}}$, if $||{\nabla _{\boldsymbol{x}}}{{\hat{L}_p}}(\boldsymbol{x},\boldsymbol{y})||^2 + ||{\nabla _{\boldsymbol{y}}}{{\hat{L}_p}}(\boldsymbol{x},\boldsymbol{y})||^2 \le \epsilon$.
\end{definition}

\setcounter{assumption}{0}
\renewcommand{\theassumption}{A.\arabic{assumption}}
\begin{assumption}\label{assumption:3}
\textbf{{{(Smoothness/Gradient Lipschitz})}} Following \citep{ji2021bilevel}, we assume that
${{\hat{L}_p}}$ has Lipschitz continuous gradients, i.e., for any $\boldsymbol{\omega}, \boldsymbol{\omega}'$, we assume that there exists $L>0$ satisfying that,
\begin{equation}
    \begin{array}{l}
||{\nabla}{{\hat{L}_p}}(\boldsymbol{\omega})  -  {\nabla}{{\hat{L}_p}}(\boldsymbol{\omega}')|| \le L||\boldsymbol{\omega}-\boldsymbol{\omega}'||.
\end{array}
\end{equation}
\end{assumption}

\begin{assumption}\label{assumption:4}
{\textbf{(Boundedness)}} Following \citep{qian2019robust}, we assume that variables have boundedness, i.e.,  $||\boldsymbol{x}||^2  \le  \beta_1$, $||\boldsymbol{y}||^2  \le  \beta_2$.
\end{assumption}

\begin{theorem}
\label{theorem: iteration complexity centralized}
(\textbf{Iteration Complexity}) Under Assumption \ref{assumption:3}, \ref{assumption:4}, and setting the step-sizes as ${\eta _{\boldsymbol{x}}} < \frac{2}{{{L}}},{\eta _{\boldsymbol{y}}} < \frac{2}{{{L}}}$, the iteration complexity (also the gradient complexity) of the proposed algorithm to obtain    $\epsilon$-stationary point is bounded by ${\cal O}(\frac{1}{\epsilon})$.
\end{theorem}

\emph{\textbf{Proof of Theorem \ref{theorem: iteration complexity centralized}:}}

According to Assumption \ref{assumption:3} and Eq. (\ref{eq:96_18}), when $t\ge T_1$, we have,
\begin{equation}
\label{eq:926_156}
\begin{array}{l}
{{\hat{L}_p}}({{\boldsymbol{x}}^{t + 1}},{{\boldsymbol{y}}^t}) \le {{\hat{L}_p}}({{\boldsymbol{x}}^t},{{\boldsymbol{y}}^t}) + \left\langle {{\nabla _{{\boldsymbol{x}}}}{{\hat{L}_p}}({{\boldsymbol{x}}^t},{{\boldsymbol{y}}^t}),{{\boldsymbol{x}}^{t + 1}} - {{\boldsymbol{x}}^t}} \right\rangle  + \frac{{{L}}}{2}||{{\boldsymbol{x}}^{t + 1}} - {{\boldsymbol{x}}^t}||^2 \vspace{1mm}\\
\qquad  \qquad \quad \; \; \, \le {{\hat{L}_p}}({{\boldsymbol{x}}^t},{{\boldsymbol{y}}^t}) - {\eta _{\boldsymbol{x}}}||{\nabla _{{\boldsymbol{x}}}}{{\hat{L}_p}}({{\boldsymbol{x}}^t},{{\boldsymbol{y}}^t})|{|^2} + \frac{{{L}{\eta _{\boldsymbol{x}}}^2}}{2}||{\nabla _{{\boldsymbol{x}}}}{{\hat{L}_p}}({{\boldsymbol{x}}^t},{{\boldsymbol{y}}^t})|{|^2}.
\end{array}
\end{equation}

Similarly, according to  Assumption \ref{assumption:3} and Eq. (\ref{eq:96_19}), we have,
\begin{equation}
\label{eq:926_157}
\begin{array}{l}
{{\hat{L}_p}}({{\boldsymbol{x}}^{t + 1}},{{\boldsymbol{y}}^{t + 1}}) \le {{\hat{L}_p}}({{\boldsymbol{x}}^{t + 1}},{{\boldsymbol{y}}^t}) + \left\langle {{\nabla _{{\boldsymbol{y}}}}{{\hat{L}_p}}({{\boldsymbol{x}}^{t + 1}},{{\boldsymbol{y}}^t}),{{\boldsymbol{y}}^{t + 1}} - {{\boldsymbol{y}}^t}} \right\rangle  + \frac{{{L}}}{2}||{{\boldsymbol{y}}^{t + 1}} - {{\boldsymbol{y}}^t}||^2 \vspace{1mm}\\
\qquad  \qquad \quad \quad \; \; \, \le {{\hat{L}_p}}({{\boldsymbol{x}}^{t + 1}},{{\boldsymbol{y}}^t}) - {\eta _{\boldsymbol{y}}}||{\nabla _{{\boldsymbol{y}}}}{{\hat{L}_p}}({{\boldsymbol{x}}^{t + 1}},{{\boldsymbol{y}}^t})|{|^2} + \frac{{{L}{\eta _{\boldsymbol{y}}}^2}}{2}||{\nabla _{{\boldsymbol{y}}}}{{\hat{L}_p}}({{\boldsymbol{x}}^{t + 1}},{{\boldsymbol{y}}^t})|{|^2}.
\end{array}
\end{equation}

Combining Eq. (\ref{eq:926_156}) with Eq. (\ref{eq:926_157}), we have,
\begin{equation}
\label{eq:926_158}
({\eta _{\boldsymbol{x}}} - \frac{{{L}{\eta _{\boldsymbol{x}}}^2}}{2})||{\nabla _{{\boldsymbol{x}}}}{{\hat{L}_p}}({{\boldsymbol{x}}^t},{{\boldsymbol{y}}^t})|{|^2} + ({\eta _{\boldsymbol{y}}} - \frac{{{L}{\eta _{\boldsymbol{y}}}^2}}{2})||{\nabla _{{\boldsymbol{y}}}}{{\hat{L}_p}}({{\boldsymbol{x}}^{t + 1}},{{\boldsymbol{y}}^t})|{|^2} \le {{\hat{L}_p}}({{\boldsymbol{x}}^t},{{\boldsymbol{y}}^t}) - {{\hat{L}_p}}({{\boldsymbol{x}}^{t + 1}},{{\boldsymbol{y}}^{t + 1}}).
\end{equation}

According to the setting of ${\eta _{\boldsymbol{x}}}$, ${\eta _{\boldsymbol{y}}}$, we have that ${\eta _{\boldsymbol{x}}} - \frac{{{L}{\eta _{\boldsymbol{x}}}^2}}{2}>0$, ${\eta _{\boldsymbol{y}}} - \frac{{{L}{\eta _{\boldsymbol{y}}}^2}}{2}>0$. And we set constant $d = \min \{ {\eta _{\boldsymbol{x}}} - \frac{{{L}{\eta _{\boldsymbol{x}}}^2}}{2},{\eta _{\boldsymbol{y}}} - \frac{{{L}{\eta _{\boldsymbol{y}}}^2}}{2}\}$, thus we can obtain that,
\begin{equation}
\label{eq:926_159}
||{\nabla _{{\boldsymbol{x}}}}{{\hat{L}_p}}({{\boldsymbol{x}}^t},{{\boldsymbol{y}}^t})|{|^2} + ||{\nabla _{{\boldsymbol{y}}}}{{\hat{L}_p}}({{\boldsymbol{x}}^{t + 1}},{{\boldsymbol{y}}^t})|{|^2} \le \frac{{{{\hat{L}_p}}({{\boldsymbol{x}}^t},{{\boldsymbol{y}}^t}) - {{\hat{L}_p}}({{\boldsymbol{x}}^{t + 1}},{{\boldsymbol{y}}^{t + 1}})}}{d}.
\end{equation}

Summing both sides of Eq. (\ref{eq:926_159}) for $t = \{ {T_1}, \cdots ,T - 1\}$, we obtain that,
\begin{equation}
\label{eq:926_160}
\frac{1}{{T - {T_1}}}\sum\limits_{t = {T_1}}^{T - 1} {(||{\nabla _{{\boldsymbol{x}}}}{{\hat{L}_p}}({{\boldsymbol{x}}^t},{{\boldsymbol{y}}^t})|{|^2} + ||{\nabla _{{\boldsymbol{y}}}}{{\hat{L}_p}}({{\boldsymbol{x}}^{t + 1}},{{\boldsymbol{y}}^t})|{|^2})}  \le \frac{{{{\hat{L}_p}}({{\boldsymbol{x}}^{{T_1}}},{{\boldsymbol{y}}^{{T_1}}}) - {{\hat{L}_p}^{*}}}}{{(T - {T_1})d}},
\end{equation}
where ${{\hat{L}_p}^{*}}=\mathop {\min } {\hat{L}_p}( {\boldsymbol{x}} , \! {\boldsymbol{y}}  )$. Combining Eq. (\ref{eq:926_160}) with Definition \ref{defi:A1}, we have that the number of iterations required by Algorithm \ref{algorithm:centralized} to return an $\epsilon$-stationary point is bounded by
\begin{equation}
\label{eq:926_161}
\mathcal{O}(\frac{{{{\hat{L}_p}}({{\boldsymbol{x}}^{{T_1}}},{{\boldsymbol{y}}^{{T_1}}}) - {{\hat{L}_p}^{*}}}}{d}\frac{1}{\epsilon} + {T_1}).
\end{equation}

\section{Proof of Theorem \ref{theorem 2}}
\label{appendix: theorem 2}
In this section, we provide complete proofs for Theorem \ref{theorem 2}. Firstly, we make some definitions about our problem.

\setcounter{definition}{0}
\renewcommand{\thedefinition}{B.\arabic{definition}}
\begin{definition}
\label{definition:A1}
Following \citep{xu2020unified}, the \textit{stationarity} \textit{gap} at $t^{{th}}$ iteration is defined as:
{\begin{equation}
\label{eq:A61}
\nabla G^t = \left[ \begin{array}{l}
\{ {\nabla _{{{\boldsymbol{x}}_i}}}{L_p}(\{ {\boldsymbol{x}_i^t}\} , \! \{ {\boldsymbol{y}_i^t}\} ,\! \boldsymbol{v}^t,\! \boldsymbol{z}^t,\! \{ {\lambda _l^t}\} ,\!\{ {\boldsymbol{\theta }_i^t}\} )\} \vspace{1ex}\\
\{ {\nabla _{{{\boldsymbol{y}}_i}}}{L_p}(\{ {\boldsymbol{x}_i^t}\} , \! \{ {\boldsymbol{y}_i^t}\} ,\! \boldsymbol{v}^t,\! \boldsymbol{z}^t,\! \{ {\lambda _l^t}\} ,\!\{ {\boldsymbol{\theta }_i^t}\} )\} \vspace{1ex}\\
\,{\nabla _{{{\boldsymbol{v}}}}}{L_p}(\{ {\boldsymbol{x}_i^t}\} , \! \{ {\boldsymbol{y}_i^t}\} ,\! \boldsymbol{v}^t,\! \boldsymbol{z}^t,\! \{ {\lambda _l^t}\} ,\!\{ {\boldsymbol{\theta }_i^t}\}) \vspace{1ex}\\
\,{\nabla _{{{\boldsymbol{z}}}}}{L_p}(\{ {\boldsymbol{x}_i^t}\} , \! \{ {\boldsymbol{y}_i^t}\} ,\! \boldsymbol{v}^t,\! \boldsymbol{z}^t,\! \{ {\lambda _l^t}\} ,\!\{ {\boldsymbol{\theta }_i^t}\})\vspace{1ex}\\
\{ {\nabla _{\lambda _l}}{L_p}(\{ {\boldsymbol{x}_i^t}\} , \! \{ {\boldsymbol{y}_i^t}\} ,\! \boldsymbol{v}^t,\! \boldsymbol{z}^t,\! \{ {\lambda _l^t}\} ,\!\{ {\boldsymbol{\theta }_i^t}\} )\} \vspace{1ex}\\
\{ {\nabla _{\boldsymbol{\theta }_i}}{L_p}(\{ {\boldsymbol{x}_i^t}\} , \! \{ {\boldsymbol{y}_i^t}\} ,\! \boldsymbol{v}^t,\! \boldsymbol{z}^t,\! \{ {\lambda _l^t}\} ,\!\{ {\boldsymbol{\theta }_i^t}\} )\}
\end{array} \right].
\end{equation}}

And we also define:
{\begin{equation}
\label{eq:A62}
\begin{array}{l}
{(\nabla G^t)_{{{\boldsymbol{x}}_i}}} ={\nabla _{{{\boldsymbol{x}}_i}}}{L_p}(\{ {\boldsymbol{x}_i^t}\} , \! \{ {\boldsymbol{y}_i^t}\} ,\! \boldsymbol{v}^t,\! \boldsymbol{z}^t,\! \{ {\lambda _l^t}\} ,\!\{ {\boldsymbol{\theta }_i^t}\} ), \vspace{1ex}\\
{(\nabla G^t)_{{{\boldsymbol{y}}_i}}} ={\nabla _{{{\boldsymbol{y}}_i}}}{L_p}(\{ {\boldsymbol{x}_i^t}\} , \! \{ {\boldsymbol{y}_i^t}\} ,\! \boldsymbol{v}^t,\! \boldsymbol{z}^t,\! \{ {\lambda _l^t}\} ,\!\{ {\boldsymbol{\theta }_i^t}\} ), \vspace{1ex}\\
{(\nabla G^t)_{{{\boldsymbol{v}}}}} ={\nabla _{{{\boldsymbol{v}}}}}{L_p}(\{ {\boldsymbol{x}_i^t}\} , \! \{ {\boldsymbol{y}_i^t}\} ,\! \boldsymbol{v}^t,\! \boldsymbol{z}^t,\! \{ {\lambda _l^t}\} ,\!\{ {\boldsymbol{\theta }_i^t}\} ), \vspace{1ex}\\
{(\nabla G^t)_{{{\boldsymbol{z}}}}} ={\nabla _{{{\boldsymbol{z}}}}}{L_p}(\{ {\boldsymbol{x}_i^t}\} , \! \{ {\boldsymbol{y}_i^t}\} ,\! \boldsymbol{v}^t,\! \boldsymbol{z}^t,\! \{ {\lambda _l^t}\} ,\!\{ {\boldsymbol{\theta }_i^t}\} ), \vspace{1ex}\\
{(\nabla G^t)_{{\lambda_l}}} ={\nabla _{\lambda_l}}{L_p}(\{ {\boldsymbol{x}_i^t}\} , \! \{ {\boldsymbol{y}_i^t}\} ,\! \boldsymbol{v}^t,\! \boldsymbol{z}^t,\! \{ {\lambda _l^t}\} ,\!\{ {\boldsymbol{\theta }_i^t}\} ), \vspace{1ex}\\
{(\nabla G^t)_{{\boldsymbol{\theta }_i}}} ={\nabla _{\boldsymbol{\theta }_i}}{L_p}(\{ {\boldsymbol{x}_i^t}\} , \! \{ {\boldsymbol{y}_i^t}\} ,\! \boldsymbol{v}^t,\! \boldsymbol{z}^t,\! \{ {\lambda _l^t}\} ,\!\{ {\boldsymbol{\theta }_i^t}\} ).
\end{array}
\end{equation}}

It follows that,
{\begin{equation}
\label{eq:A63}
||\nabla G^t|{|^2} \! =\! \sum\limits_{i = 1}^N ({||{{(\nabla G^t)}_{{{\boldsymbol{x}}_i}}}|{|^2}}   + {||{{(\nabla G^t)}_{{{\boldsymbol{y}}_i}}}|{|^2}} + {||{{(\nabla G^t)}_{{\boldsymbol{\theta }_i}}}|{|^2}})  + ||{(\nabla G^t)_{\boldsymbol{v}}}|{|^2}  + ||{(\nabla G^t)_{\boldsymbol{z}}}|{|^2}  + \sum\limits_{l = 1}^{|{\boldsymbol{\mathcal{P}}^{t}}|} {||{{(\nabla G^t)}_{{\lambda _l}}}|{|^2}}. 
\end{equation}}
\end{definition}

\begin{definition}
\label{definition:A2}
At $t^{{th}}$ iteration, the \textit{stationarity} \textit{gap} \textit{w.r.t} ${\widetilde{L}_p}(\{ {\boldsymbol{x}_i}\} ,\!\{ {\boldsymbol{y}_i}\} ,\! \boldsymbol{v},\boldsymbol{z},\!\{ {\lambda _l}\} ,\!\{ {\boldsymbol{\theta }_i}\} )$ is defined as:
\begin{equation}
\label{eq:A64}
\nabla \widetilde{G}^t = \left[ \begin{array}{l}
\{ {\nabla _{{{\boldsymbol{x}}_i}}}{\widetilde{L}_p}(\{ {\boldsymbol{x}_i^t}\} , \! \{ {\boldsymbol{y}_i^t}\} ,\! \boldsymbol{v}^t,\! \boldsymbol{z}^t,\! \{ {\lambda _l^t}\} ,\!\{ {\boldsymbol{\theta }_i^t}\} )\} \vspace{1ex}\\
\{ {\nabla _{{{\boldsymbol{y}}_i}}}{\widetilde{L}_p}(\{ {\boldsymbol{x}_i^t}\} , \! \{ {\boldsymbol{y}_i^t}\} ,\! \boldsymbol{v}^t,\! \boldsymbol{z}^t,\! \{ {\lambda _l^t}\} ,\!\{ {\boldsymbol{\theta }_i^t}\} )\} \vspace{1ex}\\
\,{\nabla _{{{\boldsymbol{v}}}}}{\widetilde{L}_p}(\{ {\boldsymbol{x}_i^t}\} , \! \{ {\boldsymbol{y}_i^t}\} ,\! \boldsymbol{v}^t,\! \boldsymbol{z}^t,\! \{ {\lambda _l^t}\} ,\!\{ {\boldsymbol{\theta }_i^t}\}) \vspace{1ex}\\
\,{\nabla _{{{\boldsymbol{z}}}}}{\widetilde{L}_p}(\{ {\boldsymbol{x}_i^t}\} , \! \{ {\boldsymbol{y}_i^t}\} ,\! \boldsymbol{v}^t,\! \boldsymbol{z}^t,\! \{ {\lambda _l^t}\} ,\!\{ {\boldsymbol{\theta }_i^t}\})\vspace{1ex}\\
\{ {\nabla _{\lambda _l}}{\widetilde{L}_p}(\{ {\boldsymbol{x}_i^t}\} , \! \{ {\boldsymbol{y}_i^t}\} ,\! \boldsymbol{v}^t,\! \boldsymbol{z}^t,\! \{ {\lambda _l^t}\} ,\!\{ {\boldsymbol{\theta }_i^t}\} )\} \vspace{1ex}\\
\{ {\nabla _{\boldsymbol{\theta }_i}}{\widetilde{L}_p}(\{ {\boldsymbol{x}_i^t}\} , \! \{ {\boldsymbol{y}_i^t}\} ,\! \boldsymbol{v}^t,\! \boldsymbol{z}^t,\! \{ {\lambda _l^t}\} ,\!\{ {\boldsymbol{\theta }_i^t}\} )\}
\end{array} \right].
\end{equation}

We further define:
{\begin{equation}
\label{eq:A65}
\begin{array}{l}
{(\nabla \widetilde{G}^t)_{{{\boldsymbol{x}}_i}}} ={\nabla _{{{\boldsymbol{x}}_i}}}{\widetilde{L}_p}(\{ {\boldsymbol{x}_i^t}\} , \! \{ {\boldsymbol{y}_i^t}\} ,\! \boldsymbol{v}^t,\! \boldsymbol{z}^t,\! \{ {\lambda _l^t}\} ,\!\{ {\boldsymbol{\theta }_i^t}\} ), \vspace{1ex}\\
{(\nabla \widetilde{G}^t)_{{{\boldsymbol{y}}_i}}} ={\nabla _{{{\boldsymbol{y}}_i}}}{\widetilde{L}_p}(\{ {\boldsymbol{x}_i^t}\} , \! \{ {\boldsymbol{y}_i^t}\} ,\! \boldsymbol{v}^t,\! \boldsymbol{z}^t,\! \{ {\lambda _l^t}\} ,\!\{ {\boldsymbol{\theta }_i^t}\} ), \vspace{1ex}\\
{(\nabla \widetilde{G}^t)_{{{\boldsymbol{v}}}}} ={\nabla _{{{\boldsymbol{v}}}}}{\widetilde{L}_p}(\{ {\boldsymbol{x}_i^t}\} , \! \{ {\boldsymbol{y}_i^t}\} ,\! \boldsymbol{v}^t,\! \boldsymbol{z}^t,\! \{ {\lambda _l^t}\} ,\!\{ {\boldsymbol{\theta }_i^t}\} ), \vspace{1ex}\\
{(\nabla \widetilde{G}^t)_{{{\boldsymbol{z}}}}} ={\nabla _{{{\boldsymbol{z}}}}}{\widetilde{L}_p}(\{ {\boldsymbol{x}_i^t}\} , \! \{ {\boldsymbol{y}_i^t}\} ,\! \boldsymbol{v}^t,\! \boldsymbol{z}^t,\! \{ {\lambda _l^t}\} ,\!\{ {\boldsymbol{\theta }_i^t}\} ), \vspace{1ex}\\
{(\nabla \widetilde{G}^t)_{{\lambda_l}}} ={\nabla _{\lambda_l}}{\widetilde{L}_p}(\{ {\boldsymbol{x}_i^t}\} , \! \{ {\boldsymbol{y}_i^t}\} ,\! \boldsymbol{v}^t,\! \boldsymbol{z}^t,\! \{ {\lambda _l^t}\} ,\!\{ {\boldsymbol{\theta }_i^t}\} ), \vspace{1ex}\\
{(\nabla \widetilde{G}^t)_{{\boldsymbol{\theta }_i}}} ={\nabla _{\boldsymbol{\theta }_i}}{\widetilde{L}_p}(\{ {\boldsymbol{x}_i^t}\} , \! \{ {\boldsymbol{y}_i^t}\} ,\! \boldsymbol{v}^t,\! \boldsymbol{z}^t,\! \{ {\lambda _l^t}\} ,\!\{ {\boldsymbol{\theta }_i^t}\} ).
\end{array}
\end{equation}}

It follows that,
{\begin{equation}
\label{eq:A66}
||\nabla \widetilde{G}^t|{|^2} \! =\! \sum\limits_{i = 1}^N ({||{{(\nabla \widetilde{G}^t)}_{{{\boldsymbol{x}}_i}}}|{|^2}}   + {||{{(\nabla \widetilde{G}^t)}_{{{\boldsymbol{y}}_i}}}|{|^2}} + {||{{(\nabla \widetilde{G}^t)}_{{\boldsymbol{\theta }_i}}}|{|^2}})  + ||{(\nabla \widetilde{G}^t)_{\boldsymbol{v}}}|{|^2}  + ||{(\nabla \widetilde{G}^t)_{\boldsymbol{z}}}|{|^2}  + \sum\limits_{l = 1}^{|{\boldsymbol{\mathcal{P}}^{t}}|} {||{{(\nabla \widetilde{G}^t)}_{{\lambda _l}}}|{|^2}}.  
\end{equation}}
\end{definition}

\begin{definition}
\label{definition:A3}
In the proposed asynchronous algorithm, for the $i^{{th}}$ worker in $t^{th}$ iteration, the last iteration where this worker was active is defined as $\hat{{t}}_i $. And the next iteration this worker  will be active is defined as $\overline{{t}_i}$. For the iteration index set which $i^{{th}}$ worker is active during $T_1+T+\tau$ iteration, it is defined as $\mathcal{V}_i(T)$. And the $j^{{th}}$ element in $\mathcal{V}_i(T)$ is defined as $\hat{v}_i(j)$.
\end{definition}

\vspace{5mm}

Then, we provide some useful lemmas used for proving the main convergence results in Theorem \ref{theorem 2}.

\begin{lemma}
Let sequences ${\eta _{\boldsymbol{x}}^t} =  {\eta _{\boldsymbol{y}}^t} = {\eta _{\boldsymbol{v}}^t} = {\eta _{\boldsymbol{z}}^t}  = \frac{2}{{L + {{\eta _{\lambda}}}|{{\boldsymbol{\mathcal{P}}^{t}}}|{L^2} + {{\eta _{\boldsymbol{\theta }}}}N{L^2} + 8(\frac{{|{{\boldsymbol{\mathcal{P}}^{t}}}|\gamma {L^2}}}{{{{\eta _{\lambda}}}({c_1^t})^2}} + \frac{{N\gamma {L^2}}}{{{{\eta _{\boldsymbol{\theta }}}}({c_2^t})^2}})}}$, suppose Assumption 1 and 2 hold, we can obtain that,
\begin{equation}
\label{eq:lemma1}
\begin{array}{l}
{L_p}(\{ \boldsymbol{x}_i^{t + 1}\} ,\!\{ \boldsymbol{y}_i^{t+1 }\} ,\!{\boldsymbol{v}^{t+1}  },\!{\boldsymbol{z}^{t+1}  },\!\{ \lambda _l^{t}  \} ,\!\{ \boldsymbol{\theta }_i^{t}  \} ) - {L_p}(\{ \boldsymbol{x}_i^{t}\} ,\!\{ \boldsymbol{y}_i^{t}\} ,\!{\boldsymbol{v}^{t}  },\!{\boldsymbol{z}^{t}  },\!\{ \lambda _l^{t}  \} ,\!\{ \boldsymbol{\theta }_i^{t}  \} ) \vspace{0.5ex}\\

  \le \sum\limits_{i  = 1}^N \!{(\frac{{L  + L^2 + 1}}{2} \!-\! \frac{1}{{{\eta _{\boldsymbol{x}}^t}}})}||{{\boldsymbol{x}}_i^{t+1}} \!-\! {{\boldsymbol{x}}_i^t}|{|^2} +
  \sum\limits_{i  = 1}^N\! {(\frac{{L  +  1}}{2} \!-\! \frac{1}{{{\eta _{\boldsymbol{y}}^t}}})}||{{\boldsymbol{y}}_i^{t+1}} \!-\! {{\boldsymbol{y}}_i^t}|{|^2}+ 3NL^2\tau{ k_1 } \sum\limits_{l = 1}^{|{\boldsymbol{\mathcal{P}}^{t}}|} {||{\lambda _l^{t+1}} \!-\! {\lambda _l^{t}}|{|^2}} \vspace{1ex} \\
  
  +(\frac{L+6NL^2\tau{ k_1 }}{2}\!-\!\frac{1}{{{\eta _{\boldsymbol{v}}^t}}})||{{\boldsymbol{v}}^{t+1}} \!-\! {{\boldsymbol{v}}^t}|{|^2}+ (\frac{L+6NL^2\tau{ k_1 }}{2}\!-\!\frac{1}{{{\eta _{\boldsymbol{z}}^t}}})||{{\boldsymbol{z}}^{t+1}} \!-\! {{\boldsymbol{z}}^t}|{|^2} .
  
\end{array}
\end{equation}
\end{lemma}

\emph{\textbf{Proof of Lemma 1:}}

Utilizing the Lipschitz properties in Assumption 1, we can obtain that, 
\begin{equation}
\label{eq:A4-1}
\begin{array}{l}
{L_p}{\rm{(\{ }}{{\boldsymbol{x}}_1^{t+1}},{{\boldsymbol{x}}_2^t}, \!\cdots\! ,{{\boldsymbol{x}}_N^t}\} ,\!\{ \boldsymbol{y}_i^{t }\} ,\!{\boldsymbol{v}^{t}  },{\boldsymbol{z}^{t}  },\!\{ \lambda _l^{t}  \} ,\!\{ \boldsymbol{\theta }_i^{t}  \} {\rm{)}} \!-\! {L_p}{\rm{(\{ }}{{\boldsymbol{x}}_i^t}\} ,\!\{ \boldsymbol{y}_i^{t }\} ,\!{\boldsymbol{v}^{t}  },{\boldsymbol{z}^{t}  },\!\{ \lambda _l^{t}  \} ,\!\{ \boldsymbol{\theta }_i^{t}  \} {\rm{)}}\vspace{1.5ex}\\
 \le \left\langle {{\nabla _{{{\boldsymbol{x}}_1}}}{L_p}{\rm{(\{ }}{{\boldsymbol{x}}_i^t}\},\!\{ \boldsymbol{y}_i^{t }\} ,\!{\boldsymbol{v}^{t}  },{\boldsymbol{z}^{t}  },\!\{ \lambda _l^{t}  \} ,\!\{ \boldsymbol{\theta }_i^{t}  \} ), {{\boldsymbol{x}}_1^{t+1}} \!-\! {{\boldsymbol{x}}_1^t}} \right\rangle  \! + \! \frac{L}{2}||{{\boldsymbol{x}}_1^{t+1}} \!-\! {{\boldsymbol{x}}_1^t}|{|^2}, \vspace{3ex}\\
 
{L_p}{\rm{(\{ }}{{\boldsymbol{x}}_1^{t+1}},{{\boldsymbol{x}}_2^{t+1}}, \!\cdots\! ,{{\boldsymbol{x}}_N^t}\},\!\{ \boldsymbol{y}_i^{t }\} ,\!{\boldsymbol{v}^{t}  },\!{\boldsymbol{z}^{t}  },\!\{ \lambda _l^{t}  \} ,\!\{ \boldsymbol{\theta }_i^{t}  \})
\!-\! {L_p}{\rm{(\{ }}{{\boldsymbol{x}}_1^{t+1}},{{\boldsymbol{x}}_2^t},\! \cdots\! ,{{\boldsymbol{x}}_N^t}\},\!\{ \boldsymbol{y}_i^{t }\} ,\!{\boldsymbol{v}^{t}  },\!{\boldsymbol{z}^{t}  },\!\{ \lambda _l^{t}  \} ,\!\{ \boldsymbol{\theta }_i^{t}  \})  \vspace{1.5ex}\\
 \le \left\langle {{\nabla _{{{\boldsymbol{x}}_2}}}{L_p}{\rm{(\{ }}{{\boldsymbol{x}}_i^t}\},\!\{ \boldsymbol{y}_i^{t }\} ,\!{\boldsymbol{v}^{t}  },\!{\boldsymbol{z}^{t}  },\!\{ \lambda _l^{t}  \} ,\!\{ \boldsymbol{\theta }_i^{t}  \} ),{{\boldsymbol{x}}_2^{t+1}} \!-\! {{\boldsymbol{x}}_2^t}} \right\rangle  \! + \! \frac{L}{2}||{{\boldsymbol{x}}_2^{t+1}} \!-\! {{\boldsymbol{x}}_2^t}|{|^2}, \vspace{2ex}\\
\qquad \qquad \qquad  \qquad \qquad \qquad \qquad \qquad \qquad \vdots  \vspace{2ex}\\
{L_p}{\rm{(\{ }}{{\boldsymbol{x}}_i^{t+1}}\},\!\{ \boldsymbol{y}_i^{t }\} ,\!{\boldsymbol{v}^{t}  },\!{\boldsymbol{z}^{t}  },\!\{ \lambda _l^{t}  \} ,\!\{ \boldsymbol{\theta }_i^{t}  \} {\rm{)}} \!-\! {L_p}{\rm{(\{ }}{{\boldsymbol{x}}_1^{t+1}}, \! \cdots \! ,{{\boldsymbol{x}}_{N \!-\! 1}^{t+1}},{{\boldsymbol{x}}_N^t}\} ,\!\{ \boldsymbol{y}_i^{t }\} ,\!{\boldsymbol{v}^{t}  },\!{\boldsymbol{z}^{t}  },\!\{ \lambda _l^{t}  \} ,\!\{ \boldsymbol{\theta }_i^{t}  \} {\rm{)}} \vspace{1.5ex}\\
 \le \left\langle {{\nabla _{{{\boldsymbol{x}}_N}}}{L_p}{\rm{(\{ }}{{\boldsymbol{x}}_i^t}\} ,\!\{ \boldsymbol{y}_i^{t }\} ,\!{\boldsymbol{v}^{t}  },\!{\boldsymbol{z}^{t}  },\!\{ \lambda _l^{t}  \} ,\!\{ \boldsymbol{\theta }_i^{t}  \} ),{{\boldsymbol{x}}_N^{t+1}} \!-\! {{\boldsymbol{x}}_N^t}} \right\rangle  \! + \! \frac{L}{2}||{{\boldsymbol{x}}_N^{t+1}} \!-\! {{\boldsymbol{x}}_N^t}|{|^2}.
\end{array}
\end{equation}

Summing up the above inequalities in Eq. (\ref{eq:A4-1}), we can obtain that,
\begin{equation}
\label{eq:A4}
\begin{array}{l}
{L_p}(\{ \boldsymbol{x}_i^{t + 1}\} ,\!\{ \boldsymbol{y}_i^{t }\} ,\!{\boldsymbol{v}^{t}  },\!{\boldsymbol{z}^{t}  },\!\{ \lambda _l^{t}  \} ,\!\{ \boldsymbol{\theta }_i^{t}  \} ) \!-\! {L_p}(\{ \boldsymbol{x}_i^{t}\} ,\!\{ \boldsymbol{y}_i^{t }\} ,\!{\boldsymbol{v}^{t}  },\!{\boldsymbol{z}^{t}  },\!\{ \lambda _l^{t}  \} ,\!\{ \boldsymbol{\theta }_i^{t}  \} )\\

 \le \sum\limits_{i  = 1}^N {\left(  \left\langle {{\nabla _{{{\boldsymbol{x}}_i}}}{L_p}{\rm{(\{ }}{{\boldsymbol{x}}_i^t}\} ,\!\{ \boldsymbol{y}_i^{t }\} ,\!{\boldsymbol{v}^{t}  },\!{\boldsymbol{z}^{t}  },\!\{ \lambda _l^{t}  \} ,\!\{ \boldsymbol{\theta }_i^{t}  \} ), {{\boldsymbol{x}}_i^{t+1}} \!-\! {{\boldsymbol{x}}_i^t}} \right\rangle  \! + \! \frac{L}{2}||{{\boldsymbol{x}}_i^{t+1}} \!-\! {{\boldsymbol{x}}_i^t}|{|^2} \right)}.
\end{array}
\end{equation}

Combining ${\nabla _{{{\boldsymbol{x}}_i}}}{L_p}(\{ {\boldsymbol{x}_i^{\hat{t}_i}}\} ,\{ {\boldsymbol{y}_i^{\hat{t}_i}}\} ,\boldsymbol{v}^{\hat{t}_i}, \boldsymbol{z}^{\hat{t}_i}, \{ {\lambda _l^{\hat{t}_i}}\} ,\{ {\boldsymbol{\theta }_i^{\hat{t}_i}}\} ) = {\nabla _{{{\boldsymbol{x}}_i}}}{\widetilde{L} _p}(\{ {\boldsymbol{x}_i^{\hat{t}_i}}\} ,\{ {\boldsymbol{y}_i^{\hat{t}_i}}\} ,\boldsymbol{v}^{\hat{t}_i}, \boldsymbol{z}^{\hat{t}_i}, \{ {\lambda _l^{\hat{t}_i}}\} ,\{ {\boldsymbol{\theta }_i^{\hat{t}_i}}\} )$ with Eq. (\ref{eq:new_25}), we have that,
\begin{equation}
\label{eq:A5-2}
\left\langle {{{\boldsymbol{x}}_i^{t+1}} \!-\! {{\boldsymbol{x}}_i^t}, {\nabla _{{{\boldsymbol{x}}_i}}}{L_p}(\{ {\boldsymbol{x}_i^{\hat{t}_i}}\} ,\!\{ {\boldsymbol{y}_i^{\hat{t}_i}}\} ,\!\boldsymbol{v}^{\hat{t}_i},\! \boldsymbol{z}^{\hat{t}_i},\! \{ {\lambda _l^{\hat{t}_i}}\} ,\!\{ {\boldsymbol{\theta }_i^{\hat{t}_i}}\} )} \right\rangle  \! = -\frac{1}{{{\eta _{\boldsymbol{x}}}}}||{{\boldsymbol{x}}_i^{t+1}} \!- {{\boldsymbol{x}}_i^t}|{|^2} \le
-\frac{1}{{{\eta _{\boldsymbol{x}}^t}}}||{{\boldsymbol{x}}_i^{t+1}} \!- {{\boldsymbol{x}}_i^t}|{|^2}.
\end{equation}

Next, combining the Cauchy-Schwarz inequality with Assumption 1, 2, we can get,
\begin{equation}
\label{eq:A5-3}
\begin{array}{l}
\left\langle {{\boldsymbol{x}}_i^{t+1} \!-\! {\boldsymbol{x}}_i^t,{\nabla _{{{\boldsymbol{x}}_i}}}{L_p}(\{ \boldsymbol{x}_i^{t}\} ,\!\{ \boldsymbol{y}_i^{t }\} ,\!{\boldsymbol{v}^{t}  },\!{\boldsymbol{z}^{t}  },\!\{ \lambda _l^{t}  \} ,\!\{ \boldsymbol{\theta }_i^{t}  \} ) \!-\! {\nabla _{{{\boldsymbol{x}}_i}}}{L_p}(\{ {\boldsymbol{x}_i^{\hat{t}_i}}\} ,\!\{ {\boldsymbol{y}_i^{\hat{t}_i}}\} ,\!\boldsymbol{v}^{\hat{t}_i},\! \boldsymbol{z}^{\hat{t}_i},\! \{ {\lambda _l^{\hat{t}_i}}\} ,\!\{ {\boldsymbol{\theta }_i^{\hat{t}_i}}\} )} \right\rangle \\
\! \le \! \frac{1}{2}||{{\boldsymbol{x}}_i^{t+1}} \!-\! {{\boldsymbol{x}}_i^t}|{|^2} \! + \! \frac{L^2}{2}(||{\boldsymbol{v}^t} \!-\! {\boldsymbol{v}^{{\hat{{t}}_j}}}|{|^2} \! + \! ||{\boldsymbol{z}^t} \!-\! {\boldsymbol{z}^{{\hat{{t}}_j}}}|{|^2} \! + \! \sum\limits_{l = 1}^{|{\boldsymbol{\mathcal{P}}^{t}}|} {||{\lambda _l^t} \!-\! {\lambda _l^{\hat{{t}}_j}}|{|^2}} )\\

\! \le \! \frac{1}{2}||{{\boldsymbol{x}}_i^{t+1}} \!-\! {{\boldsymbol{x}}_i^t}|{|^2} \! + \! \frac{{{3L^2\tau{ k_1 }}}}{2}(||{\boldsymbol{v}^{t+1}} \!-\! {\boldsymbol{v}^{t}}|{|^2} \! + \! ||{\boldsymbol{z}^{t+1}} \!-\! {\boldsymbol{z}^{{t}}}|{|^2} \! + \! \sum\limits_{l = 1}^{|{\boldsymbol{\mathcal{P}}^{t}}|} {||{\lambda _l^{t+1}} \!-\! {\lambda _l^{t}}|{|^2}} ).
\end{array}
\end{equation}

Thus, according to Eq. (\ref{eq:A4}), (\ref{eq:A5-2}) and (\ref{eq:A5-3}), we can obtain that,
\begin{equation}
\label{eq:A98_47}
\begin{array}{l}
{L_p}(\{ \boldsymbol{x}_i^{t + 1}\} ,\!\{ \boldsymbol{y}_i^{t }\} ,\!{\boldsymbol{v}^{t}  },\!{\boldsymbol{z}^{t}  },\!\{ \lambda _l^{t}  \} ,\!\{ \boldsymbol{\theta }_i^{t}  \} ) \!-\! {L_p}(\{ \boldsymbol{x}_i^{t}\} ,\!\{ \boldsymbol{y}_i^{t }\} ,\!{\boldsymbol{v}^{t}  },\!{\boldsymbol{z}^{t}  },\!\{ \lambda _l^{t}  \} ,\!\{ \boldsymbol{\theta }_i^{t}  \} )\vspace{0.3ex}\\

 \!\le\! \sum\limits_{i  = 1}^N \!{(\frac{{L  +  1}}{2} \!-\! \frac{1}{{{\eta _{\boldsymbol{x}}^t}}})}||{{\boldsymbol{x}}_i^{t+1}} \!-\! {{\boldsymbol{x}}_i^t}|{|^2} + \frac{{{3NL^2\tau{ k_1 }}}}{2}(||{\boldsymbol{v}^{t+1}} \!-\! {\boldsymbol{v}^{t}}|{|^2} \! + \! ||{\boldsymbol{z}^{t+1}} \!-\! {\boldsymbol{z}^{{t}}}|{|^2} \! + \! \sum\limits_{l = 1}^{|{\boldsymbol{\mathcal{P}}^{t}}|} {||{\lambda _l^{t+1}} \!-\! {\lambda _l^{t}}|{|^2}} ).
\end{array}
\end{equation}

Similarly, using the Lipschitz properties in Assumption 1, we have,
\begin{equation}
\label{eq:98_48}
\begin{array}{l}
{L_p}(\{ \boldsymbol{x}_i^{t + 1}\} ,\!\{ \boldsymbol{y}_i^{t+1 }\} ,\!{\boldsymbol{v}^{t}  },\!{\boldsymbol{z}^{t}  },\!\{ \lambda _l^{t}  \} ,\!\{ \boldsymbol{\theta }_i^{t}  \} ) \!-\! {L_p}(\{ \boldsymbol{x}_i^{t+1}\} ,\!\{ \boldsymbol{y}_i^{t }\} ,\!{\boldsymbol{v}^{t}  },\!{\boldsymbol{z}^{t}  },\!\{ \lambda _l^{t}  \} ,\!\{ \boldsymbol{\theta }_i^{t}  \} )\vspace{0.5ex}\\

 \le \sum\limits_{i  = 1}^N {\left(  \left\langle {{\nabla _{{{\boldsymbol{y}}_i}}}{L_p}{\rm{(\{ }}{{\boldsymbol{x}}_i^{t+1}}\} ,\!\{ \boldsymbol{y}_i^{t }\} ,\!{\boldsymbol{v}^{t}  },\!{\boldsymbol{z}^{t}  },\!\{ \lambda _l^{t}  \} ,\!\{ \boldsymbol{\theta }_i^{t}  \} ),{{\boldsymbol{y}}_i^{t+1}} \!-\! {{\boldsymbol{y}}_i^t}} \right\rangle  \! + \! \frac{L}{2}||{{\boldsymbol{y}}_i^{t+1}} \!-\! {{\boldsymbol{y}}_i^t}|{|^2} \right)}.
\end{array}
\end{equation}

Combining ${\nabla _{{{\boldsymbol{y}}_i}}}{L_p}(\{ {\boldsymbol{x}_i^{\hat{t}_i}}\} ,\{ {\boldsymbol{y}_i^{\hat{t}_i}}\} ,\boldsymbol{v}^{\hat{t}_i}, \boldsymbol{z}^{\hat{t}_i}, \{ {\lambda _l^{\hat{t}_i}}\} ,\{ {\boldsymbol{\theta }_i^{\hat{t}_i}}\} ) \!=\! {\nabla _{{{\boldsymbol{y}}_i}}}{\widetilde{L} _p}(\{ {\boldsymbol{x}_i^{\hat{t}_i}}\} ,\{ {\boldsymbol{y}_i^{\hat{t}_i}}\} ,\boldsymbol{v}^{\hat{t}_i}, \boldsymbol{z}^{\hat{t}_i}, \{ {\lambda _l^{\hat{t}_i}}\} ,\{ {\boldsymbol{\theta }_i^{\hat{t}_i}}\} )$ with Eq. (\ref{eq:new_26}), we can obtain that,
\begin{equation}
\label{eq:98_49}
\left\langle {{{\boldsymbol{y}}_i^{t+1}} \!-\! {{\boldsymbol{y}}_i^t}, {\nabla _{{{\boldsymbol{y}}_i}}}{L_p}(\{ {\boldsymbol{x}_i^{\hat{t}_i}}\} ,\!\{ {\boldsymbol{y}_i^{\hat{t}_i}}\} ,\!\boldsymbol{v}^{\hat{t}_i},\! \boldsymbol{z}^{\hat{t}_i},\! \{ {\lambda _l^{\hat{t}_i}}\} ,\!\{ {\boldsymbol{\theta }_i^{\hat{t}_i}}\} )} \right\rangle \! =  -\frac{1}{{{\eta _{\boldsymbol{y}}}}}||{{\boldsymbol{y}}_i^{t+1}} \!- {{\boldsymbol{y}}_i^t}|{|^2} \le  -\frac{1}{{{\eta _{\boldsymbol{y}}^t}}}||{{\boldsymbol{y}}_i^{t+1}} \!- {{\boldsymbol{y}}_i^t}|{|^2}.
\end{equation}

Then, combining the Cauchy-Schwarz inequality with Assumption 1, 2, we can get the following inequalities,
\begin{equation}
\label{eq:98_50}
\begin{array}{l}
\left\langle {{\boldsymbol{y}}_i^{t+1} \!-\! {\boldsymbol{y}}_i^t,{\nabla _{{{\boldsymbol{y}}_i}}}{L_p}(\{ \boldsymbol{x}_i^{t+1}\} ,\!\{ \boldsymbol{y}_i^{t }\} ,\!{\boldsymbol{v}^{t}  },\!{\boldsymbol{z}^{t}  },\!\{ \lambda _l^{t}  \} ,\!\{ \boldsymbol{\theta }_i^{t}  \} ) \!-\! {\nabla _{{{\boldsymbol{y}}_i}}}{L_p}(\{ {\boldsymbol{x}_i^{\hat{t}_i}}\} ,\!\{ {\boldsymbol{y}_i^{\hat{t}_i}}\} ,\!\boldsymbol{v}^{\hat{t}_i},\! \boldsymbol{z}^{\hat{t}_i},\! \{ {\lambda _l^{\hat{t}_i}}\} ,\!\{ {\boldsymbol{\theta }_i^{\hat{t}_i}}\} )} \right\rangle \\

\! \le \! \frac{1}{2}||{{\boldsymbol{y}}_i^{t+1}} \!-\! {{\boldsymbol{y}}_i^t}|{|^2} \! + \! \frac{L^2}{2}(||{{\boldsymbol{x}}_i^{t+1}} \!-\! {{\boldsymbol{x}}_i^t}|{|^2} \!+\! ||{\boldsymbol{v}^t} \!-\! {\boldsymbol{v}^{{\hat{{t}}_j}}}|{|^2} \! + \! ||{\boldsymbol{z}^t} \!-\! {\boldsymbol{z}^{{\hat{{t}}_j}}}|{|^2} \! + \! \sum\limits_{l = 1}^{|{\boldsymbol{\mathcal{P}}^{t}}|} {||{\lambda _l^t} \!-\! {\lambda _l^{\hat{{t}}_j}}|{|^2}} )\\

\! \le \! \frac{1}{2}||{{\boldsymbol{y}}_i^{t+1}} \!-\! {{\boldsymbol{y}}_i^t}|{|^2} \! + \! \frac{L^2}{2}||{{\boldsymbol{x}}_i^{t+1}} \!-\! {{\boldsymbol{x}}_i^t}|{|^2} \!+\!\frac{{{3L^2\tau{ k_1 }}}}{2}(||{\boldsymbol{v}^{t+1}} \!-\! {\boldsymbol{v}^{t}}|{|^2} \! + \! ||{\boldsymbol{z}^{t+1}} \!-\! {\boldsymbol{z}^{{t}}}|{|^2} \! + \! \sum\limits_{l = 1}^{|{\boldsymbol{\mathcal{P}}^{t}}|} {||{\lambda _l^{t+1}} \!-\! {\lambda _l^{t}}|{|^2}} ).
\end{array}
\end{equation}

Thus, combining Eq. (\ref{eq:98_48}), (\ref{eq:98_49}) with (\ref{eq:98_50}), we have,
\begin{equation}
\label{eq:A98_51}
\begin{array}{l}
{L_p}(\{ \boldsymbol{x}_i^{t + 1}\} ,\!\{ \boldsymbol{y}_i^{t+1 }\} ,\!{\boldsymbol{v}^{t}  },\!{\boldsymbol{z}^{t}  },\!\{ \lambda _l^{t}  \} ,\!\{ \boldsymbol{\theta }_i^{t}  \} ) - {L_p}(\{ \boldsymbol{x}_i^{t+1}\} ,\!\{ \boldsymbol{y}_i^{t }\} ,\!{\boldsymbol{v}^{t}  },\!{\boldsymbol{z}^{t}  },\!\{ \lambda _l^{t}  \} ,\!\{ \boldsymbol{\theta }_i^{t}  \} ) \vspace{0.5ex}\\

 \le \sum\limits_{i  = 1}^N {(\frac{{L  +  1}}{2} \!-\! \frac{1}{{{\eta _{\boldsymbol{y}}^t}}})}||{{\boldsymbol{y}}_i^{t+1}} \!-\! {{\boldsymbol{y}}_i^t}|{|^2} + \sum\limits_{i  = 1}^N \frac{L^2}{2}||{{\boldsymbol{x}}_i^{t+1}} \!-\! {{\boldsymbol{x}}_i^t}|{|^2}\\ 

+ \frac{{{3NL^2\tau{ k_1 }}}}{2}(||{\boldsymbol{v}^{t+1}} \!-\! {\boldsymbol{v}^{t}}|{|^2} \! + \! ||{\boldsymbol{z}^{t+1}} \!-\! {\boldsymbol{z}^{{t}}}|{|^2} \! + \! \sum\limits_{l = 1}^{|{\boldsymbol{\mathcal{P}}^{t}}|} {||{\lambda _l^{t+1}} \!-\! {\lambda _l^{t}}|{|^2}} ).
\end{array}
\end{equation}

Combining the Lipschitz properties in Assumption 1 with Eq. (\ref{eq:new_27}), we have,
\begin{equation}
\label{eq:A98_52}
\begin{array}{l}
{L_p}(\{ \boldsymbol{x}_i^{t + 1}\} ,\!\{ \boldsymbol{y}_i^{t+1 }\} ,\!{\boldsymbol{v}^{t+1}  },\!{\boldsymbol{z}^{t}  },\!\{ \lambda _l^{t}  \} ,\!\{ \boldsymbol{\theta }_i^{t}  \} ) - {L_p}(\{ \boldsymbol{x}_i^{t+1}\} ,\!\{ \boldsymbol{y}_i^{t+1 }\} ,\!{\boldsymbol{v}^{t}  },\!{\boldsymbol{z}^{t}  },\!\{ \lambda _l^{t}  \} ,\!\{ \boldsymbol{\theta }_i^{t}  \} ) \vspace{1.5ex}\\

  \le{  \left\langle {{\nabla _{{{\boldsymbol{v}}}}}{L_p}{\rm{(\{ }}{{\boldsymbol{x}}_i^{t+1}}\} ,\!\{ \boldsymbol{y}_i^{t+1 }\} ,\!{\boldsymbol{v}^{t}  },\!{\boldsymbol{z}^{t}  },\!\{ \lambda _l^{t}  \} ,\!\{ \boldsymbol{\theta }_i^{t}  \} ),{{\boldsymbol{v}}^{t+1}} \!-\! {{\boldsymbol{v}}^t}} \right\rangle  \! + \! \frac{L}{2}||{{\boldsymbol{v}}^{t+1}} \!-\! {{\boldsymbol{v}}^t}|{|^2}} \vspace{1.5ex}\\
  
  \le (\frac{L}{2}-\frac{1}{{{\eta _{\boldsymbol{v}}^t}}})||{{\boldsymbol{v}}^{t+1}} \!-\! {{\boldsymbol{v}}^t}|{|^2}.
\end{array}
\end{equation}

Similarly, combining the Lipschitz properties in Assumption 1 with Eq. (\ref{eq:new_28}), we have,
\begin{equation}
\label{eq:A98_53}
\begin{array}{l}
{L_p}(\{ \boldsymbol{x}_i^{t + 1}\} ,\!\{ \boldsymbol{y}_i^{t+1 }\} ,\!{\boldsymbol{v}^{t+1}  },\!{\boldsymbol{z}^{t+1}  },\!\{ \lambda _l^{t}  \} ,\!\{ \boldsymbol{\theta }_i^{t}  \} ) \!-\! {L_p}(\{ \boldsymbol{x}_i^{t+1}\} ,\!\{ \boldsymbol{y}_i^{t+1 }\} ,\!{\boldsymbol{v}^{t+1}  },\!{\boldsymbol{z}^{t}  },\!\{ \lambda _l^{t}  \} ,\!\{ \boldsymbol{\theta }_i^{t}  \} ) \vspace{1.5ex}\\

  \le{  \left\langle {{\nabla _{{{\boldsymbol{z}}}}}{L_p}{\rm{(\{ }}{{\boldsymbol{x}}_i^{t+1}}\} ,\!\{ \boldsymbol{y}_i^{t+1 }\} ,\!{\boldsymbol{v}^{t+1}  },\!{\boldsymbol{z}^{t}  },\!\{ \lambda _l^{t}  \} ,\!\{ \boldsymbol{\theta }_i^{t}  \} ), {{\boldsymbol{z}}^{t+1}} \!-\! {{\boldsymbol{z}}^t}} \right\rangle  \! + \! \frac{L}{2}||{{\boldsymbol{z}}^{t+1}} \!-\! {{\boldsymbol{z}}^t}|{|^2}} \vspace{1.5ex}\\
  
  \le (\frac{L}{2}-\frac{1}{{{\eta _{\boldsymbol{z}}^t}}})||{{\boldsymbol{z}}^{t+1}} \!-\! {{\boldsymbol{z}}^t}|{|^2}.
\end{array}
\end{equation}

By combining  Eq. (\ref{eq:A98_47}), (\ref{eq:A98_51}), (\ref{eq:A98_52}), (\ref{eq:A98_53}), we conclude the proof of Lemma 1.

\begin{lemma} \label{lemma 2}
Suppose Assumption 1 and 2 hold, $\forall t \ge T_1$, we have:
\begin{equation}
\label{eq:A10}
\begin{array}{l}
{L_p}(\{ \boldsymbol{x}_i^{t + 1}\} ,\!\{ \boldsymbol{y}_i^{t+1 }\} ,\!{\boldsymbol{v}^{t+1}  },\!{\boldsymbol{z}^{t+1}  },\!\{ \lambda _l^{t+1}  \} ,\!\{ \boldsymbol{\theta }_i^{t+1}  \} ) - {L_p}(\{ \boldsymbol{x}_i^{t}\} ,\!\{ \boldsymbol{y}_i^{t }\} ,\!{\boldsymbol{v}^{t}  },\!{\boldsymbol{z}^{t}  },\!\{ \lambda _l^{t}  \} ,\!\{ \boldsymbol{\theta }_i^{t}  \} )\vspace{1ex}\\

 \le  (\frac{{L  + L^2 +  1}}{2} \!-\! \frac{1}{{{\eta _{\boldsymbol{x}}^t}}} \! + \! \frac{{|{\boldsymbol{\mathcal{P}}^{t}}|{L^2}}}{{2{a_1}}} \! + \! \frac{{|{{\boldsymbol{\mathcal{Q}}}^{t  +  1}}|{L^2}}}{{2{a_3}}})\! \sum\limits_{i = 1}^N \! {||{{\boldsymbol{x}}_i^{t+1}} \!-\! {{\boldsymbol{x}}_i^t}|{|^2}} \\

+ (\frac{{L  +  1}}{2} \!-\! \frac{1}{{{\eta _{\boldsymbol{y}}^t}}} \! + \! \frac{{|{\boldsymbol{\mathcal{P}}^{t}}|{L^2}}}{{2{a_1}}} \! + \! \frac{{|{{\boldsymbol{\mathcal{Q}}}^{t  +  1}}|{L^2}}}{{2{a_3}}})\!\sum\limits_{i = 1}^N \!{||{{\boldsymbol{y}}_i^{t+1}} \!-\! {{\boldsymbol{y}}_i^t}|{|^2}} \vspace{0.5ex}\\

 +  (\frac{{L  +  6\tau{ k_1 }N{L^2}}}{2} \!-\! \frac{1}{{{\eta _{\boldsymbol{v}}^t}}} \! + \! \frac{{|{\boldsymbol{\mathcal{P}}^{t}}|{L^2}}}{{2{a_1}}} \! + \! \frac{{|{{\boldsymbol{\mathcal{Q}}}^{t  +  1}}|{L^2}}}{{2{a_3}}})||\boldsymbol{v}^{t+1} \!-\! \boldsymbol{v}^t|{|^2}
\vspace{1ex}\\

+ (\frac{{L  +  6\tau{ k_1 }N{L^2}}}{2} \!-\! \frac{1}{{{\eta _{\boldsymbol{z}}^t}}} \! + \! \frac{{|{\boldsymbol{\mathcal{P}}^{t}}|{L^2}}}{{2{a_1}}} \! + \! \frac{{|{{\boldsymbol{\mathcal{Q}}}^{t  +  1}}|{L^2}}}{{2{a_3}}})||\boldsymbol{z}^{t+1} \!-\! \boldsymbol{z}^t|{|^2} \! + \! \frac{1}{{2{{\eta _{\boldsymbol{\theta }}}}}}\sum\limits_{i = 1}^N
 \!{||{{\boldsymbol{\theta}}_i^t} \!-\! {{\boldsymbol{\theta}}_i^{t-1}}|{|^2}} \vspace{0.5ex}\\

  +  (\frac{{{a_1}  +  6\tau{ k_1 }N{L^2}}}{2} \!-\! \frac{{{c_1^{t-1}} - {c_1^t}}}{2} \! + \! \frac{1}{{2{{\eta _{\lambda}}}}})\sum\limits_{l = 1}^{|{\boldsymbol{\mathcal{P}}^{t}}|} {||{\lambda _l^{t+1}} \!-\! {\lambda _l^t}|{|^2}} \! + \! (\frac{{{a_3}}}{2} \!-\! \frac{{{c_2^{t-1}} - {c_2^t}}}{2} \! + \! \frac{1}{{2{{\eta _{\boldsymbol{\theta }}}}}}) \! \sum\limits_{i = 1}^N\! {||{{\boldsymbol{\theta}}_i^{t+1}} \!-\! {{\boldsymbol{\theta}}_i^t}|{|^2}} \vspace{0.5ex}\\
 
  +  \frac{{{c_1^{t-1}}}}{2} \! \sum\limits_{l = 1}^{|{\boldsymbol{\mathcal{P}}^{t}}|}\! {(||{\lambda _l^{t+1}}|{|^2} \!-\! ||{\lambda _l^t}|{|^2})}  \! + \! \frac{1}{{2{{\eta _{\lambda}}}}} \! \sum\limits_{l = 1}^{|{\boldsymbol{\mathcal{P}}^{t}}|}\! {||{\lambda _l^t} \!-\! {\lambda _l^{t-1}}|{|^2}} 
 
 \! +  \frac{{{c_2^{t-1}}}}{2}\sum\limits_{i = 1}^N\! {(||{{\boldsymbol{\theta}}_i^{t+1}}|{|^2} \!-\! ||{{\boldsymbol{\theta}}_i^t}|{|^2})}, 
 
\end{array}
\end{equation}
where $a_1 >0$ and $a_3 >0$ are constants.
\end{lemma}

\emph{\textbf{Proof of Lemma 2:}}

According to Eq. (\ref{eq:new_29}), in $(t+1)^{\rm{th}}$ iteration, $\forall \lambda  \in {\bf{\Lambda}} $, it follows that:
\begin{equation}
\label{eq:A11}
\left\langle {{\lambda _l^{t+1}} \!-\! {\lambda _l^t} \!-\! {{\eta _{\lambda}}}{\nabla _{{\lambda _l}}}{{\widetilde{L}}_p}(\{ \boldsymbol{x}_i^{t + 1}\} ,\!\{ \boldsymbol{y}_i^{t+1 }\} ,\!{\boldsymbol{v}^{t+1}  },\!{\boldsymbol{z}^{t+1}  },\!\{ \lambda _l^{t}  \} ,\!\{ \boldsymbol{\theta }_i^{t}  \} ),\lambda  \!-\! {\lambda _l^{t+1}}} \right\rangle  = 0.
\end{equation}

Let $\lambda  = {\lambda _l^t}$, we can obtain:
\begin{equation}
\label{eq:A12_1}
\left\langle {{\nabla _{{\lambda _l}}}{{\widetilde{L}}_p}(\{ \boldsymbol{x}_i^{t + 1}\} ,\!\{ \boldsymbol{y}_i^{t+1 }\} ,\!{\boldsymbol{v}^{t+1}  },\!{\boldsymbol{z}^{t+1}  },\!\{ \lambda _l^{t}  \} ,\!\{ \boldsymbol{\theta }_i^{t}  \} ) \!-\! \frac{1}{{{{\eta _{\lambda}}}}}({\lambda _l^{t+1}} \!-\! {\lambda _l^t}),{\lambda _l^t} \!-\! {\lambda _l^{t+1}}} \right\rangle  = 0.
\end{equation}

Likewise, in $t^{\rm{th}}$ iteration, we can obtain:
\begin{equation}
\label{eq:A12}
\left\langle {{\nabla _{{\lambda _l}}}{{\widetilde{L}}_p}(\{ \boldsymbol{x}_i^{t}\} ,\!\{ \boldsymbol{y}_i^{t}\} ,\!{\boldsymbol{v}^{t}  },\!{\boldsymbol{z}^{t}  },\!\{ \lambda _l^{t-1}  \} ,\!\{ \boldsymbol{\theta }_i^{t-1}  \} ) \!-\! \frac{1}{{{{\eta _{\lambda}}}}}({\lambda _l^t} \!-\! {\lambda _l^{t-1}}),{\lambda _l^{t+1}} \!-\! {\lambda _l^t}} \right\rangle  = 0.
\end{equation}

Since ${\widetilde{L}_p}(\{ \boldsymbol{x}_i\} ,\!\{ \boldsymbol{y}_i\} ,\!{\boldsymbol{v}  },\!{\boldsymbol{z}  },\!\{ \lambda _l  \} ,\!\{ \boldsymbol{\theta }_i  \} )$ is concave with respect to ${\lambda _l}$ and follows from Eq. (\ref{eq:A12_1}) and Eq. (\ref{eq:A12}), $\forall t \ge T_1$, we have,
\begin{equation}
\label{eq:A14}
\begin{array}{l}
{\widetilde{L}_p}(\{ \boldsymbol{x}_i^{t + 1}\} ,\!\{ \boldsymbol{y}_i^{t+1 }\} ,\!{\boldsymbol{v}^{t+1}  },\!{\boldsymbol{z}^{t+1}  },\!\{ \lambda _l^{t+1}  \} ,\!\{ \boldsymbol{\theta }_i^{t}  \} ) - {\widetilde{L}_p}(\{ \boldsymbol{x}_i^{t + 1}\} ,\!\{ \boldsymbol{y}_i^{t+1 }\} ,\!{\boldsymbol{v}^{t+1}  },\!{\boldsymbol{z}^{t+1}  },\!\{ \lambda _l^{t}  \} ,\!\{ \boldsymbol{\theta }_i^{t}  \} )\vspace{1ex}\\
\! \le \! \sum\limits_{l = 1}^{|{\boldsymbol{\mathcal{P}}^{t}}|} \! {\left\langle \! {{\nabla _{{\lambda _l}}}{{\widetilde{L} }_p}(\{ \boldsymbol{x}_i^{t + 1}\} ,\!\{ \boldsymbol{y}_i^{t+1 }\} ,\!{\boldsymbol{v}^{t+1}  },\!{\boldsymbol{z}^{t+1}  },\!\{ \lambda _l^{t}  \} ,\!\{ \boldsymbol{\theta }_i^{t}  \} ),{\lambda _l^{t+1}} - {\lambda _l^t}} \! \right\rangle } \\
\! \le \! \sum\limits_{l  =  1}^{|{\boldsymbol{\mathcal{P}}^{t}}|} \! {( \! \left\langle \! {{\nabla _{{\lambda _l}}}{{\widetilde{L} }_p}(\{ \boldsymbol{x}_i^{t + 1}\} ,\!\{ \boldsymbol{y}_i^{t+1 }\} ,\!{\boldsymbol{v}^{t+1}  },\!{\boldsymbol{z}^{t+1}  },\!\{ \lambda _l^{t}  \} ,\!\{ \boldsymbol{\theta }_i^{t}  \} ) \! -\! {\nabla _{{\lambda _l}}}{{\widetilde{L}}_p}(\{ \boldsymbol{x}_i^{t}\} ,\!\{ \boldsymbol{y}_i^{t}\} ,\!{\boldsymbol{v}^{t}  },\!{\boldsymbol{z}^{t}  },\!\{ \lambda _l^{t-1}  \} ,\!\{ \boldsymbol{\theta }_i^{t-1}  \} ),{\lambda _l^{t+1}}\! -\! {\lambda _l^t}} \! \right\rangle } \vspace{0.75ex}\\
  \qquad \quad + \frac{1}{{{{\eta _{\lambda}}}}}\left\langle {{\lambda _l^t} - {\lambda _l^{t-1}},{\lambda _l^{t+1}} - {\lambda _l^t}} \right\rangle ).
\end{array}
\end{equation}

Denoting ${{\boldsymbol{v}}_{1,l}^{t + 1}} = {\lambda _l^{t+1}} - {\lambda _l^t} - ({\lambda _l^t} - {\lambda _l^{t-1}})$, we can get the following equality,
\begin{equation}
\label{eq:A15}
\begin{array}{l}
\sum\limits_{l = 1}^{|{\boldsymbol{\mathcal{P}}^{t}}|} \! {\left\langle \! {{\nabla _{{\lambda _l}}}{{\widetilde{L}}_p}(\{ \boldsymbol{x}_i^{t + 1}\} ,\!\{ \boldsymbol{y}_i^{t+1 }\} ,\!{\boldsymbol{v}^{t+1}  },\!{\boldsymbol{z}^{t+1}  },\!\{ \lambda _l^{t}  \} ,\!\{ \boldsymbol{\theta }_i^{t}  \} )  \! -\! {\nabla _{{\lambda _l}}}{{\widetilde{L} }_p}(\{ \boldsymbol{x}_i^{t}\} ,\!\{ \boldsymbol{y}_i^{t}\} ,\!{\boldsymbol{v}^{t}  },\!{\boldsymbol{z}^{t}  },\!\{ \lambda _l^{t-1}  \} ,\!\{ \boldsymbol{\theta }_i^{t-1}  \} ),{\lambda _l^{t+1}}  \! -\! {\lambda _l^t}} \! \right\rangle } \\
 \!= \! \sum\limits_{l = 1}^{|{\boldsymbol{\mathcal{P}}^{t}}|} \! {\left\langle \! {{\nabla _{{\lambda _l}}}{{\widetilde{L}}_p}(\{ \boldsymbol{x}_i^{t + 1}\} ,\!\{ \boldsymbol{y}_i^{t+1 }\} ,\!{\boldsymbol{v}^{t+1}  },\!{\boldsymbol{z}^{t+1}  },\!\{ \lambda _l^{t}  \} ,\!\{ \boldsymbol{\theta }_i^{t}  \} )  \! -\! {\nabla _{{\lambda _l}}}{{\widetilde{L}}_p}(\{ \boldsymbol{x}_i^{t}\} ,\!\{ \boldsymbol{y}_i^{t}\} ,\!{\boldsymbol{v}^{t}  },\!{\boldsymbol{z}^{t}  },\!\{ \lambda _l^{t}  \} ,\!\{ \boldsymbol{\theta }_i^{t}  \} ) {\rm{)}},{\lambda _l^{t+1}}  \! -\! {\lambda _l^t}} \! \right\rangle } (1a)\\
  \! +\! \sum\limits_{l = 1}^{|{\boldsymbol{\mathcal{P}}^{t}}|} \! {\left\langle \! {{\nabla _{{\lambda _l}}}{{\widetilde{L}}_p}(\{ \boldsymbol{x}_i^{t}\} ,\!\{ \boldsymbol{y}_i^{t}\} ,\!{\boldsymbol{v}^{t}  },\!{\boldsymbol{z}^{t}  },\!\{ \lambda _l^{t}  \} ,\!\{ \boldsymbol{\theta }_i^{t}  \} )  \! -\! {\nabla _{{\lambda _l}}}{{\widetilde{L}}_p}(\{ \boldsymbol{x}_i^{t}\} ,\!\{ \boldsymbol{y}_i^{t}\} ,\!{\boldsymbol{v}^{t}  },\!{\boldsymbol{z}^{t}  },\!\{ \lambda _l^{t-1}  \} ,\!\{ \boldsymbol{\theta }_i^{t-1}  \} ), {{\boldsymbol{v}}_{1,l}^{t + 1}}} \right\rangle } (1b)\\
  \! +\! \sum\limits_{l = 1}^{|{\boldsymbol{\mathcal{P}}^{t}}|} \! {\left\langle {{\nabla _{{\lambda _l}}}{{\widetilde{L}}_p}(\{ \boldsymbol{x}_i^{t}\} ,\!\{ \boldsymbol{y}_i^{t}\} ,\!{\boldsymbol{v}^{t}  },\!{\boldsymbol{z}^{t}  },\!\{ \lambda _l^{t}  \} ,\!\{ \boldsymbol{\theta }_i^{t}  \} )  \! -\! {\nabla _{{\lambda _l}}}{{\widetilde{L}}_p}(\{ \boldsymbol{x}_i^{t}\} ,\!\{ \boldsymbol{y}_i^{t}\} ,\!{\boldsymbol{v}^{t}  },\!{\boldsymbol{z}^{t}  },\!\{ \lambda _l^{t-1}  \} ,\!\{ \boldsymbol{\theta }_i^{t-1}  \} ),{\lambda _l^t}  \! -\! {\lambda _l^{t-1}}} \! \right\rangle } (1c).
\end{array}
\end{equation}

\vspace{0.5ex}

First, we put attention on the ($1a$) in Eq.  (\ref{eq:A15}), ($1a$) can be expressed as follows,
\begin{equation}
\label{eq:A16}
\begin{array}{l}
\left\langle \!{{\nabla _{{\lambda _l}}}{{\widetilde{L}}_p}(\{ \boldsymbol{x}_i^{t + 1}\} ,\!\{ \boldsymbol{y}_i^{t+1 }\} ,\!{\boldsymbol{v}^{t+1}  },\!{\boldsymbol{z}^{t+1}  },\!\{ \lambda _l^{t}  \} ,\!\{ \boldsymbol{\theta }_i^{t}  \} ) \! - \! {\nabla _{{\lambda _l}}}{{\widetilde{L}}_p}(\{ \boldsymbol{x}_i^{t}\} ,\!\{ \boldsymbol{y}_i^{t}\} ,\!{\boldsymbol{v}^{t}  },\!{\boldsymbol{z}^{t}  },\!\{ \lambda _l^{t}  \} ,\!\{ \boldsymbol{\theta }_i^{t}  \} ),{\lambda _l^{t+1}} \! - \! {\lambda _l^t}}\! \right\rangle  \vspace{1ex}\\

\! = \!\left\langle {{\nabla _{{\lambda _l}}}{L_p}(\{ \boldsymbol{x}_i^{t + 1}\} ,\!\{ \boldsymbol{y}_i^{t+1 }\} ,\!{\boldsymbol{v}^{t+1}  },\!{\boldsymbol{z}^{t+1}  },\!\{ \lambda _l^{t}  \} ,\!\{ \boldsymbol{\theta }_i^{t}  \} ) \! - \! {\nabla _{{\lambda _l}}}{L_p}(\{ \boldsymbol{x}_i^{t}\} ,\!\{ \boldsymbol{y}_i^{t}\} ,\!{\boldsymbol{v}^{t}  },\!{\boldsymbol{z}^{t}  },\!\{ \lambda _l^{t}  \} ,\!\{ \boldsymbol{\theta }_i^{t}  \} ),{\lambda _l^{t+1}} \! - \! {\lambda _l^t}} \right\rangle \vspace{1ex}\\
 \! +  \frac{{{c_1^{t-1}}  -  {c_1^t}}}{2}(||{\lambda _l^{t+1}}|{|^2} \! - \! ||{\lambda _l^t}|{|^2}) \! - \! \frac{{{c_1^{t-1}}  -  {c_1^t}}}{2}||{\lambda _l^{t+1}} \! - \! {\lambda _l^t}|{|^2}.
\end{array}
\end{equation}

Combining Cauchy-Schwarz inequality with Assumption 1, we can obtain,
\begin{equation}
\label{eq:A17}
\begin{array}{l}
\left\langle {{\nabla _{{\lambda _l}}}{L_p}(\{ \boldsymbol{x}_i^{t + 1}\} ,\!\{ \boldsymbol{y}_i^{t+1 }\} ,\!{\boldsymbol{v}^{t+1}  },\!{\boldsymbol{z}^{t+1}  },\!\{ \lambda _l^{t}  \} ,\!\{ \boldsymbol{\theta }_i^{t}  \} )\! -\! {\nabla _{{\lambda _l}}}{L_p}(\{ \boldsymbol{x}_i^{t}\} ,\!\{ \boldsymbol{y}_i^{t}\} ,\!{\boldsymbol{v}^{t}  },\!{\boldsymbol{z}^{t}  },\!\{ \lambda _l^{t}  \} ,\!\{ \boldsymbol{\theta }_i^{t}  \} ),{\lambda _l^{t+1}} - {\lambda _l^t}} \right\rangle \vspace{0.5ex} \\

 \!\le \frac{{{L^2}}}{{2{a_1}}}( \sum\limits_{i = 1}^N({||{{\boldsymbol{x}}_i^{t+1}} \!-\! {{\boldsymbol{x}}_i^t}|{|^2}} + {||{{\boldsymbol{y}}_i^{t+1}} \!-\! {{\boldsymbol{y}}_i^t}|{|^2}})  \! + \! ||{\boldsymbol{v}}^{t+1} \!-\! {\boldsymbol{v}^t}|{|^2} \! + \! ||{\boldsymbol{z}}^{t+1} \!-\! {\boldsymbol{z}^t}|{|^2} ) \! + \! \frac{{{a_1}}}{2}||{\lambda _l^{t+1}} - {\lambda _l^t}|{|^2},
\end{array}
\end{equation}
where $a_1 >0$  is a constant. Combining Eq. (\ref{eq:A16}) with Eq. (\ref{eq:A17}), we can obtain that,
\begin{equation}
\label{eq:A18}
\begin{array}{l}
\sum\limits_{l = 1}^{|{\boldsymbol{\mathcal{P}}^{t}}|}\! {\left\langle \!{{\nabla _{{\lambda _l}}}{{\widetilde{L}}_p}(\{ \boldsymbol{x}_i^{t + 1}\} ,\!\{ \boldsymbol{y}_i^{t+1 }\} ,\!{\boldsymbol{v}^{t+1}  },\!{\boldsymbol{z}^{t+1}  },\!\{ \lambda _l^{t}  \} ,\!\{ \boldsymbol{\theta }_i^{t}  \} ) \! - \! {\nabla _{{\lambda _l}}}{{\widetilde{L}}_p}(\{ \boldsymbol{x}_i^{t}\} ,\!\{ \boldsymbol{y}_i^{t}\} ,\!{\boldsymbol{v}^{t}  },\!{\boldsymbol{z}^{t}  },\!\{ \lambda _l^{t}  \} ,\!\{ \boldsymbol{\theta }_i^{t}  \} ),{\lambda _l^{t+1}} \! - \! {\lambda _l^t}}\! \right\rangle  } \\

 \!\le \sum\limits_{l = 1}^{|{\boldsymbol{\mathcal{P}}^{t}}|} ( \frac{{{L^2}}}{{2{a_1}}}( \sum\limits_{i = 1}^N({||{{\boldsymbol{x}}_i^{t+1}} \!-\! {{\boldsymbol{x}}_i^t}|{|^2}} + {||{{\boldsymbol{y}}_i^{t+1}} \!-\! {{\boldsymbol{y}}_i^t}|{|^2}})  \! + \! ||{\boldsymbol{v}}^{t+1} \!-\! {\boldsymbol{v}^t}|{|^2} \! + \! ||{\boldsymbol{z}}^{t+1} \!-\! {\boldsymbol{z}^t}|{|^2} ) \! + \! \frac{{{a_1}}}{2}||{\lambda _l^{t+1}} - {\lambda _l^t}|{|^2}\vspace{1ex}\\
 
\qquad \quad + \frac{{{c_1^{t-1}}  - {c_1^t}}}{2}(||{\lambda _l^{t + 1}}|{|^2} \! - \! ||{\lambda _l^t}|{|^2}) \! - \! \frac{{{c_1^{t-1}}  -  {c_1^t}}}{2}||{\lambda _l^{t + 1}} \! - \! {\lambda _l^t}|{|^2}).
\end{array}
\end{equation}

Then, we focus on the ($1b$) in Eq. (\ref{eq:A15}). According to Cauchy-Schwarz inequality,  ($1b$) can be expressed as follows,
\begin{equation}
\label{eq:A19}
\begin{array}{l}
\sum\limits_{l = 1}^{|{\boldsymbol{\mathcal{P}}^{t}}|} \! {\left\langle \! {{\nabla _{{\lambda _l}}}{{\widetilde{L}}_p}(\{ \boldsymbol{x}_i^{t}\} ,\!\{ \boldsymbol{y}_i^{t}\} ,\!{\boldsymbol{v}^{t}  },\!{\boldsymbol{z}^{t}  },\!\{ \lambda _l^{t}  \} ,\!\{ \boldsymbol{\theta }_i^{t}  \} )  \! -\! {\nabla _{{\lambda _l}}}{{\widetilde{L}}_p}(\{ \boldsymbol{x}_i^{t}\} ,\!\{ \boldsymbol{y}_i^{t}\} ,\!{\boldsymbol{v}^{t}  },\!{\boldsymbol{z}^{t}  },\!\{ \lambda _l^{t-1}  \} ,\!\{ \boldsymbol{\theta }_i^{t-1}  \} ), {{\boldsymbol{v}}_{1,l}^{t + 1}}} \right\rangle } \\

\! \le \! \sum\limits_{l = 1}^{|{\boldsymbol{\mathcal{P}}^{t}}|}\! {( \frac{{{a_2}}}{2}||{\nabla _{{\lambda _l}}}{{\widetilde{L}}_p}(\{ \boldsymbol{x}_i^{t}\} ,\!\{ \boldsymbol{y}_i^{t}\} ,\!{\boldsymbol{v}^{t}  },\!{\boldsymbol{z}^{t}  },\!\{ \lambda _l^{t}  \} ,\!\{ \boldsymbol{\theta }_i^{t}  \} )  \! -\! {\nabla _{{\lambda _l}}}{{\widetilde{L}}_p}(\{ \boldsymbol{x}_i^{t}\} ,\!\{ \boldsymbol{y}_i^{t}\} ,\!{\boldsymbol{v}^{t}  },\!{\boldsymbol{z}^{t}  },\!\{ \lambda _l^{t-1}  \} ,\!\{ \boldsymbol{\theta }_i^{t-1}  \} )|{|^2}} \vspace{0.5mm}\\
\qquad \;\; + \frac{1}{{2{a_2}}}||{{\boldsymbol{v}}_{1,l}^{t + 1}}|{|^2}),
\end{array}
\end{equation}
where $a_2 >0$  is a constant. Next, we focus on the ($1c$) in Eq. (\ref{eq:A15}). Defining ${L_1}' = L + {c_1^0}$, according to Assumption 1 and the trigonometric inequality, $\forall \lambda _l$, we have,
\begin{equation}
\label{eq:A22}
\begin{array}{l}
||{\nabla _{{\lambda _l}}}{{\widetilde{L}}_p}(\{ \boldsymbol{x}_i^{t}\} ,\!\{ \boldsymbol{y}_i^{t}\} ,\!{\boldsymbol{v}^{t}  },\!{\boldsymbol{z}^{t}  },\!\{ \lambda _l^{t}  \} ,\!\{ \boldsymbol{\theta }_i^{t}  \} )  \! -\! {\nabla _{{\lambda _l}}}{{\widetilde{L}}_p}(\{ \boldsymbol{x}_i^{t}\} ,\!\{ \boldsymbol{y}_i^{t}\} ,\!{\boldsymbol{v}^{t}  },\!{\boldsymbol{z}^{t}  },\!\{ \lambda _l^{t-1}  \} ,\!\{ \boldsymbol{\theta }_i^{t-1}  \} )||\vspace{1ex}\\

 \!= \! ||{\nabla _{{\lambda _l}}}{{{L}}_p}(\{ \boldsymbol{x}_i^{t}\} ,\!\{ \boldsymbol{y}_i^{t}\} ,\!{\boldsymbol{v}^{t}  },\!{\boldsymbol{z}^{t}  },\!\{ \lambda _l^{t}  \} ,\!\{ \boldsymbol{\theta }_i^{t}  \} )  \! -\! {\nabla _{{\lambda _l}}}{{{L}}_p}(\{ \boldsymbol{x}_i^{t}\} ,\!\{ \boldsymbol{y}_i^{t}\} ,\!{\boldsymbol{v}^{t}  },\!{\boldsymbol{z}^{t}  },\!\{ \lambda _l^{t-1}  \} ,\!\{ \boldsymbol{\theta }_i^{t}  \} ) \!-\! c_1^{t-1}(\lambda _l^{t}\!-\! \lambda _l^{t-1})||\vspace{1ex}\\
 
 \! \le\! (L + {c_1^{t-1}})||{\lambda _l^t}  \! - \! {\lambda _l^{t-1}}{\rm{||}}\vspace{1ex}\\
 
\! \le {L_1}'||{\lambda _l^t}  \! - \! {\lambda _l^{t-1}}{\rm{||}}.
\end{array}
\end{equation}

Following from Eq. (\ref{eq:A22}) and the strong  concavity of ${\widetilde{L}}_p(\{ \boldsymbol{x}_i\} ,\!\{ \boldsymbol{y}_i\} ,\!{\boldsymbol{v} },\!{\boldsymbol{z}  },\!\{ \lambda _l  \} ,\!\{ \boldsymbol{\theta }_i \} ) $ \textit{w.r.t} ${\lambda _l}$ \citep{nesterov2003introductory,xu2020unified}, we can obtain that,
{\begin{equation}
\label{eq:A23}
\begin{array}{l}
\sum\limits_{l = 1}^{|{\boldsymbol{\mathcal{P}}^{t}}|} {\left\langle {\nabla _{{\lambda _l}}}{{\widetilde{L}}_p}(\{ \boldsymbol{x}_i^{t}\} ,\!\{ \boldsymbol{y}_i^{t}\} ,\!{\boldsymbol{v}^{t}  },\!{\boldsymbol{z}^{t}  },\!\{ \lambda _l^{t}  \} ,\!\{ \boldsymbol{\theta }_i^{t}  \} )  \! -\! {\nabla _{{\lambda _l}}}{{\widetilde{L}}_p}(\{ \boldsymbol{x}_i^{t}\} ,\!\{ \boldsymbol{y}_i^{t}\} ,\!{\boldsymbol{v}^{t}  },\!{\boldsymbol{z}^{t}  },\!\{ \lambda _l^{t-1}  \} ,\!\{ \boldsymbol{\theta }_i^{t-1}  \} ) \right\rangle } \\

\! \le\! \sum\limits_{l = 1}^{|{\boldsymbol{\mathcal{P}}^{t}}|} \! (  - \frac{1}{{{L_1}' + {c_1^{t-1}}}}||{\nabla _{{\lambda _l}}}{{\widetilde{L}}_p}(\{ \boldsymbol{x}_i^{t}\} ,\!\{ \boldsymbol{y}_i^{t}\} ,\!{\boldsymbol{v}^{t}  },\!{\boldsymbol{z}^{t}  },\!\{ \lambda _l^{t}  \} ,\!\{ \boldsymbol{\theta }_i^{t}  \} )  \! -\! {\nabla _{{\lambda _l}}}{{\widetilde{L}}_p}(\{ \boldsymbol{x}_i^{t}\} ,\!\{ \boldsymbol{y}_i^{t}\} ,\!{\boldsymbol{v}^{t}  },\!{\boldsymbol{z}^{t}  },\!\{ \lambda _l^{t-1}  \} ,\!\{ \boldsymbol{\theta }_i^{t-1}  \} )|{|^2} \vspace{0.5ex}\\

\qquad \; \; - \frac{{{c_1^{t-1}}{L_1}'}}{{{L_1}' + {c_1^{t-1}}}}||{\lambda _l^t} \! -\! {\lambda _l^{t-1}}|{|^2}).
\end{array}
\end{equation}}

In addition, the following inequality can be obtained,
{\begin{equation}
\label{eq:A24}
\begin{array}{l}
\frac{1}{{{{\eta _{\lambda}}}}}\left\langle {{\lambda _l^t} - {\lambda _l^{t-1}},{\lambda _l^{t + 1}} - {\lambda _l^t}} \right\rangle 
 \le \frac{1}{{2{{\eta _{\lambda}}}}}||{\lambda _l^{t + 1}} - {\lambda _l^t}|{|^2} - \frac{1}{{2{{\eta _{\lambda}}}}}||{{\boldsymbol{v}}_{1,l}^{t + 1}}|{|^2} + \frac{1}{{2{{\eta _{\lambda}}}}}||{\lambda _l^t} - {\lambda _l^{t-1}}|{|^2}.
\end{array}
\end{equation}}

Combining Eq. (\ref{eq:A14}), (\ref{eq:A15}), (\ref{eq:A18}), (\ref{eq:A19}), (\ref{eq:A23}), (\ref{eq:A24}), $ \frac{{{{\eta _{\lambda}}}}}{2} \le \frac{1}{{{L_1}' + c_1^0}} $, and setting  ${a_2} = {{\eta _{\lambda}}}$, we have:
{\begin{equation}
\label{eq:A25}
\begin{array}{l}
{{L}_p}(\{ \boldsymbol{x}_i^{t + 1}\} ,\!\{ \boldsymbol{y}_i^{t+1 }\} ,\!{\boldsymbol{v}^{t+1}  },\!{\boldsymbol{z}^{t+1}  },\!\{ \lambda _l^{t+1}  \} ,\!\{ \boldsymbol{\theta }_i^{t}  \} ) - {{L}_p}(\{ \boldsymbol{x}_i^{t + 1}\} ,\!\{ \boldsymbol{y}_i^{t+1 }\} ,\!{\boldsymbol{v}^{t+1}  },\!{\boldsymbol{z}^{t+1}  },\!\{ \lambda _l^{t}  \} ,\!\{ \boldsymbol{\theta }_i^{t}  \} ) \vspace{1ex}\\

\! \le \! \frac{{|{\boldsymbol{\mathcal{P}}^{t}}|{L^2}}}{{2{a_1}}}(\sum\limits_{i = 1}^N({||{{\boldsymbol{x}}_i^{t+1}} \!-\! {{\boldsymbol{x}}_i^t}|{|^2}} + {||{{\boldsymbol{y}}_i^{t+1}} \!-\! {{\boldsymbol{y}}_i^t}|{|^2}})  \! + \! ||{\boldsymbol{v}}^{t+1} \!-\! {\boldsymbol{v}^t}|{|^2} \! + \! ||{\boldsymbol{z}}^{t+1} \!-\! {\boldsymbol{z}^t}|{|^2})\\
 \!  +  (\frac{{{a_1}}}{2} \! - \! \frac{{{c_1^{t-1}}  -  {c_1^t}}}{2} \!  + \! \frac{1}{{2{{\eta _{\lambda}}}}})\sum\limits_{l = 1}^{|{\boldsymbol{\mathcal{P}}^{t}}|} {||{\lambda _l^{t+1}} \! - \! {\lambda _l^t}|{|^2}}  \!  + \! \frac{{{c_1^{t-1}}}}{2}\sum\limits_{l = 1}^{|{\boldsymbol{\mathcal{P}}^{t}}|} {(||{\lambda _l^{t+1}}|{|^2} \! - \! ||{\lambda _l^t}|{|^2})}  + \frac{1}{{2{{\eta _{\lambda}}}}}\sum\limits_{l = 1}^{|{\boldsymbol{\mathcal{P}}^{t}}|} {||{\lambda _l^t} \! - \! {\lambda _l^{t-1}}|{|^2}}.
\end{array}
\end{equation}}

According to Eq. (\ref{eq:new_30}), in $(t+1)^{\rm{th}}$ iteration, $\forall {\boldsymbol{\theta}} \in {{\boldsymbol{\Theta}}}$, it follows that,
{\begin{equation}
\label{eq:A26}
\left\langle {{{\boldsymbol{\theta}}_i^{t+1}} \! - \! {{\boldsymbol{\theta}}_i^t} - {{\eta _{\boldsymbol{\theta }}}}{\nabla _{{{\boldsymbol{\theta}}_i}}}{{\widetilde{L}}_p}(\{ \boldsymbol{x}_i^{t+1}\} ,\!\{ \boldsymbol{y}_i^{t+1}\} ,\!{\boldsymbol{v}^{t+1}  },\!{\boldsymbol{z}^{t+1}  },\!\{ \lambda _l^{t+1}  \} ,\!\{ \boldsymbol{\theta }_i^{t}  \} ),{\boldsymbol{\theta}} \! - \! {{\boldsymbol{\theta}}_i^{t+1}}} \right\rangle = 0.
\end{equation}}

Choosing ${\boldsymbol{\theta}} = {{\boldsymbol{\theta}}_i^t}$, we can obtain,
{\begin{equation}
\label{eq:A27}
\left\langle {{\nabla _{{{\boldsymbol{\theta}}_i}}}{{\widetilde{L}}_p}(\{ \boldsymbol{x}_i^{t+1}\} ,\!\{ \boldsymbol{y}_i^{t+1}\} ,\!{\boldsymbol{v}^{t+1}  },\!{\boldsymbol{z}^{t+1}  },\!\{ \lambda _l^{t+1}  \} ,\!\{ \boldsymbol{\theta }_i^{t}  \} ) \!-\! \frac{1}{{{{\eta _{\boldsymbol{\theta }}}}}}({{\boldsymbol{\theta}}_i^{t+1}}\! -\! {{\boldsymbol{\theta}}_i^t}),{{\boldsymbol{\theta}}_i^t}\! -\! {{\boldsymbol{\theta}}_i^{t+1}}} \right\rangle  = 0.
\end{equation}}

Likewise, in $t^{\rm{th}}$ iteration, we have,
{\begin{equation}
\label{eq:A28}
\left\langle {{\nabla _{{{\boldsymbol{\theta}}_i}}}{{\widetilde{L}}_p}(\{ \boldsymbol{x}_i^{t}\} ,\!\{ \boldsymbol{y}_i^{t}\} ,\!{\boldsymbol{v}^{t}  },\!{\boldsymbol{z}^{t}  },\!\{ \lambda _l^{t}  \} ,\!\{ \boldsymbol{\theta }_i^{t-1}  \} ) \!-\! \frac{1}{{{{\eta _{\boldsymbol{\theta }}}}}}({{\boldsymbol{\theta}}_i^t}\! -\! {{\boldsymbol{\theta}}_i^{t-1}}),{{\boldsymbol{\theta}}_i^{t+1}} \!-\! {{\boldsymbol{\theta}}_i^t}} \right\rangle  = 0.
\end{equation}}

Since ${{\widetilde{L}}_p}(\{ \boldsymbol{x}_i\} ,\!\{ \boldsymbol{y}_i\} ,\!{\boldsymbol{v}  },\!{\boldsymbol{z} },\!\{ \lambda _l  \} ,\!\{ \boldsymbol{\theta }_i \} )$ is concave with respect to ${{\boldsymbol{\theta}}_i}$ and follows from Eq. (\ref{eq:A28}):
{\begin{equation}
\label{eq:A29}
\begin{array}{l}
{{\widetilde{L}}_p}(\{ \boldsymbol{x}_i^{t+1}\} ,\!\{ \boldsymbol{y}_i^{t+1}\} ,\!{\boldsymbol{v}^{t+1}  },\!{\boldsymbol{z}^{t+1}  },\!\{ \lambda _l^{t+1}  \} ,\!\{ \boldsymbol{\theta }_i^{t+1}  \} ) \! - \! {{\widetilde{L}}_p}(\{ \boldsymbol{x}_i^{t+1}\} ,\!\{ \boldsymbol{y}_i^{t+1}\} ,\!{\boldsymbol{v}^{t+1}  },\!{\boldsymbol{z}^{t+1}  },\!\{ \lambda _l^{t+1}  \} ,\!\{ \boldsymbol{\theta }_i^{t}  \} ) \vspace{0.5ex}\\

\! \le \! \sum\limits_{i = 1}^N \! {\left\langle \! {{\nabla  _{{{\boldsymbol{\theta}}_i}}}{{\widetilde{L}}_p}(\{ \boldsymbol{x}_i^{t+1}\} ,\!\{ \boldsymbol{y}_i^{t+1}\} ,\!{\boldsymbol{v}^{t+1}  },\!{\boldsymbol{z}^{t+1}  },\!\{ \lambda _l^{t+1}  \} ,\!\{ \boldsymbol{\theta }_i^{t} \} ) , {{\boldsymbol{\theta}}_i^{t+1}} \! - \! {{\boldsymbol{\theta}}_i^t}}\! \right\rangle } \vspace{0.5ex} \\
 
\! \le \! \sum\limits_{i = 1}^N \! {( \left\langle \! {{\nabla \! _{{{\boldsymbol{\theta}}_i}}}{{\widetilde{L}}_p}(\{ \boldsymbol{x}_i^{t+1}\} ,\!\{ \boldsymbol{y}_i^{t+1}\} ,\!{\boldsymbol{v}^{t+1}  },\!{\boldsymbol{z}^{t+1}  },\!\{ \lambda _l^{t+1}  \} ,\!\{ \boldsymbol{\theta }_i^{t}  \} ) \! - \! {\nabla \! _{{{\boldsymbol{\theta}}_i}}}{{\widetilde{L}}_p}(\{ \boldsymbol{x}_i^{t}\} ,\!\{ \boldsymbol{y}_i^{t}\} ,\!{\boldsymbol{v}^{t}  },\!{\boldsymbol{z}^{t}  },\!\{ \lambda _l^{t}  \} ,\!\{ \boldsymbol{\theta }_i^{t-1}  \} ),  {{\boldsymbol{\theta}}_i^{t+1}} \! - \! {{\boldsymbol{\theta}}_i^t}}\! \right\rangle } \vspace{1ex}\\
\qquad \;\; +  \frac{1}{{{{\eta _{\boldsymbol{\theta }}}}}}\left\langle {{{\boldsymbol{\theta}}_i^t} \! - \! {{\boldsymbol{\theta}}_i^{t-1}},{{\boldsymbol{\theta}}_i^{t+1}} \! - \! {{\boldsymbol{\theta}}_i^t}}  \right\rangle ).
\end{array}
\end{equation}}

Denoting ${{\boldsymbol{v}}_{2,l}^{t + 1}} = {{\boldsymbol{\theta}}_i^{t+1}} - {{\boldsymbol{\theta}}_i^t} - ({{\boldsymbol{\theta}}_i^t} - {{\boldsymbol{\theta}}_i^{t-1}})$, we have that,
{\begin{equation}
\label{eq:A30}
\begin{array}{l}
\sum\limits_{i = 1}^N \! {\left\langle \! {{\nabla _{{{\boldsymbol{\theta}}_i}}}{{\widetilde{L}}_p}(\{ \boldsymbol{x}_i^{t+1}\} ,\!\{ \boldsymbol{y}_i^{t+1}\} ,\!{\boldsymbol{v}^{t+1}  },\!{\boldsymbol{z}^{t+1}  },\!\{ \lambda _l^{t+1}  \} ,\!\{ \boldsymbol{\theta }_i^{t}  \} ) \! - \! {\nabla _{{{\boldsymbol{\theta}}_i}}}{{\widetilde{L}}_p}(\{ \boldsymbol{x}_i^{t}\} ,\!\{ \boldsymbol{y}_i^{t}\} ,\!{\boldsymbol{v}^{t}  },\!{\boldsymbol{z}^{t}  },\!\{ \lambda _l^{t}  \} ,\!\{ \boldsymbol{\theta }_i^{t-1}  \} ),{{\boldsymbol{\theta}}_i^{t+1}} \! - \! {{\boldsymbol{\theta}}_i^t}} \! \right\rangle } \vspace{0.5ex}\\

 \!= \! \sum\limits_{i = 1}^N \! {\left\langle \! {{\nabla\! _{{{\boldsymbol{\theta}}_i}}}{{\widetilde{L}}_p}(\{ \boldsymbol{x}_i^{t+1}\} ,\!\{ \boldsymbol{y}_i^{t+1}\} ,\!{\boldsymbol{v}^{t+1}  },\!{\boldsymbol{z}^{t+1}  },\!\{ \lambda _l^{t+1}  \} ,\!\{ \boldsymbol{\theta }_i^{t}  \} ) \! - \! {\nabla \! _{{{\boldsymbol{\theta}}_i}}}{{\widetilde{L}}_p}(\{ \boldsymbol{x}_i^{t}\} ,\!\{ \boldsymbol{y}_i^{t}\} ,\!{\boldsymbol{v}^{t}  },\!{\boldsymbol{z}^{t}  },\!\{ \lambda _l^{t}  \} ,\!\{ \boldsymbol{\theta }_i^{t}  \} ),{{\boldsymbol{\theta}}_i^{t+1}} \! - \! {{\boldsymbol{\theta}}_i^t}} \! \right\rangle } (2a) \vspace{0.5ex}\\
 
 \! + \! \sum\limits_{i = 1}^N \! {\left\langle\! {{\nabla _{{{\boldsymbol{\theta}}_i}}}{{\widetilde{L}}_p}(\{ \boldsymbol{x}_i^{t}\} ,\!\{ \boldsymbol{y}_i^{t}\} ,\!{\boldsymbol{v}^{t}  },\!{\boldsymbol{z}^{t}  },\!\{ \lambda _l^{t}  \} ,\!\{ \boldsymbol{\theta }_i^{t}  \} ) \! - \! {\nabla _{{{\boldsymbol{\theta}}_i}}}{{\widetilde{L}}_p}(\{ \boldsymbol{x}_i^{t}\} ,\!\{ \boldsymbol{y}_i^{t}\} ,\!{\boldsymbol{v}^{t}  },\!{\boldsymbol{z}^{t}  },\!\{ \lambda _l^{t}  \} ,\!\{ \boldsymbol{\theta }_i^{t-1}  \} ),{{\boldsymbol{v}}_{2,l}^{t + 1}}} \!\right\rangle } (2b) \vspace{0.5ex}\\
 
 \! + \! \sum\limits_{i = 1}^N {\left\langle\! {{\nabla _{{{\boldsymbol{\theta}}_i}}}{{\widetilde{L}}_p}(\{ \boldsymbol{x}_i^{t}\} ,\!\{ \boldsymbol{y}_i^{t}\} ,\!{\boldsymbol{v}^{t}  },\!{\boldsymbol{z}^{t}  },\!\{ \lambda _l^{t}  \} ,\!\{ \boldsymbol{\theta }_i^{t}  \} ) \! - \! {\nabla _{{{\boldsymbol{\theta}}_i}}}{{\widetilde{L}}_p}(\{ \boldsymbol{x}_i^{t}\} ,\!\{ \boldsymbol{y}_i^{t}\} ,\!{\boldsymbol{v}^{t}  },\!{\boldsymbol{z}^{t}  },\!\{ \lambda _l^{t}  \} ,\!\{ \boldsymbol{\theta }_i^{t-1}  \} ),{{\boldsymbol{\theta}}_i^t} \! - \! {{\boldsymbol{\theta}}_i^{t-1}}}\! \right\rangle } (2c).
\end{array}
\end{equation}}

We firstly focus on the ($2a$) in Eq. (\ref{eq:A30}), we can write the ($2a$) as,
\begin{equation}
\label{eq:A31}
\begin{array}{l}
\left\langle\! {{\nabla _{{{\boldsymbol{\theta}}_i}}}{{\widetilde{L}}_p}(\{ \boldsymbol{x}_i^{t+1}\} ,\!\{ \boldsymbol{y}_i^{t+1}\} ,\!{\boldsymbol{v}^{t+1}  },\!{\boldsymbol{z}^{t+1}  },\!\{ \lambda _l^{t+1}  \} ,\!\{ \boldsymbol{\theta }_i^{t}  \} )\! - \!{\nabla _{{{\boldsymbol{\theta}}_i}}}{{\widetilde{L}}_p}(\{ \boldsymbol{x}_i^{t}\} ,\!\{ \boldsymbol{y}_i^{t}\} ,\!{\boldsymbol{v}^{t}  },\!{\boldsymbol{z}^{t}  },\!\{ \lambda _l^{t}  \} ,\!\{ \boldsymbol{\theta }_i^{t}  \} ),{{\boldsymbol{\theta}}_i^{t+1}}\! -\! {{\boldsymbol{\theta}}_i^t}}\! \right\rangle \vspace{1ex} \\

\! = \! \left\langle  {{\nabla _{{{\boldsymbol{\theta}}_i}}}{{{L}}_p}(\{ \boldsymbol{x}_i^{t+1}\} ,\!\{ \boldsymbol{y}_i^{t+1}\} ,\!{\boldsymbol{v}^{t+1}  },\!{\boldsymbol{z}^{t+1}  },\!\{ \lambda _l^{t+1}  \} ,\!\{ \boldsymbol{\theta }_i^{t}  \} ) \!-\! {\nabla _{{{\boldsymbol{\theta}}_i}}}{{{L}}_p}(\{ \boldsymbol{x}_i^{t}\} ,\!\{ \boldsymbol{y}_i^{t}\} ,\!{\boldsymbol{v}^{t}  },\!{\boldsymbol{z}^{t}  },\!\{ \lambda _l^{t}  \} ,\!\{ \boldsymbol{\theta }_i^{t}  \} ),{{\boldsymbol{\theta}}_i^{t+1}}\! -\! {{\boldsymbol{\theta}}_i^t}} \right\rangle \vspace{1ex}
 \vspace{1ex}\\
   +  \frac{{{c_2^{t-1}}  -  {c_2^t}}}{2}(||{{\boldsymbol{\theta}}_i^{t+1}}|{|^2} \! - \! ||{{\boldsymbol{\theta}}_i^t}|{|^2}) \! - \! \frac{{{c_2^{t-1}}  -  {c_2^t}}}{2}||{{\boldsymbol{\theta}}_i^{t+1}} \! - \! {{\boldsymbol{\theta}}_i^t}|{|^2}).
\end{array}
\end{equation}

And combining the Cauchy-Schwarz inequality with Assumption 1, we can obtain,
\begin{equation}
\label{eq:A32}
\begin{array}{l}
\left\langle  {{\nabla _{{{\boldsymbol{\theta}}_i}}}{{{L}}_p}(\{ \boldsymbol{x}_i^{t+1}\} ,\!\{ \boldsymbol{y}_i^{t+1}\} ,\!{\boldsymbol{v}^{t+1}  },\!{\boldsymbol{z}^{t+1}  },\!\{ \lambda _l^{t+1}  \} ,\!\{ \boldsymbol{\theta }_i^{t}  \} ) \!-\! {\nabla _{{{\boldsymbol{\theta}}_i}}}{{{L}}_p}(\{ \boldsymbol{x}_i^{t}\} ,\!\{ \boldsymbol{y}_i^{t}\} ,\!{\boldsymbol{v}^{t}  },\!{\boldsymbol{z}^{t}  },\!\{ \lambda _l^{t}  \} ,\!\{ \boldsymbol{\theta }_i^{t}  \} ),{{\boldsymbol{\theta}}_i^{t+1}}\! -\! {{\boldsymbol{\theta}}_i^t}} \right\rangle  \vspace{1.5ex}\\

\! =\!  \left\langle  {{\nabla _{{{\boldsymbol{\theta}}_i}}}{{{L}}_p}(\{ \boldsymbol{x}_i^{t+1}\} ,\!\{ \boldsymbol{y}_i^{t+1}\} ,\!{\boldsymbol{v}^{t+1}  },\!{\boldsymbol{z}^{t+1}  },\!\{ \lambda _l^{t}  \} ,\!\{ \boldsymbol{\theta }_i^{t}  \} ) \!-\! {\nabla _{{{\boldsymbol{\theta}}_i}}}{{{L}}_p}(\{ \boldsymbol{x}_i^{t}\} ,\!\{ \boldsymbol{y}_i^{t}\} ,\!{\boldsymbol{v}^{t}  },\!{\boldsymbol{z}^{t}  },\!\{ \lambda _l^{t}  \} ,\!\{ \boldsymbol{\theta }_i^{t}  \} ),{{\boldsymbol{\theta}}_i^{t+1}}\! -\! {{\boldsymbol{\theta}}_i^t}} \right\rangle  \vspace{0.5ex}\\

\! \le \! \frac{{{L^2}}}{{2{a_3}}}(\sum\limits_{i = 1}^N({||{{\boldsymbol{x}}_i^{t+1}} \!-\! {{\boldsymbol{x}}_i^t}|{|^2}} + {||{{\boldsymbol{y}}_i^{t+1}} \!-\! {{\boldsymbol{y}}_i^t}|{|^2}})  \! + \! ||{\boldsymbol{v}}^{t+1} \!-\! {\boldsymbol{v}^t}|{|^2} \! + \! ||{\boldsymbol{z}}^{t+1} \!-\! {\boldsymbol{z}^t}|{|^2}) + \frac{{{a_3}}}{2}||{{\boldsymbol{\theta}}_i^{t+1}}\! -\! {{\boldsymbol{\theta}}_i^t}|{|^2},
\end{array}
\end{equation}
where $a_3 >0$ is a constant. Thus, we can get the upper bound of ($2a$) by combining  Eq. (\ref{eq:A31}) with Eq. (\ref{eq:A32}), that is,
\begin{equation}
\label{eq:A33}
\begin{array}{l}
\sum\limits_{i = 1}^N \! \left\langle\! {{\nabla _{{{\boldsymbol{\theta}}_i}}}{{\widetilde{L}}_p}(\{ \boldsymbol{x}_i^{t+1}\} ,\!\{ \boldsymbol{y}_i^{t+1}\} ,\!{\boldsymbol{v}^{t+1}  },\!{\boldsymbol{z}^{t+1}  },\!\{ \lambda _l^{t+1}  \} ,\!\{ \boldsymbol{\theta }_i^{t}  \} )\! - \!{\nabla _{{{\boldsymbol{\theta}}_i}}}{{\widetilde{L}}_p}(\{ \boldsymbol{x}_i^{t}\} ,\!\{ \boldsymbol{y}_i^{t}\} ,\!{\boldsymbol{v}^{t}  },\!{\boldsymbol{z}^{t}  },\!\{ \lambda _l^{t}  \} ,\!\{ \boldsymbol{\theta }_i^{t}  \} ),{{\boldsymbol{\theta}}_i^{t+1}}\! -\! {{\boldsymbol{\theta}}_i^t}}\! \right\rangle \vspace{0.2ex}\\

\! \le \! \sum\limits_{i \in {\boldsymbol{\mathcal{Q}}^{t + 1}}}^{} \! {( \frac{{{L^2}}}{{2{a_3}}}(\sum\limits_{i = 1}^N({||{{\boldsymbol{x}}_i^{t+1}} \!-\! {{\boldsymbol{x}}_i^t}|{|^2}} + {||{{\boldsymbol{y}}_i^{t+1}} \!-\! {{\boldsymbol{y}}_i^t}|{|^2}})  \! + \! ||{\boldsymbol{v}}^{t+1} \!-\! {\boldsymbol{v}^t}|{|^2} \! + \! ||{\boldsymbol{z}}^{t+1} \!-\! {\boldsymbol{z}^t}|{|^2}) \! + \! \frac{{{a_3}}}{2}||{{\boldsymbol{\theta}}_i^{t+1}} \! - \! {{\boldsymbol{\theta}}_i^t}|{|^2}} \vspace{1ex}\\

 \qquad \qquad +  \frac{{{c_2^{t-1}}  -  {c_2^t}}}{2}(||{{\boldsymbol{\theta}}_i^{t+1}}|{|^2} \! - \! ||{{\boldsymbol{\theta}}_i^t}|{|^2}) \! - \! \frac{{{c_2^{t-1}}  -  {c_2^t}}}{2}||{{\boldsymbol{\theta}}_i^{t+1}} \! - \! {{\boldsymbol{\theta}}_i^t}|{|^2}).
\end{array}
\end{equation}

Next we focus on the ($2b$) in Eq. (\ref{eq:A30}). According to Cauchy-Schwarz inequality we can write ($2b$) as,
\begin{equation}
\label{eq:A34}
\begin{array}{l}
\sum\limits_{i = 1}^N \left\langle\! {{\nabla _{{{\boldsymbol{\theta}}_i}}}{{\widetilde{L}}_p}(\{ \boldsymbol{x}_i^{t}\} ,\!\{ \boldsymbol{y}_i^{t}\} ,\!{\boldsymbol{v}^{t}  },\!{\boldsymbol{z}^{t}  },\!\{ \lambda _l^{t}  \} ,\!\{ \boldsymbol{\theta }_i^{t}  \} ) \! - \! {\nabla _{{{\boldsymbol{\theta}}_i}}}{{\widetilde{L}}_p}(\{ \boldsymbol{x}_i^{t}\} ,\!\{ \boldsymbol{y}_i^{t}\} ,\!{\boldsymbol{v}^{t}  },\!{\boldsymbol{z}^{t}  },\!\{ \lambda _l^{t}  \} ,\!\{ \boldsymbol{\theta }_i^{t-1}  \} ),{{\boldsymbol{v}}_{2,l}^{t + 1}}} \!\right\rangle \\

\! \le \! \sum\limits_{i = 1}^N  (\frac{{{a_4}}}{2}||{\nabla _{{{\boldsymbol{\theta}}_i}}}{{\widetilde{L}}_p}(\{ \boldsymbol{x}_i^{t}\} ,\!\{ \boldsymbol{y}_i^{t}\} ,\!{\boldsymbol{v}^{t}  },\!{\boldsymbol{z}^{t}  },\!\{ \lambda _l^{t}  \} ,\!\{ \boldsymbol{\theta }_i^{t}  \} ) \! - \! {\nabla _{{{\boldsymbol{\theta}}_i}}}{{\widetilde{L}}_p}(\{ \boldsymbol{x}_i^{t}\} ,\!\{ \boldsymbol{y}_i^{t}\} ,\!{\boldsymbol{v}^{t}  },\!{\boldsymbol{z}^{t}  },\!\{ \lambda _l^{t}  \} ,\!\{ \boldsymbol{\theta }_i^{t-1}  \} )|{|^2}\\

\qquad \; + \frac{1}{{2{a_4}}}||{\boldsymbol{v}}_{2,l}^{t+1}|{|^2}),
\end{array}
\end{equation}
where $a_4>0$ is a constant. Then, we focus on the ($2c$) in Eq. (\ref{eq:A30}). Defining ${L_2}' = L  +  {c_2^0}$, according to Assumption 1 and the trigonometric inequality, we have,
\begin{equation}
\label{eq:A36}
\begin{array}{l}
||{\nabla _{{{\boldsymbol{\theta}}_i}}}{{\widetilde{L}}_p}(\{ \boldsymbol{x}_i^{t}\} ,\!\{ \boldsymbol{y}_i^{t}\} ,\!{\boldsymbol{v}^{t}  },\!{\boldsymbol{z}^{t}  },\!\{ \lambda _l^{t}  \} ,\!\{ \boldsymbol{\theta }_i^{t}  \} ) \! - \! {\nabla _{{{\boldsymbol{\theta}}_i}}}{{\widetilde{L}}_p}(\{ \boldsymbol{x}_i^{t}\} ,\!\{ \boldsymbol{y}_i^{t}\} ,\!{\boldsymbol{v}^{t}  },\!{\boldsymbol{z}^{t}  },\!\{ \lambda _l^{t}  \} ,\!\{ \boldsymbol{\theta }_i^{t-1}  \} )|| \vspace{1.5mm} \\ 

 \le {L_2}'||{{\boldsymbol{\theta}}_i^t} \!-\! {{\boldsymbol{\theta}}_i^{t-1}}{\rm{||}}.
\end{array}
\end{equation}

Following Eq. (\ref{eq:A36}) and the strong  concavity of ${{\widetilde{L}}_p}(\{ \boldsymbol{x}_i\} ,\!\{ \boldsymbol{y}_i\} ,\!{\boldsymbol{v} },\!{\boldsymbol{z}  },\!\{ \lambda _l  \} ,\!\{ \boldsymbol{\theta }_i \} )$ \textit{w.r.t} ${\boldsymbol{\theta}}_i$, the upper bound of ($2c$) can be obtained, that is,
\begin{equation}
\label{eq:A37}
\begin{array}{l}
\sum\limits_{i = 1}^N {\left\langle\! {{\nabla _{{{\boldsymbol{\theta}}_i}}}{{\widetilde{L}}_p}(\{ \boldsymbol{x}_i^{t}\} ,\!\{ \boldsymbol{y}_i^{t}\} ,\!{\boldsymbol{v}^{t}  },\!{\boldsymbol{z}^{t}  },\!\{ \lambda _l^{t}  \} ,\!\{ \boldsymbol{\theta }_i^{t}  \} ) \! - \! {\nabla _{{{\boldsymbol{\theta}}_i}}}{{\widetilde{L}}_p}(\{ \boldsymbol{x}_i^{t}\} ,\!\{ \boldsymbol{y}_i^{t}\} ,\!{\boldsymbol{v}^{t}  },\!{\boldsymbol{z}^{t}  },\!\{ \lambda _l^{t}  \} ,\!\{ \boldsymbol{\theta }_i^{t-1}  \} ),{{\boldsymbol{\theta}}_i^t} \! - \! {{\boldsymbol{\theta}}_i^{t-1}}}\! \right\rangle } \vspace{0.5ex}\\

\! \le \! \sum\limits_{i = 1}^N \! {( - \frac{1}{{{L_2}' + {c_2^{t-1}}}}||{\nabla _{{{\boldsymbol{\theta}}_i}}}{{\widetilde{L}}_p}(\{ \boldsymbol{x}_i^{t}\} ,\!\{ \boldsymbol{y}_i^{t}\} ,\!{\boldsymbol{v}^{t}  },\!{\boldsymbol{z}^{t}  },\!\{ \lambda _l^{t}  \} ,\!\{ \boldsymbol{\theta }_i^{t}  \} ) \! - \! {\nabla _{{{\boldsymbol{\theta}}_i}}}{{\widetilde{L}}_p}(\{ \boldsymbol{x}_i^{t}\} ,\!\{ \boldsymbol{y}_i^{t}\} ,\!{\boldsymbol{v}^{t}  },\!{\boldsymbol{z}^{t}  },\!\{ \lambda _l^{t}  \} ,\!\{ \boldsymbol{\theta }_i^{t-1}  \} )|{|^2}} \vspace{1ex}\\

 \qquad \;\; - \frac{{{c_2^{t-1}}{L_2}'}}{{{L_2}' + {c_2^{t-1}}}}||{{\boldsymbol{\theta}}_i^t} \!-\! {{\boldsymbol{\theta}}_i^{t-1}}|{|^2}).
\end{array}
\end{equation}

In addition, the following inequality can also be obtained,
\begin{equation}
\label{eq:A38}
\begin{array}{l}
\sum\limits_{i = 1}^N {\frac{1}{{{{\eta _{\boldsymbol{\theta }}}}}}\left\langle {{{\boldsymbol{\theta}}_i^t} \!-\! {{\boldsymbol{\theta}}_i^{t-1}},{{\boldsymbol{\theta}}_i^{t+1}} \!-\! {{\boldsymbol{\theta}}_i^t}} \right\rangle } 
 \le \sum\limits_{i = 1}^N {(\frac{1}{{2{{\eta _{\boldsymbol{\theta }}}}}}||{{\boldsymbol{\theta}}_i^{t+1}} \!-\! {{\boldsymbol{\theta}}_i^t}|{|^2} - \frac{1}{{2{{\eta _{\boldsymbol{\theta }}}}}}||{{\boldsymbol{v}}_{2,l}^{t+1}}|{|^2} + \frac{1}{{2{{\eta _{\boldsymbol{\theta }}}}}}||{{\boldsymbol{\theta}}_i^t} - {{\boldsymbol{\theta}}_i^{t-1}}|{|^2})}.
\end{array}
\end{equation}

Combining Eq. (\ref{eq:A29}), (\ref{eq:A30}), (\ref{eq:A33}), (\ref{eq:A34}), (\ref{eq:A37}), (\ref{eq:A38}),  $ \frac{{{{\eta _{\boldsymbol{\theta }}}}}}{2} \le \frac{1}{{{L_2}' + c_2^0}}$, and setting ${a_4} = {{\eta _{\boldsymbol{\theta }}}}$, we have,
{\begin{equation}
\label{eq:A39}
\begin{array}{l}
{{{L}}_p}(\{ \boldsymbol{x}_i^{t+1}\} ,\!\{ \boldsymbol{y}_i^{t+1}\} ,\!{\boldsymbol{v}^{t+1}  },\!{\boldsymbol{z}^{t+1}  },\!\{ \lambda _l^{t+1}  \} ,\!\{ \boldsymbol{\theta }_i^{t+1}  \} ) \! - \! {{{L}}_p}(\{ \boldsymbol{x}_i^{t+1}\} ,\!\{ \boldsymbol{y}_i^{t+1}\} ,\!{\boldsymbol{v}^{t+1}  },\!{\boldsymbol{z}^{t+1}  },\!\{ \lambda _l^{t+1}  \} ,\!\{ \boldsymbol{\theta }_i^{t}  \} ) \vspace{1ex}\\

\! \le \! \frac{{|{\boldsymbol{\mathcal{Q}}^{t + 1}}|{L^2}}}{{2{a_3}}}(\sum\limits_{i = 1}^N({||{{\boldsymbol{x}}_i^{t+1}} \!-\! {{\boldsymbol{x}}_i^t}|{|^2}} + {||{{\boldsymbol{y}}_i^{t+1}} \!-\! {{\boldsymbol{y}}_i^t}|{|^2}})  \! + \! ||{\boldsymbol{v}}^{t+1} \!-\! {\boldsymbol{v}^t}|{|^2} \! + \! ||{\boldsymbol{z}}^{t+1} \!-\! {\boldsymbol{z}^t}|{|^2})\vspace{0.5ex}\\

 \! +  (\frac{{{a_3}}}{2} \! - \! \frac{{{c_2^{t-1}}  -  {c_2^t}}}{2} \! + \! \frac{1}{{2{{\eta _{\boldsymbol{\theta }}}}}})\! \sum\limits_{i = 1}^N\! {||{{\boldsymbol{\theta}}_i^{t+1}} \! - \! {{\boldsymbol{\theta}}_i^t}|{|^2}}  \! + \! \frac{{{c_2^{t-1}}}}{2}\! \sum\limits_{i = 1}^N\! {(||{{\boldsymbol{\theta}}_i^{t+1}}|{|^2} \! - \! ||{{\boldsymbol{\theta}}_i^t}|{|^2})} 
 \! + \! \frac{1}{{2{{\eta _{\boldsymbol{\theta }}}}}}\! \sum\limits_{i = 1}^N \! {||{{\boldsymbol{\theta}}_i^t} \! - \! {{\boldsymbol{\theta}}_i^{t-1}}|{|^2}}. 
\end{array}
\end{equation}}

By combining Lemma 1 with Eq. (\ref{eq:A25}) and Eq. (\ref{eq:A39}), we conclude the proof of Lemma \ref{lemma 2}.

\begin{lemma} \label{lemma3}
Firstly,  we denote ${S_1^{t+1}}$, ${S_2^{t+1}}$ and ${F^{t+1}}$  as,
\begin{equation}
\label{eq:A40}
{S_1^{t+1}} = \frac{4}{{{{\eta _{\lambda}}}^2{c_1^{t+1}}}}\sum\limits_{l = 1}^{|{\boldsymbol{\mathcal{P}}^{t}}|} {||{\lambda _l^{t+1}} \! - \! {\lambda _l^t}|{|^2}}  \! - \! \frac{4}{{{{\eta _{\lambda}}}}}(\frac{{{c_1^{t-1}}}}{{{c_1^t}}} \! - \! 1)\sum\limits_{l = 1}^{|{\boldsymbol{\mathcal{P}}^{t}}|} {||{\lambda _l^{t+1}}|{|^2}},
\end{equation}
\begin{equation}
\label{eq:A41}
{S_2^{t+1}} = \frac{4}{{{{\eta _{\boldsymbol{\theta }}}}^2{c_2^{t+1}}}}\sum\limits_{i = 1}^N {||{{\boldsymbol{\theta}}_i^{t+1}} \! - \! {{\boldsymbol{\theta}}_i^t}|{|^2}}  \! - \! \frac{4}{{{{\eta _{\boldsymbol{\theta }}}}}}(\frac{{{c_2^{t-1}}}}{{{c_2^t}}} \! - \! 1)\sum\limits_{i = 1}^N {||{{\boldsymbol{\theta}}_i^{t+1}}|{|^2}}, 
\end{equation}
\begin{equation}
\label{eq:A42}
\begin{array}{l}
F^{t+1} = {L_p}{\rm{(\{ }}{{\boldsymbol{x}}_i^{t+1}}{\rm{\} }}, \{{{\boldsymbol{y}}_i^{t+1}}\},{\boldsymbol{z}}^{t+1},h^{t+1},\{ {\lambda _l^{t+1}}\} ,\{ {{\boldsymbol{\theta}}_i^{t+1}}\} {\rm{)}}  +  {S_1^{t+1}}  +  {S_2^{t+1}}\vspace{1ex}\\
 \quad  \quad   \quad \; \;    -  \frac{7}{{2{{\eta _{\lambda}}}}}\!\sum\limits_{l = 1}^{|{\boldsymbol{\mathcal{P}}^{t}}|}\! {||{\lambda _l^{t+1}} \! - \! {\lambda _l^t}|{|^2}} 
 \! - \! \frac{c_1^t}{{2}}\!\sum\limits_{l = 1}^{|{\boldsymbol{\mathcal{P}}^{t}}|}\! {||{\lambda _l^{t+1}}|{|^2}} 
 
 \! - \! \frac{7}{{2{{\eta _{\boldsymbol{\theta }}}}}}\! \sum\limits_{i = 1}^N\! {||{{\boldsymbol{\theta}}_i^{t+1}} \! - \! {{\boldsymbol{\theta}}_i^t}|{|^2}}
 
  \! - \! \frac{c_2^t}{{2}} \! \sum\limits_{i = 1}^N\! {||{{\boldsymbol{\theta}}_i^{t+1}}|{|^2}}.
\end{array}
\end{equation}
Defining $a_5 = \max\{1, 1 + L^2, 6\tau{ k_1 }N{L^2}\}$, $\forall t \ge T_1$, we have,
{\begin{equation}
\label{eq:A43}
\begin{array}{l}
F^{t+1} \! - \! F^{t}\\
 \!\le\! (\frac{{L + a_5}}{2} \! - \! \frac{1}{{{\eta _{\boldsymbol{x}}^t}}} \! +\! \frac{{{{\eta _{\lambda}}}|{\boldsymbol{\mathcal{P}}^{t}}|{L^2}}}{2} \! +\! \frac{{{{\eta _{\boldsymbol{\theta }}}}|{\boldsymbol{\mathcal{Q}}^{t + 1}}|{L^2}}}{2} \! +\! \frac{{8|{\boldsymbol{\mathcal{P}}^{t}}|{L^2}}}{{{{\eta _{\lambda}}}({c_1^t})^2}} \! +\! \frac{{8N{L^2}}}{{{{\eta _{\boldsymbol{\theta }}}}({c_2^t})^2}})\sum\limits_{i = 1}^N {||{{\boldsymbol{x}}_i^{t+1}} \! - \! {{\boldsymbol{x}}_i^t}|{|^2}} \\
 
\! + (\frac{{L + a_5}}{2} \! - \! \frac{1}{{{\eta _{\boldsymbol{y}}^t}}} \! +\! \frac{{{{\eta _{\lambda}}}|{\boldsymbol{\mathcal{P}}^{t}}|{L^2}}}{2} \! +\! \frac{{{{\eta _{\boldsymbol{\theta }}}}|{\boldsymbol{\mathcal{Q}}^{t + 1}}|{L^2}}}{2} \! +\! \frac{{8|{\boldsymbol{\mathcal{P}}^{t}}|{L^2}}}{{{{\eta _{\lambda}}}({c_1^t})^2}} \! +\! \frac{{8N{L^2}}}{{{{\eta _{\boldsymbol{\theta }}}}({c_2^t})^2}})\sum\limits_{i = 1}^N {||{{\boldsymbol{y}}_i^{t+1}} \! - \! {{\boldsymbol{y}}_i^t}|{|^2}} \\
 
 \! + (\frac{{L + a_5}}{2} \! - \! \frac{1}{{{\eta _{\boldsymbol{v}}^t}}} \! +\! \frac{{{{\eta _{\lambda}}}|{\boldsymbol{\mathcal{P}}^{t}}|{L^2}}}{2} \! +\! \frac{{{{\eta _{\boldsymbol{\theta }}}}|{\boldsymbol{\mathcal{Q}}^{t + 1}}|{L^2}}}{2} \! +\! \frac{{8|{\boldsymbol{\mathcal{P}}^{t}}|{L^2}}}{{{{\eta _{\lambda}}}({c_1^t})^2}} \! +\! \frac{{8N{L^2}}}{{{{\eta _{\boldsymbol{\theta }}}}({c_2^t})^2}})||\boldsymbol{v}^{t+1} \! - \! \boldsymbol{v}^t|{|^2}\\
 
 \! + (\frac{{L + a_5}}{2} \! - \! \frac{1}{{{\eta _{\boldsymbol{z}}^t}}} \! +\! \frac{{{{\eta _{\lambda}}}|{\boldsymbol{\mathcal{P}}^{t}}|{L^2}}}{2} \! +\! \frac{{{{\eta _{\boldsymbol{\theta }}}}|{\boldsymbol{\mathcal{Q}}^{t + 1}}|{L^2}}}{2} \! +\! \frac{{8|{\boldsymbol{\mathcal{P}}^{t}}|{L^2}}}{{{{\eta _{\lambda}}}({c_1^t})^2}} \! +\! \frac{{8N{L^2}}}{{{{\eta _{\boldsymbol{\theta }}}}({c_2^t})^2}})||\boldsymbol{z}^{t+1} \! - \! \boldsymbol{z}^t|{|^2}\\
 
 \! -  (\frac{1}{{10{{\eta _{\lambda}}}}} \! - \! \frac{{6\tau{ k_1 }N{L^2}}}{2})\sum\limits_{l = 1}^{|{\boldsymbol{\mathcal{P}}^{t}}|} {||{\lambda _l^{t+1}} \! - \! {\lambda _l^t}|{|^2}}  \! - \! \frac{1}{{10{{\eta _{\boldsymbol{\theta }}}}}}\sum\limits_{i = 1}^N {||{{\boldsymbol{\theta}}_i^{t+1}} \! - \! {{\boldsymbol{\theta}}_i^t}|{|^2}} 
 \! +\! \frac{{{c_1^{t-1}}  -  {c_1^t}}}{2}\sum\limits_{l = 1}^{|{\boldsymbol{\mathcal{P}}^{t}}|} {||{\lambda _l^{t+1}}|{|^2}} \\
 
 \! + \frac{{{c_2^{t-1}}  -  {c_2^t}}}{2}\sum\limits_{i = 1}^N {||{{\boldsymbol{\theta}}_i^{t+1}}|{|^2}} 
 \! +\! \frac{4}{{{{\eta _{\lambda}}}}}(\frac{{{c_1^{t-2}}}}{{{c_1^{t-1}}}} \! - \! \frac{{{c_1^{t-1}}}}{{{c_1^t}}})\sum\limits_{l = 1}^{|{\boldsymbol{\mathcal{P}}^{t}}|} {||{\lambda _l^t}|{|^2}}  \! +\! \frac{4}{{{{\eta _{\boldsymbol{\theta }}}}}}(\frac{{{c_2^{t-2}}}}{{{c_2^{t-1}}}} \! - \! \frac{{{c_2^{t-1}}}}{{{c_2^t}}})\sum\limits_{i = 1}^N {||{{\boldsymbol{\theta}}_i^t}|{|^2}}.
\end{array}
\end{equation}}
\end{lemma}

\emph{\textbf{Proof of Lemma \ref{lemma3}}:} 

Let ${a_1} = \frac{1}{{{{\eta _{\lambda}}}}}$, ${a_3} = \frac{1}{{{{\eta _{\boldsymbol{\theta }}}}}}$ and substitute them into the Lemma \ref{lemma 2}, $\forall t \ge T_1$, we have,
\begin{equation}
\label{eq:A44}
\begin{array}{l}
{L_p}(\{ \boldsymbol{x}_i^{t + 1}\} ,\!\{ \boldsymbol{y}_i^{t+1 }\} ,{\boldsymbol{v}^{t+1}  },{\boldsymbol{z}^{t+1}  },\!\{ \lambda _l^{t+1}  \} ,\!\{ \boldsymbol{\theta }_i^{t+1}  \} ) \!-\! {L_p}(\{ \boldsymbol{x}_i^{t}\} ,\!\{ \boldsymbol{y}_i^{t }\} ,{\boldsymbol{v}^{t}  },{\boldsymbol{z}^{t}  },\!\{ \lambda _l^{t}  \} ,\!\{ \boldsymbol{\theta }_i^{t}  \} )\vspace{1ex}\\

 \le  (\frac{{L + L^2 +  1}}{2} \!-\! \frac{1}{{{\eta _{\boldsymbol{x}}^t}}} \! + \! \frac{{\eta _{\lambda}|{\boldsymbol{\mathcal{P}}^{t}}|{L^2}+\eta _{\boldsymbol{\theta}}|{{\boldsymbol{\mathcal{Q}}}^{t \! + \! 1}}|{L^2}}}{{2}})\!\sum\limits_{i = 1}^N\! {||{{\boldsymbol{x}}_i^{t+1}} \!-\! {{\boldsymbol{x}}_i^t}|{|^2}} \\

+ (\frac{{L  +  1}}{2} \!-\! \frac{1}{{{\eta _{\boldsymbol{y}}^t}}} \! + \! \frac{{\eta _{\lambda}|{\boldsymbol{\mathcal{P}}^{t}}|{L^2}+\eta _{\boldsymbol{\theta}}|{{\boldsymbol{\mathcal{Q}}}^{t \! + \! 1}}|{L^2}}}{{2}})\!\sum\limits_{i = 1}^N \!{||{{\boldsymbol{y}}_i^{t+1}} \!-\! {{\boldsymbol{y}}_i^t}|{|^2}} \\

 +  (\frac{{L  +  6\tau{ k_1 }N{L^2}}}{2} \!-\! \frac{1}{{{\eta _{\boldsymbol{v}}^t}}} \! + \! \frac{{\eta _{\lambda}|{\boldsymbol{\mathcal{P}}^{t}}|{L^2}+\eta _{\boldsymbol{\theta}}|{{\boldsymbol{\mathcal{Q}}}^{t \! + \! 1}}|{L^2}}}{{2}})||\boldsymbol{v}^{t+1} \!-\! \boldsymbol{v}^t|{|^2} \! + \! \frac{1}{{2{{\eta _{\lambda}}}}} \! \sum\limits_{l = 1}^{|{\boldsymbol{\mathcal{P}}^{t}}|}\! {||{\lambda _l^t} \!-\! {\lambda _l^{t-1}}|{|^2}} \\

+ (\frac{{L  +  6\tau{ k_1 }N{L^2}}}{2} \!-\! \frac{1}{{{\eta _{\boldsymbol{z}}^t}}} \! + \! \frac{{\eta _{\lambda}|{\boldsymbol{\mathcal{P}}^{t}}|{L^2}+\eta _{\boldsymbol{\theta}}|{{\boldsymbol{\mathcal{Q}}}^{t \! + \! 1}}|{L^2}}}{{2}})||\boldsymbol{z}^{t+1} \!-\! \boldsymbol{z}^t|{|^2}  \! + \! \frac{1}{{2{{\eta _{\boldsymbol{\theta }}}}}}\sum\limits_{i = 1}^N {||{{\boldsymbol{\theta}}_i^t} \!-\! {{\boldsymbol{\theta}}_i^{t-1}}|{|^2}}  \\

  +  (\frac{{ 6\tau{ k_1 }N{L^2}}}{2} \!-\! \frac{{{c_1^{t-1}} - {c_1^t}}}{2} \! + \! \frac{1}{{{{\eta _{\lambda}}}}})\sum\limits_{l = 1}^{|{\boldsymbol{\mathcal{P}}^{t}}|} {||{\lambda _l^{t+1}} \!-\! {\lambda _l^t}|{|^2}} \! + \! (\frac{1}{{{{\eta _{\boldsymbol{\theta }}}}}} \!-\! \frac{{{c_2^{t-1}} - {c_2^t}}}{2} ) \! \sum\limits_{i = 1}^N\! {||{{\boldsymbol{\theta}}_i^{t+1}} \!-\! {{\boldsymbol{\theta}}_i^t}|{|^2}}  \vspace{0.5ex}\\
 
 \! +  \frac{{{c_1^{t-1}}}}{2} \! \sum\limits_{l = 1}^{|{\boldsymbol{\mathcal{P}}^{t}}|}\! {(||{\lambda _l^{t+1}}|{|^2} \!-\! ||{\lambda _l^t}|{|^2})} 
 \! +  \frac{{{c_2^{t-1}}}}{2}\sum\limits_{i = 1}^N {(||{{\boldsymbol{\theta}}_i^{t+1}}|{|^2} \!-\! ||{{\boldsymbol{\theta}}_i^t}|{|^2})}.
 
\end{array}
\end{equation}

According to Eq. (\ref{eq:new_29}), in $(t+1)^{\rm{th}}$ iteration, it follows that:
{\begin{equation}
\label{eq:A45}
\left\langle {{\lambda _l^{t+1}}\! -\! {\lambda _l^t}\! -\! {{\eta _{\lambda}}}{\nabla _{{\lambda _l}}}{{\widetilde{L}}_p}(\{ \boldsymbol{x}_i^{t + 1}\} ,\!\{ \boldsymbol{y}_i^{t+1 }\} ,\!{\boldsymbol{v}^{t+1}  },\!{\boldsymbol{z}^{t+1}  },\!\{ \lambda _l^{t}  \} ,\!\{ \boldsymbol{\theta }_i^{t}  \} ),{\lambda _l^t}\! -\! {\lambda _l^{t+1}}} \right\rangle = 0.
\end{equation}}

Similar to Eq. (\ref{eq:A45}), in $t^{\rm{th}}$ iteration, we have,
{\begin{equation}
\label{eq:A46}
\left\langle {{\lambda _l^t}\! -\! {\lambda _l^{t-1}} \!-\! {{\eta _{\lambda}}}{\nabla _{{\lambda _l}}}{{\widetilde{L}}_p}(\{ \boldsymbol{x}_i^{t}\} ,\!\{ \boldsymbol{y}_i^{t}\} ,\!{\boldsymbol{v}^{t}  },\!{\boldsymbol{z}^{t}  },\!\{ \lambda _l^{t-1}  \} ,\!\{ \boldsymbol{\theta }_i^{t-1}  \} ),{\lambda _l^{t+1}} \!-\! {\lambda _l^t}} \right\rangle  = 0.
\end{equation}}

Thus, $\forall t \ge T_1$, by combining Eq. (\ref{eq:A45}) with Eq. (\ref{eq:A46}), we can obtain that, 
{\begin{equation}
\label{eq:A47}
\begin{array}{l}
\frac{1}{{{{\eta _{\lambda}}}}}\left\langle {{{\boldsymbol{v}}_{1,l}^{t + 1}},{\lambda _l^{t+1}} \! - \! {\lambda _l^t}} \right\rangle \vspace{1ex} \\

\! = \! \left\langle \! {{\nabla _{{\lambda _l}}}{{\widetilde{L}}_p}(\{ \boldsymbol{x}_i^{t + 1}\} ,\!\{ \boldsymbol{y}_i^{t+1 }\} ,\!{\boldsymbol{v}^{t+1}  },\!{\boldsymbol{z}^{t+1}  },\!\{ \lambda _l^{t}  \} ,\!\{ \boldsymbol{\theta }_i^{t}  \} ) \! - \! {\nabla _{{\lambda _l}}}{{\widetilde{L}}_p}(\{ \boldsymbol{x}_i^{t}\} ,\!\{ \boldsymbol{y}_i^{t}\} ,\!{\boldsymbol{v}^{t}  },\!{\boldsymbol{z}^{t}  },\!\{ \lambda _l^{t-1}  \} ,\!\{ \boldsymbol{\theta }_i^{t-1}  \} ),{\lambda _l^{t+1}} \! - \! {\lambda _l^t}} \! \right\rangle \vspace{1ex} \\

\! = \! \left\langle \!  {{\nabla _{{\lambda _l}}}{{\widetilde{L}}_p}(\{ \boldsymbol{x}_i^{t + 1}\} ,\!\{ \boldsymbol{y}_i^{t+1 }\} ,\!{\boldsymbol{v}^{t+1}  },\!{\boldsymbol{z}^{t+1}  },\!\{ \lambda _l^{t}  \} ,\!\{ \boldsymbol{\theta }_i^{t}  \} ) \! - \! {\nabla _{{\lambda _l}}}{{\widetilde{L}}_p}(\{ \boldsymbol{x}_i^{t}\} ,\!\{ \boldsymbol{y}_i^{t}\} ,\!{\boldsymbol{v}^{t}  },\!{\boldsymbol{z}^{t}  },\!\{ \lambda _l^{t}  \} ,\!\{ \boldsymbol{\theta }_i^{t}  \} ),{\lambda _l^{t+1}} \! - \! {\lambda _l^t}} \! \right\rangle \vspace{1ex} \\

 \! + \! \left\langle \! {{\nabla _{{\lambda _l}}}{{\widetilde{L}}_p}(\{ \boldsymbol{x}_i^{t}\} ,\!\{ \boldsymbol{y}_i^{t}\} ,\!{\boldsymbol{v}^{t}  },\!{\boldsymbol{z}^{t}  },\!\{ \lambda _l^{t}  \} ,\!\{ \boldsymbol{\theta }_i^{t}  \} ) \! - \! {\nabla _{{\lambda _l}}}{{\widetilde{L}}_p}(\{ \boldsymbol{x}_i^{t}\} ,\!\{ \boldsymbol{y}_i^{t}\} ,\!{\boldsymbol{v}^{t}  },\!{\boldsymbol{z}^{t}  },\!\{ \lambda _l^{t-1}  \} ,\!\{ \boldsymbol{\theta }_i^{t-1}  \} ),{{\boldsymbol{v}}_{1,l}^{t + 1}}} \! \right\rangle \vspace{1ex} \\
 
 \! + \! \left\langle \! {{\nabla _{{\lambda _l}}}{{\widetilde{L}}_p}(\{ \boldsymbol{x}_i^{t}\} ,\!\{ \boldsymbol{y}_i^{t}\} ,\!{\boldsymbol{v}^{t}  },\!{\boldsymbol{z}^{t}  },\!\{ \lambda _l^{t}  \} ,\!\{ \boldsymbol{\theta }_i^{t}  \} ) \! - \! {\nabla _{{\lambda _l}}}{{\widetilde{L}}_p}(\{ \boldsymbol{x}_i^{t}\} ,\!\{ \boldsymbol{y}_i^{t}\} ,\!{\boldsymbol{v}^{t}  },\!{\boldsymbol{z}^{t}  },\!\{ \lambda _l^{t-1}  \},\{ {{\boldsymbol{\theta}}_i^{t-1}}\} {\rm{)}},{\lambda _l^t} \! - \! {\lambda _l^{t-1}}} \! \right\rangle . 
\end{array}
\end{equation}}

Since we have that, 
\begin{equation}
\label{eq:A48}
\begin{array}{l}
\frac{1}{{{{\eta _{\lambda}}}}}\left\langle {{{\boldsymbol{v}}_{1,l}^{t + 1}},{\lambda _l^{t+1}} \!-\! {\lambda _l^t}} \right\rangle 
 = \frac{1}{{2{{\eta _{\lambda}}}}}||{\lambda _l^{t+1}} \!-\! {\lambda _l^t}|{|^2} \! + \! \frac{1}{{2{{\eta _{\lambda}}}}}||{{\boldsymbol{v}}_{1,l}^{t + 1}}|{|^2} - \frac{1}{{2{{\eta _{\lambda}}}}}||{\lambda _l^t} - {\lambda _l^{t-1}}|{|^2},
\end{array}
\end{equation}
it follows from Eq. (\ref{eq:A47}) and Eq. (\ref{eq:A48}) that,
{\begin{equation}
\label{eq:A49}
\begin{array}{l}
\frac{1}{{2{{\eta _{\lambda}}}}}||{\lambda _l^{t+1}} \! - \! {\lambda _l^t}|{|^2} \! + \! \frac{1}{{2{{\eta _{\lambda}}}}}||{{\boldsymbol{v}}_{1,l}^{t + 1}}|{|^2} \! - \! \frac{1}{{2{{\eta _{\lambda}}}}}||{\lambda _l^t} \! - \! {\lambda _l^{t-1}}|{|^2} \vspace{1ex}\\
\! = \! \frac{{{L^2}}}{{2{b_1^t}}}(\sum\limits_{i = 1}^N({||{{\boldsymbol{x}}_i^{t+1}} \!-\! {{\boldsymbol{x}}_i^t}|{|^2}} + {||{{\boldsymbol{y}}_i^{t+1}} \!-\! {{\boldsymbol{y}}_i^t}|{|^2}})  \! + \! ||{\boldsymbol{v}}^{t+1} \!-\! {\boldsymbol{v}^t}|{|^2} \! + \! ||{\boldsymbol{z}}^{t+1} \!-\! {\boldsymbol{z}^t}|{|^2}) \! + \! \frac{{{b_1^t}}}{2}||{\lambda _l^{t+1}} \! - \! {\lambda _l^t}|{|^2} \vspace{1ex}\\
 \! +  \frac{{{c_1^{t-1}}  -  {c_1^t}}}{2}(||{\lambda _l^{t+1}}|{|^2} \! - \! ||{\lambda _l^t}|{|^2}) \! - \! \frac{{{c_1^{t-1}}  -  {c_1^t}}}{2}||{\lambda _l^{t+1}} \! - \! {\lambda _l^t}|{|^2} \vspace{1.5ex}\\
 
 \! +  \frac{{{{\eta _{\lambda}}}}}{2}||{\nabla _{{\lambda _l}}}{\widetilde{L}_p}(\{ \boldsymbol{x}_i^{t}\} ,\!\{ \boldsymbol{y}_i^{t}\} ,\!{\boldsymbol{v}^{t}  },\!{\boldsymbol{z}^{t}  },\!\{ \lambda _l^{t}  \} ,\!\{ \boldsymbol{\theta }_i^{t}  \} ) \! - \! {\nabla _{{\lambda _l}}}{\widetilde{L}_p}(\{ \boldsymbol{x}_i^{t}\} ,\!\{ \boldsymbol{y}_i^{t}\} ,\!{\boldsymbol{v}^{t}  },\!{\boldsymbol{z}^{t}  },\!\{ \lambda _l^{t-1}  \} ,\!\{ \boldsymbol{\theta }_i^{t-1}  \} ) |{|^2} \! + \! \frac{1}{{2{{\eta _{\lambda}}}}}\!||{{\boldsymbol{v}}_{1,l}^{t + 1}}|{|^2} \vspace{1.5ex}\\
 
\! - \frac{1}{{{L_1}'  +  {c_1^{t-1}}}}||{\nabla _{{\lambda _l}}}{\widetilde{L}_p}(\{ \boldsymbol{x}_i^{t}\} ,\!\{ \boldsymbol{y}_i^{t}\} ,\!{\boldsymbol{v}^{t}  },\!{\boldsymbol{z}^{t}  },\!\{ \lambda _l^{t}  \} ,\!\{ \boldsymbol{\theta }_i^{t}  \} ) \! - \! {\nabla _{{\lambda _l}}}{\widetilde{L}_p}(\{ \boldsymbol{x}_i^{t}\} ,\!\{ \boldsymbol{y}_i^{t}\} ,\!{\boldsymbol{v}^{t}  },\!{\boldsymbol{z}^{t}  },\!\{ \lambda _l^{t-1}  \} ,\!\{ \boldsymbol{\theta }_i^{t-1}  \} ) |{|^2} \vspace{1.5ex}\\

\! - \frac{{{c_1^{t-1}}{L_1}'}}{{{L_1}'  +  {c_1^{t-1}}}}||{\lambda _l^t} \! - \! {\lambda _l^{t-1}}|{|^2},
\end{array}
\end{equation}}

\noindent where ${b_1^t} > 0$. According to the setting that ${c_1^0} \le {L_1}'$, we have $- \frac{{{c_1^{t-1}}{L_1}'}}{{{L_1}' + {c_1^{t-1}}}} \le  - \frac{{{c_1^{t-1}}{L_1}'}}{{2{L_1}'}} =  - \frac{{{c_1^{t-1}}}}{2} \le  - \frac{{{c_1^t}}}{2}$. Multiplying both sides of Eq. (\ref{eq:A49}) by $\frac{8}{{{{\eta _{\lambda}}}{c_1^t}}}$, we have,
{\begin{equation}
\label{eq:A51}
\begin{array}{l}
\frac{4}{{{{\eta _{\lambda}}}^2{c_1^t}}}||{\lambda _l^{t+1}} - {\lambda _l^t}|{|^2} - \frac{4}{{{{\eta _{\lambda}}}}}(\frac{{{c_1^{t-1}} - {c_1^t}}}{{{c_1^t}}})||{\lambda _l^{t+1}}|{|^2} \vspace{1ex}\\
 \le \frac{4}{{{{\eta _{\lambda}}}^2{c_1^t}}}||{\lambda _l^t} - {\lambda _l^{t-1}}|{|^2} - \frac{4}{{{{\eta _{\lambda}}}}}(\frac{{{c_1^{t-1}} - {c_1^t}}}{{{c_1^t}}})||{\lambda _l^t}|{|^2}  + \frac{{4{b_1^t}}}{{{{\eta _{\lambda}}}{c_1^t}}}||{\lambda _l^{t+1}} - {\lambda _l^t}|{|^2} - \frac{4}{{{{\eta _{\lambda}}}}}||{\lambda _l^t} - {\lambda _l^{t-1}}|{|^2} \vspace{1ex}\\
 \! + \frac{{4{L^2}}}{{{{\eta _{\lambda}}}{c_1^t}{b_1^t}}}(\sum\limits_{i = 1}^N({||{{\boldsymbol{x}}_i^{t+1}} \!-\! {{\boldsymbol{x}}_i^t}|{|^2}} + {||{{\boldsymbol{y}}_i^{t+1}} \!-\! {{\boldsymbol{y}}_i^t}|{|^2}})  \! + \! ||{\boldsymbol{v}}^{t+1} \!-\! {\boldsymbol{v}^t}|{|^2} \! + \! ||{\boldsymbol{z}}^{t+1} \!-\! {\boldsymbol{z}^t}|{|^2}).
\end{array}
\end{equation}}

Setting ${b_1^t} = \frac{{{c_1^t}}}{2}$ in Eq. (\ref{eq:A51}) and using the definition of ${S_1^t}$, $\forall t \ge T_1$, we have,
{\begin{equation}
\label{eq:A52}
\begin{array}{l}
{S_1^{t+1}} \! - \! {S_1^t} \vspace{1ex}\\
\! \le \!\sum\limits_{l = 1}^{|{\boldsymbol{\mathcal{P}}^{t}}|} {\frac{4}{{{{\eta _{\lambda}}}}}(\frac{{{c_1^{t-2}}}}{{{c_1^{t-1}}}} \! - \! \frac{{{c_1^{t-1}}}}{{{c_1^t}}})||{\lambda _l^t}|{|^2}}  \! +\! \sum\limits_{l = 1}^{|{\boldsymbol{\mathcal{P}}^{t}}|} {(\frac{2}{{{{\eta _{\lambda}}}}} \! +\! \frac{4}{{{\eta _{\lambda}}^2}}(\frac{1}{{{c_1^{t+1}}}} \! - \! \frac{1}{{{c_1^t}}}))||{\lambda _l^{t+1}} \! - \! {\lambda _l^t}|{|^2}} \vspace{1ex}\\
\! - \sum\limits_{l = 1}^{|{\boldsymbol{\mathcal{P}}^{t}}|} {\frac{4}{{{{\eta _{\lambda}}}}}||{\lambda _l^t} \! - \! {\lambda _l^{t-1}}|{|^2} \vspace{1ex}}
\! +\! \frac{{8|{\boldsymbol{\mathcal{P}}^{t}}|{L^2}}}{{{{\eta _{\lambda}}}({c_1^t}){^2}}}(\sum\limits_{i = 1}^N({||{{\boldsymbol{x}}_i^{t+1}} \!-\! {{\boldsymbol{x}}_i^t}|{|^2}} + {||{{\boldsymbol{y}}_i^{t+1}} \!-\! {{\boldsymbol{y}}_i^t}|{|^2}})  \! + \! ||{\boldsymbol{v}}^{t+1} \!-\! {\boldsymbol{v}^t}|{|^2} \! + \! ||{\boldsymbol{z}}^{t+1} \!-\! {\boldsymbol{z}^t}|{|^2}).
\end{array}
\end{equation}}

Similarly, according to Eq. (\ref{eq:new_30}), it follows that,
{\begin{equation}
\label{eq:A53}
\begin{array}{l}
\frac{1}{{{{\eta _{\boldsymbol{\theta }}}}}}\left\langle {{{\boldsymbol{v}}_{2,l}^{t + 1}},{{\boldsymbol{\theta}}_i^{t+1}} \! - \! {{\boldsymbol{\theta}}_i^t}} \right\rangle \vspace{1ex}\\

\! = \! \left\langle\! {{\nabla _{{{\boldsymbol{\theta}}_i}}}{{\widetilde{L}}_p}(\{ \boldsymbol{x}_i^{t+1}\} ,\!\{ \boldsymbol{y}_i^{t+1}\} ,\!{\boldsymbol{v}^{t+1}  },\!{\boldsymbol{z}^{t+1}  },\!\{ \lambda _l^{t+1}  \} ,\!\{ \boldsymbol{\theta }_i^{t}  \} ) \!-\! {\nabla _{{{\boldsymbol{\theta}}_i}}}{{\widetilde{L}}_p}(\{ \boldsymbol{x}_i^{t}\} ,\!\{ \boldsymbol{y}_i^{t}\} ,\!{\boldsymbol{v}^{t}  },\!{\boldsymbol{z}^{t}  },\!\{ \lambda _l^{t}  \} ,\!\{ \boldsymbol{\theta }_i^{t-1} \} ),{{\boldsymbol{\theta}}_i^{t+1}} \! - \! {{\boldsymbol{\theta}}_i^t}} \! \right\rangle \vspace{1ex}\\

\! = \! \left\langle\! {{\nabla _{{{\boldsymbol{\theta}}_i}}}{{\widetilde{L}}_p}(\{ \boldsymbol{x}_i^{t+1}\} ,\!\{ \boldsymbol{y}_i^{t+1}\} ,\!{\boldsymbol{v}^{t+1}  },\!{\boldsymbol{z}^{t+1}  },\!\{ \lambda _l^{t+1}  \} ,\!\{ \boldsymbol{\theta }_i^{t}  \} ) \!-\! {\nabla _{{{\boldsymbol{\theta}}_i}}}{{\widetilde{L}}_p}(\{ \boldsymbol{x}_i^{t}\} ,\!\{ \boldsymbol{y}_i^{t}\} ,\!{\boldsymbol{v}^{t}  },\!{\boldsymbol{z}^{t}  },\!\{ \lambda _l^{t}  \} ,\!\{ \boldsymbol{\theta }_i^{t} \} ),{{\boldsymbol{\theta}}_i^{t+1}} \! - \! {{\boldsymbol{\theta}}_i^t}} \!\right\rangle \vspace{1ex}\\

 \! + \! \left\langle\! {{\nabla _{{{\boldsymbol{\theta}}_i}}}{{\widetilde{L}}_p}(\{ \boldsymbol{x}_i^{t}\} ,\!\{ \boldsymbol{y}_i^{t}\} ,\!{\boldsymbol{v}^{t}  },\!{\boldsymbol{z}^{t}  },\!\{ \lambda _l^{t}  \} ,\!\{ \boldsymbol{\theta }_i^{t}  \} ) \!-\! {\nabla _{{{\boldsymbol{\theta}}_i}}}{{\widetilde{L}}_p}(\{ \boldsymbol{x}_i^{t}\} ,\!\{ \boldsymbol{y}_i^{t}\} ,\!{\boldsymbol{v}^{t}  },\!{\boldsymbol{z}^{t}  },\!\{ \lambda _l^{t}  \} ,\!\{ \boldsymbol{\theta }_i^{t-1} \} ),{{\boldsymbol{v}}_{2,l}^{t + 1}}} \!\right\rangle \vspace{1ex}\\
 
 \! + \! \left\langle \! {{\nabla _{{{\boldsymbol{\theta}}_i}}}{{\widetilde{L}}_p}(\{ \boldsymbol{x}_i^{t}\} ,\!\{ \boldsymbol{y}_i^{t}\} ,\!{\boldsymbol{v}^{t}  },\!{\boldsymbol{z}^{t}  },\!\{ \lambda _l^{t}  \} ,\!\{ \boldsymbol{\theta }_i^{t}  \} ) \!-\! {\nabla _{{{\boldsymbol{\theta}}_i}}}{{\widetilde{L}}_p}(\{ \boldsymbol{x}_i^{t}\} ,\!\{ \boldsymbol{y}_i^{t}\} ,\!{\boldsymbol{v}^{t}  },\!{\boldsymbol{z}^{t}  },\!\{ \lambda _l^{t}  \} ,\!\{ \boldsymbol{\theta }_i^{t-1} \} ),{{\boldsymbol{\theta}}_i^t} \! - \! {{\boldsymbol{\theta}}_i^{t-1}}} \! \right\rangle .
\end{array}
\end{equation}}

In addition, since
{\begin{equation}
\label{eq:A54}
\begin{array}{l}
\frac{1}{{{{\eta _{\boldsymbol{\theta }}}}}}\left\langle {{{\boldsymbol{v}}_{2,l}^{t+1}},{{\boldsymbol{\theta}}_i^{t+1}} - {{\boldsymbol{\theta}}_i^t}} \right\rangle 
 = \frac{1}{{2{{\eta _{\boldsymbol{\theta }}}}}}||{{\boldsymbol{\theta}}_i^{t+1}} - {{\boldsymbol{\theta}}_i^t}|{|^2} \! +\! \frac{1}{{2{{\eta _{\boldsymbol{\theta }}}}}}||{{\boldsymbol{v}}_{2,l}^{t+1}}|{|^2} - \frac{1}{{2{{\eta _{\boldsymbol{\theta }}}}}}||{{\boldsymbol{\theta}}_i^t} - {{\boldsymbol{\theta}}_i^{t-1}}|{|^2},
\end{array}
\end{equation}}
it follows that,
{\begin{equation}
\label{eq:A55}
\begin{array}{l}
\frac{1}{{2{{\eta _{\boldsymbol{\theta }}}}}}||{{\boldsymbol{\theta}}_i^{t+1}} \! - \! {{\boldsymbol{\theta}}_i^t}|{|^2} \! +\! \frac{1}{{2{{\eta _{\boldsymbol{\theta }}}}}}||{{\boldsymbol{v}}_{2,l}^{t+1}}|{|^2} \! - \! \frac{1}{{2{{\eta _{\boldsymbol{\theta }}}}}}||{{\boldsymbol{\theta}}_i^t} \! - \! {{\boldsymbol{\theta}}_i^{t-1}}|{|^2} \vspace{1ex}\\

 \! =  \! \frac{{{L^2}}}{{2{b_2^t}}}(\sum\limits_{i = 1}^N({||{{\boldsymbol{x}}_i^{t+1}} \!-\! {{\boldsymbol{x}}_i^t}|{|^2}} + {||{{\boldsymbol{y}}_i^{t+1}} \!-\! {{\boldsymbol{y}}_i^t}|{|^2}})  \! + \! ||{\boldsymbol{v}}^{t+1} \!-\! {\boldsymbol{v}^t}|{|^2} \! + \! ||{\boldsymbol{z}}^{t+1} \!-\! {\boldsymbol{z}^t}|{|^2}) \! +\! \frac{{{b_2^t}}}{2}||{{\boldsymbol{\theta}}_i^{t+1}} \! - \! {{\boldsymbol{\theta}}_i^t}|{|^2} \vspace{1ex}\\
 
 \! + \frac{{{c_2^{t-1}}  -  {c_2^t}}}{2}(||{{\boldsymbol{\theta}}_i^{t+1}}|{|^2} \! - \! ||{{\boldsymbol{\theta}}_i^t}|{|^2}) \! - \! \frac{{{c_2^{t-1}}  -  {c_2^t}}}{2}||{{\boldsymbol{\theta}}_i^{t+1}} \! - \! {{\boldsymbol{\theta}}_i^t}|{|^2}  -  \frac{{{c_2^{t-1}}L_2'}}{{L_2'  + {c_2^{t-1}}}}||{{\boldsymbol{\theta}}_i^t} \! - \! {{\boldsymbol{\theta}}_i^{t-1}}|{|^2}\! +\! \frac{1}{{2{{\eta _{\boldsymbol{\theta }}}}}}||{{\boldsymbol{v}}_{2,l}^{t+1}}|{|^2} \vspace{1.5ex}\\
 
 \! + \frac{{{{\eta _{\boldsymbol{\theta }}}}}}{2}||{\nabla _{{{\boldsymbol{\theta}}_i}}}{{\widetilde{L}}_p}(\{ \boldsymbol{x}_i^{t}\} ,\!\{ \boldsymbol{y}_i^{t}\} ,\!{\boldsymbol{v}^{t}  },\!{\boldsymbol{z}^{t}  },\!\{ \lambda _l^{t}  \} ,\!\{ \boldsymbol{\theta }_i^{t}  \} ) \!-\! {\nabla _{{{\boldsymbol{\theta}}_i}}}{{\widetilde{L}}_p}(\{ \boldsymbol{x}_i^{t}\} ,\!\{ \boldsymbol{y}_i^{t}\} ,\!{\boldsymbol{v}^{t}  },\!{\boldsymbol{z}^{t}  },\!\{ \lambda _l^{t}  \} ,\!\{ \boldsymbol{\theta }_i^{t-1} \} )|{|^2}  \vspace{1.5ex}\\
 
 \! -  \frac{1}{{L_2'  + {c_2^{t-1}}}}||{\nabla _{{{\boldsymbol{\theta}}_i}}}{{\widetilde{L}}_p}(\{ \boldsymbol{x}_i^{t}\} ,\!\{ \boldsymbol{y}_i^{t}\} ,\!{\boldsymbol{v}^{t}  },\!{\boldsymbol{z}^{t}  },\!\{ \lambda _l^{t}  \} ,\!\{ \boldsymbol{\theta }_i^{t}  \} ) \!-\! {\nabla _{{{\boldsymbol{\theta}}_i}}}{{\widetilde{L}}_p}(\{ \boldsymbol{x}_i^{t}\} ,\!\{ \boldsymbol{y}_i^{t}\} ,\!{\boldsymbol{v}^{t}  },\!{\boldsymbol{z}^{t}  },\!\{ \lambda _l^{t}  \} ,\!\{ \boldsymbol{\theta }_i^{t-1} \} )|{|^2}.
\end{array}
\end{equation}}

According to the setting ${c_2^0} \le {L_2}'$, we have $- \frac{{{c_2^{t-1}}{L_2}'}}{{{L_2}' + {c_2^{t-1}}}} \le  - \frac{{{c_2^{t-1}}{L_2}'}}{{2{L_2}'}} =  - \frac{{{c_2^{t-1}}}}{2} \le  - \frac{{{c_2^t}}}{2}$.  Multiplying both sides of Eq. (\ref{eq:A55})  by $\frac{8}{{{{\eta _{\boldsymbol{\theta }}}}{c_2^t}}}$, we have,
{\begin{equation}
\label{eq:A57}
\begin{array}{l}
\frac{4}{{{{\eta _{\boldsymbol{\theta }}}}^2{c_2^t}}}||{{\boldsymbol{\theta}}_i^{t+1}} \! - \! {{\boldsymbol{\theta}}_i^t}|{|^2} \! - \! \frac{4}{{{{\eta _{\boldsymbol{\theta }}}}}}(\frac{{{c_2^{t-1}}  -  {c_2^t}}}{{{c_2^t}}})||{{\boldsymbol{\theta}}_i^{t+1}}|{|^2} \vspace{1ex}\\
\! \le \! \frac{4}{{{{\eta _{\boldsymbol{\theta }}}}^2{c_2^t}}}||{{\boldsymbol{\theta}}_i^t} \! - \! {{\boldsymbol{\theta}}_i^{t-1}}|{|^2} \! - \! \frac{4}{{{{\eta _{\boldsymbol{\theta }}}}}}(\frac{{{c_2^{t-1}}  -  {c_2^t}}}{{{c_2^t}}})||{{\boldsymbol{\theta}}_i^t}|{|^2} \! +\! \frac{{4{b_2^t}}}{{{{\eta _{\boldsymbol{\theta }}}}{c_2^t}}}||{{\boldsymbol{\theta}}_i^{t+1}} \! - \! {{\boldsymbol{\theta}}_i^t}|{|^2} \! - \! \frac{4}{{{{\eta _{\boldsymbol{\theta }}}}}}||{{\boldsymbol{\theta}}_i^t} \! - \! {{\boldsymbol{\theta}}_i^{t-1}}|{|^2} \vspace{1ex}\\
 \! + \frac{{4{L^2}}}{{{{\eta _{\boldsymbol{\theta }}}}{c_2^t}{b_2^t}}}(\sum\limits_{i = 1}^N({||{{\boldsymbol{x}}_i^{t+1}} \!-\! {{\boldsymbol{x}}_i^t}|{|^2}} + {||{{\boldsymbol{y}}_i^{t+1}} \!-\! {{\boldsymbol{y}}_i^t}|{|^2}})  \! + \! ||{\boldsymbol{v}}^{t+1} \!-\! {\boldsymbol{v}^t}|{|^2} \! + \! ||{\boldsymbol{z}}^{t+1} \!-\! {\boldsymbol{z}^t}|{|^2}).
\end{array}
\end{equation}}

Setting ${b_2^t} = \frac{{{c_2^t}}}{2}$ in Eq. (\ref{eq:A57}) and utilizing the definition of ${S_2^t}$, we have that,
{\begin{equation}
\label{eq:A58}
\begin{array}{l}
{S_2^{t+1}} \! - \! {S_2^t} \vspace{1ex}\\

\! \le \! \sum\limits_{i = 1}^N \! {\frac{4}{{{{\eta _{\boldsymbol{\theta }}}}}}(\frac{{{c_2^{t-2}}}}{{{c_2^{t-1}}}} \! - \! \frac{{{c_2^{t-1}}}}{{{c_2^t}}})||{{\boldsymbol{\theta}}_i^t}|{|^2}}
 \! +\! \sum\limits_{i = 1}^N \!{(\frac{2}{{{{\eta _{\boldsymbol{\theta }}}}}} \! +\! \frac{4}{{{{\eta _{\boldsymbol{\theta }}}}^2}}(\frac{1}{{{c_2^{t+1}}}} \! - \! \frac{1}{{{c_2^t}}}))||{{\boldsymbol{\theta}}_i^{t+1}} \! - \! {{\boldsymbol{\theta}}_i^t}|{|^2}} \vspace{1ex}\\
 \! - \! \sum\limits_{i = 1}^N \! {\frac{4}{{{{\eta _{\boldsymbol{\theta }}}}}}||{{\boldsymbol{\theta}}_i^t} \! - \! {{\boldsymbol{\theta}}_i^{t-1}}|{|^2}}  \! +\! \frac{{8N{L^2}}}{{{{\eta _{\boldsymbol{\theta }}}}({c_2^t})^2}}(\sum\limits_{i = 1}^N({||{{\boldsymbol{x}}_i^{t+1}} \!-\! {{\boldsymbol{x}}_i^t}|{|^2}} + {||{{\boldsymbol{y}}_i^{t+1}} \!-\! {{\boldsymbol{y}}_i^t}|{|^2}})  \! + \! ||{\boldsymbol{v}}^{t+1} \!-\! {\boldsymbol{v}^t}|{|^2} \! + \! ||{\boldsymbol{z}}^{t+1} \!-\! {\boldsymbol{z}^t}|{|^2}).
\end{array}
\end{equation}}

Based on the setting of ${c_1^t}$ and ${c_2^t}$, we can obtain that $\frac{{{{\eta _{\lambda}}}}}{{10}} \ge \frac{1}{{{c_1^{t+1}}}} - \frac{1}{{{c_1^t}}},
\frac{{{{\eta _{\boldsymbol{\theta }}}}}}{{10}} \ge \frac{1}{{{c_2^{t+1}}}} - \frac{1}{{{c_2^t}}},{\rm{    }}\forall t \ge T_1$. Defining $a_5 = \max\{1, 1+ L^2, 6\tau{ k_1 }N{L^2}\}$.
Combining the definition of $F^{t+1}$ with Eq. (\ref{eq:A52}) and Eq. (\ref{eq:A58}), $\forall t \ge T_1$, we can obtain that,
\begin{equation}
\label{eq:A59}
\begin{array}{l}
F^{t+1} \! - \! F^{t}\\
 \!\le\! (\frac{{L  + a_5}}{2} \! - \! \frac{1}{{{\eta _{\boldsymbol{x}}^t}}} \! +\! \frac{{{{\eta _{\lambda}}}|{\boldsymbol{\mathcal{P}}^{t}}|{L^2}}}{2} \! +\! \frac{{{{\eta _{\boldsymbol{\theta }}}}|{\boldsymbol{\mathcal{Q}}^{t + 1}}|{L^2}}}{2} \! +\! \frac{{8|{\boldsymbol{\mathcal{P}}^{t}}|{L^2}}}{{{{\eta _{\lambda}}}({c_1^t})^2}} \! +\! \frac{{8N{L^2}}}{{{{\eta _{\boldsymbol{\theta }}}}({c_2^t})^2}})\sum\limits_{i = 1}^N {||{{\boldsymbol{x}}_i^{t+1}} \! - \! {{\boldsymbol{x}}_i^t}|{|^2}} \\
 
\! + (\frac{{L  + a_5}}{2} \! - \! \frac{1}{{{\eta _{\boldsymbol{y}}^t}}} \! +\! \frac{{{{\eta _{\lambda}}}|{\boldsymbol{\mathcal{P}}^{t}}|{L^2}}}{2} \! +\! \frac{{{{\eta _{\boldsymbol{\theta }}}}|{\boldsymbol{\mathcal{Q}}^{t + 1}}|{L^2}}}{2} \! +\! \frac{{8|{\boldsymbol{\mathcal{P}}^{t}}|{L^2}}}{{{{\eta _{\lambda}}}({c_1^t})^2}} \! +\! \frac{{8N{L^2}}}{{{{\eta _{\boldsymbol{\theta }}}}({c_2^t})^2}})\sum\limits_{i = 1}^N {||{{\boldsymbol{y}}_i^{t+1}} \! - \! {{\boldsymbol{y}}_i^t}|{|^2}} \\
 
 \! + (\frac{{L  + a_5}}{2} \! - \! \frac{1}{{{\eta _{\boldsymbol{v}}^t}}} \! +\! \frac{{{{\eta _{\lambda}}}|{\boldsymbol{\mathcal{P}}^{t}}|{L^2}}}{2} \! +\! \frac{{{{\eta _{\boldsymbol{\theta }}}}|{\boldsymbol{\mathcal{Q}}^{t + 1}}|{L^2}}}{2} \! +\! \frac{{8|{\boldsymbol{\mathcal{P}}^{t}}|{L^2}}}{{{{\eta _{\lambda}}}({c_1^t})^2}} \! +\! \frac{{8N{L^2}}}{{{{\eta _{\boldsymbol{\theta }}}}({c_2^t})^2}})||\boldsymbol{v}^{t+1} \! - \! \boldsymbol{v}^t|{|^2}\\
 
 \! + (\frac{{L  + a_5}}{2} \! - \! \frac{1}{{{\eta _{\boldsymbol{z}}^t}}} \! +\! \frac{{{{\eta _{\lambda}}}|{\boldsymbol{\mathcal{P}}^{t}}|{L^2}}}{2} \! +\! \frac{{{{\eta _{\boldsymbol{\theta }}}}|{\boldsymbol{\mathcal{Q}}^{t + 1}}|{L^2}}}{2} \! +\! \frac{{8|{\boldsymbol{\mathcal{P}}^{t}}|{L^2}}}{{{{\eta _{\lambda}}}({c_1^t})^2}} \! +\! \frac{{8N{L^2}}}{{{{\eta _{\boldsymbol{\theta }}}}({c_2^t})^2}})||\boldsymbol{z}^{t+1} \! - \! \boldsymbol{z}^t|{|^2}\\
 
 \! -  (\frac{1}{{10{{\eta _{\lambda}}}}} \! - \! \frac{{6\tau{ k_1 }N{L^2}}}{2})\sum\limits_{l = 1}^{|{\boldsymbol{\mathcal{P}}^{t}}|} {||{\lambda _l^{t+1}} \! - \! {\lambda _l^t}|{|^2}}  \! - \! \frac{1}{{10{{\eta _{\boldsymbol{\theta }}}}}}\sum\limits_{i = 1}^N {||{{\boldsymbol{\theta}}_i^{t+1}} \! - \! {{\boldsymbol{\theta}}_i^t}|{|^2}} 
 \! +\! \frac{{{c_1^{t-1}}  -  {c_1^t}}}{2}\sum\limits_{l = 1}^{|{\boldsymbol{\mathcal{P}}^{t}}|} {||{\lambda _l^{t+1}}|{|^2}} \\
 
 \! + \frac{{{c_2^{t-1}}  -  {c_2^t}}}{2}\sum\limits_{i = 1}^N {||{{\boldsymbol{\theta}}_i^{t+1}}|{|^2}} 
 \! +\! \frac{4}{{{{\eta _{\lambda}}}}}(\frac{{{c_1^{t-2}}}}{{{c_1^{t-1}}}} \! - \! \frac{{{c_1^{t-1}}}}{{{c_1^t}}})\sum\limits_{l = 1}^{|{\boldsymbol{\mathcal{P}}^{t}}|} {||{\lambda _l^t}|{|^2}}  \! +\! \frac{4}{{{{\eta _{\boldsymbol{\theta }}}}}}(\frac{{{c_2^{t-2}}}}{{{c_2^{t-1}}}} \! - \! \frac{{{c_2^{t-1}}}}{{{c_2^t}}})\sum\limits_{i = 1}^N {||{{\boldsymbol{\theta}}_i^t}|{|^2}}.
\end{array}
\end{equation}
which concludes the proof of Lemma 3.

\vspace{5mm}

\emph{\textbf{Proof of Theorem 1:}}

First, we set that,
\begin{equation}
\label{eq:A68_11_16}
{a_6^t} =  \frac{{4|{\boldsymbol{\mathcal{P}}^{t}}|(\gamma-2){L^2}}}{{{{\eta _{\lambda}}}({c_1^t})^2}}+
\frac{{4N(\gamma-2){L^2}}}{{{{\eta _{\boldsymbol{\theta }}}}({c_2^t})^2}} +\frac{{{{\eta _{\boldsymbol{\theta }}}}(N-|{\boldsymbol{\mathcal{Q}}^{t + 1}}|){L^2}}}{2}  - \frac{a_5}{2},
\end{equation}
where constant $\gamma$ satisfies that $\gamma>2$ and $\frac{{4(\gamma-2){L^2}}}{{{{\eta _{\lambda}}}({c_1^0})^2}}+
\frac{{4N(\gamma-2){L^2}}}{{{{\eta _{\boldsymbol{\theta }}}}({c_2^0})^2}} > \frac{a_5}{2}$, thus we have that $a_6^t>0,\forall t$. According to the setting of ${\eta _{\boldsymbol{x}}^t}$, ${\eta _{\boldsymbol{y}}^t}$, ${\eta _{\boldsymbol{v}}^t}$, ${\eta _{\boldsymbol{z}}^t}$ and $c_1^t$, $c_2^t$, we have,
\begin{equation}
\label{eq:A68}
\frac{L\! +\!a_5}{2} - \frac{1}{{{\eta _{\boldsymbol{x}}^t}}} \! +\! \frac{{{{\eta _{\lambda}}}|{\boldsymbol{\mathcal{P}}^{t}}|{L^2}}}{2} \! +\! \frac{{{{\eta _{\boldsymbol{\theta }}}}|{\boldsymbol{\mathcal{Q}}^{t + 1}}|{L^2}}}{2} \! +\! \frac{{8|{\boldsymbol{\mathcal{P}}^{t}}|{L^2}}}{{{{\eta _{\lambda}}}({c_1^t})^2}} \! +\! \frac{{8N{L^2}}}{{{{\eta _{\boldsymbol{\theta }}}}({c_2^t})^2}} =  - {a_6^t},
\end{equation}

\begin{equation}
\label{eq:A69}
\frac{L\! +\!a_5}{2} - \frac{1}{{{\eta _{\boldsymbol{y}}^t}}} \! +\! \frac{{{{\eta _{\lambda}}}|{\boldsymbol{\mathcal{P}}^{t}}|{L^2}}}{2} \! +\! \frac{{{{\eta _{\boldsymbol{\theta }}}}|{\boldsymbol{\mathcal{Q}}^{t + 1}}|{L^2}}}{2} \! +\! \frac{{8|{\boldsymbol{\mathcal{P}}^{t}}|{L^2}}}{{{{\eta _{\lambda}}}({c_1^t})^2}} \! +\! \frac{{8N{L^2}}}{{{{\eta _{\boldsymbol{\theta }}}}({c_2^t})^2}} =  - {a_6^t},
\end{equation}

\begin{equation}
\label{eq:A70}
\frac{L\! +\!a_5}{2} - \frac{1}{{{\eta _{\boldsymbol{v}}^t}}} \! +\! \frac{{{{\eta _{\lambda}}}|{\boldsymbol{\mathcal{P}}^{t}}|{L^2}}}{2} \! +\! \frac{{{{\eta _{\boldsymbol{\theta }}}}|{\boldsymbol{\mathcal{Q}}^{t + 1}}|{L^2}}}{2} \! +\! \frac{{8|{\boldsymbol{\mathcal{P}}^{t}}|{L^2}}}{{{{\eta _{\lambda}}}({c_1^t})^2}} \! +\! \frac{{8N{L^2}}}{{{{\eta _{\boldsymbol{\theta }}}}({c_2^t})^2}} =  - {a_6^t},
\end{equation}

\begin{equation}
\label{eq:A70_1}
\frac{L\! +\!a_5}{2} - \frac{1}{{{\eta _{\boldsymbol{z}}^t}}} \! +\! \frac{{{{\eta _{\lambda}}}|{\boldsymbol{\mathcal{P}}^{t}}|{L^2}}}{2} \! +\! \frac{{{{\eta _{\boldsymbol{\theta }}}}|{\boldsymbol{\mathcal{Q}}^{t + 1}}|{L^2}}}{2} \! +\! \frac{{8|{\boldsymbol{\mathcal{P}}^{t}}|{L^2}}}{{{{\eta _{\lambda}}}({c_1^t})^2}} \! +\! \frac{{8N{L^2}}}{{{{\eta _{\boldsymbol{\theta }}}}({c_2^t})^2}} =  - {a_6^t}.
\end{equation}

Combining Eq. (\ref{eq:A68}), (\ref{eq:A69}), (\ref{eq:A70}), (\ref{eq:A70_1})  with Lemma \ref{lemma3}, $\forall t \ge T_1$, we can obtain that,
{\begin{equation}
\label{eq:A71}
\begin{array}{l}
{a_6^t}\sum\limits_{i = 1}^N \! {(||{{\boldsymbol{x}}_i^{t+1}} - {{\boldsymbol{x}}_i^t}|{|^2}} + {||{{\boldsymbol{y}}_i^{t+1}} - {{\boldsymbol{y}}_i^t}|{|^2})} \! +\! {a_6^t}||\boldsymbol{v}^{t+1} - \boldsymbol{v}^t|{|^2} \! +\! {a_6^t}||\boldsymbol{z}^{t+1} - \boldsymbol{z}^t|{|^2} \\
 \! + {(\frac{1}{{10{{\eta _{\lambda}}}}} - \frac{{6\tau{ k_1 }N{L^2}}}{2})  \sum\limits_{l = 1}^{|{\boldsymbol{\mathcal{P}}^{t}}|}||{\lambda _l^{t+1}} - {\lambda _l^t}|{|^2}}   \! +\! \frac{1}{{10{{\eta _{\boldsymbol{\theta }}}}}}\sum\limits_{i = 1}^N {||{{\boldsymbol{\theta}}_i^{t+1}} - {{\boldsymbol{\theta}}_i^t}|{|^2}} \\
 \le F^{t} - F^{t+1} \! +\! \frac{{{c_1^{t-1}} - {c_1^t}}}{2}\sum\limits_{l = 1}^{|{\boldsymbol{\mathcal{P}}^{t}}|} {||{\lambda _l^{t+1}}|{|^2}}  \! +\! \frac{{{c_2^{t-1}} - {c_2^t}}}{2}\sum\limits_{i = 1}^N {||{{\boldsymbol{\theta}}_i^{t+1}}|{|^2}} \\
  +  {\frac{4}{{{{\eta _{\lambda}}}}}(\frac{{{c_1^{t-2}}}}{{{c_1^{t-1}}}} - \frac{{{c_1^{t-1}}}}{{{c_1^t}}})\sum\limits_{l = 1}^{|{\boldsymbol{\mathcal{P}}^{t}}|}||{\lambda _l^t}|{|^2}}  \! +\!  {\frac{4}{{{{\eta _{\boldsymbol{\theta }}}}}}(\frac{{{c_2^{t-2}}}}{{{c_2^{t-1}}}} - \frac{{{c_2^{t-1}}}}{{{c_2^t}}})\sum\limits_{i = 1}^N||{{\boldsymbol{\theta}}_i^t}|{|^2}}.
\end{array}
\end{equation}}

Utilizing the definition of ${(\nabla \widetilde{G}^t)_{{{\boldsymbol{x}}_i}}}$ and combining it with trigonometric inequality, Cauchy-Schwarz inequality and Assumption 1 and 2, we can obtain that,
\begin{equation}
\label{eq:A72}
\begin{array}{l}
||{(\nabla {\widetilde{G}}^t)_{{{\boldsymbol{x}}_i}}}|{|^2}
 \le \frac{2}{{\eta _{\boldsymbol{x}}}^2}||{{\boldsymbol{x}}_i^{\overline{{t}_i}}} \!-\! {{\boldsymbol{x}}_i^t}|{|^2} \!+\! {{6L^2\tau{ k_1 }}}(||{\boldsymbol{v}^{t+1}} \!-\! {\boldsymbol{v}^{t}}|{|^2} \! + \! ||{\boldsymbol{z}^{t+1}} \!-\! {\boldsymbol{z}^{{t}}}|{|^2} \! + \! \sum\limits_{l = 1}^{|{\boldsymbol{\mathcal{P}}^{t}}|} {||{\lambda _l^{t+1}} \!-\! {\lambda _l^{t}}|{|^2}} ).
\end{array}
\end{equation}

Utilizing the definition of ${(\nabla \widetilde{G}^t)_{{{\boldsymbol{y}}_i}}}$ and combining it with trigonometric inequality and Cauchy-Schwarz inequality, it follows that,
\begin{equation}
\label{eq:A72_2}
\begin{array}{l}
||{(\nabla {\widetilde{G}}^t)_{{{\boldsymbol{y}}_i}}}|{|^2} \\
\! \le \! \frac{2}{{\eta _{\boldsymbol{y}}}^2}||{{\boldsymbol{y}}_i^{\overline{{t}_i}}} \!-\! {{\boldsymbol{y}}_i^t}|{|^2} \!+\!  {{6L^2\tau{ k_1 }}}(||{\boldsymbol{v}^{t+1}} \!-\! {\boldsymbol{v}^{t}}|{|^2} \! + \! ||{\boldsymbol{z}^{t+1}} \!-\! {\boldsymbol{z}^{{t}}}|{|^2} \! + \! \sum\limits_{l = 1}^{|{\boldsymbol{\mathcal{P}}^{t}}|} \! {||{\lambda _l^{t+1}} \!-\! {\lambda _l^{t}}|{|^2}} ).
\end{array}
\end{equation}

Utilizing the definition of ${(\nabla {\widetilde{G}}^t)_{\boldsymbol{v}}}$ and combining it with trigonometric inequality and Cauchy-Schwarz inequality, we have that,
\begin{equation}
\label{eq:A73}
\begin{array}{l}
||{(\nabla {\widetilde{G}}^t)_{\boldsymbol{v}}}|{|^2}

\! \le \! 2{L^2}\sum\limits_{i = 1}^N ({||{{\boldsymbol{x}}_i^{t+1}} - {{\boldsymbol{x}}_i^t}|{|^2} \!+\!  ||{{\boldsymbol{y}}_i^{t+1}} - {{\boldsymbol{y}}_i^t}|{|^2})}  \! +\! \frac{2}{{\eta _{\boldsymbol{v}}}^2}||\boldsymbol{v}^{t+1} - \boldsymbol{v}^t|{|^2}.
\end{array}
\end{equation}

Using the definition of ${(\nabla {\widetilde{G}}^t)_{\boldsymbol{z}}}$ and combining it with trigonometric inequality and Cauchy-Schwarz inequality, it follows that,
\begin{equation}
\label{eq:A73_2}
\begin{array}{l}
||{(\nabla {\widetilde{G}}^t)_{\boldsymbol{z}}}|{|^2}

\! \le \! 2{L^2}\left(\sum\limits_{i = 1}^N ({||{{\boldsymbol{x}}_i^{t+1}} - {{\boldsymbol{x}}_i^t}|{|^2} \!+\!  ||{{\boldsymbol{y}}_i^{t+1}} - {{\boldsymbol{y}}_i^t}|{|^2})} \!+\! ||\boldsymbol{v}^{t+1} - \boldsymbol{v}^t|{|^2} \right)  \! +\! \frac{2}{{\eta _{\boldsymbol{z}}}^2}||\boldsymbol{z}^{t+1} - \boldsymbol{z}^t|{|^2}.
\end{array}
\end{equation}

Using the definition of ${(\nabla {\widetilde{G}}^t)_{{\lambda _l}}}$ and combining it with trigonometric inequality and Cauchy-Schwarz inequality, we can obtain the following inequality,
{\begin{equation}
\label{eq:A76}
\begin{array}{l}
||{(\nabla {\widetilde{G}}^t)_{{\lambda _l}}}|{|^2}
\! \le \! \frac{3}{{{\eta _{\lambda}}^2}}||{\lambda _l^{t+1}} \! -\! {\lambda _l^t}|{|^2}
  + 3{(({c_1^{t-1}})^2  - ({c_1^t})^2)}||{\lambda _l^t}|{|^2}\\
 \qquad \qquad \qquad + 3{L^2}\left(\sum\limits_{i = 1}^N ({||{{\boldsymbol{x}}_i^{t+1}} - {{\boldsymbol{x}}_i^t}|{|^2} \!+\!  ||{{\boldsymbol{y}}_i^{t+1}} - {{\boldsymbol{y}}_i^t}|{|^2})} \!+\! ||\boldsymbol{v}^{t+1} - \boldsymbol{v}^t|{|^2} \!+\! ||\boldsymbol{z}^{t+1} - \boldsymbol{z}^t|{|^2} \right).
\end{array}
\end{equation}}

Combining the definition of ${(\nabla {\widetilde{G}}^t)_{{{\boldsymbol{\theta }_i}}}}$ with Cauchy-Schwarz inequality and Assumption 2, we have,
{\begin{equation}
\label{eq:A77}
\begin{array}{l}
||{(\nabla {\widetilde{G}}^t)_{{{\boldsymbol{\theta }_i}}}}|{|^2}\\

\! \le \! \frac{3}{{{\eta _{\boldsymbol{\theta}}}^2}}||{{\boldsymbol{\theta }_i^{\overline{{t}_i}}}} \! -\! {{\boldsymbol{\theta }_i^t}}|{|^2} \! +\! 3{L^2}\left(\sum\limits_{i = 1}^N\! {(||{{\boldsymbol{x}}_i^{\overline{{t}_i}}} \! -\! {{\boldsymbol{x}}_i^t}|{|^2}\!+\!||{{\boldsymbol{y}}_i^{\overline{{t}_i}}} \! -\! {{\boldsymbol{y}}_i^t}|{|^2})}  \! +\! ||{\boldsymbol{v}}^{\overline{{t}_i}} \! -\! {\boldsymbol{v}^t}|{|^2}\right)
 \! +\! 3{({c_2^{\hat{{t}}_i \! -\! 1}} \! -\! {c_2^{\overline{{t}_i}\! -\!1}})^2}||{{\boldsymbol{\theta }_i^t}}|{|^2}\\
 
\! \le\! \frac{3}{{{\eta _{\boldsymbol{\theta}}}^2}}||{{\boldsymbol{\theta }_i^{\overline{{t}_i}}}} \! -\! {{\boldsymbol{\theta }_i^t}}|{|^2} \! +\! 3{L^2}\sum\limits_{i = 1}^N(||{{\boldsymbol{x}}_i^{\overline{{t}_i}}} \! -\! {{\boldsymbol{x}}_i^t}|{|^2}\!+\!||{{\boldsymbol{y}}_i^{\overline{{t}_i}}} \! -\! {{\boldsymbol{y}}_i^t}|{|^2})\\
+ {{3L^2\tau{ k_1 }}}(||{\boldsymbol{v}^{t+1}} \!-\! {\boldsymbol{v}^{t}}|{|^2} \! + \! ||{\boldsymbol{z}^{t+1}} \!-\! {\boldsymbol{z}^{{t}}}|{|^2} \! + \! \sum\limits_{l = 1}^{|{\boldsymbol{\mathcal{P}}^{t}}|} {||{\lambda _l^{t+1}} \!-\! {\lambda _l^{t}}|{|^2}} )  + 3{(({c_2^{\hat{{t}}_i \! -\! 1}})^2 \! -\! ({c_2^{\overline{{t}_i}\! -\!1}})^2)}||{{\boldsymbol{\theta }_i^t}}|{|^2}.
\end{array}
\end{equation}}

In sight of the Definition \ref{definition:A2} as well as Eq. (\ref{eq:A72}), (\ref{eq:A72_2}), (\ref{eq:A73}), (\ref{eq:A73_2}), (\ref{eq:A76}) and Eq. (\ref{eq:A77}), we can obtain that,
{\begin{equation}
\label{eq:A78}
\begin{array}{l}
||\nabla \widetilde{G}^t|{|^2}\\  = \sum\limits_{i = 1}^N ({||{{(\nabla \widetilde{G}^t)}_{{{\boldsymbol{x}}_i}}}|{|^2}}  \! +\! {||{{(\nabla \widetilde{G}^t)}_{{{\boldsymbol{y}}_i}}}|{|^2}} \!+\! {||{{(\nabla \widetilde{G}^t)}_{{\boldsymbol{\theta }_i}}}|{|^2}})  \!+\! ||{(\nabla \widetilde{G}^t)_{\boldsymbol{v}}}|{|^2}  \!+\! ||{(\nabla \widetilde{G}^t)_{\boldsymbol{z}}}|{|^2} \! + \! \sum\limits_{l = 1}^{|{\boldsymbol{\mathcal{P}}^{t}}|}\! {||{{(\nabla \widetilde{G}^t)}_{{\lambda _l}}}|{|^2}} \\
 
 \le (\frac{2}{{\eta _{\boldsymbol{x}}}^2} \! +\! 3NL^2)\sum\limits_{i = 1}^N {||{{\boldsymbol{x}}_i^{\overline{{t}_i}}} - {{\boldsymbol{x}}_i^t}|{|^2}}\! +\! (\frac{2}{{\eta _{\boldsymbol{y}}}^2}\! +\! 3NL^2)\sum\limits_{i = 1}^N {||{{\boldsymbol{y}}_i^{\overline{{t}_i}}} - {{\boldsymbol{y}}_i^t}|{|^2}}\\

  + (4 \! +\! 3|{\boldsymbol{\mathcal{P}}^{t}}|{L^2})\sum\limits_{i = 1}^N {||{{\boldsymbol{x}}_i^{t+1}} - {{\boldsymbol{x}}_i^t}|{|^2}} + (4 \! +\! 3|{\boldsymbol{\mathcal{P}}^{t}}|{L^2})\sum\limits_{i = 1}^N {||{{\boldsymbol{y}}_i^{t+1}} - {{\boldsymbol{y}}_i^t}|{|^2}} \\

 + (\frac{2}{{\eta _{\boldsymbol{v}}}^2} \! +\! (2 \! +\! 15\tau{ k_1 }N \! +\! 3|{\boldsymbol{\mathcal{P}}^{t}}|){L^2})||\boldsymbol{v}^{t+1} \!-\! \boldsymbol{v}^t|{|^2}
 \! +\! (\frac{2}{{\eta _{\boldsymbol{z}}}^2} \! +\! (15\tau{ k_1 }N \! +\! 3|{\boldsymbol{\mathcal{P}}^{t}}|){L^2})||\boldsymbol{z}^{t+1}\! -\! \boldsymbol{z}^t|{|^2} \\ 
 
 + \sum\limits_{l = 1}^{|{\boldsymbol{\mathcal{P}}^{t}}|} {(\frac{3}{{{\eta _{\lambda}}^2}} \! +\! 15\tau{ k_1 }N{L^2})||{\lambda _l^{t+1}} - {\lambda _l^t}|{|^2}}  \! +\! \sum\limits_{l = 1}^{|{\boldsymbol{\mathcal{P}}^{t}}|} {3{{(({c_1^{t-1}})^2 - ({c_1^t})^2)}}||{\lambda _l^t}|{|^2}} \\

 + \sum\limits_{i = 1}^N {\frac{3}{{{\eta _{\boldsymbol{\theta}}}^2}}||{{\boldsymbol{\theta }_i^{\overline{{t}_i}}}} - {{\boldsymbol{\theta }_i^t}}|{|^2}}  \! +\! \sum\limits_{i = 1}^N {3{(({c_2^{\hat{{t}}_i - 1}})^2 - ({c_2^{\overline{{t}_i}-1}})^2)}||{{\boldsymbol{\theta }_i^t}}|{|^2}} .
\end{array}
\end{equation}}

Let constant $\underline{a_6}$ denote the lower bound of $a_6^t$ ($\underline{a_6}>0$), and we set constants $d_1, d_2, d_3, d_4$ that,
\begin{equation}
\label{eq:A79}{d_1} = \frac{{2 k_{\tau} \tau \! +\! (4 \! +\! 3M \! +\! 3 k_{\tau}  \tau N){L^2}{\eta _{\boldsymbol{x}}}^2}}{{{\eta _{\boldsymbol{x}}}^2({\underline{a_6}}){^2}}} \ge \frac{{2 k_{\tau}  \tau \! +\! (4 \! +\! 3|{\boldsymbol{\mathcal{P}}^{t}}| \! +\! 3k_{\tau}   \tau N){L^2}{\eta _{\boldsymbol{x}}}^2}}{{{\eta _{\boldsymbol{x}}}^2({a_6^t}){^2}}},
\end{equation}
\begin{equation}
\label{eq:A79_2}{d_2} = \frac{{2 k_{\tau} \tau \! +\! (4 \! +\! 3M \! +\! 3 k_{\tau}  \tau N){L^2}{\eta _{\boldsymbol{y}}}^2}}{{{\eta _{\boldsymbol{y}}}^2({\underline{a_6}}){^2}}} \ge \frac{{2 k_{\tau}  \tau \! +\! (4 \! +\! 3|{\boldsymbol{\mathcal{P}}^{t}}| \! +\! 3k_{\tau}   \tau N){L^2}{\eta _{\boldsymbol{y}}}^2}}{{{\eta _{\boldsymbol{y}}}^2({a_6^t}){^2}}},
\end{equation}
\begin{equation}
\label{eq:A80}
{d_3} = \frac{{2 \! +\! (2 \! +\!{15\tau{ k_1 }N} \! +\! 3M){L^2}{\eta _{\boldsymbol{v}}}^2}}{{{\eta _{\boldsymbol{v}}}^2({\underline{a_6}}){^2}}} \ge \frac{{2 \! +\! (2 \! +\!{15\tau{ k_1 }N} \! +\! 3|{\boldsymbol{\mathcal{P}}^{t}}|){L^2}{\eta _{\boldsymbol{v}}}^2}}{{{\eta _{\boldsymbol{v}}}^2({a_6^t}){^2}}},
\end{equation}
\begin{equation}
\label{eq:A81}
{d_4} = \frac{{2 \! +\! ({15\tau{ k_1 }N} \! +\! 3M){L^2}{\eta _{\boldsymbol{z}}}^2}}{{{\eta _{\boldsymbol{z}}}^2({\underline{a_6}}){^2}}} \ge \frac{{2 \! +\! ({15\tau{ k_1 }N} \! +\! 3|{\boldsymbol{\mathcal{P}}^{t}}|){L^2}{\eta _{\boldsymbol{z}}}^2}}{{{\eta _{\boldsymbol{z}}}^2({a_6^t}){^2}}},
\end{equation}

\vspace{1ex}

\noindent where  $k_{\tau}$ is a positive constant. Thus, combining Eq. (\ref{eq:A78}) with Eq. (\ref{eq:A79}), Eq. (\ref{eq:A79_2}), (\ref{eq:A80}), (\ref{eq:A81}), we can obtain,
{\begin{equation}
\label{eq:A82}
\begin{array}{l}
||\nabla {\widetilde{G}}^t|{|^2} 
\! \le \! \sum\limits_{i = 1}^N {{d_1}({a_6^t}){^2}||{{\boldsymbol{x}}_i^{t+1}} - {{\boldsymbol{x}}_i^t}|{|^2}} + \sum\limits_{i = 1}^N{{d_2}({a_6^t}){^2}||{{\boldsymbol{y}}_i^{t+1}} - {{\boldsymbol{y}}_i^t}|{|^2}} \vspace{0.5ex}\\

\quad \quad \quad \; \; \; \! + {d_3}({a_6^t}){^2}||{\boldsymbol{v}}^{t + 1} - {\boldsymbol{v}^t}|{|^2} \! +\! {d_4}({a_6^t}){^2}||{\boldsymbol{z}}^{t + 1} - {\boldsymbol{z}^t}|{|^2}\! +\! \sum\limits_{i = 1}^N  \! {\frac{3}{{{\eta _{\boldsymbol{\theta}}}^2}}||{{\boldsymbol{\theta }_i}^{\overline{{t}_i}}} - {{\boldsymbol{\theta }_i^t}}|{|^2}}\\

\quad \quad \quad \; \; \; \! +\! \sum\limits_{l = 1}^{|{\boldsymbol{\mathcal{P}}^{t}}|}\! {(\frac{3}{{{\eta _{\lambda}}^2}} \! +\! 15\tau{ k_1 }N{L^2})||{\lambda _l^{t+1}}\! -\! {\lambda _l^t}|{|^2}}  \! +\! \sum\limits_{l = 1}^{|{\boldsymbol{\mathcal{P}}^{t}}|} {3(({c_1^{t - 1}}){^2} \!-\! ({c_1^t})^2)||{\lambda _l^t}|{|^2}}   \\ 
 
\quad \quad \quad \; \; \;  \! +\! \sum\limits_{i = 1}^N {3{(({c_2^{\hat{{t}}_i - 1}})^2 - ({c_2^{\overline{{t}_i}-1}})^2)}||{{\boldsymbol{\theta }_i^t}}|{|^2}} \! +\! (\frac{2}{{\eta _{\boldsymbol{x}}}^2} \! +\! 3NL^2)\sum\limits_{i = 1}^N {||{{\boldsymbol{x}}_i^{\overline{{t}_i}}} - {{\boldsymbol{x}}_i^t}|{|^2}} \\
 
\quad \quad \quad \; \; \; \! - (\frac{2 k_{\tau} \tau}{{\eta _{\boldsymbol{x}}}^2} \! +\! 3 k_{\tau} \tau NL^2)\sum\limits_{i = 1}^N {||{{\boldsymbol{x}}_i^{t+1}} \! - \! {{\boldsymbol{x}}_i^t}|{|^2}} \! +\! (\frac{2}{{\eta _{\boldsymbol{y}}}^2} \! +\! 3NL^2)\sum\limits_{i = 1}^N {||{{\boldsymbol{y}}_i^{\overline{{t}_i}}} - {{\boldsymbol{y}}_i^t}|{|^2}} \\

\quad \quad \quad \; \; \; \! - (\frac{2 k_{\tau} \tau}{{\eta _{\boldsymbol{y}}}^2} \! +\! 3 k_{\tau} \tau NL^2)\sum\limits_{i = 1}^N {||{{\boldsymbol{y}}_i^{t+1}} \! - \! {{\boldsymbol{y}}_i^t}|{|^2}}.
\end{array}
\end{equation}}

Let $d_5^t$ denote a nonnegative sequence, \textit{i.e.}, ${d_5^t} = \frac{1}{{\max \{ {d_1}{a_6^t},{d_2}{a_6^t},{d_3}{a_6^t}, {d_4}{a_6^t},\frac{{\frac{{30}}{{{{\eta _{\lambda}}}}} + 150{{\eta _{\lambda}}}\tau{ k_1 }N{L^2}}}{{1 - 30{{\eta _{\lambda}}}\tau{ k_1 }N{L^2}}},\frac{{30 \tau }}{{{{\eta _{\boldsymbol{\theta }}}}}}\} }}$. We denote the upper and lower bound of $d_5^t$ as $\overline{d_5}$ and $\underline{d_5}$, respectively.  And we set the constant $k_{\tau}$ satisfies $k_{\tau} \ge \max \{ \frac{{{\overline{d_5}}}(\frac{2}{{\eta _{\boldsymbol{y}}}^2} + 3NL^2) }{{\underline{d_5}}(\frac{2}{\overline{\eta _{\boldsymbol{y}}}^2} + 3NL^2)}, \frac{{{\overline{d_5}}}(\frac{2}{{\eta _{\boldsymbol{x}}}^2} + 3NL^2) }{{\underline{d_5}}(\frac{2}{\overline{\eta _{\boldsymbol{x}}}^2} + 3NL^2)} \}$, where ${\overline{\eta _{\boldsymbol{x}}}}$ and ${\overline{\eta _{\boldsymbol{y}}}} $ are the upper bounds of ${\eta _{\boldsymbol{x}}^t}$ and ${\eta _{\boldsymbol{y}}^t}$, respectively.  We can obtain the following inequality by combining Eq. (\ref{eq:A82}) with the definition of $d_5^t$:
\begin{equation}
\label{eq:A84}
\begin{array}{l}
{d_5^t}||\nabla {\widetilde{G}}^t|{|^2}
 \!\le\! {a_6^t} \! \sum\limits_{i = 1}^N \!( {||{{\boldsymbol{x}}_i^{t+1}}\! -\! {{\boldsymbol{x}}_i^t}|{|^2}}  \! +\! {||{{\boldsymbol{y}}_i^{t+1}}\! -\! {{\boldsymbol{y}}_i^t}|{|^2}} )  \! +\! {a_6^t}||{\boldsymbol{v}}^{t + 1}\! -\! {\boldsymbol{v}^t}|{|^2} \! +\! {a_6^t}||{\boldsymbol{z}}^{t + 1}\! -\! {\boldsymbol{z}^t}|{|^2}\\
 
\quad \quad \quad \quad \; \; \;  \! +  {(\frac{1}{{10{{\eta _{\lambda}}}}}\! -\! \frac{{6\tau{ k_1 }N{L^2}}}{2})\sum\limits_{l = 1}^{|{\boldsymbol{\mathcal{P}}^{t}}|}||{\lambda _l^{t+1}} - {\lambda _l^t}|{|^2}}   + \frac{1}{{10\tau {{\eta _{\boldsymbol{\theta }}}}}}\sum\limits_{i = 1}^N {||{{\boldsymbol{\theta }_i}^{\overline{{t}_i}}} - {{\boldsymbol{\theta }_i^t}}|{|^2}} \\

\quad \quad \quad \quad \; \; \; \! + \sum\limits_{l = 1}^{|{\boldsymbol{\mathcal{P}}^{t}}|}3{d_5^t}(({c_1^{t-1}}){^2} - ({c_1^t}){^2}) {||{\lambda _l^t}|{|^2}}   + \sum\limits_{i = 1}^N 3{d_5^t}(({c_2^{\hat{{t}}_i - 1}}){^2} - ({c_2^{\overline{{t}_i}-1}}){^2}) {||{{\boldsymbol{\theta }_i^t}}|{|^2}} \\

\quad \quad \quad \quad  \; \; \; \! +{d_5^t} (\frac{2}{{\eta _{\boldsymbol{x}}}^2} \! +\! 3NL^2)\sum\limits_{i = 1}^N {||{{\boldsymbol{x}}_i^{\overline{{t}_i}}}\! -\! {{\boldsymbol{x}}_i^t}|{|^2}} - {d_5^t}(\frac{2 k_{\tau}  \tau}{{\eta _{\boldsymbol{x}}}^2} \! +\! 3 k_{\tau}   \tau NL^2)\sum\limits_{i = 1}^N {||{{\boldsymbol{x}}_i^{t+1}}\! -\! {{\boldsymbol{x}}_i^t}|{|^2}}\\

\quad \quad \quad \quad  \; \; \; \! +{d_5^t} (\frac{2}{{\eta _{\boldsymbol{y}}}^2} \! +\! 3NL^2)\sum\limits_{i = 1}^N {||{{\boldsymbol{y}}_i^{\overline{{t}_i}}}\! -\! {{\boldsymbol{y}}_i^t}|{|^2}} - {d_5^t}(\frac{2 k_{\tau}  \tau}{{\eta _{\boldsymbol{y}}}^2} \! +\! 3 k_{\tau}   \tau NL^2)\sum\limits_{i = 1}^N {||{{\boldsymbol{y}}_i^{t+1}}\! -\! {{\boldsymbol{y}}_i^t}|{|^2}}.

\end{array}
\end{equation}

Combining  the definition of $d_5^t$ with Eq. (\ref{eq:A71}) and  according to the setting $||{\lambda _l^t}|{|^2} \le \alpha_3$, $||{{\boldsymbol{\theta }_i^t}}|{|^2} \le \alpha_4$ and ${\overline{d_5}} \ge {d_5^t} \ge \underline{d_5}, \forall t  \ge T_1$, thus, we have,
\begin{equation}
\label{eq:A85}
\begin{array}{l}
{d_5^t}||\nabla {\widetilde{G}}^t|{|^2} \vspace{0.5ex} \\
  \le F^{t} - F^{t+1} + \frac{{{c_1^{t-1}} - {c_1^t}}}{2}M\alpha_3  + \frac{{{c_2^{t-1}} - {c_2^t}}}{2}N\alpha_4 
 + \frac{4}{{{{\eta _{\lambda}}}}}(\frac{{{c_1^{t-2}}}}{{{c_1^{t-1}}}} - \frac{{{c_1^{t-1}}}}{{{c_1^t}}})M\alpha_3 \vspace{1ex}\\
 + \frac{4}{{{{\eta _{\boldsymbol{\theta }}}}}}(\frac{{{c_2^{t-2}}}}{{{c_2^{t-1}}}} - \frac{{{c_2^{t-1}}}}{{{c_2^t}}})N\alpha_4 
 + 3{\overline{d_5}}(({c_1^{t-1}}){^2} - ({c_1^t}){^2})M\alpha_3 + 3{\overline{d_5}}\sum\limits_{i = 1}^N(({c_2^{\hat{{t}}_i - 1}}){^2} - ({c_2^{\overline{{t}_i}-1}}){^2})\alpha_4 \vspace{0.5ex}\\
 
  + \frac{1}{{10\tau {{\eta _{\boldsymbol{\theta }}}}}}\sum\limits_{i = 1}^N {||{{\boldsymbol{\theta }_i}^{\overline{{t}_i}}} - {{\boldsymbol{\theta }_i^t}}|{|^2}} - \frac{1}{{10{{\eta _{\boldsymbol{\theta }}}}}}\sum\limits_{i = 1}^N {||{{\boldsymbol{\theta }_i}^{t+1}} - {{\boldsymbol{\theta }_i^t}}|{|^2}}\\
 
  +{\overline{d_5}} (\frac{2}{{\eta _{\boldsymbol{x}}}^2} + 3NL^2)\sum\limits_{i = 1}^N {||{{\boldsymbol{x}}_i^{\overline{{t}_i}}} - {{\boldsymbol{x}}_i^t}|{|^2}} - \underline{d_5}(\frac{2 k_{\tau}  \tau}{{\eta _{\boldsymbol{x}}}^2} + 3 k_{\tau}   \tau NL^2)\sum\limits_{i = 1}^N {||{{\boldsymbol{x}}_i^{t+1}} - {{\boldsymbol{x}}_i^t}|{|^2}}\\
 
  +{\overline{d_5}} (\frac{2}{{\eta _{\boldsymbol{y}}}^2} + 3NL^2)\sum\limits_{i = 1}^N {||{{\boldsymbol{y}}_i^{\overline{{t}_i}}} - {{\boldsymbol{y}}_i^t}|{|^2}} - \underline{d_5}(\frac{2 k_{\tau}  \tau}{{\eta _{\boldsymbol{y}}}^2} + 3 k_{\tau}   \tau NL^2)\sum\limits_{i = 1}^N {||{{\boldsymbol{y}}_i^{t+1}} - {{\boldsymbol{y}}_i^t}|{|^2}}.

\end{array}
\end{equation}

Denoting $\widetilde{T}(\epsilon )$ as $\widetilde{T}(\epsilon ) = \min \{ t \ | \; ||\nabla \widetilde{G}^{T_1+t}||^2 \le \frac{\epsilon}{4}, t\ge 2 \}$. Summing up Eq. (\ref{eq:A85}) from $t=T_1+2$ to $t =T_1+{{\widetilde{T}} (\epsilon )} $, we have,
\begin{equation}
\label{eq:A86}
\begin{array}{l}
\sum\limits_{t = T_1 + 2}^{T_1 + {\widetilde{T}} (\epsilon )} {{d_5^t}||\nabla {\widetilde{G}}^t|{|^2}} \vspace{1ex}\\
 
 \le F^{T_1 + 2} - \mathop L\limits_ - 
 + \frac{4}{{{{\eta _{\lambda}}}}}(\frac{{{c_1^0}}}{{{c_1^1}}} + \frac{{{c_1^1}}}{{{c_1^2}}})M {\alpha_3}  + \frac{{{c_1^1}}}{2}M {\alpha_3}  + \frac{7}{{2{{\eta _{\lambda}}}}}M {{\sigma _3}^2}  + 3{\overline{d_5}} ({{c_1^1}){^2}M {\alpha_3} } \vspace{0.5ex}\\
 + \frac{4}{{{{\eta _{\boldsymbol{\theta }}}}}}(\frac{{{c_1^0}}}{{{c_1^1}}} + \frac{{{c_1^1}}}{{{c_1^2}}})N {\alpha_4}  + \frac{{{c_2^1}}}{2}N {\alpha_4}  + \frac{7}{{2{{\eta _{\boldsymbol{\theta }}}}}}N {{\sigma _4}^2}  + \sum\limits_{i = 1}^N \sum\limits_{t = T_1 + 2}^{T_1 + {\widetilde{T}} (\epsilon )}3{\overline{d_5}}(({c_2^{\hat{{t}}_i - 1}}){^2} - ({c_2^{\overline{{t}_i}-1}}){^2})\alpha_4 \vspace{0.5ex}\\

+ \frac{c_1^{T_1+2}}{2}M {{\sigma _3}^2} + \frac{c_2^{T_1+2}}{2}N {{\sigma _4}^2} + \frac{1}{{10\tau {{\eta _{\boldsymbol{\theta }}}}}}\sum\limits_{t = T_1 + 2}^{T_1 + {\widetilde{T}} (\epsilon )}\sum\limits_{i = 1}^N {||{{\boldsymbol{\theta }_i}^{\overline{{t}_i}}} - {{\boldsymbol{\theta }_i^t}}|{|^2}} - \frac{1}{{10{{\eta _{\boldsymbol{\theta }}}}}}\sum\limits_{t = T_1 + 2}^{T_1 + {\widetilde{T}} (\epsilon )}\sum\limits_{i = 1}^N {||{{\boldsymbol{\theta }_i}^{t+1}} - {{\boldsymbol{\theta }_i^t}}|{|^2}}\\
 
  +{\overline{d_5}} (\frac{2}{{\eta _{\boldsymbol{x}}}^2} + 3NL^2)\sum\limits_{t = T_1 + 2}^{T_1 + {\widetilde{T}} (\epsilon )}\sum\limits_{i = 1}^N {||{{\boldsymbol{x}}_i^{\overline{{t}_i}}} - {{\boldsymbol{x}}_i^t}|{|^2}} - \underline{d_5}(\frac{2 k_{\tau} \tau}{\overline{\eta _{\boldsymbol{x}}}^2} + 3 k_{\tau} \tau NL^2)\sum\limits_{t = T_1 + 2}^{T_1 + {\widetilde{T}} (\epsilon )}\sum\limits_{i = 1}^N {||{{\boldsymbol{x}}_i^{t+1}} - {{\boldsymbol{x}}_i^t}|{|^2}}\\
 
  +{\overline{d_5}} (\frac{2}{{\eta _{\boldsymbol{y}}}^2} + 3NL^2)\sum\limits_{t = T_1 + 2}^{T_1 + {\widetilde{T}} (\epsilon )}\sum\limits_{i = 1}^N {||{{\boldsymbol{y}}_i^{\overline{{t}_i}}} - {{\boldsymbol{y}}_i^t}|{|^2}} - \underline{d_5}(\frac{2 k_{\tau} \tau}{\overline{\eta _{\boldsymbol{y}}}^2} + 3 k_{\tau} \tau NL^2)\sum\limits_{t = T_1 + 2}^{T_1 + {\widetilde{T}} (\epsilon )}\sum\limits_{i = 1}^N {||{{\boldsymbol{y}}_i^{t+1}} - {{\boldsymbol{y}}_i^t}|{|^2}},  
\end{array}
\end{equation}
where ${\sigma _3} \!=\! \max \{  ||{\lambda _1} - {\lambda _2}|| \}$, ${\sigma _4} = \max \{ ||{\boldsymbol{\theta}_1} - {\boldsymbol{\theta}_2}||\} $ and  $\mathop L\limits_ -  \! =  \! \mathop {\min } {L_p}(\{ {\boldsymbol{x}_i^t}\} , \! \{ {\boldsymbol{y}_i^t}\} ,\! \boldsymbol{v}^t,\! \boldsymbol{z}^t,\! \{ {\lambda _l^t}\} ,\!\{ {\boldsymbol{\theta }_i^t}\} )$, which satisfy that, $\forall t \ge T_1 + 2$,
{\begin{equation}
\label{eq:A87}
F^{t} \ge \mathop L\limits_ -   - \frac{4}{{{{\eta _{\lambda}}}}}\frac{{{c_1^1}}}{{{c_1^2}}}M {\alpha_3}  - \frac{4}{{{{\eta _{\boldsymbol{\theta }}}}}}\frac{{{c_2^1}}}{{{c_2^2}}}N {\alpha_4}  - \frac{7}{{2{{\eta _{\lambda}}}}}M {{\sigma _3}^2}  - \frac{7}{{2{{\eta _{\boldsymbol{\theta }}}}}}N {{\sigma _4}^2} - \frac{c_1^{T_1+2}}{2}M {{\sigma _3}^2} - \frac{c_2^{T_1+2}}{2}N {{\sigma _4}^2}.
\end{equation}}

\vspace{2ex}

For each worker $i$, we have that $\overline{{t}_i} -\hat{{t}}_i \le \tau $, thus,
{\begin{equation}
\label{eq:A87-0}
\begin{array}{l}
\sum\limits_{t = T_1 + 2}^{T_1 + {\widetilde{T}} (\epsilon )}3{\overline{d_5}}(({c_2^{\hat{{t}}_i - 1}}){^2} - ({c_2^{\overline{{t}_i}-1}}){^2})\alpha_4 \vspace{1ex}\\
\le \tau \sum\limits_{\scriptstyle{{\hat v}_i}(j) \in \mathcal{V}_i({\widetilde{T}}(\epsilon )),\hfill\atop 
\scriptstyle T_1+2 \le {{\hat v}_i}(j) \le T_1+{\widetilde{T}} (\epsilon ) \hfill} 3{\overline{d_5}}(({c_2^{{\hat v}_i(j) - 1}}){^2} - ({c_2^{{\hat v}_i(j+1)-1}}){^2})\alpha_4
\vspace{1ex}\\

\le 3\tau {\overline{d_5}}({c_2^1}){^2}\alpha_4.
\end{array}
\end{equation}}

Since the idle workers do not update their variables in each master iteration, for any $t$ that satisfies ${\hat v_i}(j - 1) \le t < {\hat v_i}(j)$, we have ${\boldsymbol{\theta}_i^t} = {\boldsymbol{\theta}_i^{{\hat v_i}(j) - 1}}$. And for $t \notin \mathcal{V}_i(T)$, we have ${||{{\boldsymbol{\theta }_i^t}} - {\boldsymbol{\theta }_i^{t-1}}|{|^2}}=0$. Combining with ${\hat v_i}(j) - {\hat v_i}(j - 1) \le \tau $, we can obtain that, 
{\begin{equation}
\label{eq:A87-1}
\begin{array}{l}
\sum\limits_{t = T_1 + 2}^{T_1 + {\widetilde{T}} (\epsilon )}\! {\sum\limits_{i = 1}^N\! {||{{\boldsymbol{\theta }_i}^{\overline {{t_j}}}} \!-\! {{\boldsymbol{\theta }_i^t}}|{|^2}} } 

\!\le\! \tau \sum\limits_{\scriptstyle{{\hat v}_i}(j) \in \mathcal{V}_i({\widetilde{T}}(\epsilon )),\hfill\atop
\scriptstyle T_1+3 \le {{\hat v}_i}(j)\hfill} {\sum\limits_{i = 1}^N {||{\boldsymbol{\theta}_i^{{{\hat v}_i}(j)}} - {\boldsymbol{\theta}_i^{{{\hat v}_i}(j) - 1}}|{|^2}} } 
\vspace{0.5ex}\\

\quad \quad \quad \quad \quad \quad  \quad \quad \quad \;  = \tau \! \sum\limits_{t = T_1 + 2}^{T_1 + {\widetilde{T}} (\epsilon )}\! {\sum\limits_{i = 1}^N\! {||{{\boldsymbol{\theta }_i}^{t+1}}\! -\! {{\boldsymbol{\theta }_i^t}}|{|^2}} }  + \tau \!\sum\limits_{t = {\widetilde{T}}(\epsilon ) + 1}^{{\widetilde{T}}(\epsilon ) + \tau  - 1} {\sum\limits_{i = 1}^N \!{||{{\boldsymbol{\theta }_i}^{t+1}} \!-\! {{\boldsymbol{\theta }_i^t}}|{|^2}} } \vspace{0.5ex}\\

\quad \quad \quad \quad \quad \quad  \quad \quad \quad \; \le \tau\! \sum\limits_{t = T_1 + 2}^{T_1 + {\widetilde{T}} (\epsilon )}\! {\sum\limits_{i = 1}^N\! {||{{\boldsymbol{\theta }_i}^{t+1}} \!-\! {{\boldsymbol{\theta }_i^t}}|{|^2}} }  + 4\tau (\tau  - 1)N\alpha_4.
\end{array}
\end{equation}}

Similarly, for any $t$ that satisfies ${\hat v_i}(j - 1) \le t < {\hat v_i}(j)$, we have ${\boldsymbol{x}_i^t} = {\boldsymbol{x}_i^{{\hat v_i}(j) - 1}}$, ${\boldsymbol{y}_i^t} = {\boldsymbol{y}_i^{{\hat v_i}(j) - 1}}$. And for $t \notin \mathcal{V}_i(T)$, we have ${||{{\boldsymbol{x}}_i^t} - {{\boldsymbol{x}}_i^{t-1}}|{|^2}}=0$, ${||{{\boldsymbol{y}}_i^t} - {{\boldsymbol{y}}_i^{t-1}}|{|^2}}=0$. Combining with ${\hat v_i}(j) - {\hat v_i}(j - 1) \le \tau $, we can get that,
{\begin{equation}
\label{eq:A87-2}
\begin{array}{l}
\sum\limits_{t = T_1 + 2}^{T_1 + {\widetilde{T}} (\epsilon )}\! {\sum\limits_{i = 1}^N\! {||{{\boldsymbol{x}}_i^{\overline {{t_j}}}} \!-\! {{\boldsymbol{x}}_i^t}|{|^2}} } 

\!\le\! \tau\! \sum\limits_{\scriptstyle{{\hat v}_i}(j) \in \mathcal{V}_i({\widetilde{T}}(\epsilon )),\hfill\atop
\scriptstyle T_1+3 \le {{\hat v}_i}(j)\hfill}\! {\sum\limits_{i = 1}^N\! {||{\boldsymbol{x}_i^{{{\hat v}_i}(j)}} - {\boldsymbol{x}_i^{{{\hat v}_i}(j) - 1}}|{|^2}} } 
\vspace{0.5ex}\\

 \quad \quad \quad \quad \quad \quad  \quad \qquad\,  = \tau \! \sum\limits_{t = T_1 + 2}^{T_1 + {\widetilde{T}} (\epsilon )}\! {\sum\limits_{i = 1}^N\! {||{{\boldsymbol{x}}_i^{t+1}}\! -\! {{\boldsymbol{x}}_i^t}|{|^2}} } \! +\! \tau\! \sum\limits_{t = {\widetilde{T}}(\epsilon )\! +\! 1}^{{\widetilde{T}}(\epsilon ) + \tau  - 1} {\sum\limits_{i = 1}^N\! {||{{\boldsymbol{x}}_i^{t+1}} \!-\! {{\boldsymbol{x}}_i^t}|{|^2}} } \vspace{0.5ex}\\

\quad \quad \quad \quad \quad \quad  \quad \qquad\,   \le \tau \! \sum\limits_{t = T_1 + 2}^{T_1 + {\widetilde{T}} (\epsilon )}\! {\sum\limits_{i = 1}^N\! {||{{\boldsymbol{x}}_i^{t+1}} \!-\! {{\boldsymbol{x}}_i^t}|{|^2}} }  + 4\tau (\tau  \!- \!1)N{\alpha _1}.
\end{array}
\end{equation}}

{\begin{equation}
\label{eq:A87-3}
\begin{array}{l}
\sum\limits_{t = T_1 + 2}^{T_1 + {\widetilde{T}} (\epsilon )}\! {\sum\limits_{i = 1}^N\! {||{{\boldsymbol{y}}_i^{\overline {{t_j}}}} \!-\! {{\boldsymbol{y}}_i^t}|{|^2}} } 

\!\le\! \tau\! \sum\limits_{\scriptstyle{{\hat v}_i}(j) \in \mathcal{V}_i({\widetilde{T}}(\epsilon )),\hfill\atop
\scriptstyle3 \le {{\hat v}_i}(j)\hfill}\! {\sum\limits_{i = 1}^N\! {||{\boldsymbol{y}_i^{{{\hat v}_i}(j)}} - {\boldsymbol{y}_i^{{{\hat v}_i}(j) - 1}}|{|^2}} } 
\vspace{0.5ex}\\

 \quad \quad \quad \quad \quad \quad  \quad \quad \quad \, = \tau \! \sum\limits_{t = T_1 + 2}^{T_1 + {\widetilde{T}} (\epsilon )}\! {\sum\limits_{i = 1}^N\! {||{{\boldsymbol{y}}_i^{t+1}}\! -\! {{\boldsymbol{y}}_i^t}|{|^2}} } \! +\! \tau\! \sum\limits_{t = {\widetilde{T}}(\epsilon )\! +\! 1}^{{\widetilde{T}}(\epsilon ) + \tau  - 1} {\sum\limits_{i = 1}^N\! {||{{\boldsymbol{y}}_i^{t+1}} \!-\! {{\boldsymbol{y}}_i^t}|{|^2}} } \vspace{0.5ex}\\

\quad \quad \quad \quad \quad \quad  \quad \quad \quad \,  \le \tau \! \sum\limits_{t = T_1 + 2}^{T_1 + {\widetilde{T}} (\epsilon )}\! {\sum\limits_{i = 1}^N\! {||{{\boldsymbol{y}}_i^{t+1}} \!-\! {{\boldsymbol{y}}_i^t}|{|^2}} }  + 4\tau (\tau  \!- \!1)N{\alpha _2}.
\end{array}
\end{equation}}

It follows from Eq. (\ref{eq:A86}), (\ref{eq:A87-0}), (\ref{eq:A87-1}), (\ref{eq:A87-2}) that,
\begin{equation}
\label{eq:A87-3}
\begin{array}{l}
\sum\limits_{t = T_1 + 2}^{T_1 + {\widetilde{T}} (\epsilon )} {{d_5^t}||\nabla {\widetilde{G}}^t|{|^2}} \vspace{1ex}\\
 
 \le F^{T_1 + 2} - \mathop L\limits_ -  
 + \frac{4}{{{{\eta _{\lambda}}}}}(\frac{{{c_1^0}}}{{{c_1^1}}} + \frac{{{c_1^1}}}{{{c_1^2}}})M {\alpha_3}  + \frac{{{c_1^1}}}{2}M {\alpha_3}  + \frac{7}{{2{{\eta _{\lambda}}}}}M {{\sigma _3}^2}  + 3{\overline{d_5}} {({c_1^1}){^2}M {\alpha_3} } \vspace{1ex}\\
 
 + \frac{4}{{{{\eta _{\boldsymbol{\theta }}}}}}(\frac{{{c_1^0}}}{{{c_1^1}}} + \frac{{{c_1^1}}}{{{c_1^2}}})N {\alpha_4}  + \frac{{{c_2^1}}}{2}N {\alpha_4}  + \frac{7}{{2{{\eta _{\boldsymbol{\theta }}}}}}N {{\sigma _4}^2}  + 3\tau{\overline{d_5}} {({c_2^1}){^2}N {\alpha_4} } + \frac{c_1^{T_1+2}}{2}M {{\sigma _3}^2} + \frac{c_2^{T_1+2}}{2}N {{\sigma _4}^2} \vspace{2ex}\\
 
 + (\frac{{2N\alpha_4}}{{5{{\eta _{\boldsymbol{\theta }}}}}} + 4{\overline{d_5}}(\frac{2}{{\eta _{\boldsymbol{x}}}^2} + 3N{L^2})N{\alpha _1}\tau +  4{\overline{d_5}}(\frac{2}{{\eta _{\boldsymbol{y}}}^2} + 3N{L^2})N{\alpha _2}\tau) (\tau  - 1) \vspace{0.5ex} \\
 = \mathop d\limits^ -  + k_d\tau(\tau -1),
\end{array}
\end{equation}
where $\mathop d\limits^ -$ and $k_d$ are constants. Constant $d_6$ is given by, 
{\begin{equation}
\label{eq:A88}
\begin{array}{l}
{d_6} = \max \{ {d_1},{d_2},{d_3},{d_4},\frac{{\frac{{30}}{{{{\eta _{\lambda}}}}} + 150{{\eta _{\lambda}}}\tau{ k_1 }N{L^2}}}{({1 - 30{{\eta _{\lambda}}}\tau{ k_1 }N{L^2}}){\underline{a_6}}},\frac{{30\tau}}{{{{\eta _{\boldsymbol{\theta }}}}{\underline{a_6}}}}\} \vspace{1ex}\\
 \ge \max \{ {d_1},{d_2},{d_3},{d_4},\frac{{\frac{{30}}{{{{\eta _{\lambda}}}}} + 150{{\eta _{\lambda}}}\tau{ k_1 }N{L^2}}}{({1 - 30{{\eta _{\lambda}}}\tau{ k_1 }N{L^2}}){a_6^t}},\frac{{30\tau}}{{{{\eta _{\boldsymbol{\theta }}}}{a_6^t}}}\} \vspace{1ex}\\
 = \frac{1}{{{d_5^t}{a_6^t}}}.
\end{array}
\end{equation}}

Thus, we can obtain that,
{\begin{equation}
\label{eq:A89}
\sum\limits_{t = T_1 + 2}^{T_1 + {\widetilde{T}} (\epsilon )} {\frac{1}{{{d_6}{a_6^t}}}}||\nabla \widetilde{G}^{T_1+\widetilde{T}(\epsilon )}|{|^2}  \le \sum\limits_{t = T_1 + 2}^{T_1 + {\widetilde{T}} (\epsilon )} {\frac{1}{{{d_6}{a_6^t}}}}||\nabla \widetilde{G}^t|{|^2}  \le \sum\limits_{t = T_1 + 2}^{T_1 + {\widetilde{T}} (\epsilon )} {{d_5^t}||\nabla \widetilde{G}^t|{|^2}}  \le \mathop d\limits^ -  + k_d\tau(\tau -1) .
\end{equation}}

And it follows from Eq. (\ref{eq:A89}) that,
{\begin{equation}
\label{eq:A90}
||\nabla \widetilde{G}^{T_1+\widetilde{T}(\epsilon )}|{|^2} \le \frac{{(\mathop d\limits^ -  + k_d\tau(\tau -1) )  {d_6}}}{{\sum\limits_{t = T_1 + 2}^{T_1 + {\widetilde{T}} (\epsilon )} {\frac{1}{{{a_6^t}}}} }}.  
\end{equation}}

According to the setting of ${c_1^t}$, ${c_2^t}$, we have,

{\begin{equation}
\label{eq:A91}
\frac{1}{{{a_6^t}}} \ge \frac{1}{{{{4(\gamma  - 2){L^2}(M{{\eta _{\lambda}}} + N{{\eta _{\boldsymbol{\theta }}}}){(t+1)^{\frac{1}{2}}} + \frac{{{{\eta _{\boldsymbol{\theta }}}}(N - S){L^2}}}{2}}}}}.
\end{equation}}

Summing up $\frac{1}{{{a_6^t}}}$ from ${t = T_1 +  2}$ to ${t = T_1 + {{\widetilde{T}} (\epsilon) }}$, it follows that,
{\begin{equation}
\label{eq:A92}
{\begin{array}{l}
\sum\limits_{t = T_1 + 2}^{T_1 + {\widetilde{T}} (\epsilon )} {\frac{1}{{{a_6^t}}}} 
 \ge \sum\limits_{t = T_1 + 2}^{T_1 + {\widetilde{T}} (\epsilon )} {\frac{1}{{{{4(\gamma  - 2){L^2}(M{{\eta _{\lambda}}} + N{{\eta _{\boldsymbol{\theta }}}}){(t+1)^{\frac{1}{2}}} + \frac{{{{\eta _{\boldsymbol{\theta }}}}(N - S){L^2}}}{2}}}}}} \vspace{1ex}\\
 \quad \quad \quad \; \; \; \; \;\, \ge \sum\limits_{t = T_1 + 2}^{T_1 + {\widetilde{T}} (\epsilon )} {\frac{1}{{{{4(\gamma  - 2){L^2}(M{{\eta _{\lambda}}} + N{{\eta _{\boldsymbol{\theta }}}}){(t+1)^{\frac{1}{2}}} + \frac{{{{\eta _{\boldsymbol{\theta }}}}(N - S){L^2}}}{2}{(t+1)^{\frac{1}{2}}}}}}}} \vspace{1ex}\\
 
\quad \quad \quad \; \; \; \; \;\, \ge \frac{(T_1+{\widetilde{T}} {(\epsilon )})^{^{\frac{1}{2}}} - (T_1+2)^{^{\frac{1}{2}}}}{{{{4(\gamma  - 2){L^2}(M{{\eta _{\lambda}}} + N{{\eta _{\boldsymbol{\theta }}}}) + \frac{{{{\eta _{\boldsymbol{\theta }}}}(N - S){L^2}}}{2}}}}}.
\end{array}}
\end{equation}}

The second inequality in Eq. (\ref{eq:A92}) is due to that $\forall t \ge T_1+2$, we have,
{\begin{equation}
\label{eq:A93}
{4(\gamma  \!-\! 2){L^2}(M{{\eta _{\lambda}}} \!+\! N{{\eta _{\boldsymbol{\theta }}}}){(t\!+\!1)^{\frac{1}{2}}}\! +\! \frac{{{{\eta _{\boldsymbol{\theta }}}}(N \!-\! S){L^2}}}{2} \!\le\! (4(\gamma \! - \!2){L^2}(M{{\eta _{\lambda}}}\! +\! N{{\eta _{\boldsymbol{\theta }}}}) \!+\! \frac{{{{\eta _{\boldsymbol{\theta }}}}(N \!-\! S){L^2}}}{2}){(t\!+\!1)^{\frac{1}{2}}}.}
\end{equation}}

The last inequality in Eq. (\ref{eq:A92}) follows from the fact that $\sum\limits_{t = T_1 + 2}^{T_1 + {\widetilde{T}} (\epsilon )} {\frac{1}{{{(t+1)^{\frac{1}{2}}}}}}  \!\ge\! (T_1+{\widetilde{T}} {(\epsilon )})^{^{\frac{1}{2}}} - (T_1+2)^{^{\frac{1}{2}}}$.

\vspace{1ex}

Thus, plugging Eq. (\ref{eq:A92}) into Eq. (\ref{eq:A90}), we can obtain:
{\begin{equation}
\label{eq:A94}
||\nabla {\widetilde{G}}^{T_1+{\widetilde{T}} (\epsilon )} |{|^2}  \le \frac{{\mathop (\mathop d\limits^ -  + k_d\tau(\tau\! -\!1))   {d_6}}}{{\sum\limits_{t = T_1 + 2}^{T_1 + {\widetilde{T}} (\epsilon )} {\frac{1}{{{a_6^t}}}} }}  \le  \frac{{{{(4(\gamma  \!-\! 2){L^2}(M{{\eta _{\lambda}}} \!+\! N{{\eta _{\boldsymbol{\theta }}}}) + \frac{{{{\eta _{\boldsymbol{\theta }}}}(N - S){L^2}}}{2})}}(\mathop d\limits^ -  + k_d\tau(\tau\! -\! 1))   {d_6}}}{(T_1+{\widetilde{T}} {(\epsilon )})^{^{\frac{1}{2}}} \!-\! (T_1+2)^{^{\frac{1}{2}}}}.
\end{equation}}

Let constant $d_7 = {{4(\gamma  - 2){L^2}(M{{\eta _{\lambda}}} + N{{\eta _{\boldsymbol{\theta }}}})}}$, and according to the definition of ${\widetilde{T}(\epsilon )}$, we have:
{\begin{equation}
\label{eq:A95}
T_1 + {\widetilde{T}} (\epsilon ) \ge {(\frac{{4{{(d_7 + \frac{{{{\eta _{\boldsymbol{\theta }}}}(N - S){L^2}}}{2})}}(\mathop d\limits^ -  + k_d\tau(\tau -1))   {d_6}}}{{{\epsilon}}} + (T_1 + 2)^{\frac{1}{2}})^2}.
\end{equation}}

Combining the definition of $\nabla G^t$ and $\nabla {\widetilde{G}}^t$ with trigonometric inequality, we then get:
{\begin{equation}
\label{eq:A96}
||\nabla G^t|| - ||\nabla {\widetilde{G}}^t|| \le ||\nabla G^t - \nabla {\widetilde{G}}^t|| \le \sqrt {\sum\limits_{l = 1}^{|{\boldsymbol{\mathcal{P}}^{t}}|} {||{c_1^{t-1}}{\lambda _l^t}|{|^2}}  + \sum\limits_{i = 1}^N {||{c_2^{t-1}}{{\boldsymbol{\theta}}_i^t}|{|^2}} }.
\end{equation}}

If $t \ge {(\frac{{4M\alpha_3}}{{{{\eta _{\lambda}}}^2}} + \frac{{4N\alpha_4}}{{{{\eta _{\boldsymbol{\theta }}}}^2}})^2}\frac{1}{{{\epsilon ^2}}}$, then we have $\sqrt {\sum\limits_{l = 1}^{|{\boldsymbol{\mathcal{P}}^{t}}|} {||{c_1^{t-1}}{\lambda _l^t}|{|^2}}  + \sum\limits_{i = 1}^N {||{c_2^{t-1}}{{\boldsymbol{\theta}}_i^t}|{|^2}} }  \le \frac{\sqrt{\epsilon} }{2}$. Combining it with Eq. (\ref{eq:A95}), we can conclude that there exists a
\begin{equation}
\label{eq:A97}
T( \epsilon ) \! \sim  \! \mathcal{O}(\max  \{ {(\frac{{4M\alpha_3}}{{{{\eta _{\lambda}}}^2}} \!+\! \frac{{4N\alpha_4}}{{{{\eta _{\boldsymbol{\theta }}}}^2}})^2}\frac{1}{{{\epsilon ^2}}}, 
{(\frac{{4{{{(d_7 + \frac{{{{\eta _{\boldsymbol{\theta }}}}(N  -  S){{L}^2}}}{2})}}} (\mathop d\limits^ -  +  k_d\tau(\tau  -  1))  {d_6}}}{{{\epsilon}}} + (T_1 + 2)^{\frac{1}{2}})^2}\}), 
\end{equation}
such that $||\nabla G^t||^2  \le \epsilon $, which concludes our proof.

\vspace{5mm}

\section{Proof of Theorem \ref{theorem: convergence}}
\label{appendix:theorem1}
Assuming that there are cutting planes added every $k$ iteration, \textit{i.e.}, 
\begin{equation}
\label{eq:19_new}
    {{\boldsymbol{\mathcal{P}}}^0} \supseteq {{\boldsymbol{\mathcal{P}}}^k} \supseteq  \cdots  \supseteq {{\boldsymbol{\mathcal{P}}}^{nk}}.
\end{equation}

Let $\mathcal{R}^k$ denote the feasible region of problem in Eq. (\ref{eq:new_22}) in $k^{\rm{th}}$ iteration, and let $\mathcal{R}'$ denote the feasible region of problem in Eq. (\ref{eq:new_21}), we have that,
\begin{equation}
\label{eq:917_135}
    \mathcal{R}^0 \supseteq \mathcal{R}^k \supseteq  \cdots  \supseteq \mathcal{R}^{nk} \supseteq \mathcal{R}'.
\end{equation}

Let $F(\{ {\boldsymbol{x}_i^{k*}}\} ,\{ {\boldsymbol{y}_i^{k*}}\} ,\boldsymbol{v}^{k*},\boldsymbol{z}^{k*})$ denote the optimal objective value of the problem in Eq. (\ref{eq:new_22}) in $k^{\rm{th}}$ iteration and let $F^*$ denote the optimal objective value of the problem in Eq. (\ref{eq:new_21}). According to Eq. (\ref{eq:917_135}), we have that,
\begin{equation}
\label{eq:20_new}
   F(\{ {\boldsymbol{x}_i^{0*}}\} ,\!\{ {\boldsymbol{y}_i^{0*}}\} ,\!\boldsymbol{v}^{0*},\!\boldsymbol{z}^{0*}) \!\le\! F(\{ {\boldsymbol{x}_i^{k*}}\} ,\!\{ {\boldsymbol{y}_i^{k*}}\} ,\!\boldsymbol{v}^{k*},\!\boldsymbol{z}^{k*}) \!\le\!  \cdots \! \le\! F(\{ {\boldsymbol{x}_i^{nk*}}\} ,\!\{ {\boldsymbol{y}_i^{nk*}}\} ,\!\boldsymbol{v}^{nk*},\!\boldsymbol{z}^{nk*}).
\end{equation}

And we can obtain that,
\begin{equation}
\label{eq:21_new}
    \frac{{F^*}}{F(\{ {\boldsymbol{x}_i^{0*}}\} ,\!\{ {\boldsymbol{y}_i^{0*}}\} ,\!\boldsymbol{v}^{0*},\!\boldsymbol{z}^{0*})} \!\ge\!  \frac{{F^*}}{F(\{ {\boldsymbol{x}_i^{k*}}\} ,\!\{ {\boldsymbol{y}_i^{k*}}\} ,\!\boldsymbol{v}^{k*},\!\boldsymbol{z}^{k*})} \!\ge\!  \cdots \! \ge \! \frac{{F^*}}{F(\{ {\boldsymbol{x}_i^{nk*}}\} ,\!\{ {\boldsymbol{y}_i^{nk*}}\} ,\!\boldsymbol{v}^{nk*},\!\boldsymbol{z}^{nk*})}  \!\ge\! \beta.
\end{equation}

It is seen from Eq. (\ref{eq:21_new}) that the sequence $\{\frac{{F^*}}{F(\{ {\boldsymbol{x}_i^{k*}}\} ,\{ {\boldsymbol{y}_i^{k*}}\} ,\boldsymbol{v}^{k*},\boldsymbol{z}^{k*})}\}$ is monotonically non-increasing. When $nk \to \infty$, the optimal objective value of the problem in Eq. (\ref{eq:new_22}) monotonically converges to $\beta$ ($\beta \ge 1$).

\renewcommand\arraystretch{1.3}
\renewcommand\tabcolsep{15pt}
\begin{table*}[t]
\centering
\renewcommand{\thetable}{\arabic{table}}
\caption{Step-sizes of all variables in the experiments.}
{
\scalebox{0.95}{
\begin{tabular}{l|c|c|c|c|c|c}
\toprule
Datasets    & ${\eta _{\boldsymbol{x}}}$     & ${\eta _{\boldsymbol{y}}}$ & ${\eta _{\boldsymbol{v}}}$  & ${\eta _{\boldsymbol{z}}}$  &  ${\eta _{\lambda}}$   &  ${\eta _{\boldsymbol{\theta}}}$     \\ \hline

MNIST & 0.001  & 0.02 & 0.001 & 0.02 & 0.1 & 0.001 \\ 
Fashion MNIST & 0.001  & 0.02 & 0.001 & 0.02 & 0.1 & 0.001\\ 
CIFAR-10 & 0.001  & 0.02 & 0.001 & 0.02 & 0.1 & 0.001 \\ 
Covertype & 0.01  & 0.02 & 0.01 & 0.02 & 0.1 & 0.01 \\ 
IJCNN1 & 0.01  & 0.005 & 0.01 & 0.005 & 0.1 & 0.01 \\
Australian & 0.001  & 0.02 & 0.001 & 0.02 & 5 & 0.001 \\
\bottomrule  
\end{tabular}}
\label{tab:step-sizes}}
\end{table*}

\newpage

\section{Details of experiments}

\subsection{Additional results}
In this section, additional experiment results on CIFAR-10 \citep{krizhevsky2009learning} and Australian \citep{quinlan1987simplifying} datasets are reported in Figure \ref{fig:cifar-10} and Figure \ref{fig:Australian}. It is seen from Figure \ref{fig:cifar-10} and Figure \ref{fig:Australian} that the proposed ADBO also achieves faster convergence rate.

\subsection{Details of experiments}\label{detail experiment}
In this section, we provide more details of the experimental setup in this work. In data hyper-cleaning task, experiments are carried out on MNIST, Fashion MNIST and CIFAR-10 datasets. Following \citep{ji2021bilevel}, we utilize the same model in data-hypercleaning task for MNIST, Fashion MNIST and CIFAR-10 datasets, and SGD optimizer is utilized. And the step-sizes are summarized in Table \ref{tab:step-sizes}. In MNIST and Fashion MNIST datasets, we set $N=18$, $S=9$, $\tau=15$. And in CIFAR-10 dataset, we set $N=18$, $S=9$, $\tau=5$. We set that the (communication + computation) delays of each worker obey log-normal distribution ${\rm{LN}}(3.5,1)$.

In regularization coefficient optimization task, experiments are carried out on Covertype, IJCNN1 and Australian datasets. Following \citep{chen2022single}, we utilize the same logistic regression model, and SGD optimizer is used. And the step-sizes are summarized in Table \ref{tab:step-sizes}. In Covertype dataset, we set $N = 18$, $S = 9$, $\tau =15$; in IJCNN1 dataset, we set $N = 24$, $S = 12$, $\tau =15$; and in Australian dataset, we set $N = 4$, $S = 2$, $\tau =5$.  In the experiments that consider straggler problems, three
stragglers are set in the distributed system, and the mean of (communication + computation) delay of stragglers is four times the delay of normal workers.

{\makeatletter\def\@captype{figure}\makeatother 
\centering    
\subfigure[test accuracy vs time] 
{\begin{minipage}[t]{0.38\linewidth}
	\centering      
	\includegraphics[scale=0.32]{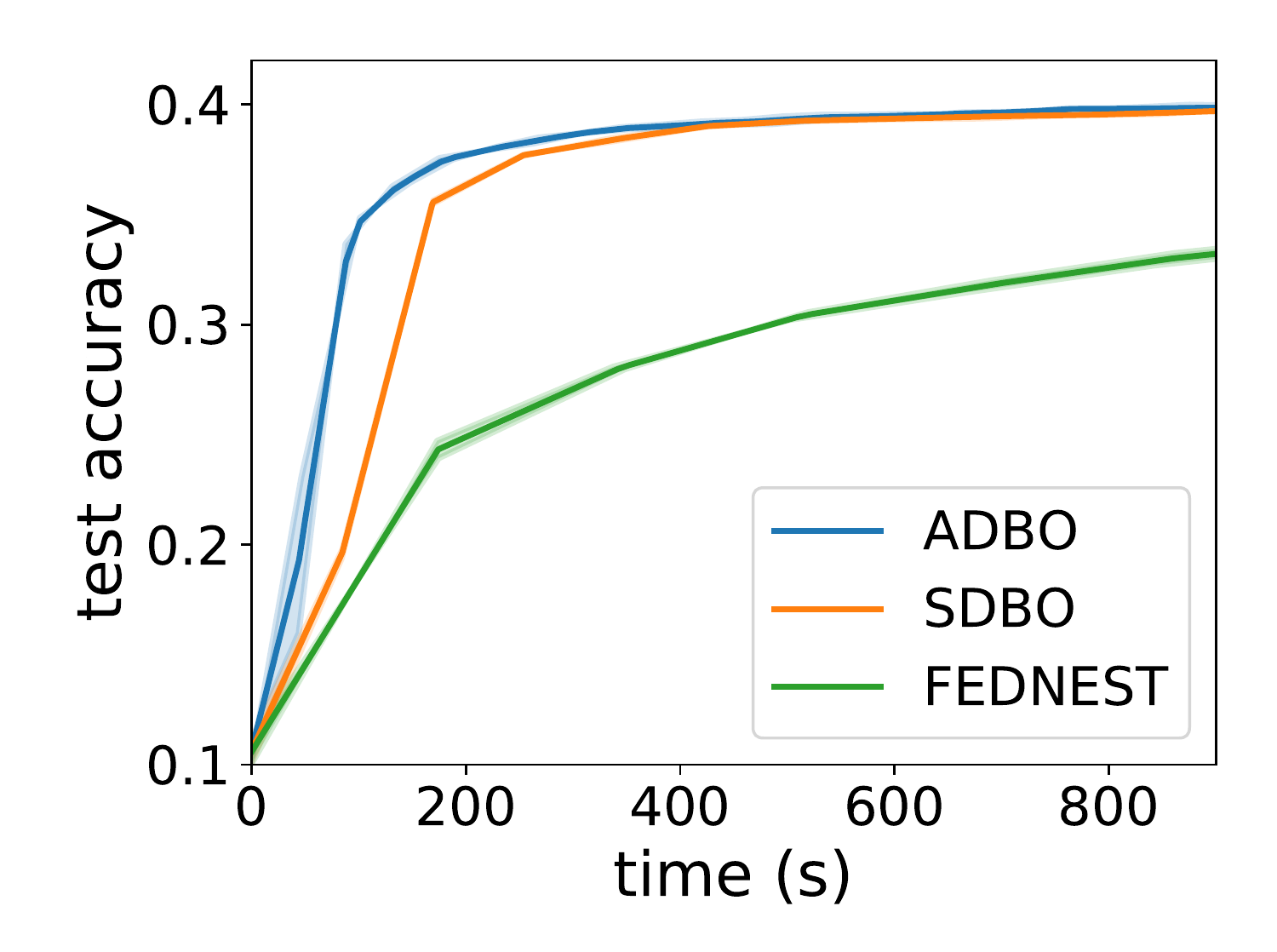}   
	\end{minipage}}
\subfigure[test loss vs time] 
{\begin{minipage}[t]{0.38\linewidth}
	\centering      
	\includegraphics[scale=0.32]{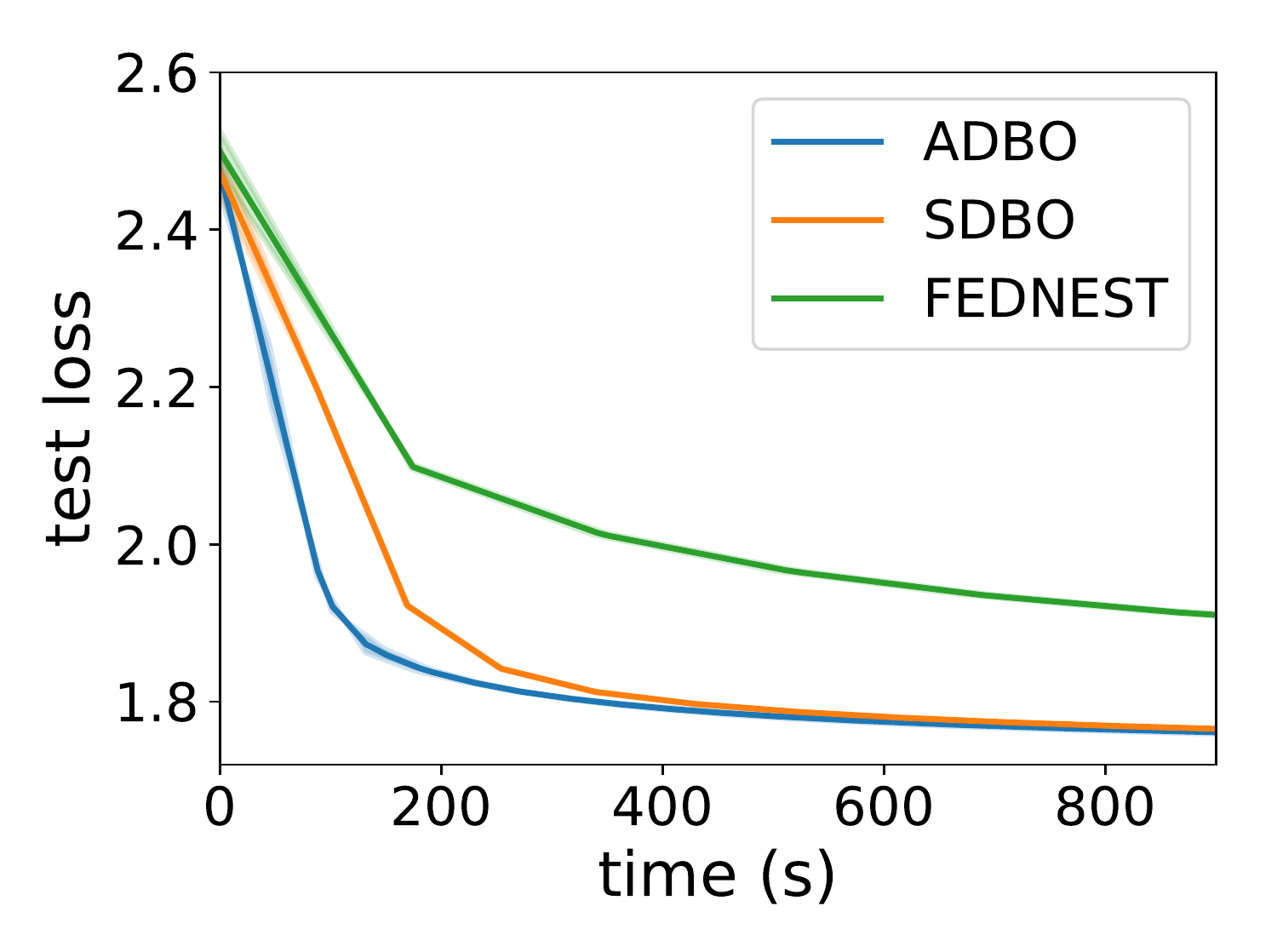}   
	\end{minipage}}
\caption{(a) Test accuracy vs time and (b) Test loss vs time on CIFAR-10 dataset on distributed data hyper-cleaning task.} 
\label{fig:cifar-10}  
}

{\makeatletter\def\@captype{figure}\makeatother 
\centering    
\subfigure[test accuracy vs time] 
{\begin{minipage}[t]{0.38\linewidth}
	\centering      
	\includegraphics[scale=0.32]{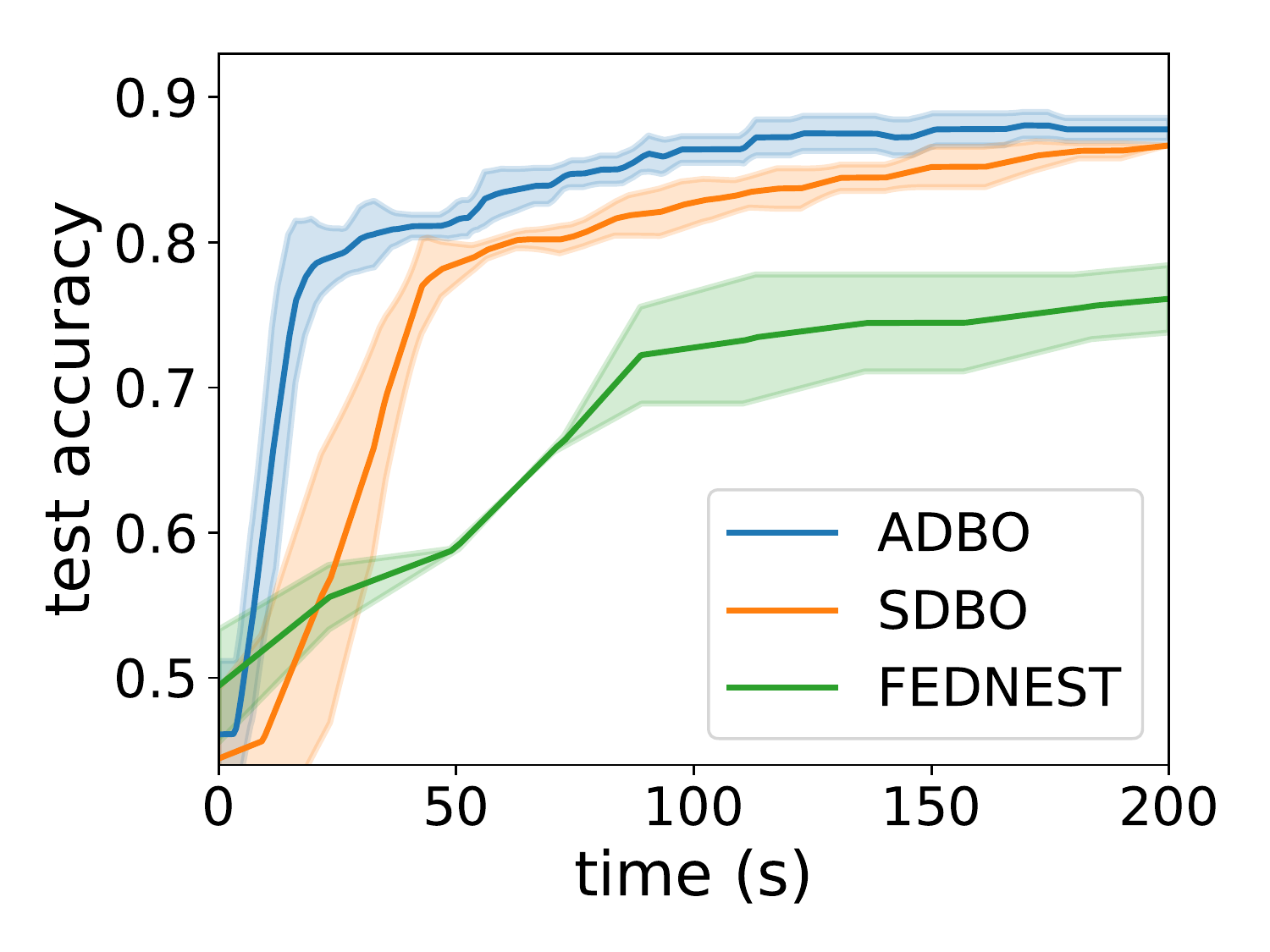}   
	\end{minipage}}
\subfigure[test loss vs time] 
{\begin{minipage}[t]{0.38\linewidth}
	\centering      
	\includegraphics[scale=0.32]{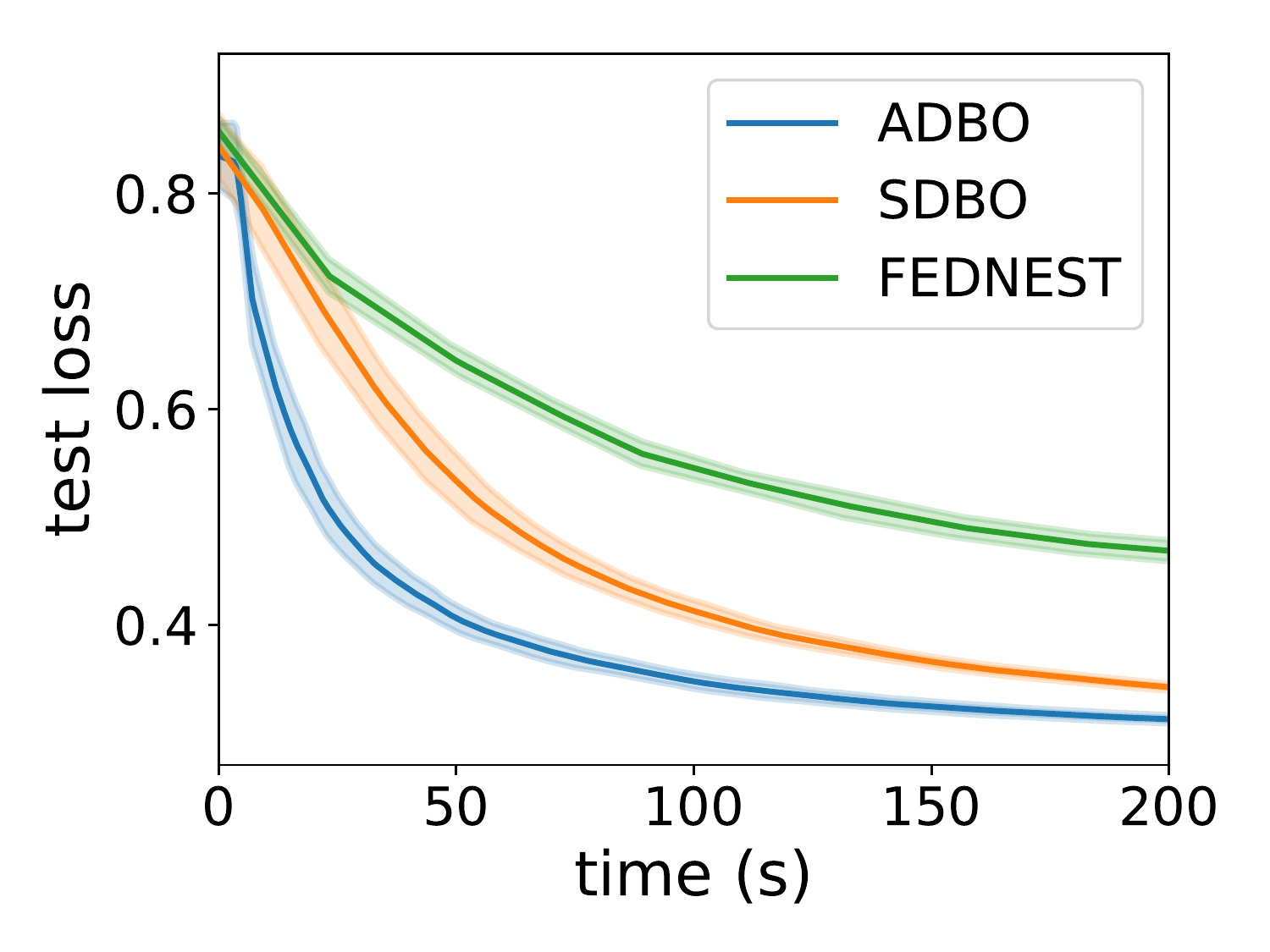}   
	\end{minipage}}
\caption{(a) Test accuracy vs time and (b) Test loss vs time on Australian dataset on  distributed regularization coefficient optimization task.} 
\label{fig:Australian}  
}

\textcolor{black}{Codes are available in \url{https://github.com/ICLR23Submission6251/adbo}.}

\newpage

\section{Parameter Server Architecture}
\begin{figure}[h]  
\begin{center}
\includegraphics[height=0.46\textwidth,scale=1]{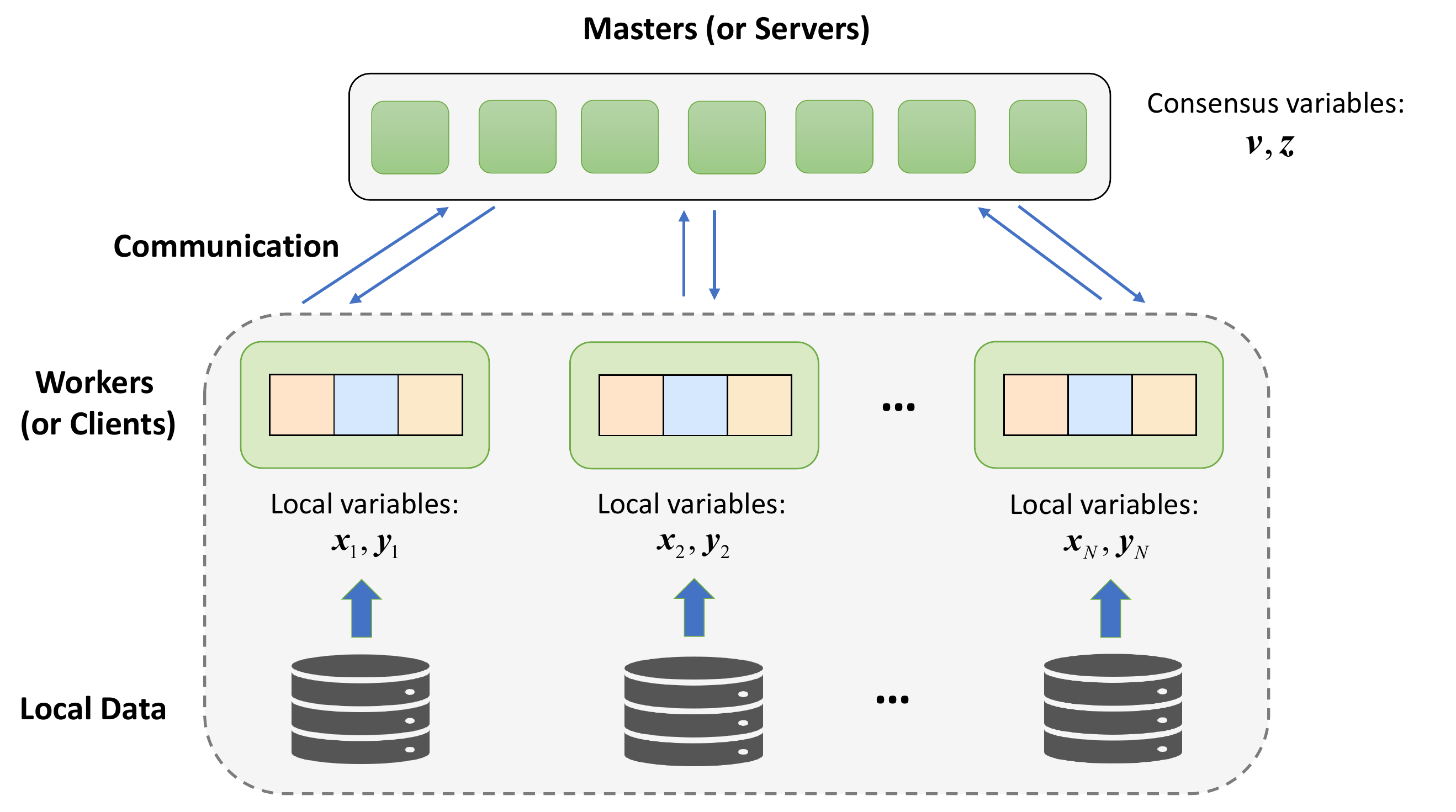} 
\caption{The illustration of parameter server architecture.} 
\label{fig:PS}
\end{center}
\end{figure}

In this section, we give the illustration of the parameter server architecture, which is shown in Figure \ref{fig:PS}. In parameter server architecture, the
communication is centralized around a set of master nodes (or servers) that constitute the hubs of a star network, and worker nodes (or clients) pull the
shared parameters from and send their updates to the master nodes.

\end{document}